\documentclass[10pt,twocolumn,letterpaper]{article}

\usepackage{iccv}
\usepackage{times}
\usepackage{epsfig}
\usepackage{graphicx}
\usepackage{amsmath}
\usepackage{amssymb}
\usepackage{multirow}
\usepackage{gensymb}
\usepackage{subcaption}
\usepackage{makecell}
\usepackage{animate}
\usepackage{xspace}
\usepackage[accsupp]{axessibility}  % Improves PDF readability for those with disabilities

\newcommand{\zhpcui}[1]{#1}
\newcommand{\yz}[1]{#1}
\newcommand{\zc}[1]{#1}

\usepackage[colorlinks=true,allcolors=black,urlcolor=blue,pagebackref=true,breaklinks=true,bookmarks=false]{hyperref}
\let\orgautoref\autoref

\renewcommand{\autoref}[1]{\def\figureautorefname{Fig.}\def\sectionautorefname{Sec.}\def\subsectionautorefname{Sec.}\def\equationautorefname{Equation}\def\tableautorefname{Tab.}\orgautoref{#1}}

\iccvfinalcopy % *** Uncomment this line for the final submission

\def\iccvPaperID{7881} % *** Enter the ICCV Paper ID here

\ificcvfinal\fi

\begin{document}

%%%%%%%%% TITLE
\title{DeepPanoContext: Panoramic 3D Scene Understanding with \\ Holistic Scene Context Graph and Relation-based Optimization}

\author{
    \zc{Cheng Zhang$^{1}$ \ Zhaopeng Cui$^{2}$\thanks{Corresponding author} \ Cai Chen$^{1}$ \ Shuaicheng Liu$^{1}$\footnotemark[1] \ Bing Zeng$^{1}$ \ Hujun Bao$^{2}$ \ Yinda Zhang$^{3}$\footnotemark[1]}\\
    \zc{$^{1}$ University of Electronic Science and Technology of China \\
    \zc{$^{2}$ State Key Lab of CAD \& CG, Zhejiang University} \quad $^{3}$ Google} \\
}

\maketitle
\ificcvfinal\thispagestyle{empty}\fi

\begin{abstract}
Panorama images have a much larger field-of-view thus naturally encode enriched scene context information compared to standard perspective images, which however is not well exploited in the previous scene understanding methods. 
In this paper, we propose a novel method for panoramic 3D scene understanding which recovers the 3D room layout and the shape, pose, position, and semantic category for each object from a single full-view panorama image. In order to fully utilize the rich context information, we design a novel 
\yz{graph neural network based context model}
to predict the relationship among objects and room layout, and a differentiable relationship-based optimization module to optimize object arrangement with well-designed objective functions on-the-fly.
Realizing the existing data are either with incomplete ground truth or overly-simplified scene, we present a new synthetic dataset with 
\yz{good diversity in room layout and furniture placement, and realistic image quality}
for total panoramic 3D scene understanding. Experiments demonstrate that our method outperforms existing methods on panoramic scene understanding 
in terms of 
both geometry accuracy and object arrangement.  
Code is available at \href{https://chengzhag.github.io/publication/dpc/}{https://chengzhag.github.io/publication/dpc}.
\end{abstract}

\vspace{-1em}
\section{Introduction}

Image-based holistic 3D indoor scene understanding is a long-lasting challenging problem in computer vision, due to scene clutter and 3D ambiguity in perspective geometry.
Over decades, the scene context\yz{,}
\yz{which}
encodes high-order relations across multiple objects following certain design rules\yz{,} has been 
\yz{widely}
utilized to improve the scene understanding \cite{zhang2014panocontext, chen2019holistic++}.
However, it is still arguable and unclear if the top-down context is more or less important than bottom-up local appearance-based approaches for the scene parsing task, especially with the rapidly emerging deep learning methods that have achieved great success on object classification and detection.
One possible reason could be that the field of view of a \zhpcui{standard} camera photo is \zhpcui{normally} less than 60\degree, and thus only limited context can be utilized 
\yz{among} a small number of objects 
\yz{co-existing} in the image.
Zhang \etal \cite{zhang2014panocontext} proposed a 3D scene parsing method that takes a 360\degree~full-view panorama as the input, where almost all \yz{major} objects are visible.
They showed that the context became significantly stronger with more objects in the same image, which 
\yz{enables} accurate 3D scene understanding even 
\yz{with less engineered local features.}

\begin{figure}[t]
    \vspace{-0.5em}
	\centering
	\begin{subfigure}[t]{0.23\textwidth}
		\includegraphics[width=\textwidth]  
		{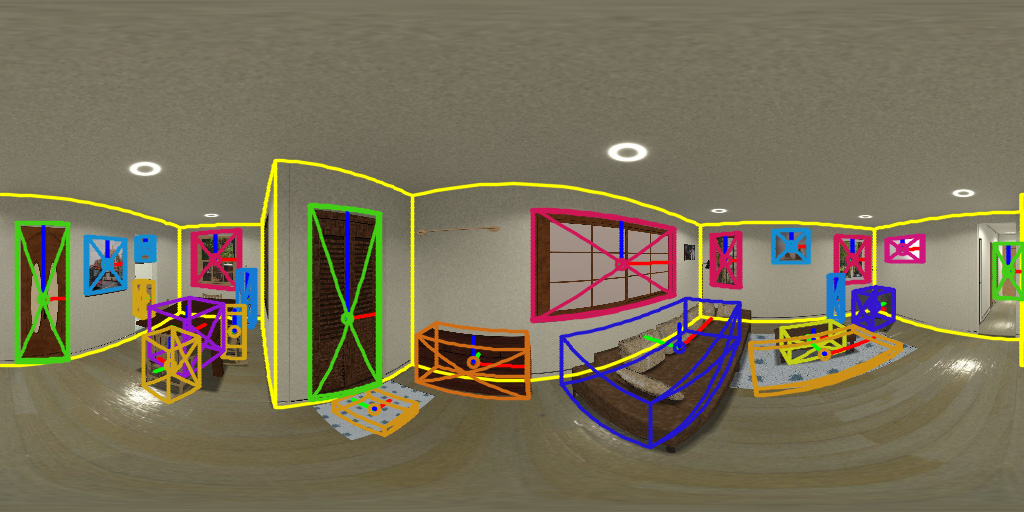}
		\includegraphics[width=\textwidth]
		{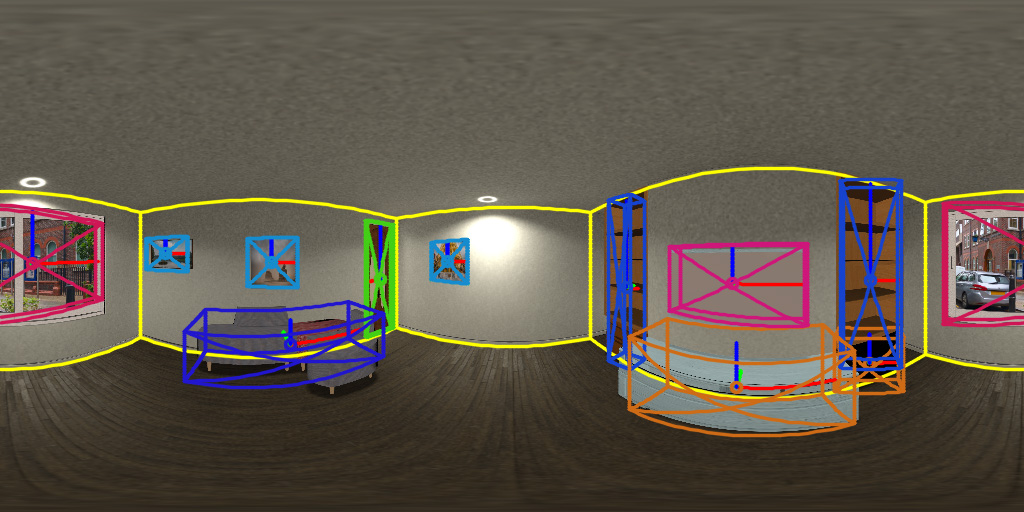}
        	\end{subfigure}
	\begin{subfigure}[t]{0.23\textwidth}
		\includegraphics[width=\textwidth]  
		{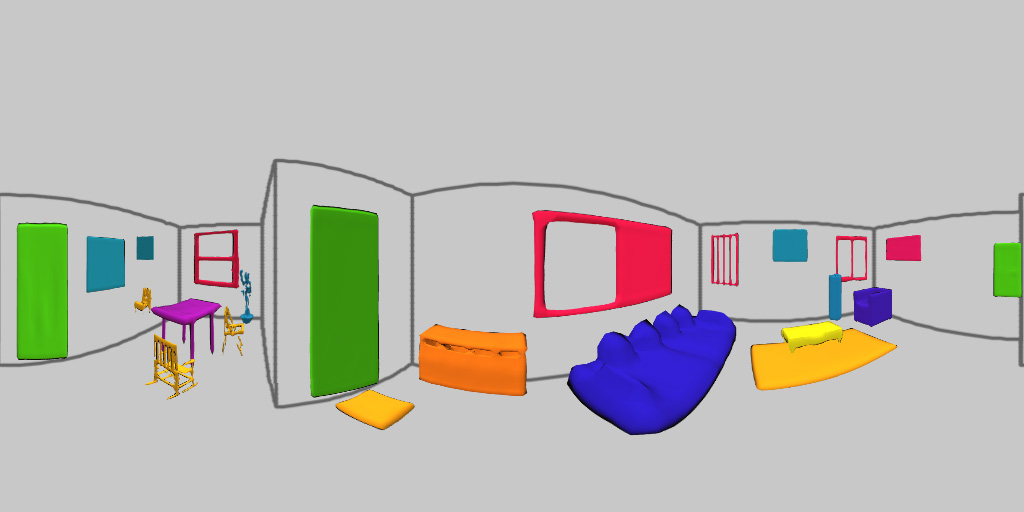}
		\includegraphics[width=\textwidth]
		{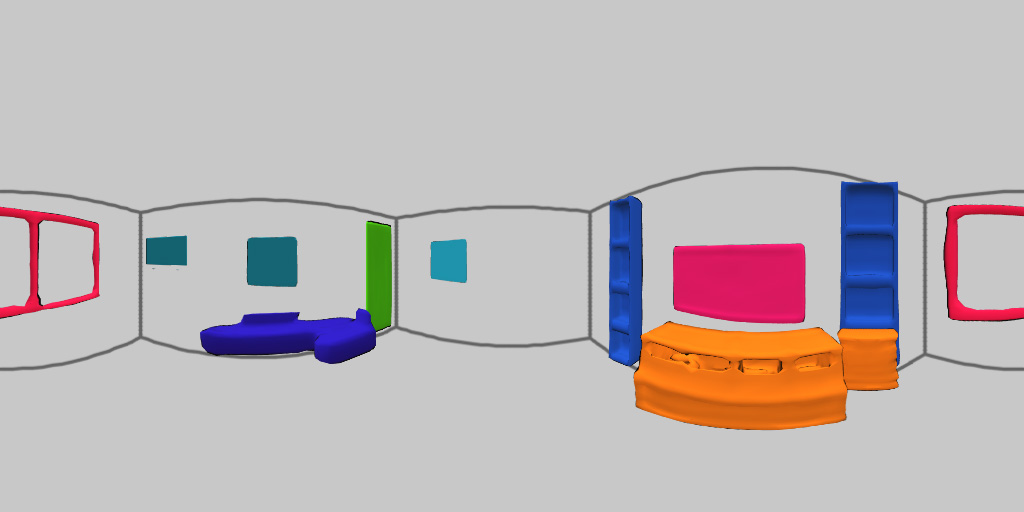}
        	\end{subfigure}
	\vspace{-0.5em}
	\caption{From a single panorama image as input, our proposed pipeline estimates layout and object poses, then reconstructs the scene with object reconstruction, to achieve total scene understanding.}
	\label{fig:teaser}
\end{figure}

In this paper, we 
\yz{empower} the panoramic scene understanding task with stronger 3D perception capability and aim to predict the objects' shapes, 3D poses, and semantic categories as well as the room layout by taking a single color full-view panorama image as the input.
To fulfill this goal, we propose a novel deep learning based framework that leverages both local image information and global context for panoramic 3D scene understanding.
Specifically, we first extract room layout and object hypothesis from local image regions with \zhpcui{the} algorithms customized for panorama images, and rely on a global graph-based context model to effectively refine the initial estimations. 
Overall, our method achieves phenomenal performance on both geometry accuracy and object arrangement for 3D panoramic scene understanding.

Besides renovating the predecessor \cite{zhang2014panocontext} with a more advanced deep learning algorithm, the key to the significant performance gain is 
\yz{\textbf{a novel context model that predicts relations across objects and room layout including supporting, attaching, relative orientation, \etc, which are then fed into an optimization to adjust the object arrangement.}}
This is inspired by the common sense that we, humans, tend to place objects tightly against the wall, \eg, beds, or side-by-side \yz{with consistent orientation}, \eg, nightstands, and these relations could provide critical information to fix the object arrangement errors that may be minor \yz{in traditional metrics} but obviously wrong judged by human perception.
To leverage 
\yz{the} predicted relations, we propose a novel differentiable optimization with carefully designed objective functions to adjust the initial object arrangement \wrt the predicted relations, which further enables joint training of relation prediction and object arrangement.
\yz{The optimization is fully differentiable, which can be attached with our graph based context model, and conceptually any neural network, for joint training.}

Unfortunately, the panoramic scene datasets for holistic 3D scene parsing are still missing in the literature. 
Existing panorama datasets are either with overly simplified scenes \cite{zhang2014panocontext}, purely 2D-based \cite{xiao2012recognizing}, or missing important 3D ground truth such as object poses \cite{armeni2017joint,chang2017matterport3d}.
Since annotating real data with accurate 3D shapes is extremely challenging,
we resort to synthetic data and create a new dataset for holistic panoramic 3D scene understanding.
The dataset provides high-quality ground truth for object location, pose, shape, and pairwise relations, and serves well for training and rigorous evaluation.
Though purely synthetic, we find the learned context model, which relies mainly on 
\yz{indoor scene context} but not heavily on the image appearance, can be naturally generalized to real images by retraining bottom-up models 
\yz{that provide} the initialization.

In summary, our contributions are as follows.
We propose the first deep learning based pipeline for holistic 3D scene understanding that recovers 3D room layout and detailed shape, pose, location for objects in the scene from a single color full-view panorama image.
To fully exploit context, we design a novel context model that predicts the relationship among objects and room layout, followed by a new differentiable relationship-based optimization module to refine the initial results.
To learn and evaluate our model, a new dataset is created for total panoramic 3D scene understanding. Our model achieves the state-of-the-art performance on both geometry accuracy and 3D object arrangement.

\section{Related Work}

\textbf{3D Scene Understanding}
Scene understanding in 3D world is a trending topic in the vision community. 
The task includes a volley of interesting sub-tasks including layout estimation, 3D object detection and pose estimation, and shape reconstruction. 
Various methods estimate the layout by adopting 
\zc{Manhattan World assumption}~\cite{ramalingam2013manhattan, coughlan1999manhattan, sun2019horizonnet, zou2018layoutnet, yang2019dula} or cuboid assumption~\cite{dasgupta2016delay, mallya2015learning, lee2009geometric, hedau2009recovering}. 
\yz{3D bounding boxes and object poses can be predicted from 2D representation with CNN-based methods~\cite{chen2019holistic++, huang2018cooperative, du2018learning, wald2019rio, tremblay2018deep, bregier2017symmetry}.}
Object shapes can also be recovered by matching similar models, with geometrical or implicit representations~\cite{genova2020local, li2017grass, lin2018learning, wu20153d, groueix2018papier, izadinia2017im2cad, hueting2017seethrough, huang2018holistic}. 

Total3D~\cite{nie2020total3dunderstanding} \yz{is the first work to jointly solve multiple scene understanding tasks, including estimating} the scene layout, object poses, and shapes.
Recently, Zhang~\etal~\cite{zhang2021holistic} improves the performance of all three tasks via the implicit function and scene graph neural network. However, they still suffer from the insufficient exploitation of relationships among objects in the scene. 
In this work, we study the problem \yz{using} panorama images which contain rich context information compared with the perspective ones \yz{with limited field of views}.

\textbf{Context for Scene Understanding}
Context priors can be employed for the scene understanding, \eg, a bed is placed on the floor and aligned with the wall. 
\zc{For perspective images,} some methods~\cite{del2012bayesian, del2013understanding} adopt explicit constraints to avoid object overlaps.
\zc{Zhang \etal~\cite{zhang2017deepcontext} proposes to exploit scene context with a 3D context network.}
\zc{Recently, panoramic images have been exploited by}
optimization-based methods~\cite{pintore2016omnidirectional, fukano2016room, xu2017pano2cad, yang2016efficient, yang2018automatic} designed over geometric or semantic cues, and learning-based methods~\cite{zou2018layoutnet, lee2017roomnet, yang2019dula, sun2019horizonnet} with drastically advantageous representation of local context. 
\zc{Zhang \etal~\cite{zhang2014panocontext} achieves several tasks of 3D scene understanding by generating 3D hypotheses based on contextual constraints to exploit  rich context information provided from large field of view (FOV).}
However, 
\yz{none of them provides a complete understanding of the scene.}
\yz{Instead,}
we propose a learning-based framework to jointly predict object
\yz{shapes, 3D poses, semantic categories, and the room layout from a single panorama image, which takes full advantages of the scene context.}

\textbf{Panoramic Dataset}
For real-world scenes, the first panorama dataset is published by \zhpcui{Xiao \etal}~\cite{xiao2012recognizing}, namely SUN360, and is later annotated for indoor scene understanding by Zhang \etal~\cite{zhang2014panocontext}. It contains high-resolution color panoramas with diverse objects, layout, and axis-aligned object boxes. However, it lacks object poses as well as shapes and only includes 700 images which is not adequate for the neural network training. 2D-3D-S~\cite{armeni2017joint} and Matterport3D~\cite{chang2017matterport3d} are also real-world datasets with more data and richer annotations, but poses are absent. Some datasets~\cite{chou2020360, yang2018object} with the bounding FOV annotations are published for the purpose of panoramic object detection. Recently, a large photo-realistic dataset is proposed for structured 3D modeling, namely Structured3D~\cite{zheng2019structured3d}, but mesh ground truths are not published. 
\yz{A panoramic scene dataset contains complete ground truth, including shape, object arrangement, and room layout is still missing.}

\begin{figure*}
    \vspace{-1em}
	\centering
	\includegraphics[width=\textwidth]  
		{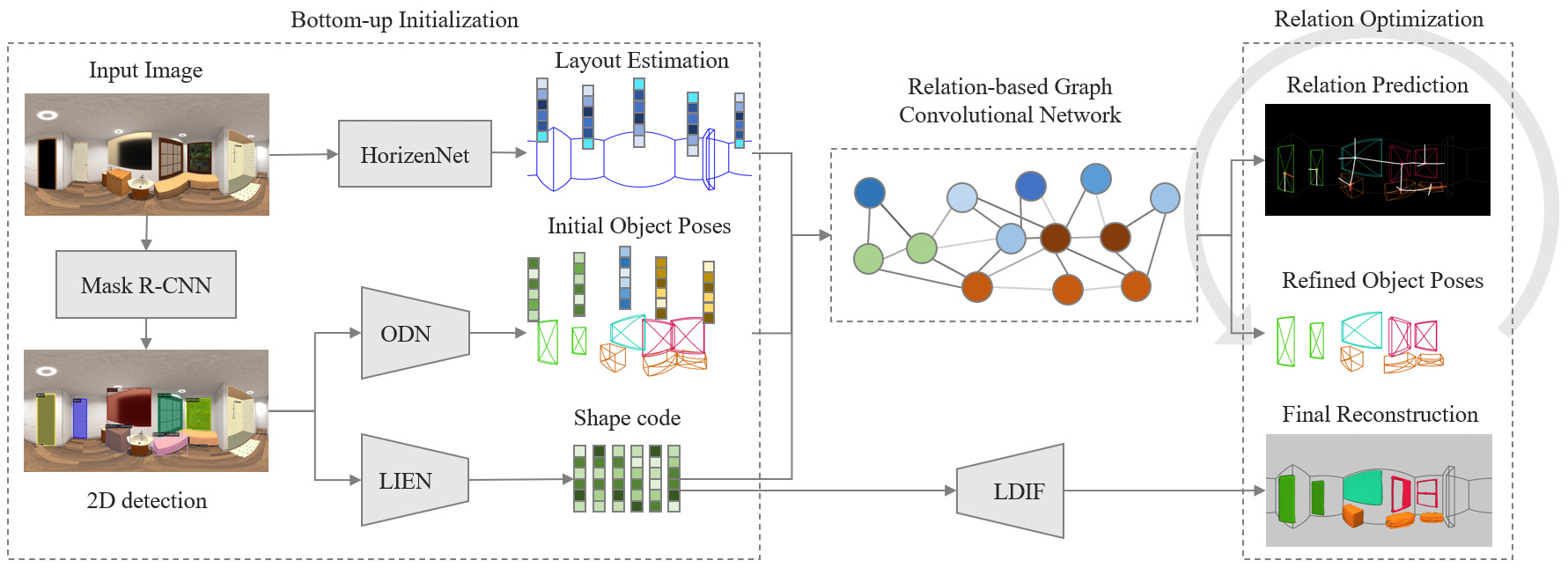}
		\vspace{-1.8em}
	\caption{
	\zc{Our proposed pipeline. We first do a bottom-up initialization with several SoTA methods \cite{zhang2021holistic, nie2020total3dunderstanding, he2017mask, sun2019horizonnet} and provide}
	\yz{various features, including geometric, semantic, and appearance features of objects and layout. These are then fed into our proposed RGCN network to refine the initial object pose and estimate the relation among objects and layout.}
    \yz{A relation optimization is adopted afterward to further adjust the 3D object arrangement to align with the 2D observation, conform with the predicted relation, and resolve physical collision.}
    }
	\vspace{-1.2em}
	\label{fig:pipeline}
\end{figure*}

\section{Method}
In this section, we introduce our method for 3D panoramic scene understanding. As shown in \autoref{fig:pipeline}, we first extract \zhpcui{the} whole-room layout under Manhattan World assumption and the initial object estimates including locations, sizes, poses, semantic categories, and latent shape codes. 
These, along with extracted features, are then fed into the Relation-based Graph Convolutional Network (RGCN) for refinement and \zhpcui{to} estimate relations among objects and layout simultaneously.
Then, a differentiable Relation Optimization (RO) based on physical violation, observation, and relation is \zhpcui{proposed} to resolve collisions and adjust object poses.
Finally, the 3D shape is recovered by feeding the latent shape code into Local Implicit Deep Function (LDIF) \cite{genova2020local}, \zhpcui{and} combined with object pose and room layout to achieve total scene understanding. 

\subsection{Bottom-up Initialization}
\zhpcui{We first estimate the room layout, initial objects' poses and shape codes}
for the panoramic scene from local image appearance. 
Similar to 
Zhang \etal \cite{zhang2021holistic}, we run a Mask R-CNN to detect 2D objects, 
an Object Detection Network (ODN) \cite{nie2020total3dunderstanding} to generate initial pose, 
and a Local Implicit Embedding Network (LIEN) \cite{zhang2021holistic} to embed implicit \yz{3D} representation for each object.
All the networks are retrained or customized for \zc{equirectangular panorama images}.

Specifically, we first fine-tune the Mask R-CNN on our data such that it learns to handle the distortion and runs directly on panorama.
We then fit a bounding box for 
each detected object \yz{mask} represented as a Bounding FoV (BFoV) \cite{chou2020360, yang2018object}, which is defined with the latitude and longitude of the center and the horizontal and vertical field of view.
Since the left and right borders \yz{of full-view panorama images} are actually connected, we extend the panorama by half of the width (concatenating the left half to the right) before feeding into the detector, then offset the detections of the extended part back to the left, following a standard non-maximum suppression (NMS) to merge overlapped or cross-border object detection.
Images in each BFoV are then projected to the perspective view and fed into ODN and LIEN for 3D pose and latent shape representation.
\zhpcui{Note that for simplicity we assume that the object only rotates around $y$ axis and ODN predicts the yaw angle of the object in the cropped perspective image coordinate as the object rotation.}
\yz{We empirically found this representation benefit the pose estimation performance,}
and the result can easily be converted to panorama (\ie, world) coordinates.
Regarding the room layout, we use the SoTA HorizonNet \cite{sun2019horizonnet}.

\subsection{Relation-based Graph Convolutional Network}
\label{sec:rgcn}
After having the initial estimation, similar to Zhang \etal \cite{zhang2021holistic}, we model the whole scene with a graph and refine the results via a Graph R-CNN \cite{yang2018graph}.
Thanks to the full-view panorama, our GCN 
\yz{can now model} all the objects in the room, which is able to encode and leverage stronger context than that in a perspective view \cite{zhang2021holistic}.
Different than Zhang \etal \cite{zhang2021holistic}, our model not only 
\yz{refines object poses} but also predicts relations between objects and room layouts. Therefore we call our model Relation-based Graph Convolutional Network (RGCN).

\noindent \textbf{Graph Construction}
Besides modeling each object as a node as in Zhang \etal \cite{zhang2021holistic}, we further represent each wall in the 
\yz{estimated room layout via HorizonNet }
into a cuboid with a certain thickness and model them as separate nodes.
This \yz{facilitates the learning of the}
relation between objects to each wall without additional complexity.
For each pair of wall/object nodes, we connect them with an undirected edge to form a complete graph with self circles.
Then two relation nodes with directed edges are added to connect the wall/object nodes.
Each node\yz{,} including wall, object, and relation\yz{,} is embedded with a 
\yz{latent} vector, which is updated by GCN through message passing \cite{zhang2021holistic,yang2018graph}.

\begin{figure}[t]
	\centering
	\includegraphics[width=0.33\textwidth]
		{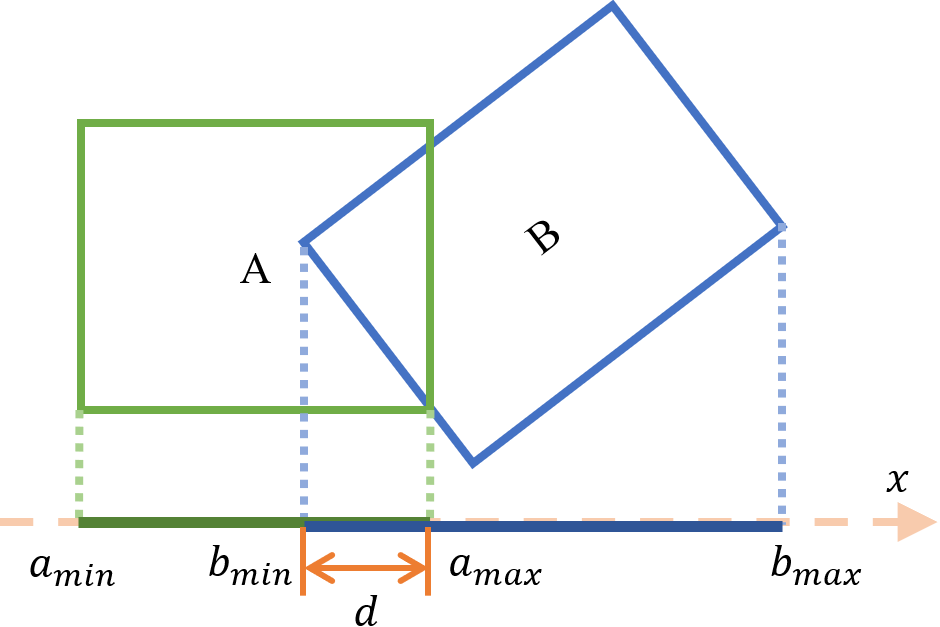}
	\vspace{-0.5em}
	\caption{Object-object collision \zc{term} defined with Separating Axis Theorem. We calculate separation distance $d$ on all the separation axis $x$ of object A and B.}
	\label{fig:sat}
\end{figure}

\noindent \textbf{Input Features}
For each type of node, we collect different features from various sources, concatenate, and embed them with Multi-Layer Perceptron (MLP) into initial node 
\yz{latent} vectors.
Following \cite{zhang2021holistic}, we take 
bounding box parameters for object/wall nodes, category/analytic code and blob centers of LDIF in the world frame \cite{zhang2021holistic, genova2020local} for object nodes, and geometry feature of 2D bounding boxes \cite{hu2018relation, vaswani2017attention} for relation nodes.
Besides, we propose to further take geometric features from the \yz{room} layout and the initial 3D object pose estimations to favor the relation estimation. \zhpcui{Specifically speaking, on relation nodes, we } 
\yz{add} rotation (same definition as object-object rotation) and separation distance (to be further discussed in \autoref{sec:rel_optim}) between each pair of object/wall 3D bounding boxes.
On object nodes, we 
\yz{add} height differences between object 3D box corners and the floor/ceiling plane, and the 2D distances from bounding box corners to the layout polygon.

\noindent \textbf{Relation Estimation}
\label{sec:rel_estimation}
Besides 
\yz{refining initial object pose} objects' poses, our RGCN also outputs the relations between objects and the layout.
The purpose of the relation estimation is to learn valuable context information 
\yz{which may have not been captured by the pose refinement branch.}
Specifically, we design two categories of relations between a pair of elements: object-object and object-layout.
For object-object (including walls since they are also represented as nodes) relation, we define 1) the \zc{relative rotation} between the front face of two objects; 2) whether the two 3D bounding boxes contact with a certain tolerance; and 3) if the center of one object's 3D bounding box is further than that of the other \yz{\wrt the camera center}.
For object-layout relation, we define 1) whether the object is supported by the floor or contacts with ceiling; and 2) if the 3D bounding box is fully inside the room.
\yz{The later one is required to}
\zc{disable certain terms in relation optimization (to be further discussed in \autoref{sec:rel_optim}) for objects visible but outside the room.}
Motivated by \cite{avetisyan2020scenecad}, we design relation estimation as binary classification tasks for binary relations.
For the angular differences, we formulate it into multi-class classification by making a decision on one of the 8 discretized bins in 360\degree~\zhpcui{considering that most furniture in the room is well arranged.}
All the relations are estimated by an additional MLP that takes the node representation as inputs.

\subsection{Relation Optimization}
\label{sec:rel_optim}
While RGCN refines object poses, some numerically tiny errors may severely violate the context and thus be obvious in human perception, such as physical collision, flying objects, or small gaps to the wall.
To fix these, we propose a differentiable optimization to update the refined poses \wrt the predicted relation as introduced in \autoref{sec:rel_estimation}.
Specifically, we use a \zc{gradient descend} to minimize a loss function including three major components measuring physical collision, conformity to relation, and consistency with bottom-up observations.

\vspace{-0.5em}
\subsubsection{Collision \zc{Term}}
At first, we define collision \zc{terms}, which measures the amount of collision between objects, walls, ceiling, and floor. Two types of collision \zc{terms} are 
\yz{defined} according to the node types.

\vspace{0.5em}
\noindent \textbf{Object-Object Collision}
\zc{Since the object pose is represented by a cuboid, }
we \yz{use}
Separating Axis Theorem (SAT) \cite{liang2015research, huynh2009separating}, which measures the collision between convex polygons \yz{to penalize the collision between two objects}.
As explained in \autoref{fig:sat}, two oriented bounding boxes A and B collide with each other if their projections overlap along all \zc{separating axes (directions perpendicular to edges)}.
\zc{Specifically, the projection of bounding box A on separating axis $x$ can be defined as $a_{min} = min\{ c \cdot x|c \in \mathbb{C}_{\rm{A}} \}$ and $a_{max} = max\{ c \cdot x|c \in \mathbb{C}_{\rm{A}} \}$, where $\mathbb{C}_{\rm{A}}$ is the set of corners of the bounding box A and $x$ is represented as vector.}
\zc{Thus} the sum of overlaps $d$ \zc{on every separating axis of A and B} can be treated as a measurement for \zc{their} collision.
\zc{It is also true in 3D space for convex polyhedrons with separating axes defined as the directions perpendicular to faces.}
Based on this, we define the object-object collision \zc{term} between object $i$ and $j$ as:
\vspace{-1.5mm}
\begin{equation}
\begin{aligned}
{{e}_{{i}{j}}^{oc}} = &
\begin{cases}
\sum_{{x} \in \mathbb{S}_{{i}{j}}} {d}_{x}, & \text{if ${i}$, ${j}$ have collision}\\
0, & \text{otherwise}
\end{cases}
\end{aligned},
\vspace{-1mm}
\end{equation}
where $\mathbb{S}_{{i}{j}}$ is the set of separating axes, and $d_x = min(|a_{max} - b_{min}|, |a_{min} - b_{max}|)$ is the separation distance along axis $x$.

\noindent \textbf{Object-Layout Collision}
Since the room layout is under Manhattan World assumption,
\zc{we define 1) object-wall collision ${e}^{wc}$ of each object as the sum of the distances between its bounding box corners and the layout floor map;}
and 2) object-floor/ceiling collision, ${e}^{fc}$ and ${e}^{cc}$, as the distance between the lower/upper surface of the bounding box and the floor/ceiling.
All of these terms are zero if no collision happens.
As mentioned in \autoref{sec:rgcn}, some objects may still be visible even they are outside the room, which should not be considered for our task.
Therefore, we weight ${e}^{wc}$ with in-room likelihood ${\l^{in}}$ to avoid pulling out-room objects inside.

The scene collision \zc{term} with objects $\mathbb{O}$ can be defined as:
\begin{equation}
\begin{aligned}
{E}^{c} = & \sum_{{i}, {j} \in \mathbb{O}, {i} \neq {j}} {\lambda^{oc}} {{e}_{{i}{j}}^{oc}} 
+ \sum_{{i} \in \mathbb{O}} ({\lambda^{wc}} {{l}_{i}^{in}} {{e}_{i}^{wc}} + {\lambda^{fc}} {{e}_{i}^{fc}} + {\lambda^{cc}} {{e}_{i}^{cc}}),
\end{aligned}
\end{equation}
where {${\lambda^{*}}$} are preset weights.
\vspace{-0.8em}
\subsubsection{Relation \zc{Term}}
We then define relation \zc{terms} to measure the conformity of object poses with regard to the predicted relations from \yz{RGCN} in \autoref{sec:rgcn}.

For the relative rotation, we define the \zc{term ${e}^{rr}$} as the absolute error between the observed and predicted relative angle.
For the \zc{object} attachment relation (\ie, contact), we define the \zc{term ${e}^{oa}$} similar to ${e}^{oc}$ but only penalize sum of separation distances when there is no collision. 
\zc{The terms ${e}^{fa}$ and ${e}^{ca}$ are defined as the distance from the lower/upper surface of the bounding box to the floor/ceiling, and are respectively set to zero}
\yz{if the object is already attaching with the floor/ceiling.}
For relative distance, we calculate a view distance for each object as the distance from camera center to object center, 
and define the \zc{term} ${e}^{rd}$ as the difference between view distance if their relative order disobeys with the prediction and zero otherwise.
Overall, the relation \zc{term} is defined as:
\begin{equation}
\begin{aligned}
{E}^{r} = & \sum_{{i} \in \mathbb{O}} {\lambda^{rr}} {{e}_{i}^{rr}}
+ \sum_{{i} \in \mathbb{O}, {j} \in {\mathbb{O} \cup \mathbb{W}}, {i} \neq {j}} {{\lambda^{oa}} {l}_{{i}{j}}^{oa}} {{e}_{{i}{j}}^{oa}}\\
& + \sum_{{x} \in \{ft, ct, rd\}} \sum_{{i} \in \mathbb{O}} {\lambda^{x}} {{l}_{i}^{x}} {{e}_{i}^{x}}
\end{aligned}
\end{equation}
where \yz{$\mathbb{W}$ is the set of walls}, {${l}^{*}$} are the relation labels predicted by RGCN, and {${\lambda^{*}}$} are weights for each term.
\vspace{-0.8em}

\subsubsection{Observation \zc{Term}}
Not only abide to the predicted relation and physics, the object pose refinement should also respect the initial predictions observed from the input image.

We first define a loss term that measures the consistency with the raw image observation.
For each object, we fit a 2D bounding box to the projection of the 3D cuboid on the tangent plane centered at cuboid center, and compare it with the results from Mask-RCNN.
We define ${e}^{bp}$ as the intersection over union between two boxes.
We \yz{then} define a loss term to measure the consistency between the optimized cuboid with the
\zc{initial estimation,} which is the \zc{L1 loss of the cuboid parameters,}
\zc{including the offset from the 2D detection center to the cuboid center projection ${\delta}$, distance from camera center to the cuboid center ${d}$, size ${s}$, and orientation ${\theta}$ as defined in previous work \cite{nie2020total3dunderstanding, huang2018holistic}.}
The total scene observation \zc{term} is then defined as:
\begin{equation}
\begin{aligned}
{E}^{o} = & \sum_{{x} \in \{bp, {\delta}, {d}, {s}, {\theta}\}} \sum_{{i} \in \mathbb{O}} {\lambda^{x}} {{e}_{i}^{x}}.
\end{aligned}
\end{equation}

\vspace{-1em}
\subsubsection{Optimization}
We minimize the sum of the three \zc{terms}:
\begin{equation}
\min E({\delta}, {d}, {s}, {\theta}) = {E}^{c} + {E}^{r} + {E}^{o}.
\end{equation}
are chosen according to the confidence of estimated relation and bottom-up observations. More details can be found in Supp. Materials.
Note that the optimization can be achieved via \zc{gradient decent} such that is differentiable and can be added to the RGCN for joint training.

\subsection{Loss Function}
We adopt the loss from Nie \etal \cite{nie2020total3dunderstanding} to train the ODN:
\begin{equation}
\begin{aligned}
\mathcal{L}_{ODN} = & \sum_{x \in \{  {\delta}, {d}, {s}, \theta  \} }\lambda_{x}\mathcal{L}_{x},
\end{aligned}
\end{equation}
where {$\mathcal{L}_{*}$} are the classification and regression loss for the object pose parameters.
To train RGCN, we first train pose refinement branch with $\mathcal{L}_{ODN}$, then add the losses for the relation branch:
\begin{equation}
\begin{aligned}
\mathcal{L}_{RGCN} = & \mathcal{L}_{ODN} + & \sum_{x \in \{ {rr}, {oa}, {fa}, {ca}, {rd} \} }\lambda_{x}\mathcal{L}_{x},
\end{aligned}
\end{equation}
where $\mathcal{L}_{rr}$ is 8-class cross-entropy loss of rotation classification, and $\mathcal{L}_{x}, x \in {{oa}, {fa}, {ca}, {rd}}$ are binary cross-entropy loss. 
When training ODN, RGCN with \yz{RO} end-to-end, we define the joint loss as:
\begin{equation}
\begin{aligned}
\mathcal{L} = & \mathcal{L}_{ODN} + \mathcal{L}_{RGCN}
+ & \sum_{x \in \{ {\delta}, {d}, {s}, \theta \} }\lambda_{x}^{\prime}\mathcal{L}_{x}^{\prime},
\end{aligned}
\end{equation}
where $\mathcal{L}_{x}^{\prime}$ is the $L_1$ loss of the optimized pose parameters.

\begin{table*}[ht]
    \vspace{-1em}
	\begin{center}
	    \resizebox{2.095\columnwidth}{!}{
    		\begin{tabular}{|l|c|c|c|c|c|c|c|c|c|c|c|c|c|c|}
    			\hline
    			Method & chair & sofa & table & fridge & sink & door & floor lamp & bottom cabinet & top cabinet & sofa chair & dryer & mAP\\
    			\hline
    			\yz{Total3D-Pers}  & 13.71 & 68.06 & 30.55 & 36.02 & 69.84 & 11.88 & 12.57 & 35.56 & 19.19 & 64.29 & 41.36 & 36.64\\     			\yz{Total3D-Pano}  & 20.84 & 69.65 & 31.79 & 43.13 & 68.42 & 10.27 & 16.42 & 34.42 & 20.83 & 62.38 & 33.78 & 37.45 \\     			\yz{Im3D-Pers}     & 30.23	& \textbf{75.23} & 44.16 & 52.56 & 76.46 & 14.91 & 9.99 & 45.51 & 23.37 & \textbf{80.11} & 53.28 & 45.98  \\     			\yz{Im3D-Pano}     & 33.08 & 72.15 & 37.43 & 70.45 & 75.20 & 11.58 & 6.06 & 43.28 & 18.99 & 78.46 & 41.02 & 44.34 \\     			Ours (w/o. RO)  & \textbf{33.57} & 75.18 & 38.65 & 71.97 & \textbf{80.66} & 19.94 & 18.29 & 50.67 & 29.05 & 79.42 & \textbf{60.07}  & 50.68 \\     			Ours (Full)       & 27.78 & 73.96 & \textbf{46.85} & \textbf{74.22} & 75.29 & \textbf{21.43} & \textbf{20.69} & \textbf{52.03}  & \textbf{50.39} & 77.09 & 59.91 & \textbf{52.69} \\     			\hline
    		\end{tabular}
        }    	
	\end{center}
 	\vspace{-1.7em}
	\caption{3D object detection. Following \cite{nie2020total3dunderstanding,huang2018cooperative}, we \zc{evaluate on common object categories and} use mean average precision (mAP) with the threshold of 3D bounding box IoU set at 0.15 as the evaluation metric. \zc{Please refer to supplementary material for evaluation on full 57 categories.}}
	\label{tbl:3d_detection}
	\vspace{-0.9em}
\end{table*}

\begin{table*}[ht]
    	\begin{center}
	    \resizebox{2.095\columnwidth}{!}{
    		\begin{tabular}{|l|c|c|c|c|c|c|c|c|c|c|c|c|c|c|}
    			\hline
    			Method & background & bed & painting & window & mirror & desk & wardrobe & tv & door & chair & sofa & cabinet & \yz{mIoU}\\
    			\hline
    			PanoContext & 86.90 & \textbf{78.58} & 38.70 & 35.58 & 38.15 & 29.55 & 27.44 & 34.81 & 19.40 & 9.61 & 11.10 & 5.46 & 31.38\\
    			Ours (Full) & \textbf{87.48} & 62.99 & \textbf{56.33} & \textbf{65.36} & \textbf{40.48} & \textbf{52.86} & \textbf{53.50} & \textbf{46.88} & \textbf{49.70} & \textbf{34.21} & \textbf{48.59} & \textbf{10.36} & \textbf{50.73} \\     			\hline
    		\end{tabular}
        }    	
	\end{center}
	\vspace{-1.7em}
	\caption{Semantic segmentation IoU. Following \cite{zhang2014panocontext} \zc{we calculate IoU with} uniformly sampled points on sphere surface.}
	\label{tbl:sem_seg}
	\vspace{-1.2em}
\end{table*}

\subsection{Panoramic Datasets}
As there 
\yz{is no panorama dataset with complete ground truth for room layout, object poses, and object shapes,}
we propose to synthesize a panoramic dataset that provides the detailed 3D shapes, poses, positions, semantics of objects as well as the room layout by utilizing the latest simulation environment   
iGibson \cite{shen2020igibson}. iGibson contains 500+ objects of 57 categories, and 15 fully interactive scenes with 100+ rooms in total, and 75 objects on average.
Before rendering, we run a physical simulation \cite{shen2020igibson} to resolve bad placement (\eg, floating objects) and randomly replace objects with models from the same semantic category for each scene.
Then we set the cameras with height of 1.6m looking at random directions in the horizontal plane.
By building a 2D occupation map of objects, we avoid setting cameras inside, over, or too close to objects.
Finally, we render 1,500 panorama images with semantic/instance segmentation, depth images, room layout, and the oriented 3D object bounding boxes from the physical simulator.
Among 15 provided scenes, we use 10 for training and 5 for testing, generating 100 images per scene.

We crop each object separately 
\yz{to train LIEN and LDIF}.
In total, we collect 19,245 object crops from the training set and 7,753 from the test set.
Besides these, we also render extra object-centric images, which contain 51,285 for training and 5,715 for testing.
To generate implicit signed distance field ground truth, we process the 3D object CAD model 
\cite{mescheder2019occupancy,genova2020local} to make sure the objects are watertight. 
Please refer to supplementary material for examples of our synthesized dataset.

\section{Experiments}

To our best knowledge, we are the first to achieve total 3D scene understanding on panorama images with scene level reconstruction.
Thus to make comparisons with the SoTA methods Total3D \cite{nie2020total3dunderstanding} and Im3D \cite{zhang2021holistic} which work with perspective cameras, we divide the panorama camera into a set of cameras with horizontal FoV of 60\degree.
\zc{We then retrieve the detection results on panorama from our 2D detector and group them by camera splits then feed them into Total3D and Im3D.
The results of object pose
and 
\yz{shape} are transformed from camera 
\yz{coordinates}
to world 
\yz{coordinates}
to make the final results \yz{(\yz{Total3D-Pers} and \yz{Im3D-Pers})}.
\yz{Besides the perspective version, we also extend Total3D and Im3D to work directly on panorama images \zc{(\yz{Total3D-Pano} and \yz{Im3D-Pano})}. Specifically, we change}
the representation of 2D bounding box into BFoV and input object detection results as a whole to provide richer scene context information.}
Since Total3D and Im3D are designed to do cuboid layout estimation, for a fair comparison, we replaced their layout estimation network with HorizonNet and only compare with them on 3D object detection and scene reconstruction.
\zc{All the models are fine-tuned on our proposed dataset following the same process.}
Please refer to supplementary material for 
\yz{more} details.

\begin{figure*}[!ht]
    \vspace{-0.4em}
	\centering
	\scriptsize
	\newcommand{\rgb}[1]{\raisebox{-0.5\height}{\includegraphics[width=.147\linewidth]{#1}}}
	\newcommand{\bird}[1]{\raisebox{-0.5\height}{\includegraphics[width=.147\linewidth,clip,trim=100 80 90 60]{#1}}}
	\newcommand{\rot}[1]{\rotatebox[origin=c]{90}{#1}}
	\def\arraystretch{0.5}	\begin{tabular}{c|c*{6}{c@{\hspace{1px}}}}
            \rule{0pt}{5px}&\rot{Input}
                        & \rgb{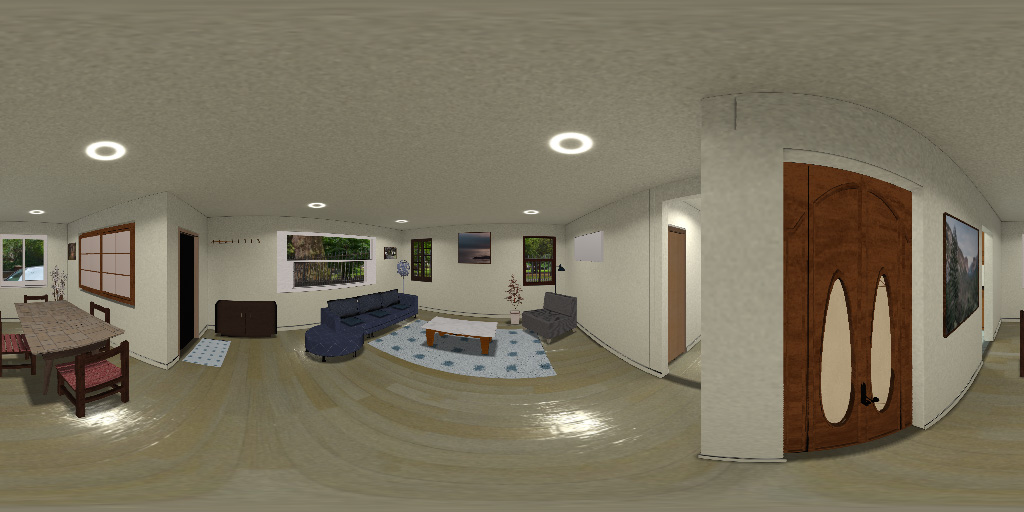}
    	    & \rgb{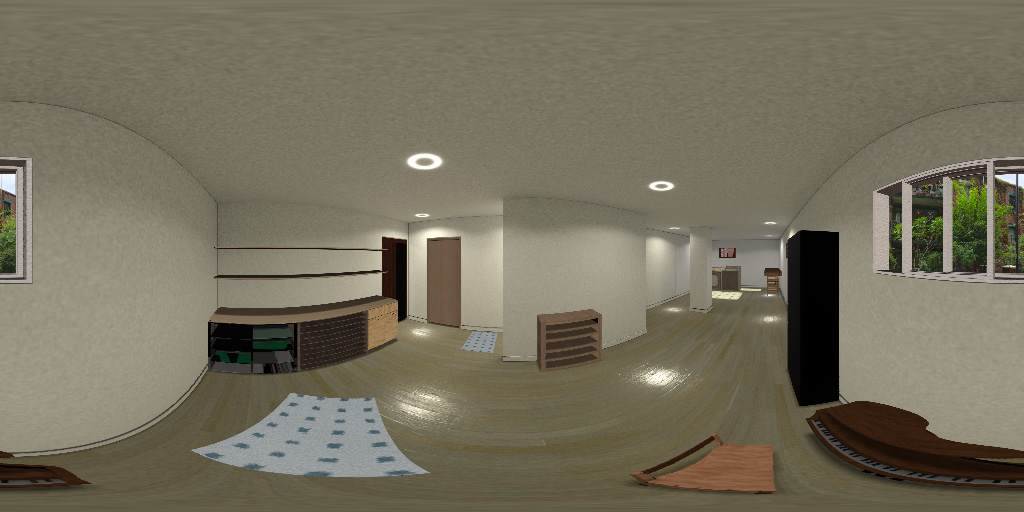}
    	       	    & \rgb{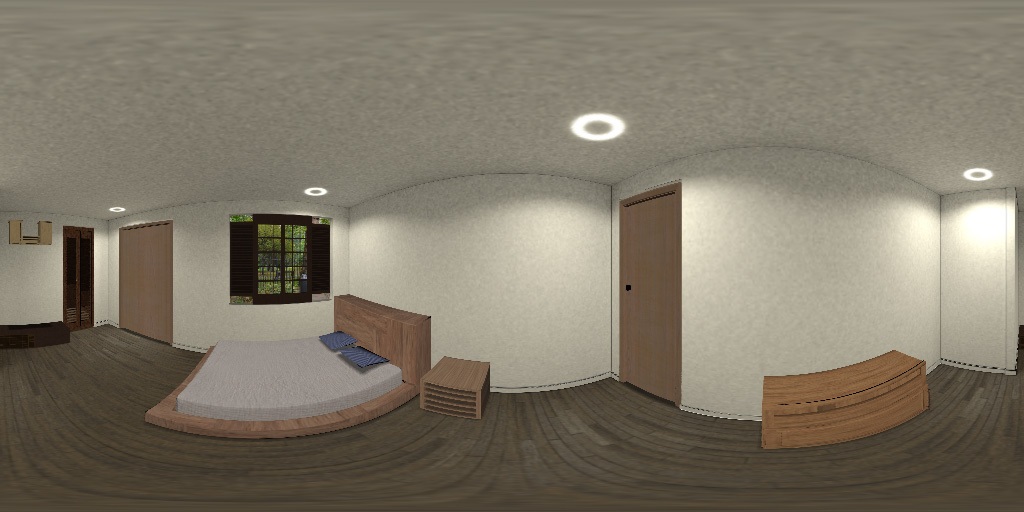}
    	    & \rgb{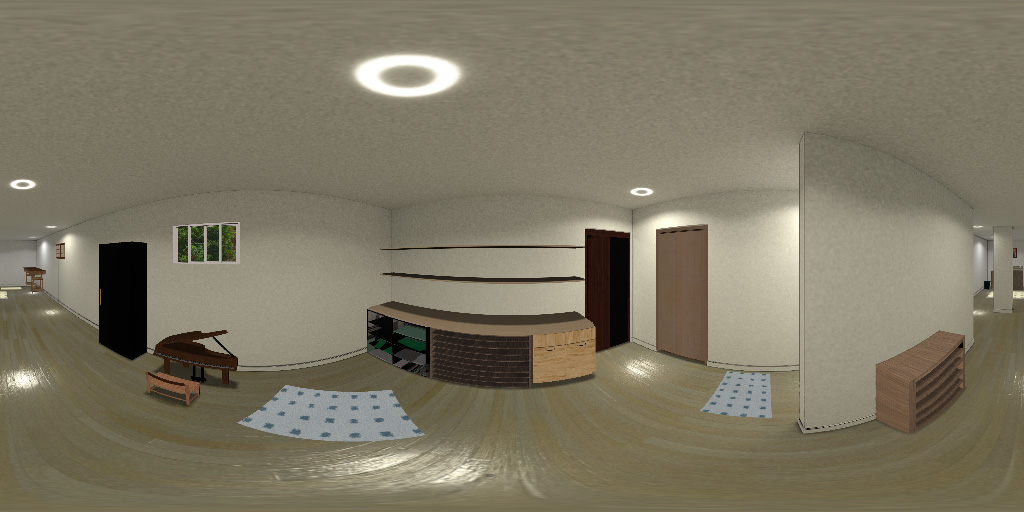}
    	    & \rgb{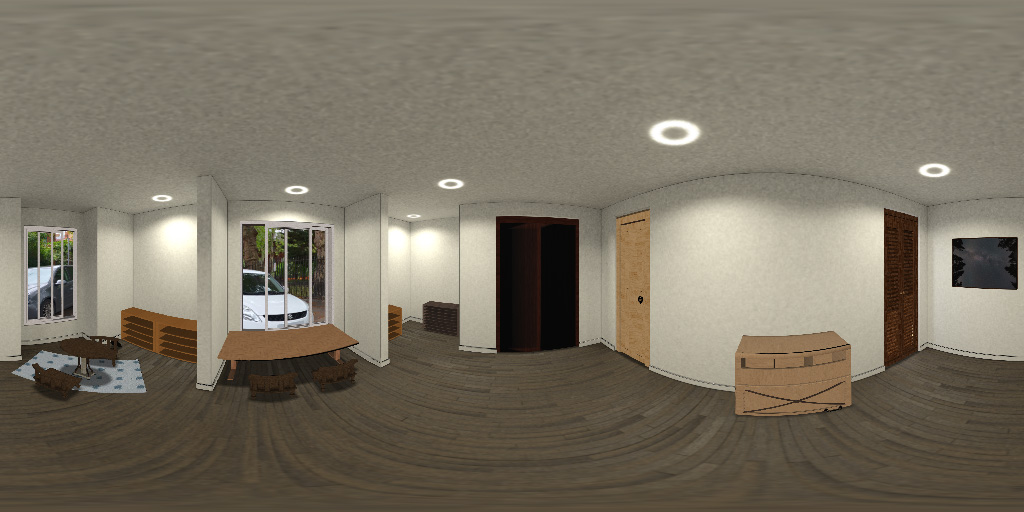}
    	       	    & \rgb{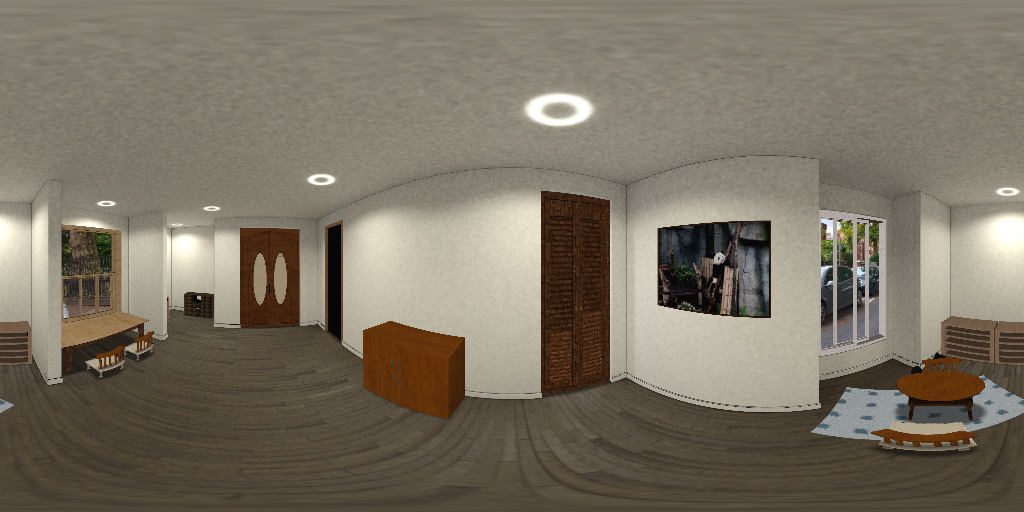}
    	    \\
    	\hline
    	    \multirow{4}{*}[-27ex]{\rot{Bird's Eye View}}
    	    &\rot{Total3D}
                                & \bird{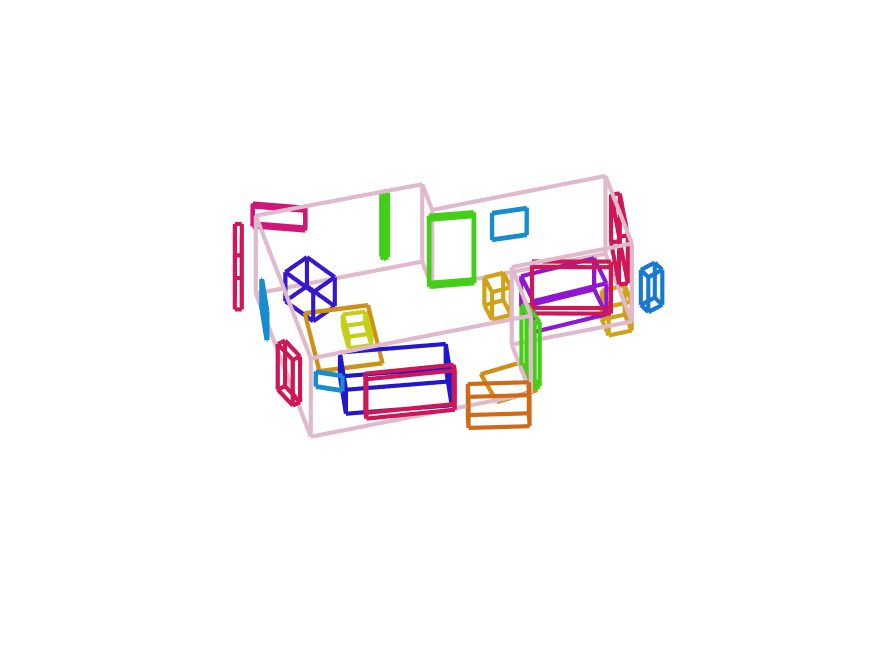}
        	    & \raisebox{-0.5\height}{\includegraphics[width=.147\linewidth,clip,trim=120 110 110 90]{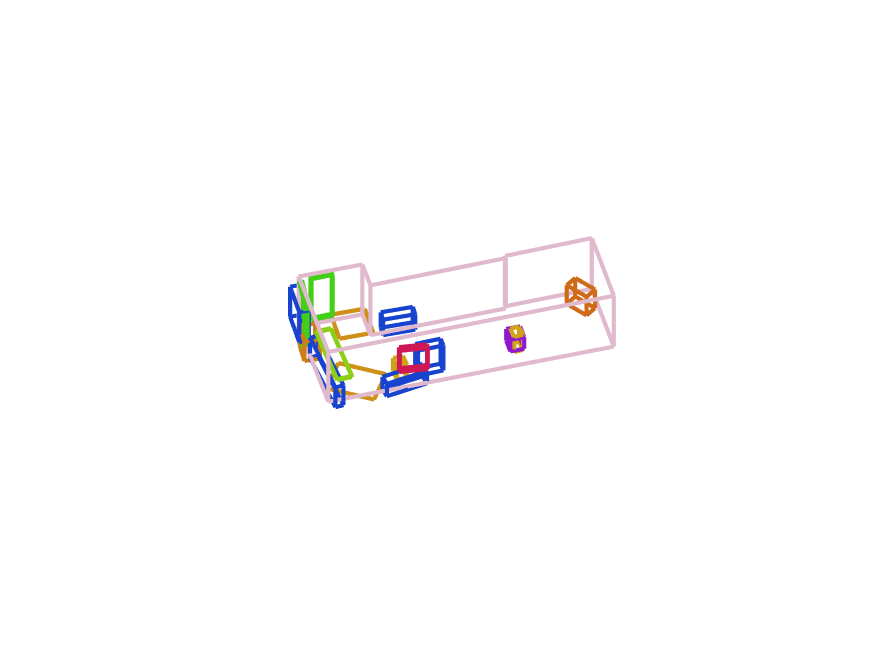}}
        	           	    & \bird{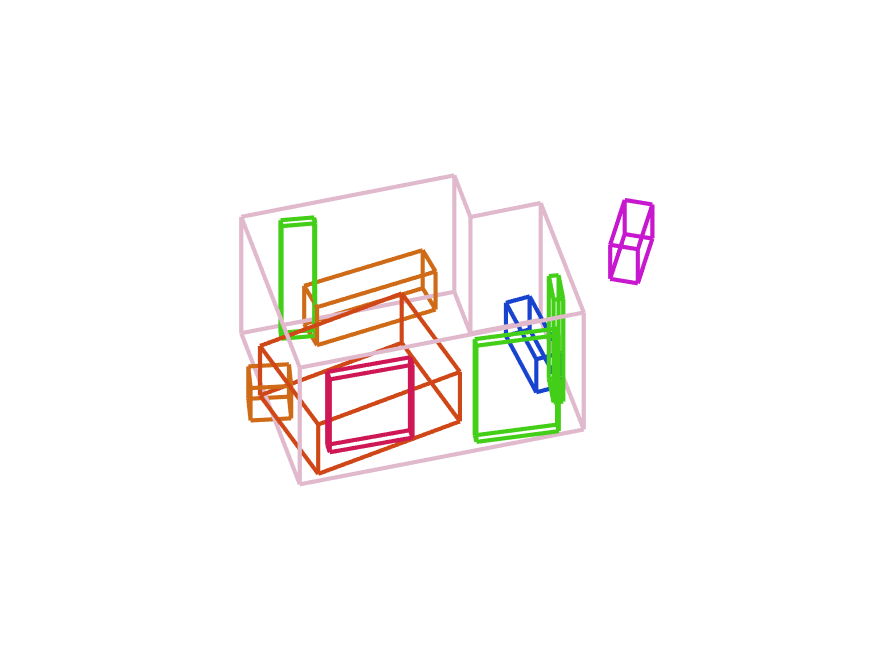}
        	    & \bird{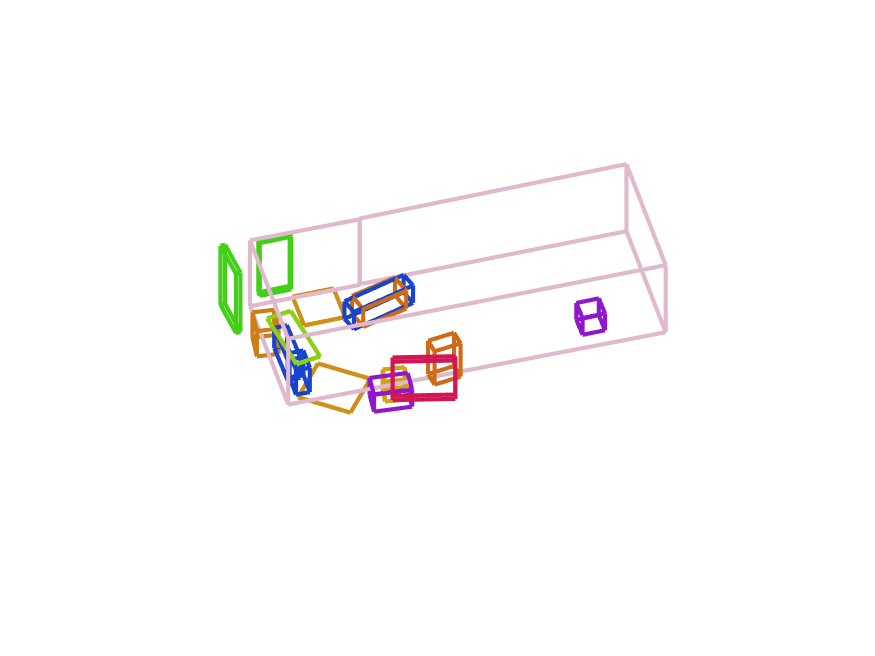}
        	    & \bird{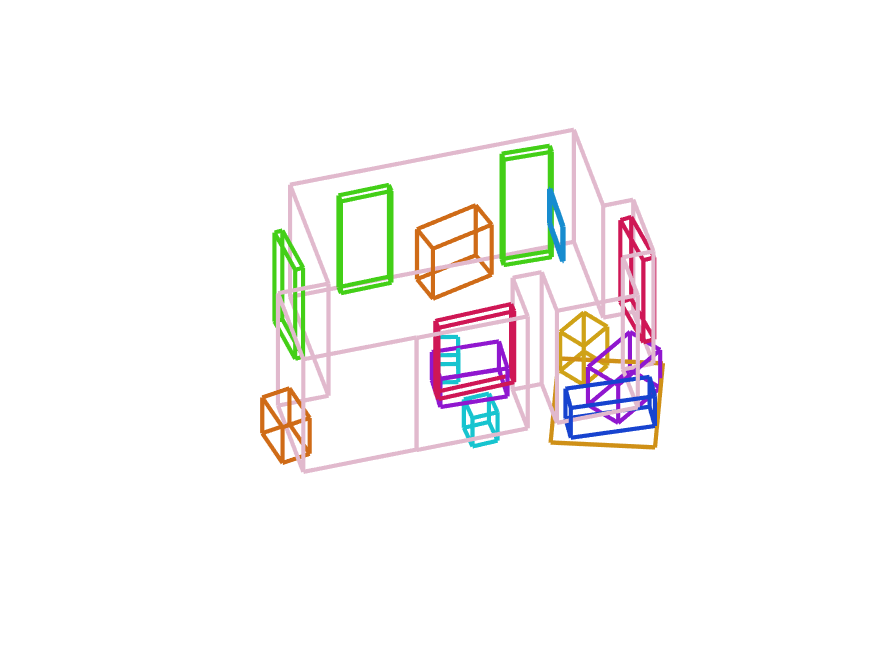}
        	           	    & \raisebox{-0.5\height}{\includegraphics[width=.147\linewidth,clip,trim=115 95 105 80]{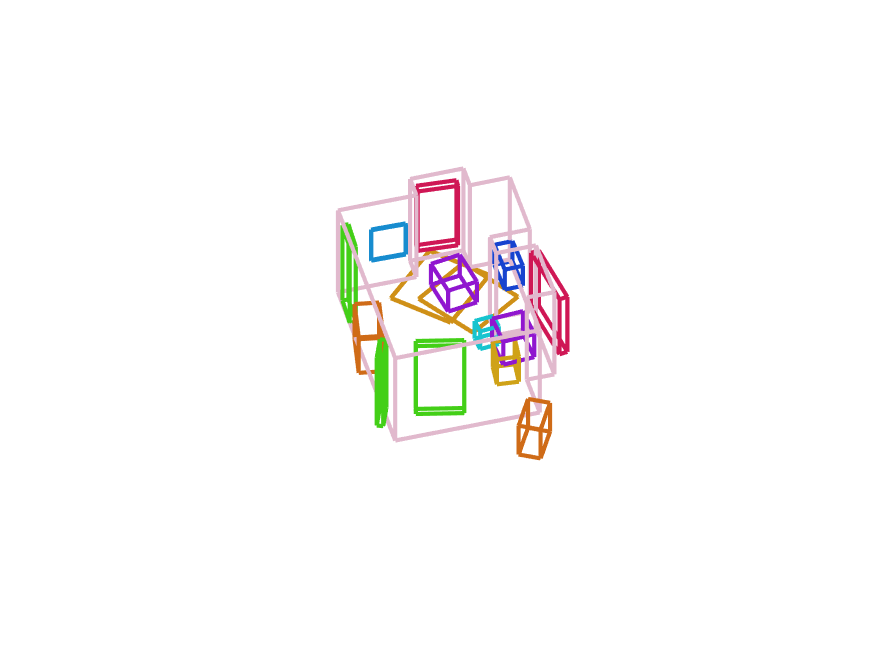}}
        	    \\
    	    &\rot{Im3D}
                                & \bird{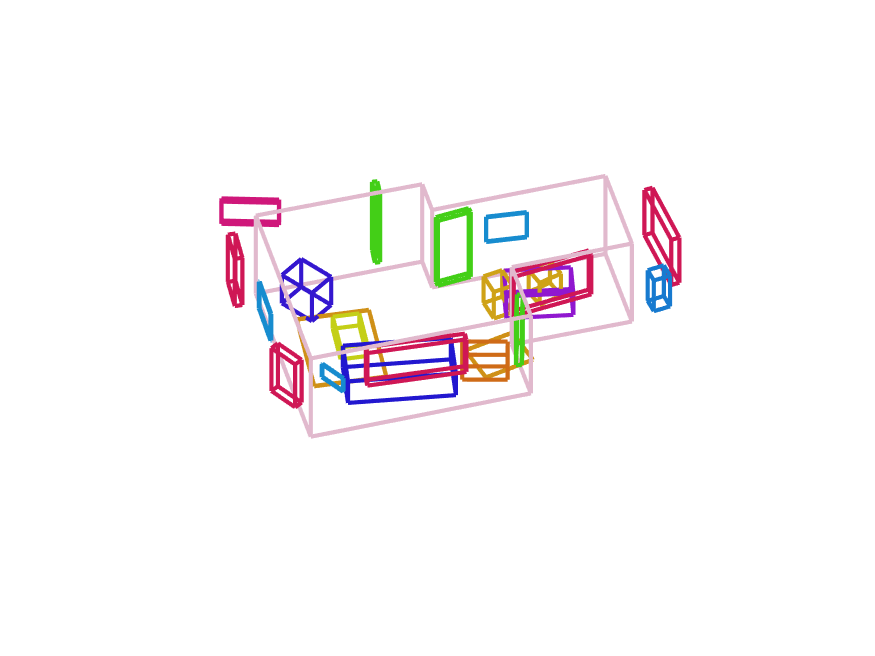}
        	    & \raisebox{-0.5\height}{\includegraphics[width=.147\linewidth,clip,trim=120 110 110 90]{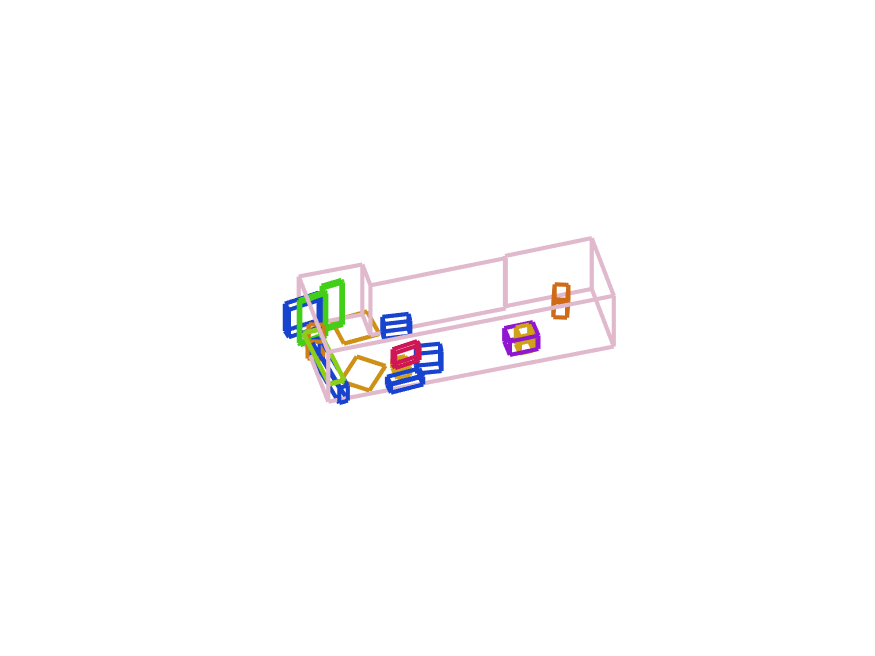}}
        	           	    & \bird{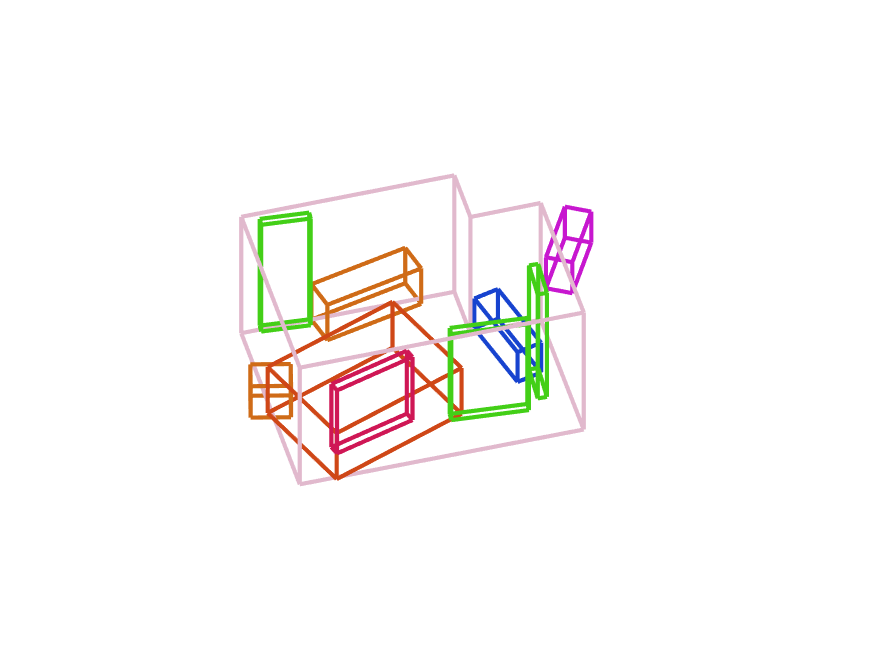}
        	    & \bird{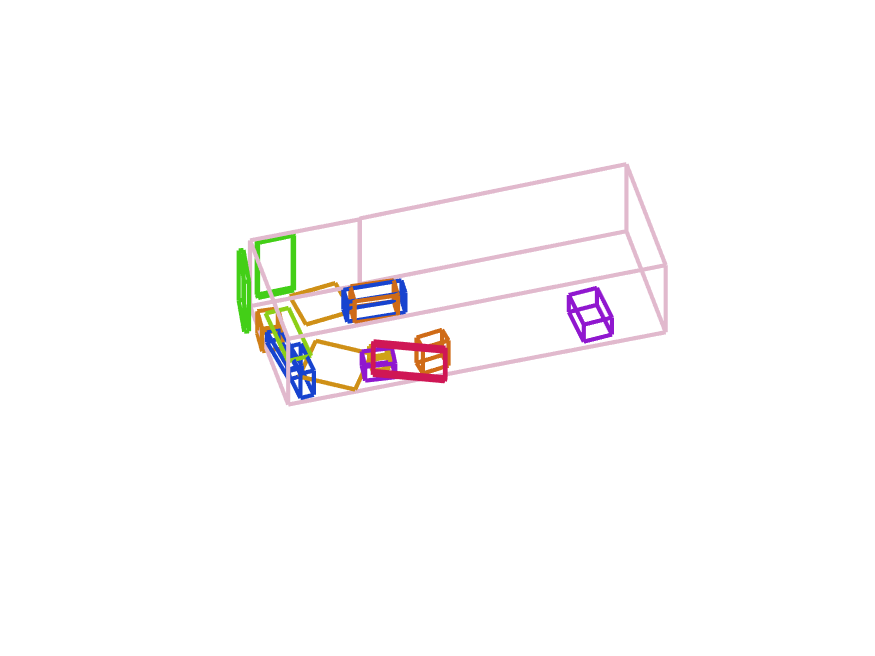}
        	    & \bird{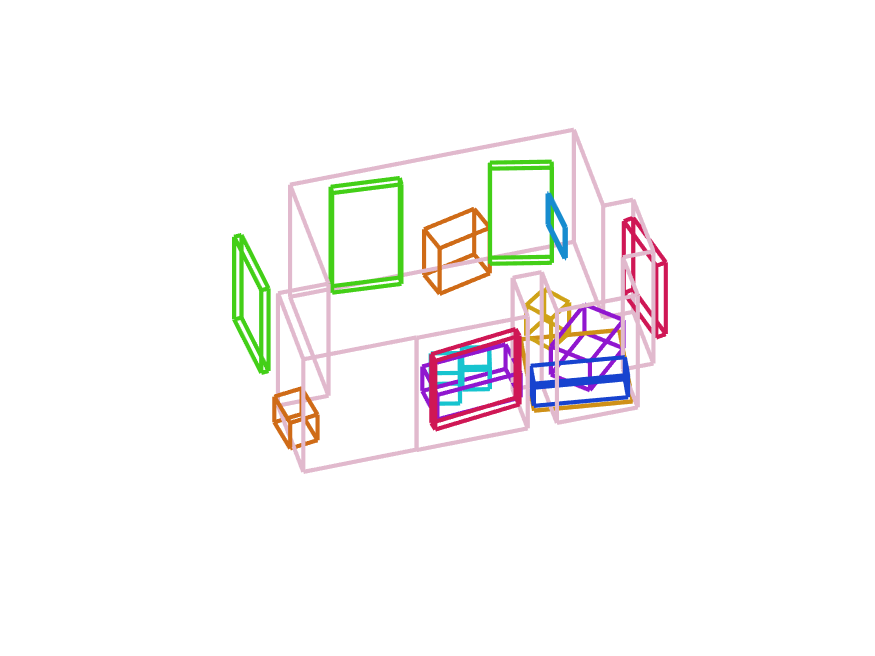}
        	           	    & \raisebox{-0.5\height}{\includegraphics[width=.147\linewidth,clip,trim=115 95 105 80]{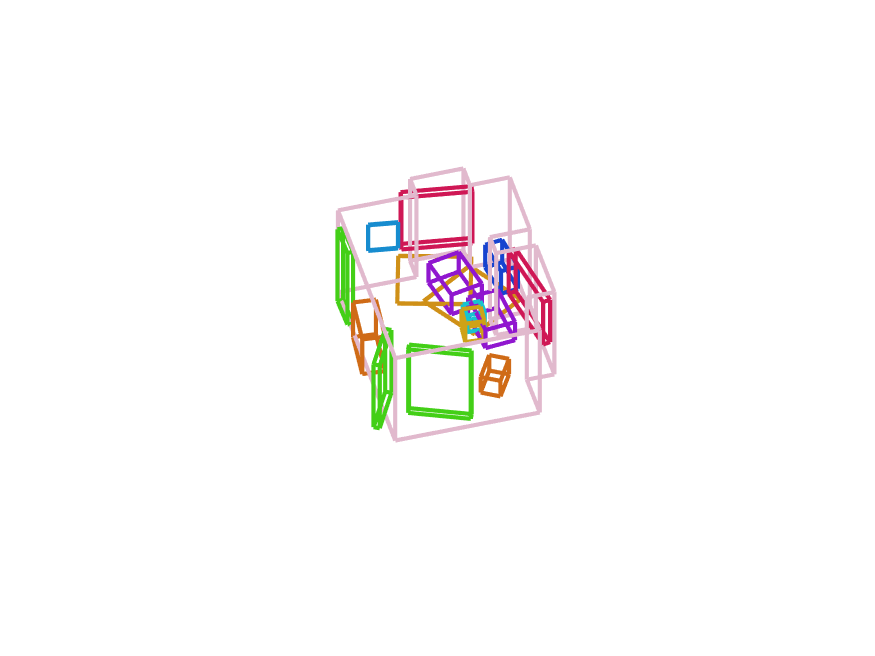}}
        	    \\
        	& \rot{Ours}
                                & \bird{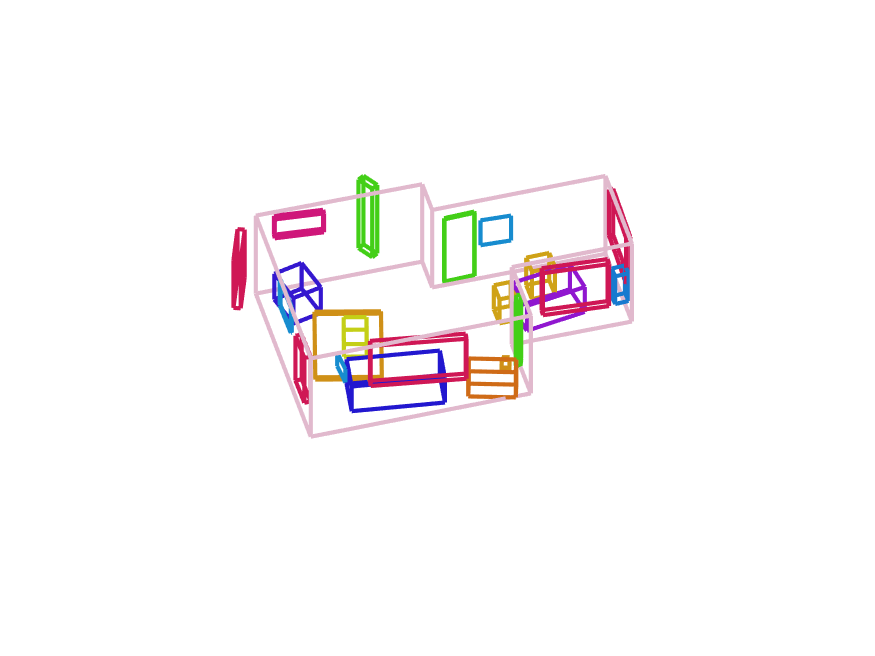}
        	    & \raisebox{-0.5\height}{\includegraphics[width=.147\linewidth,clip,trim=120 110 110 90]{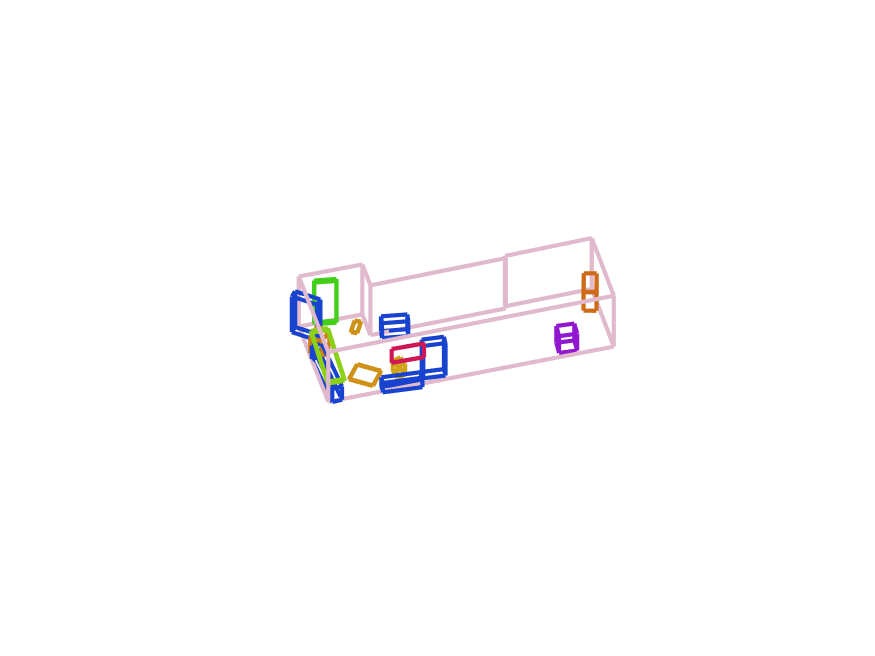}}
        	           	    & \bird{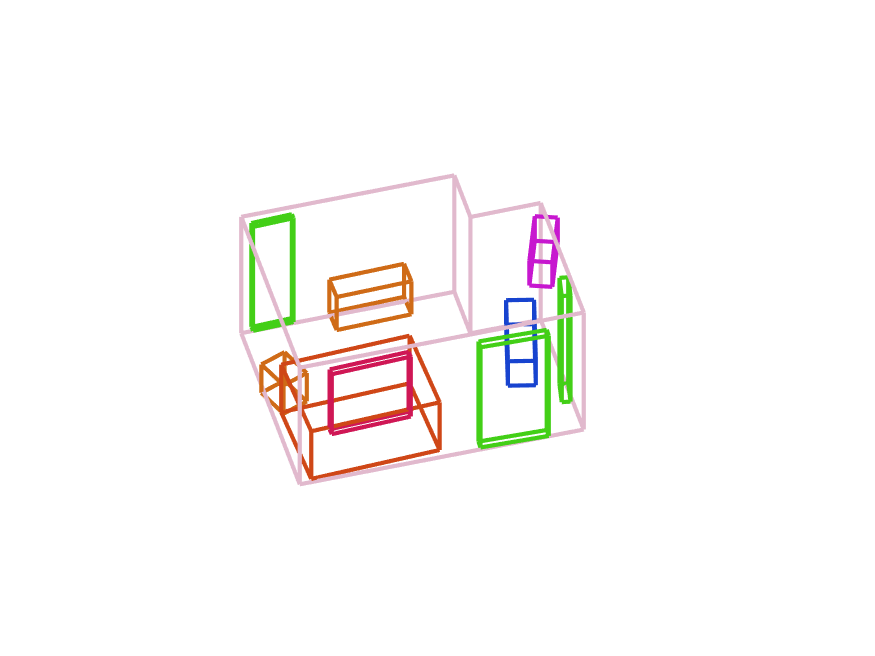}
        	    & \bird{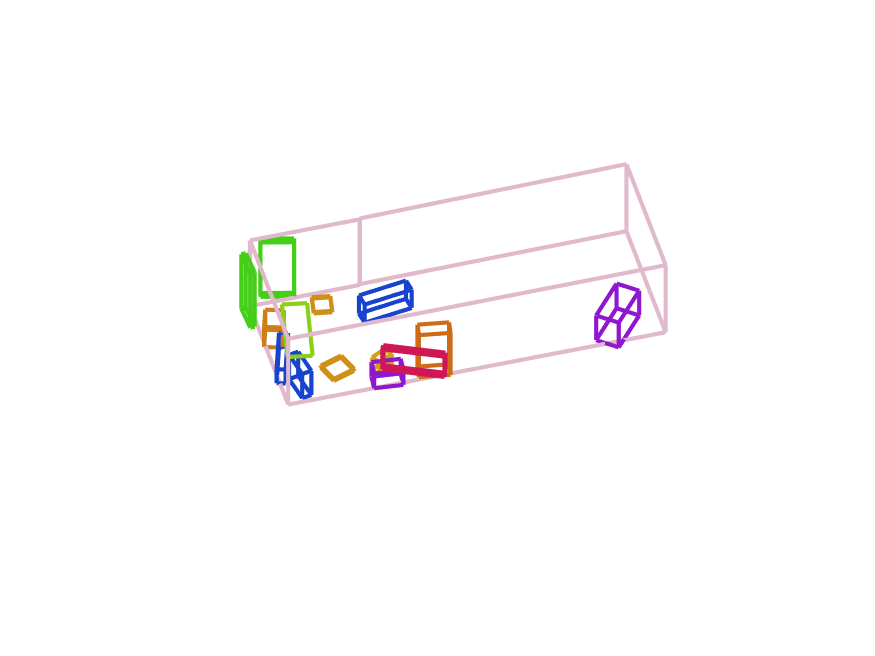}
        	    & \bird{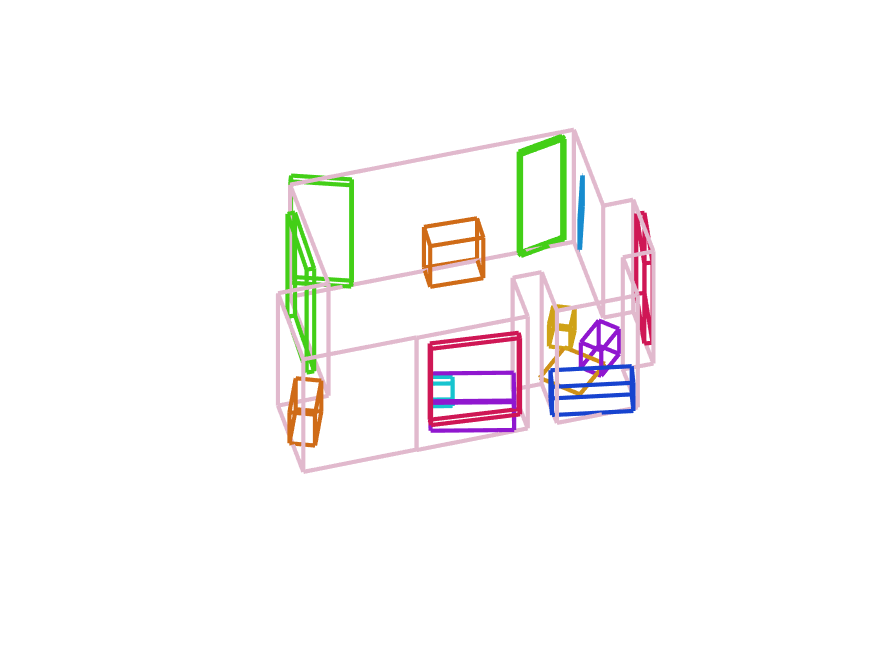}
        	           	    & \raisebox{-0.5\height}{\includegraphics[width=.147\linewidth,clip,trim=115 95 105 80]{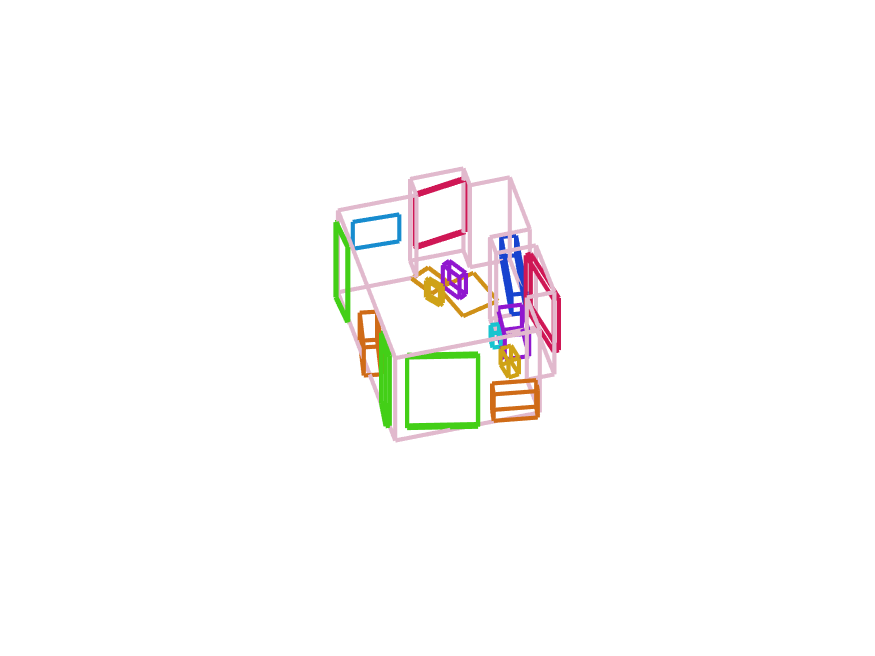}}
        	    \\
        	&\rot{GT}
                                & \bird{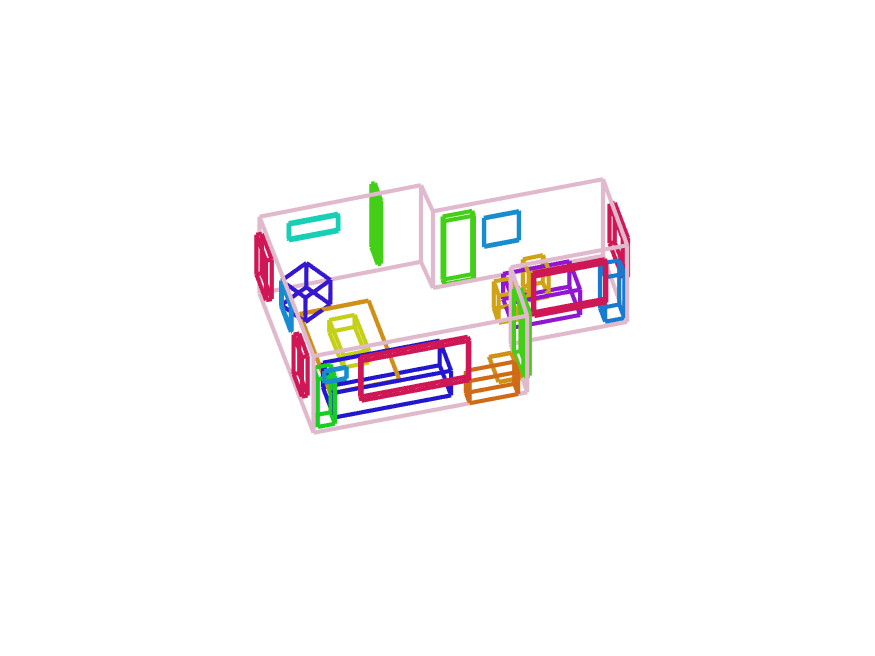}
        	    & \raisebox{-0.5\height}{\includegraphics[width=.147\linewidth,clip,trim=120 110 110 90]{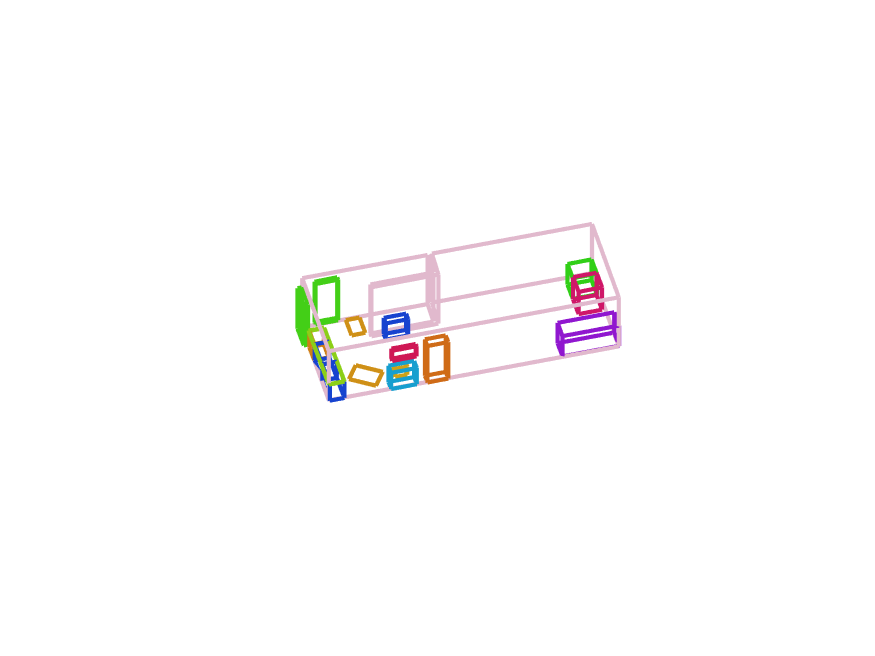}}
        	           	    & \bird{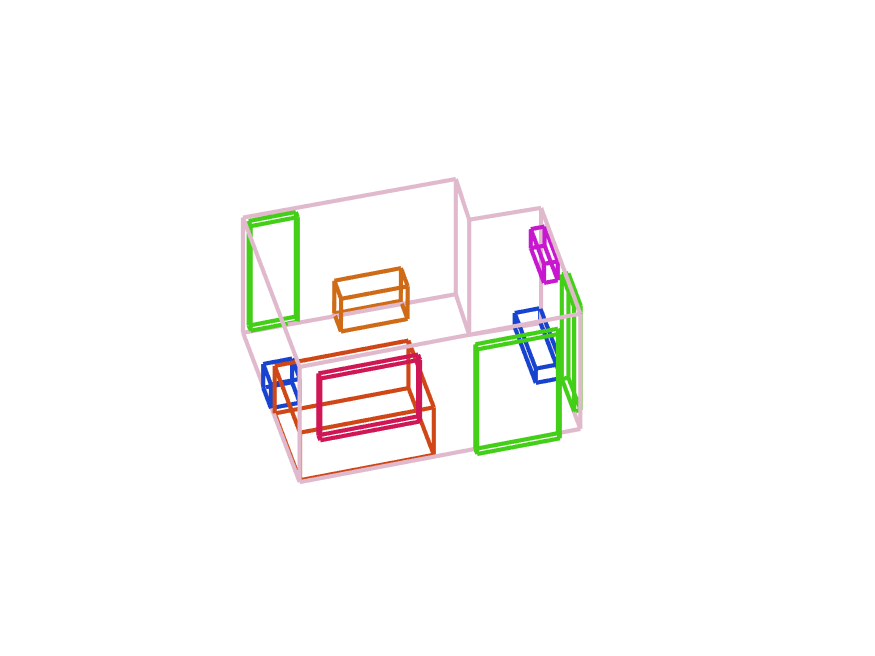}
        	    & \bird{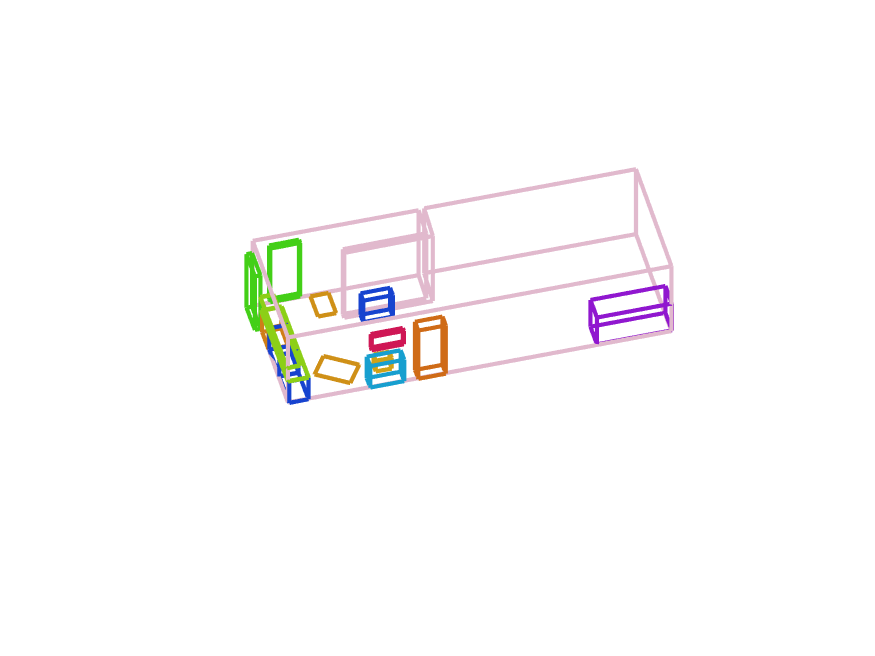}
        	    & \bird{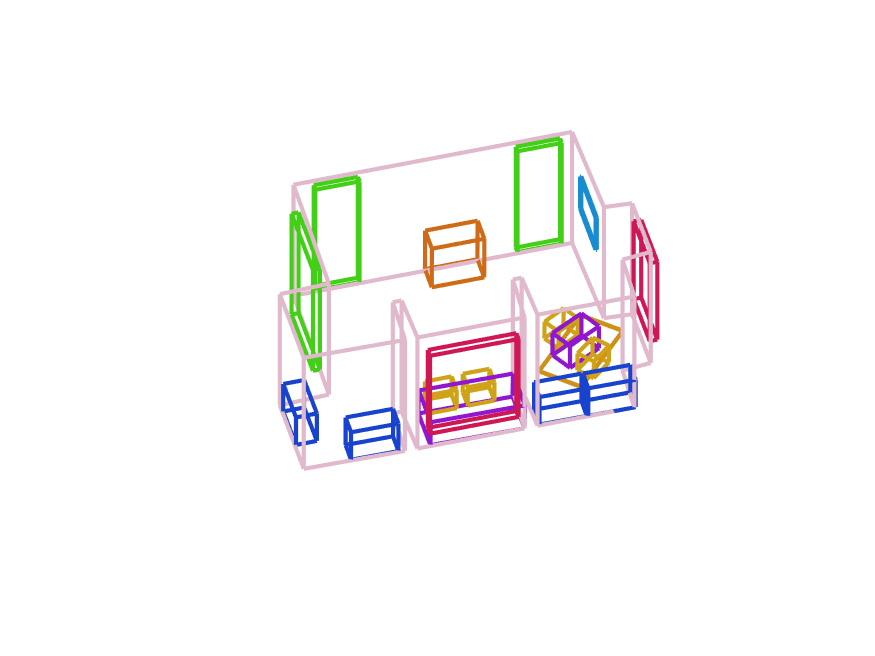}
        	           	    & \raisebox{-0.5\height}{\includegraphics[width=.147\linewidth,clip,trim=115 95 105 80]{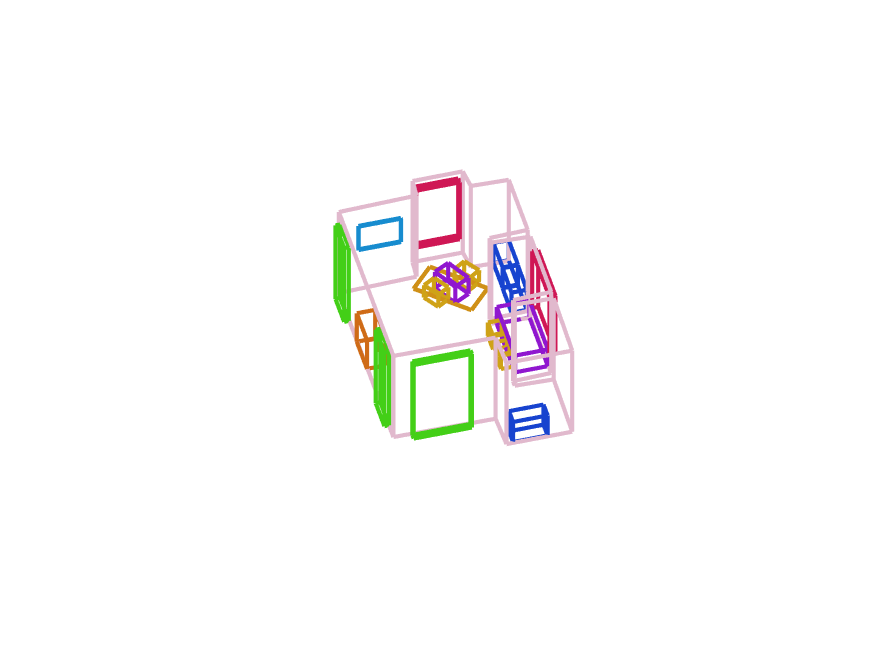}}
        	    \\
    	\hline
    	    \multirow{4}{*}[-10ex]{\rot{Panorama View}}
    	    &\rot{Total3D}
                                & \rgb{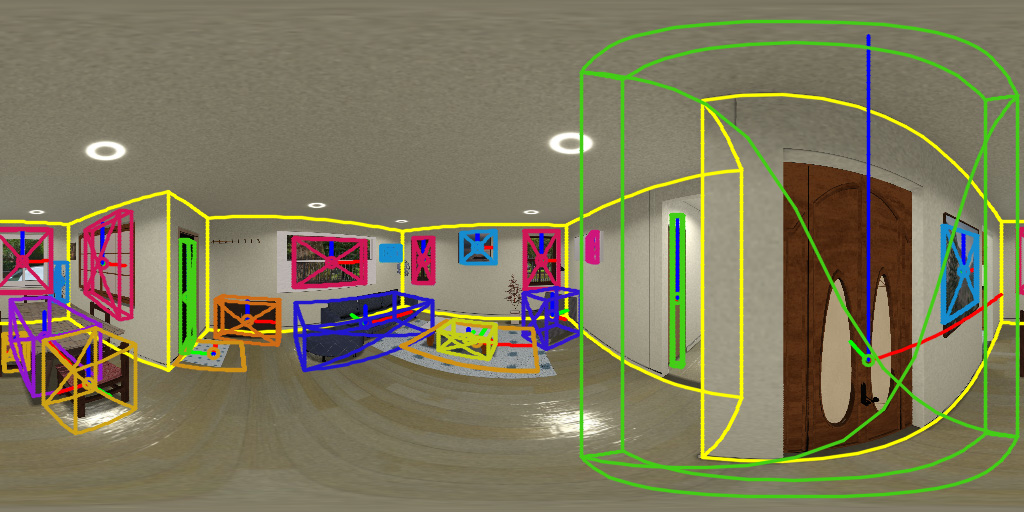}
        	    & \rgb{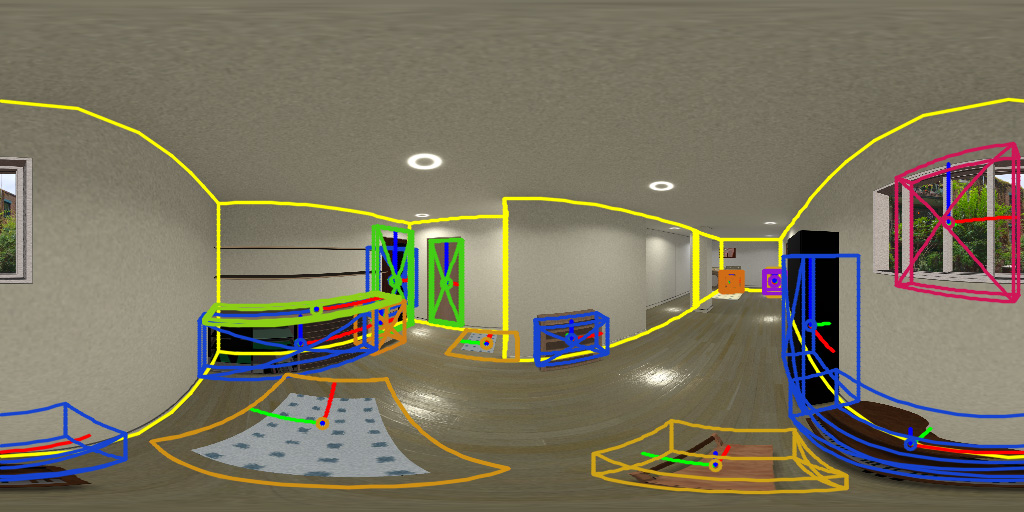}
        	           	    & \rgb{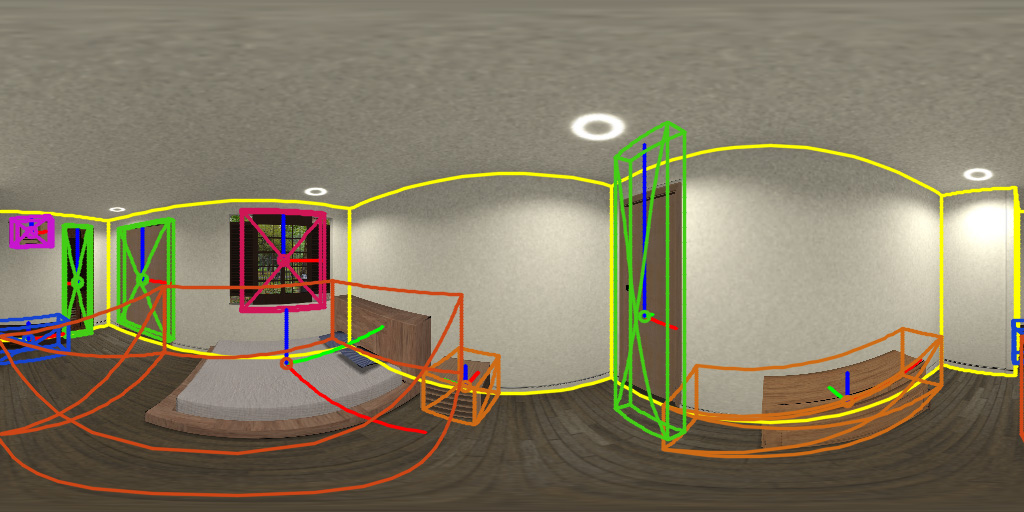}
        	    & \rgb{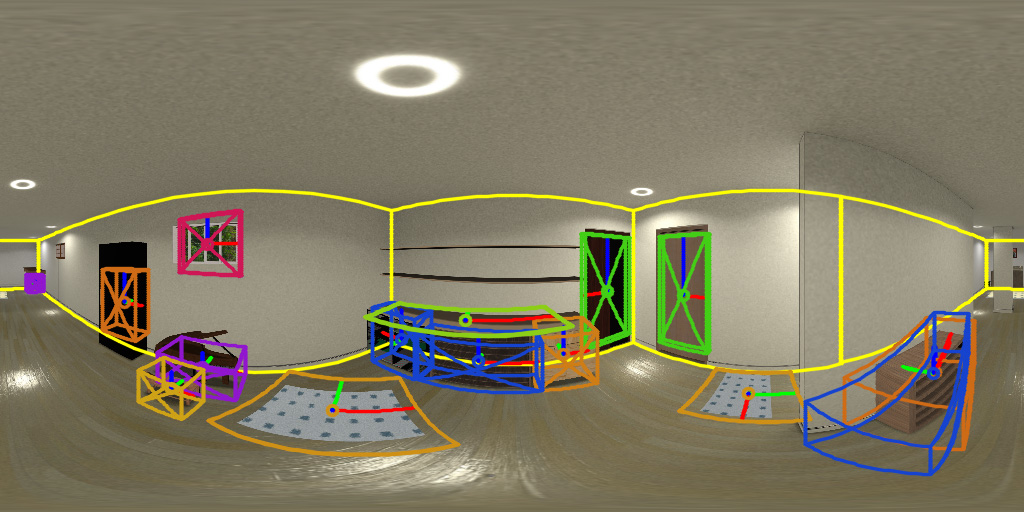}
        	    & \rgb{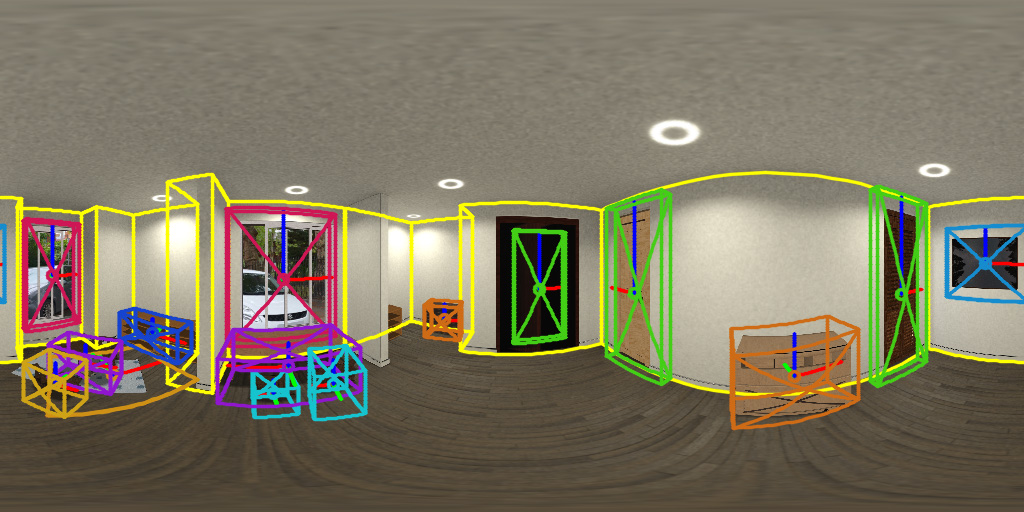}
        	           	    & \rgb{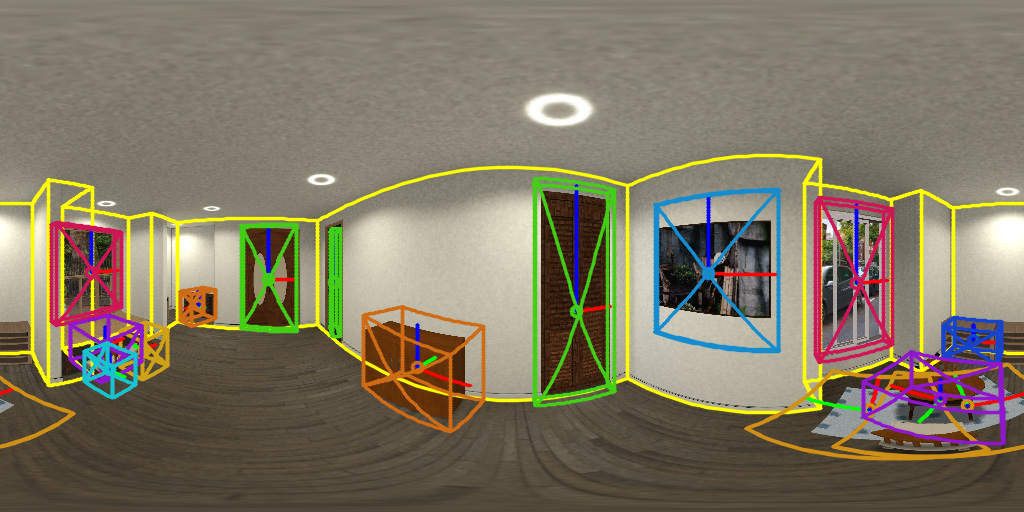}
        	    \\
    	    &\rot{Im3D}
                                & \rgb{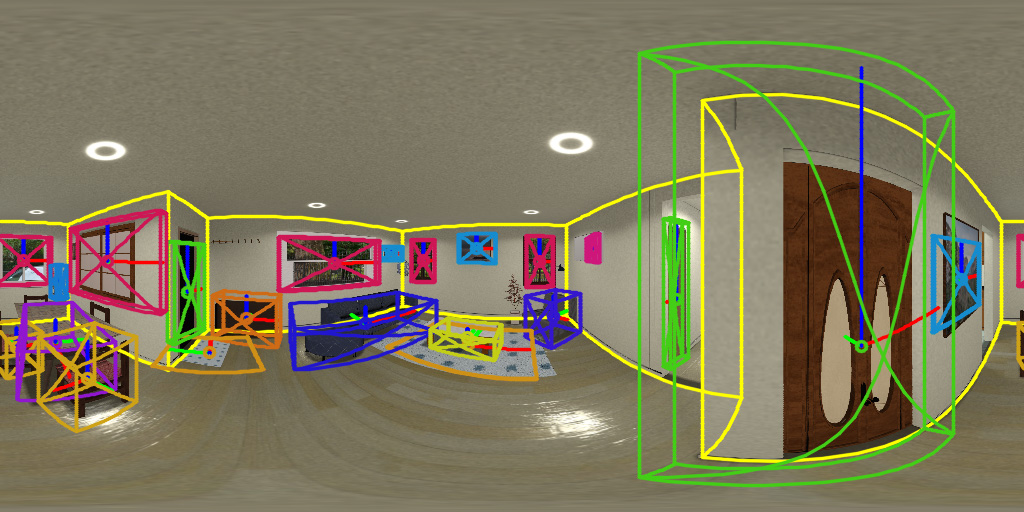}
        	    & \rgb{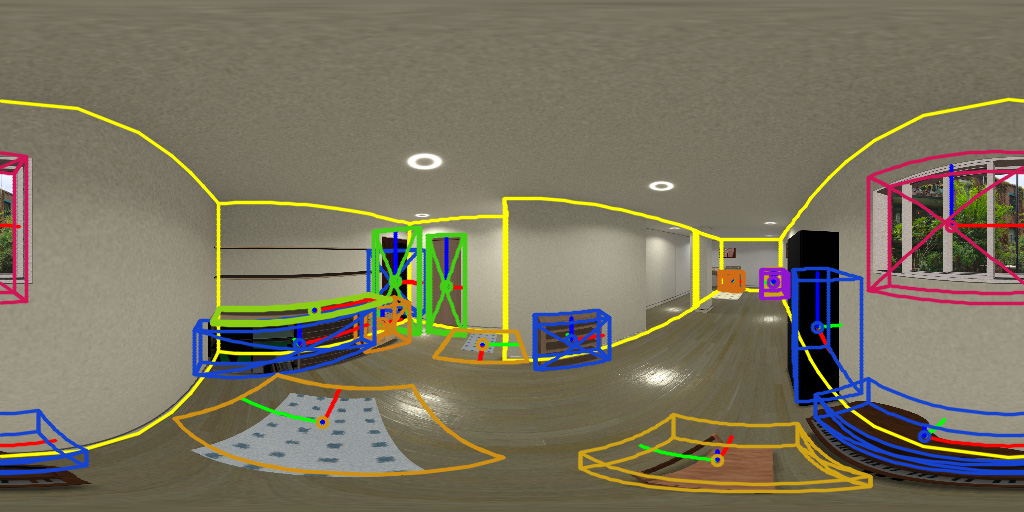}
        	           	    & \rgb{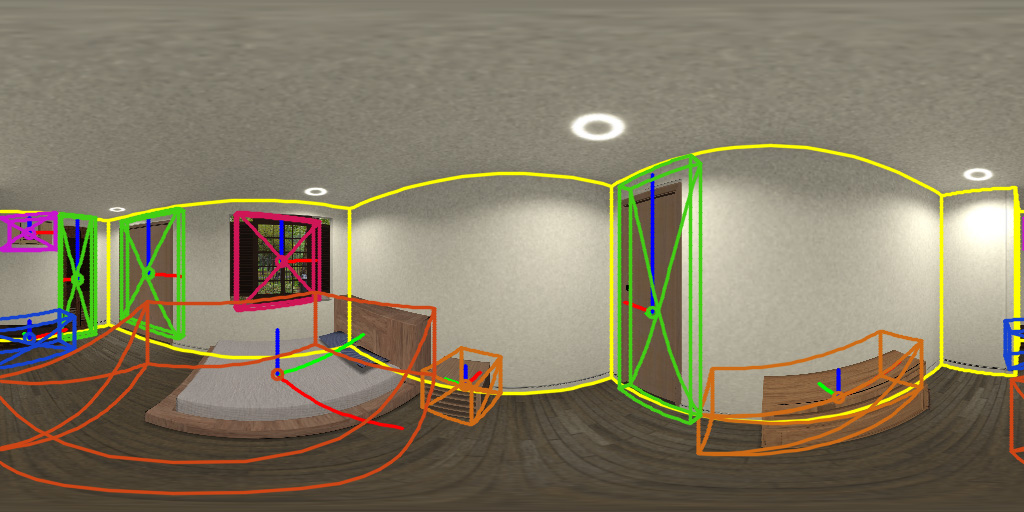}
        	    & \rgb{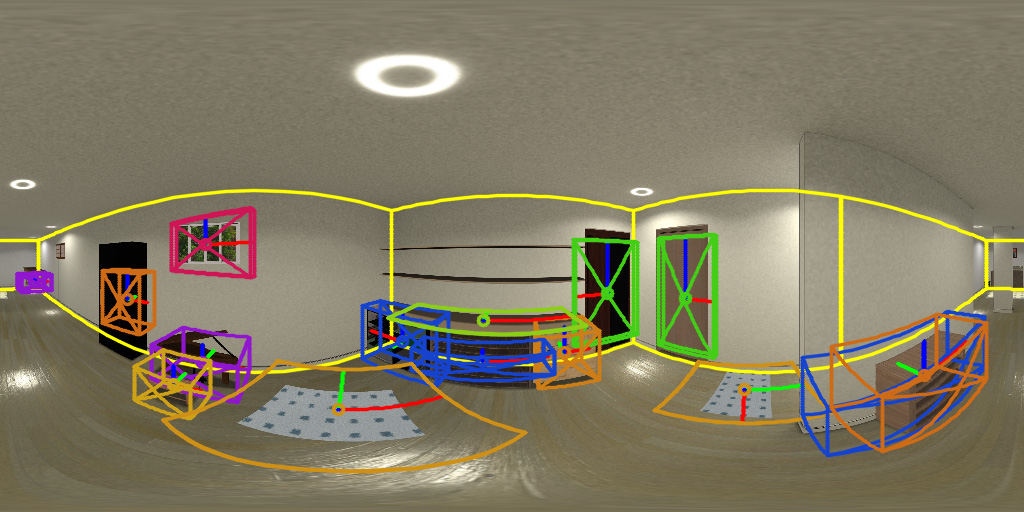}
        	    & \rgb{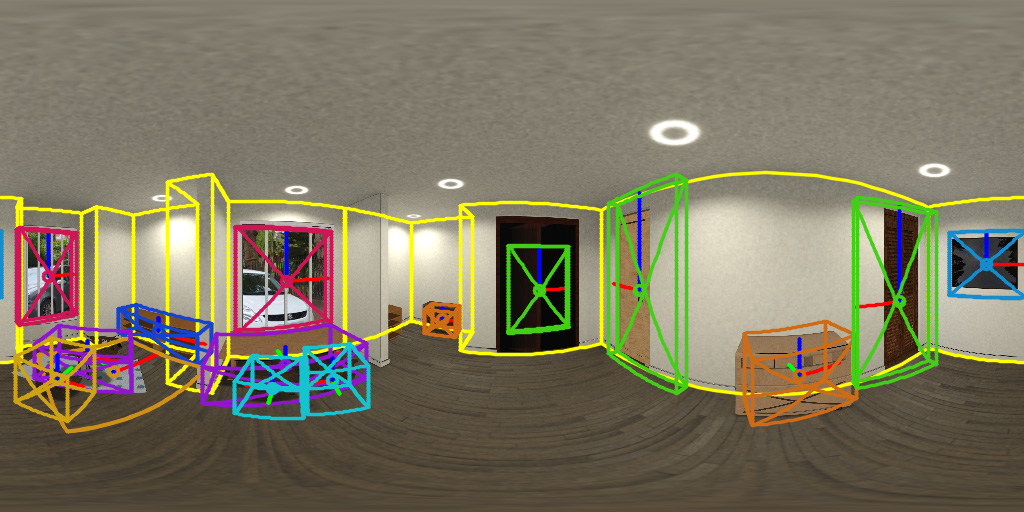}
        	           	    & \rgb{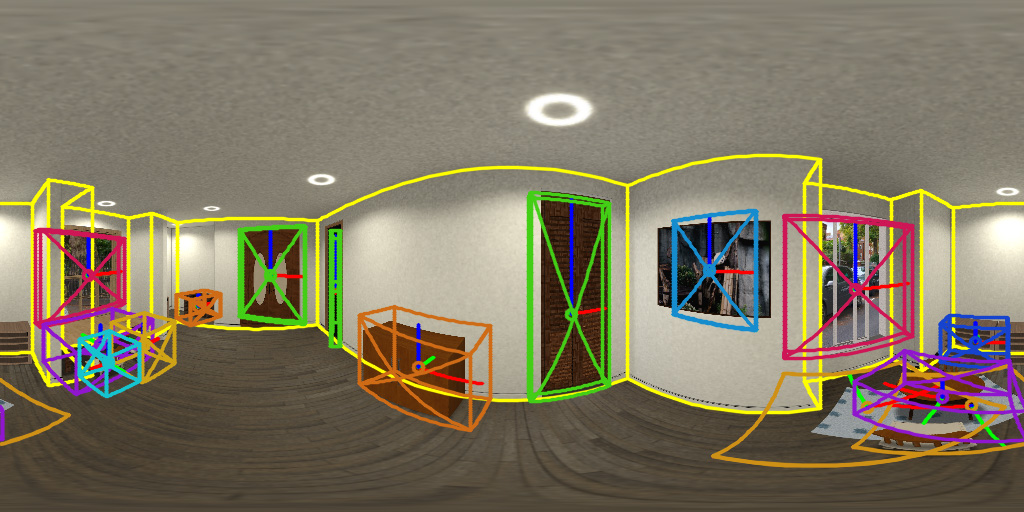}
        	    \\
            &\rot{Ours}
                                & \rgb{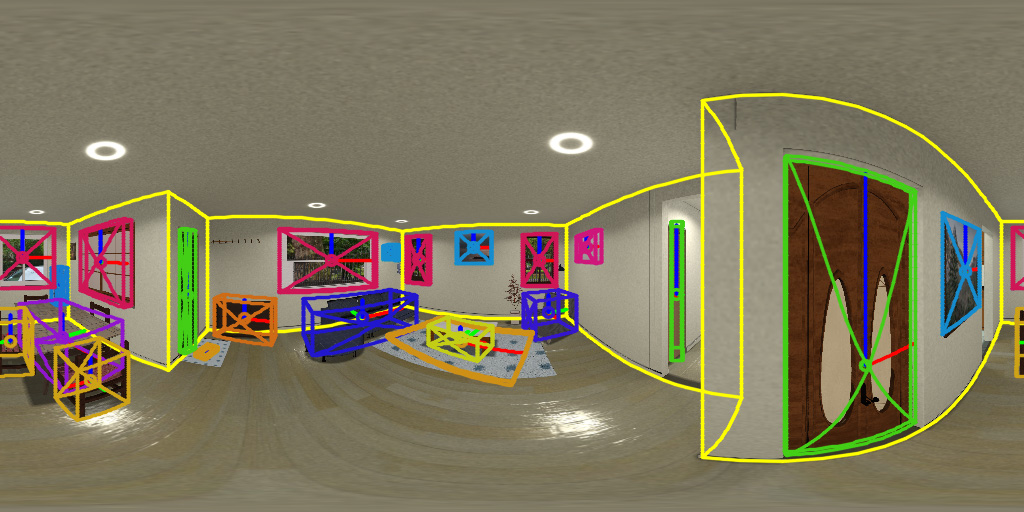}
        	    & \rgb{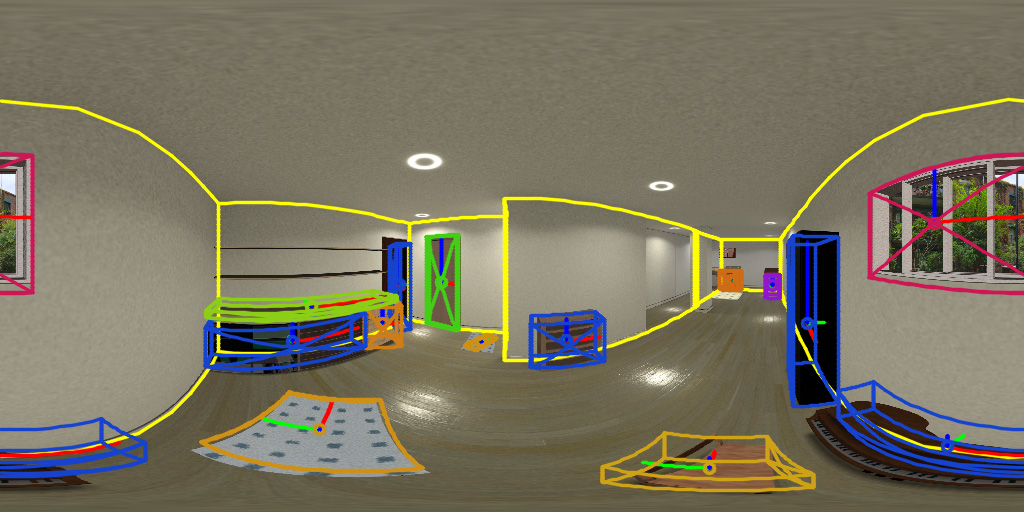}
        	           	    & \rgb{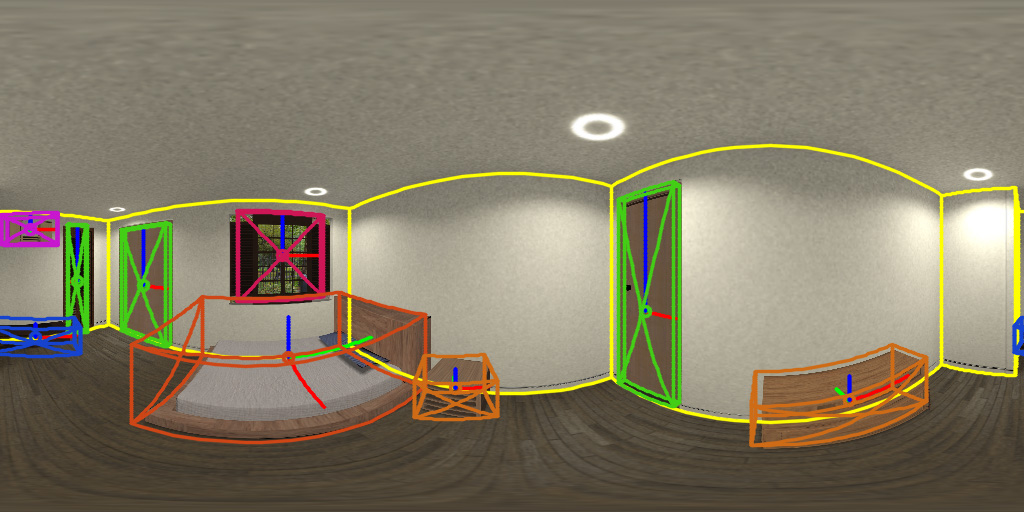}
        	    & \rgb{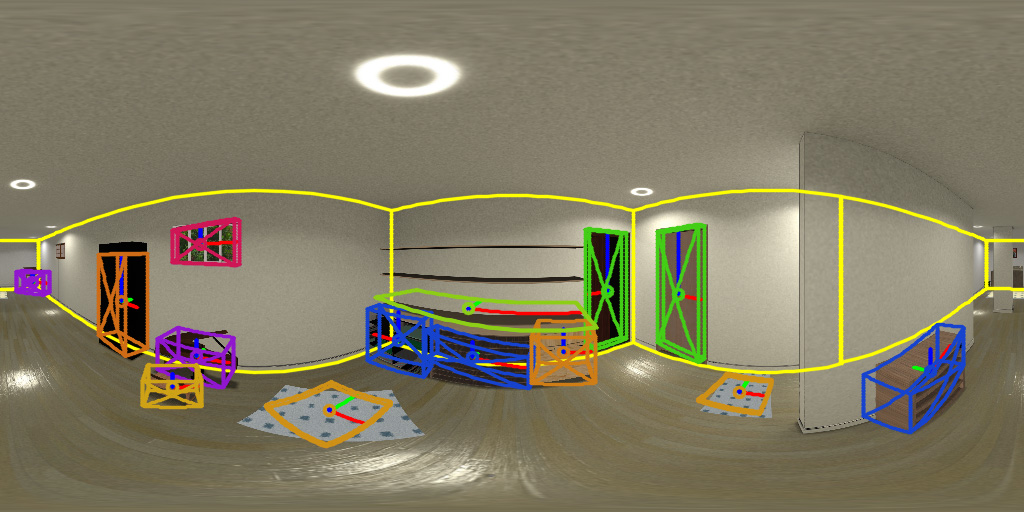}
        	    & \rgb{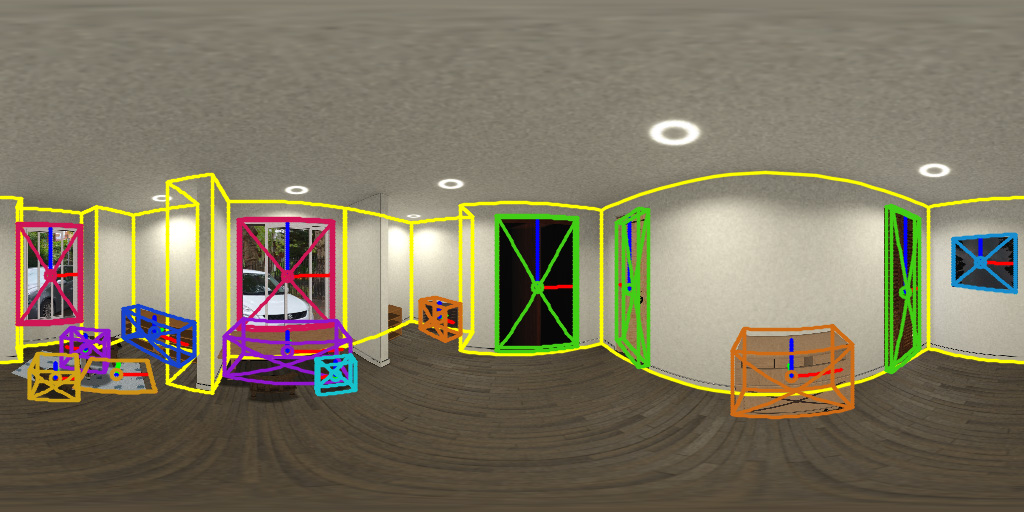}
        	           	    & \rgb{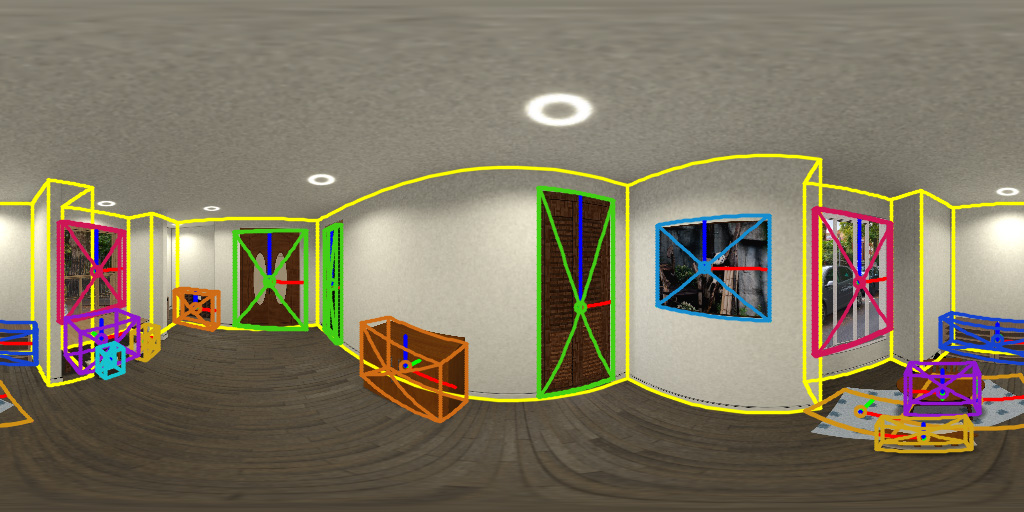}
        	    \\
        	&\rot{GT}
                                & \rgb{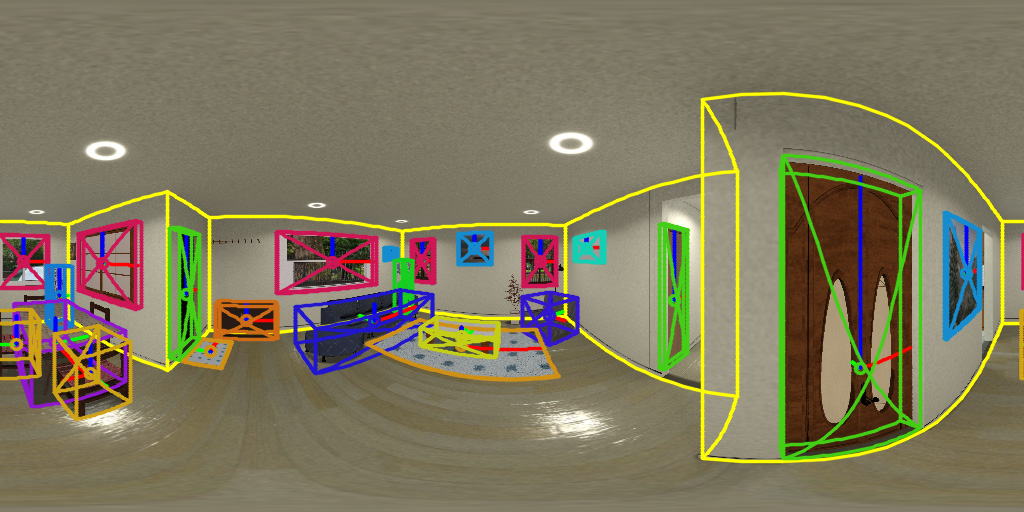}
        	    & \rgb{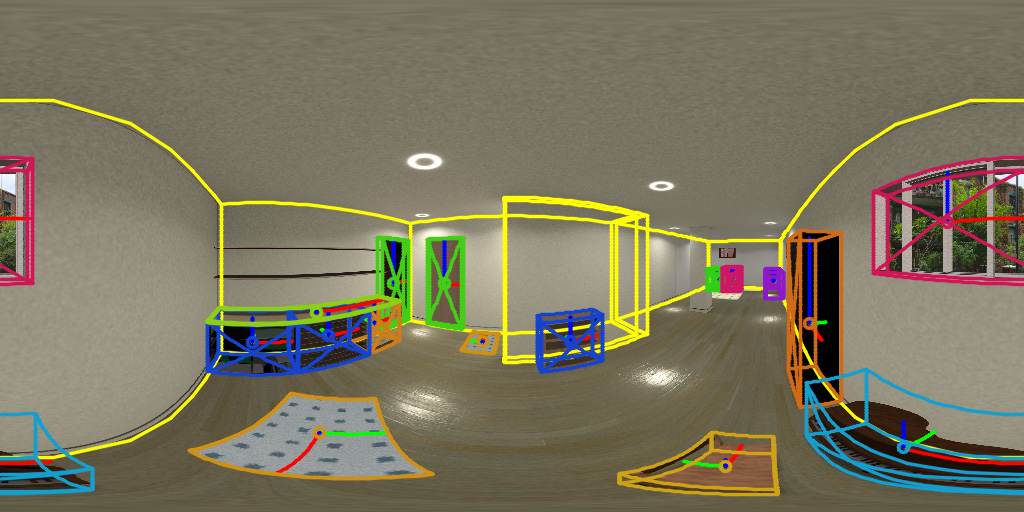}
        	           	    & \rgb{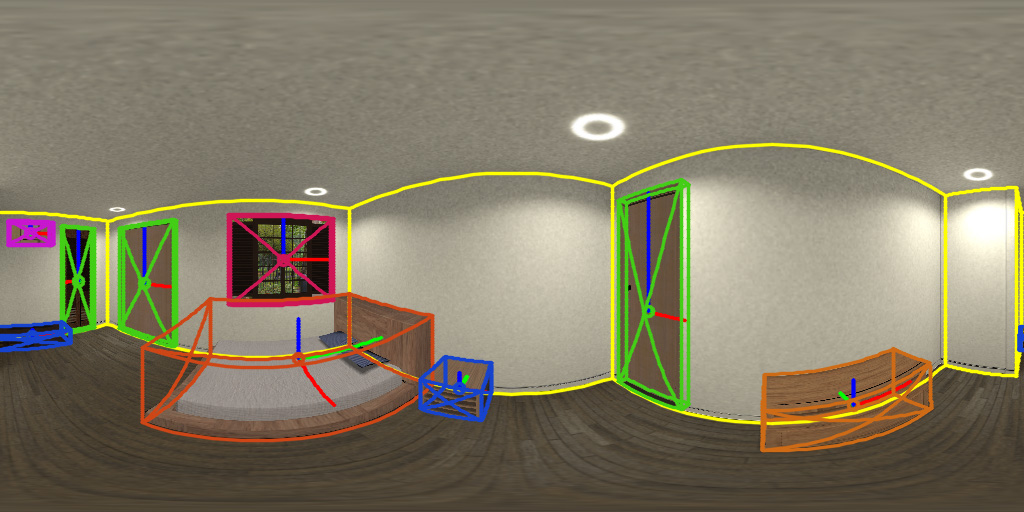}
        	    & \rgb{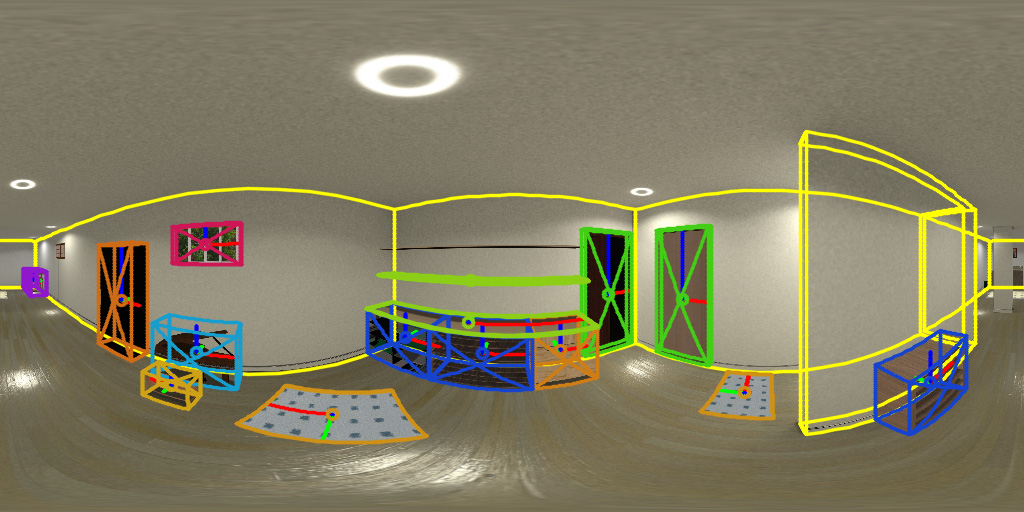}
        	    & \rgb{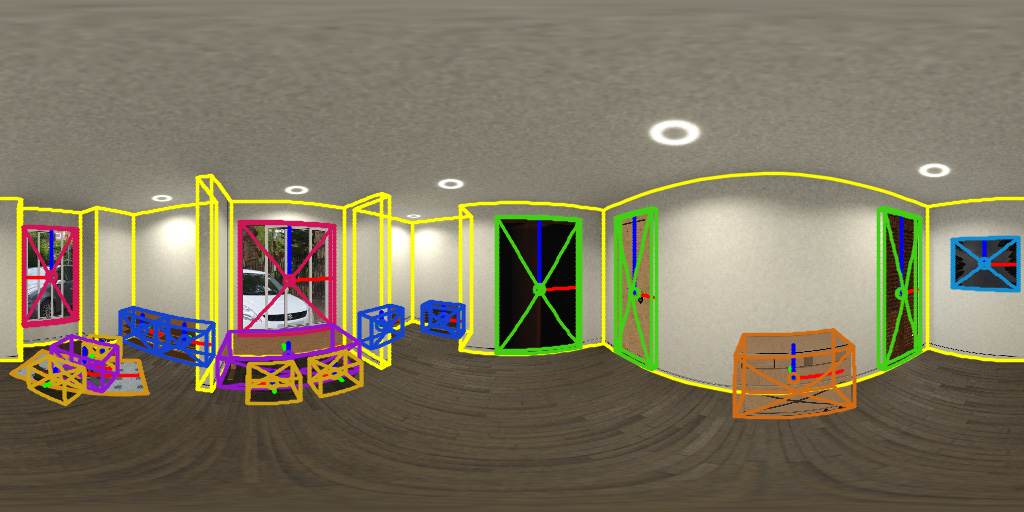}
        	           	    & \rgb{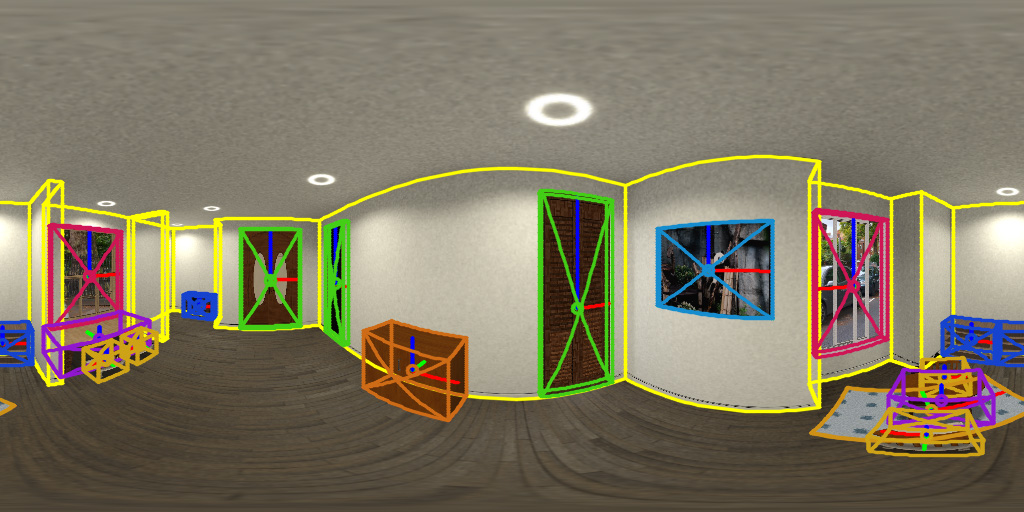}
        	    \\
    	\hline
    	    \multirow{4}{*}[-10ex]{\rot{Scene Reconstruction}}
    	    &\rot{Total3D}
                                & \rgb{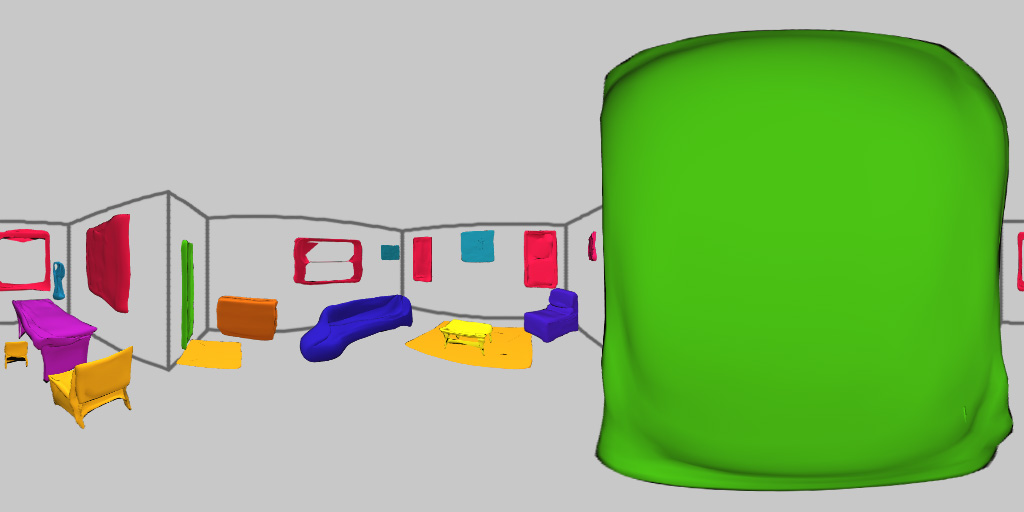}
        	    & \rgb{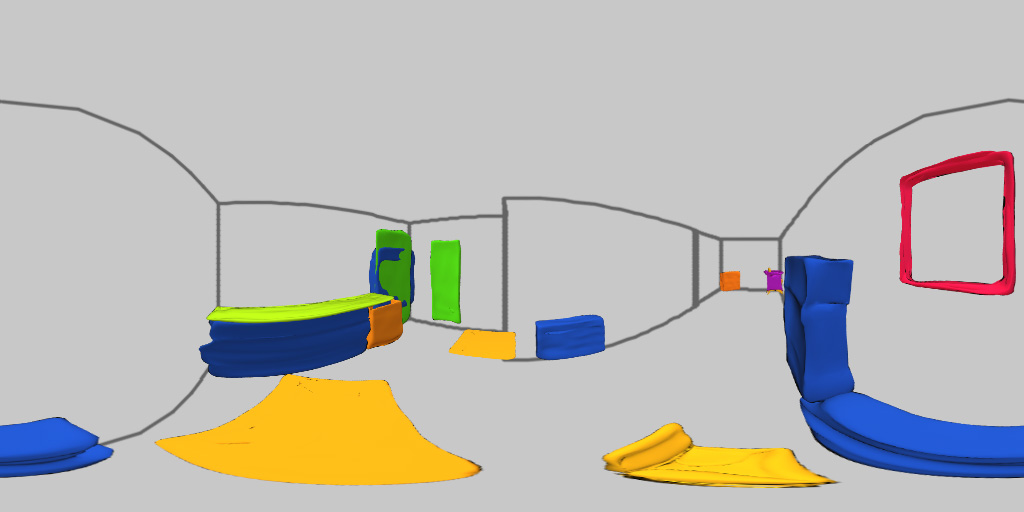}
        	           	    & \rgb{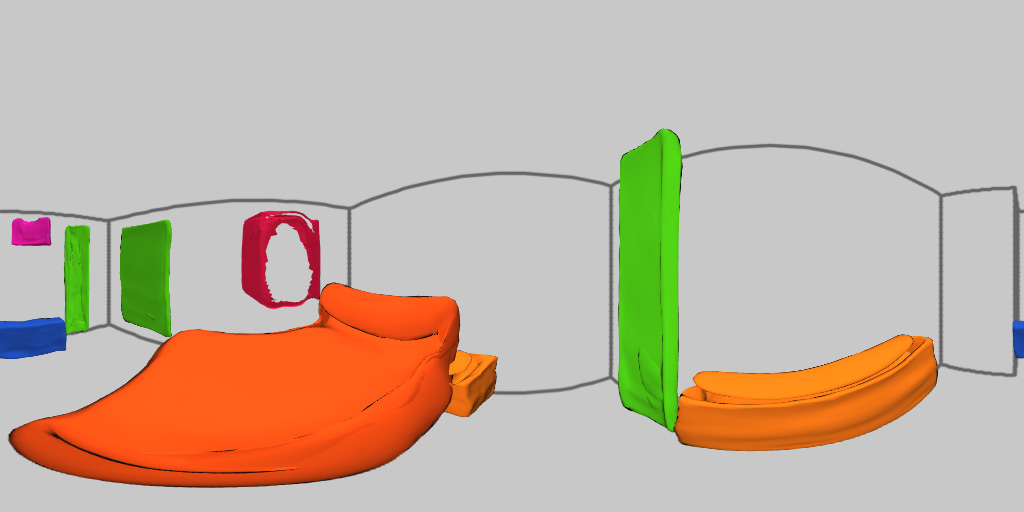}
        	    & \rgb{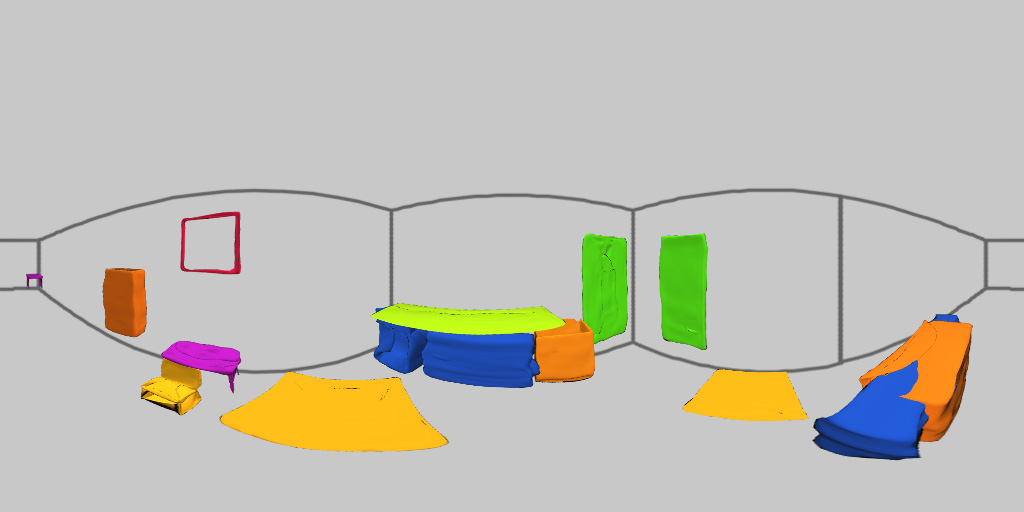}
        	    & \rgb{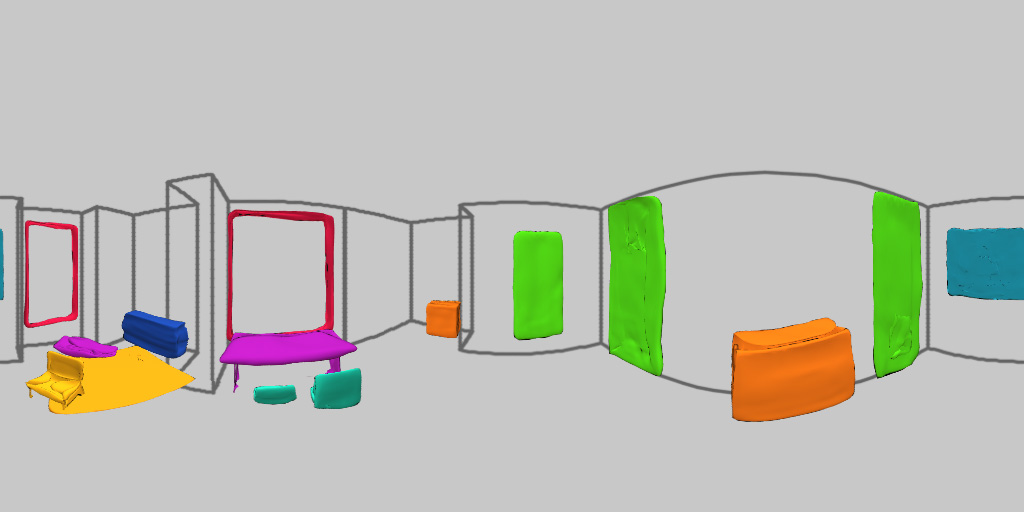}
        	           	    & \rgb{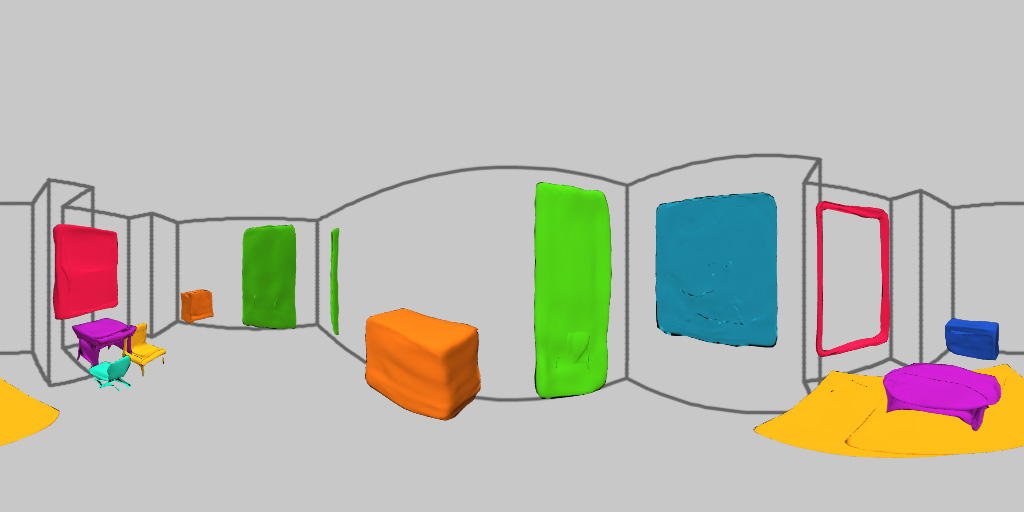}
        	    \\
        	&\rot{Im3D}
                                & \rgb{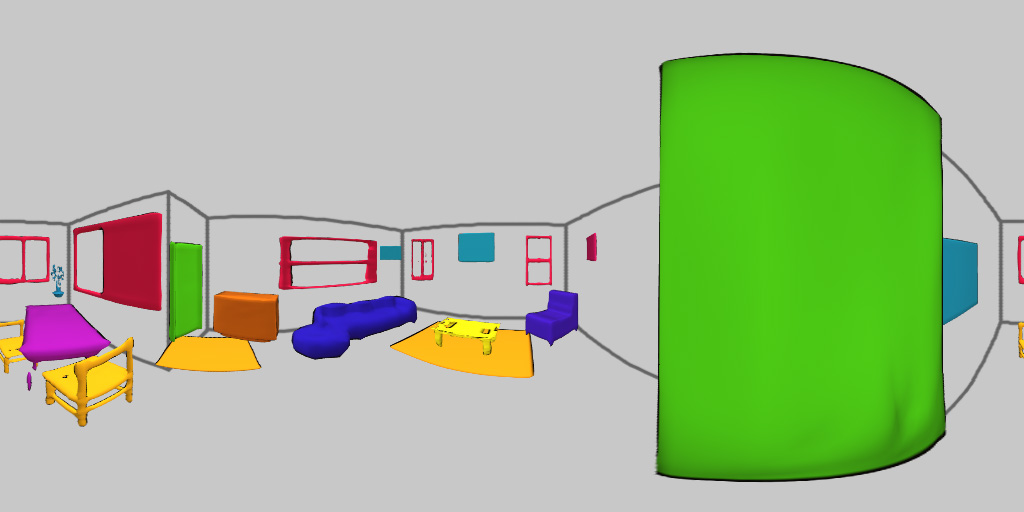}
        	    & \rgb{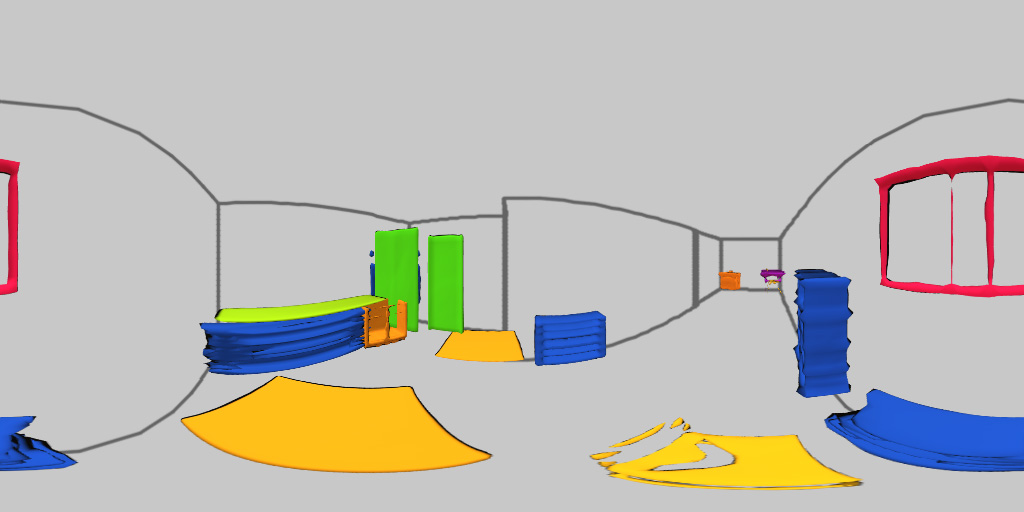}
        	           	    & \rgb{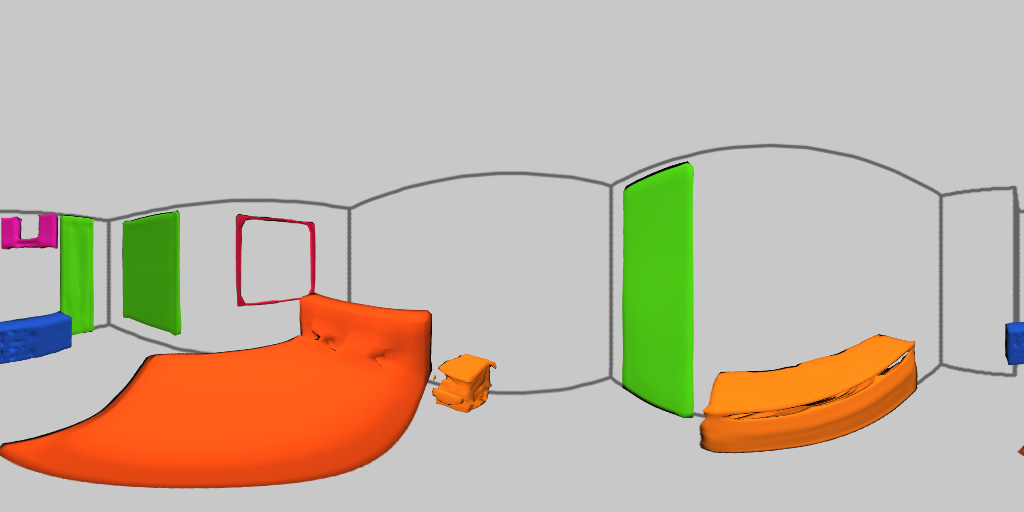}
        	    & \rgb{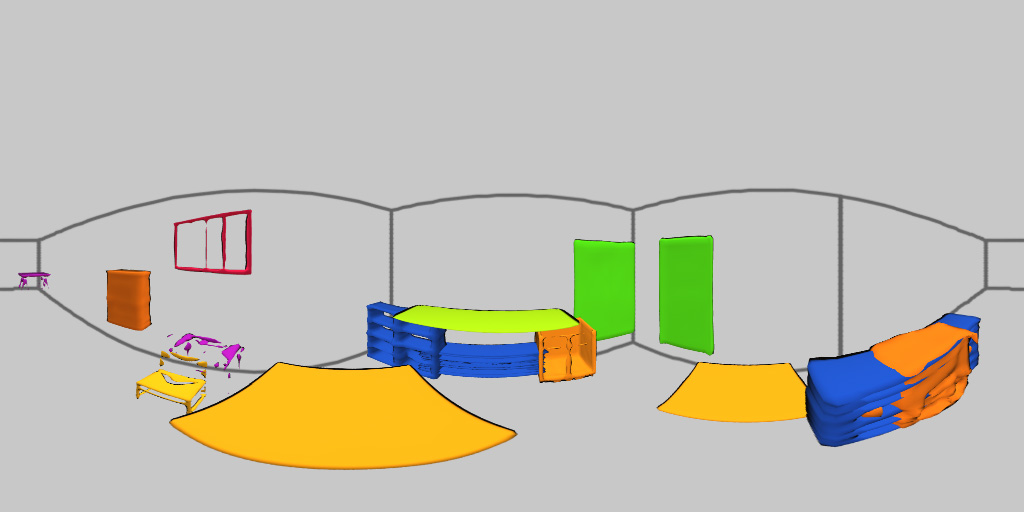}
        	    & \rgb{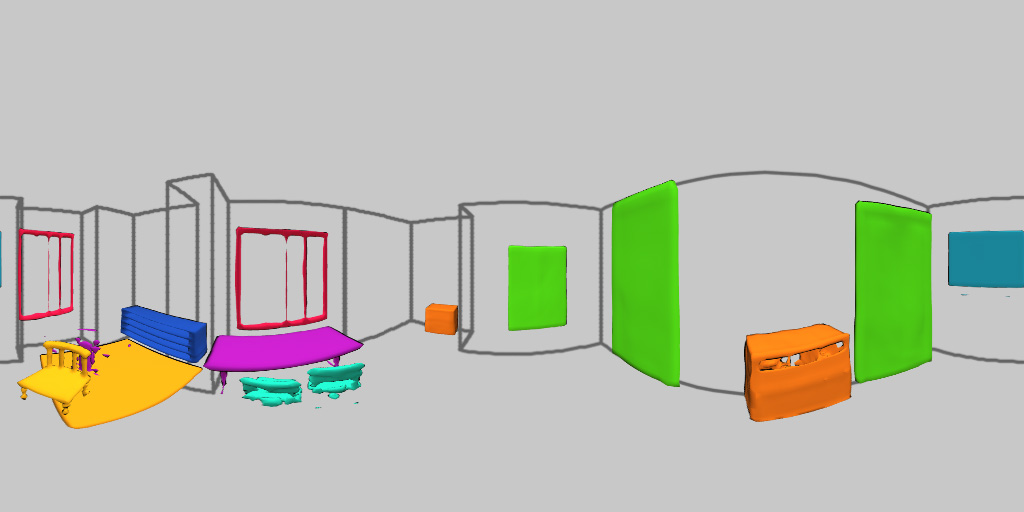}
        	           	    & \rgb{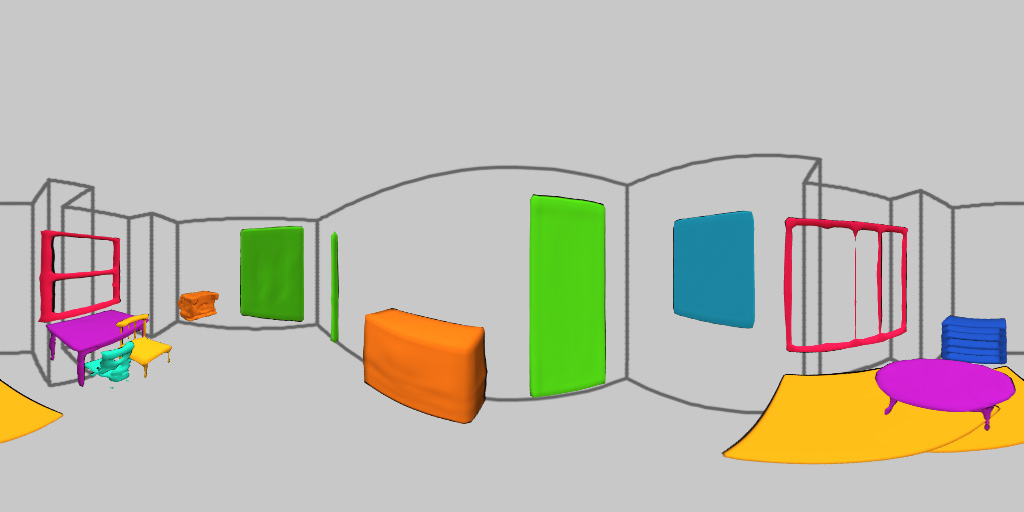}
        	    \\
            &\rot{Ours}
                                & \rgb{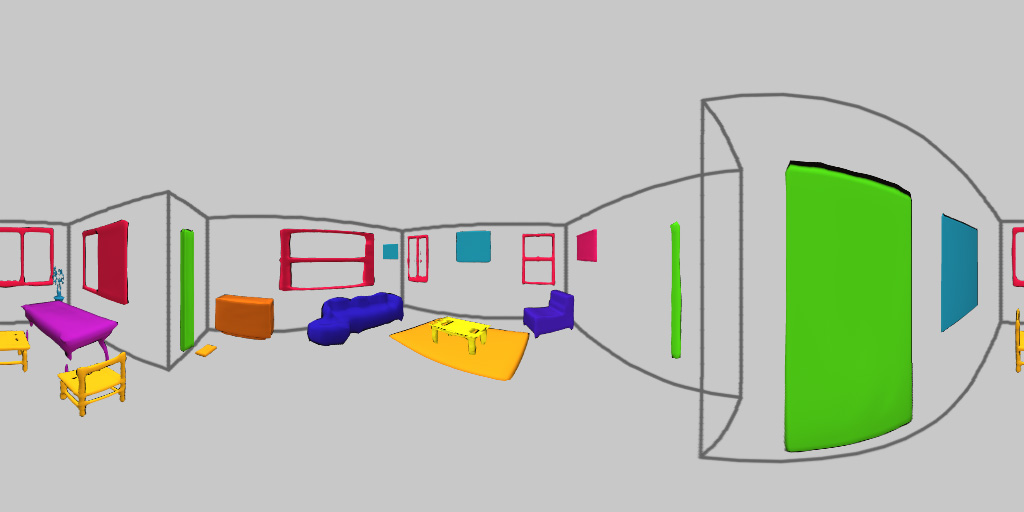}
        	    & \rgb{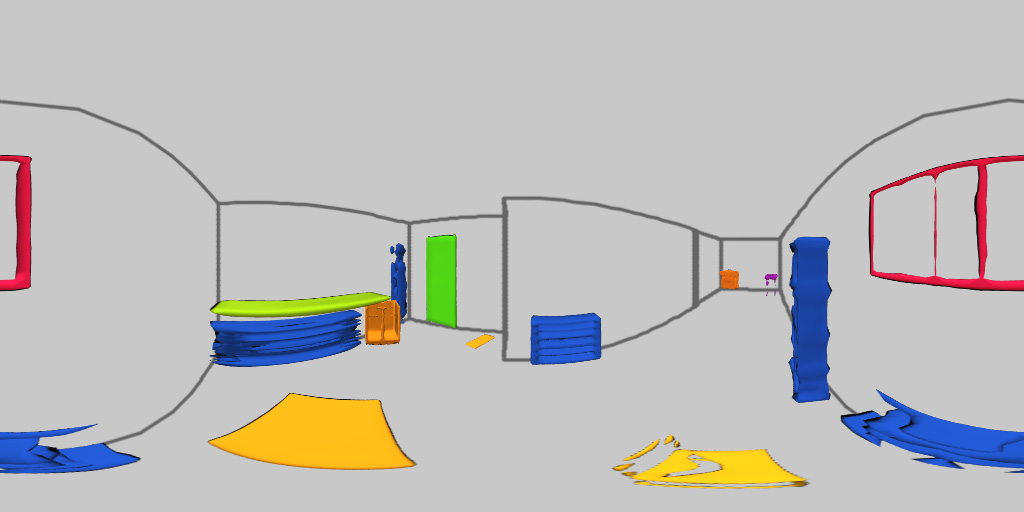}
        	           	    & \rgb{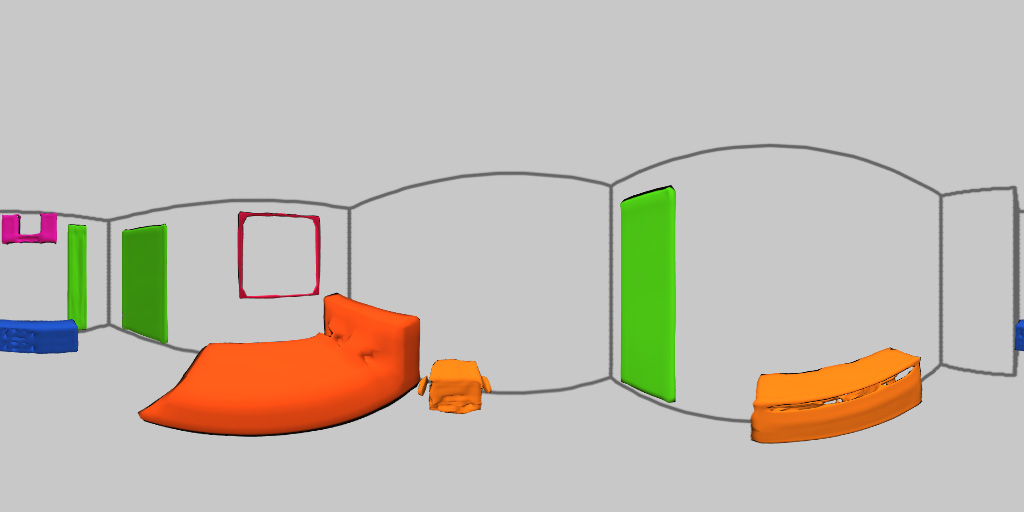}
        	    & \rgb{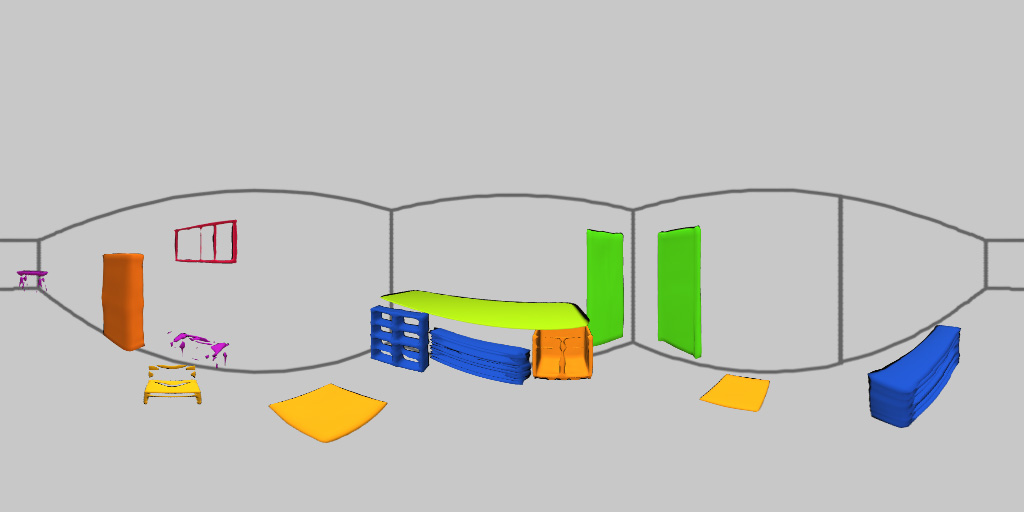}
        	    & \rgb{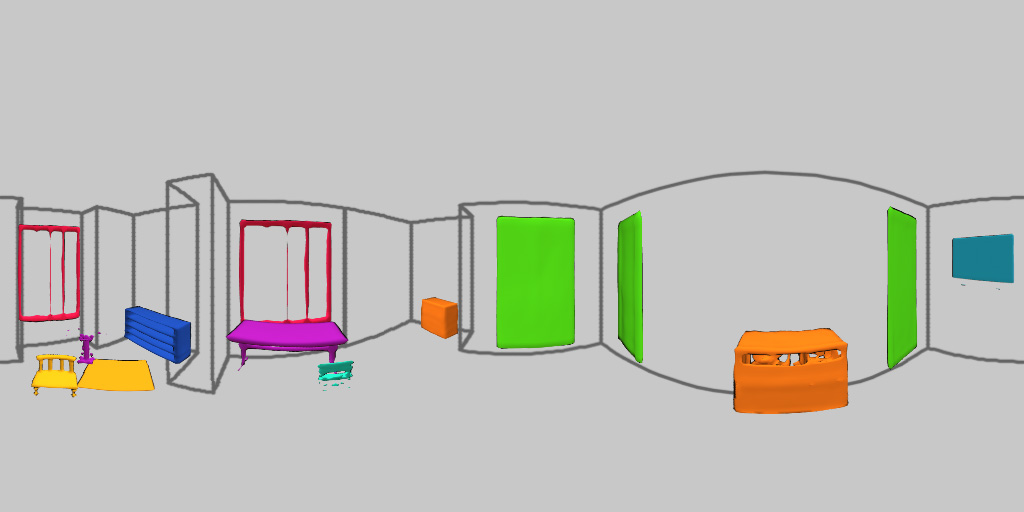}
        	           	    & \rgb{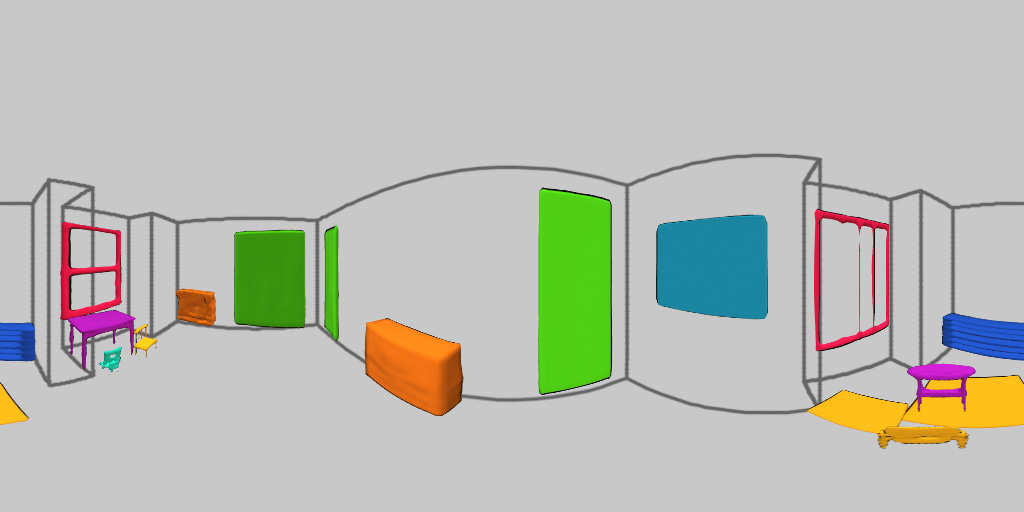}
        	    \\
        	&\rot{GT}
                                & \rgb{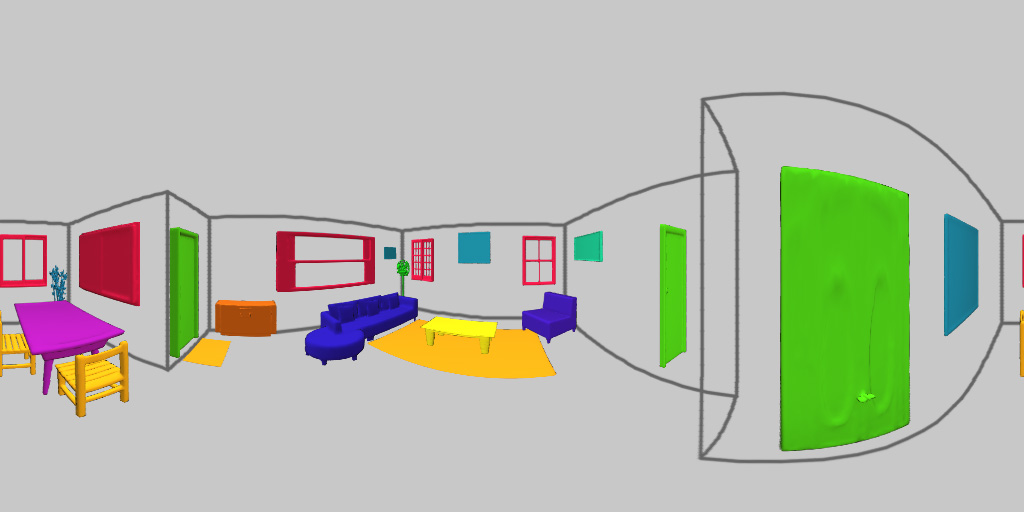}
        	    & \rgb{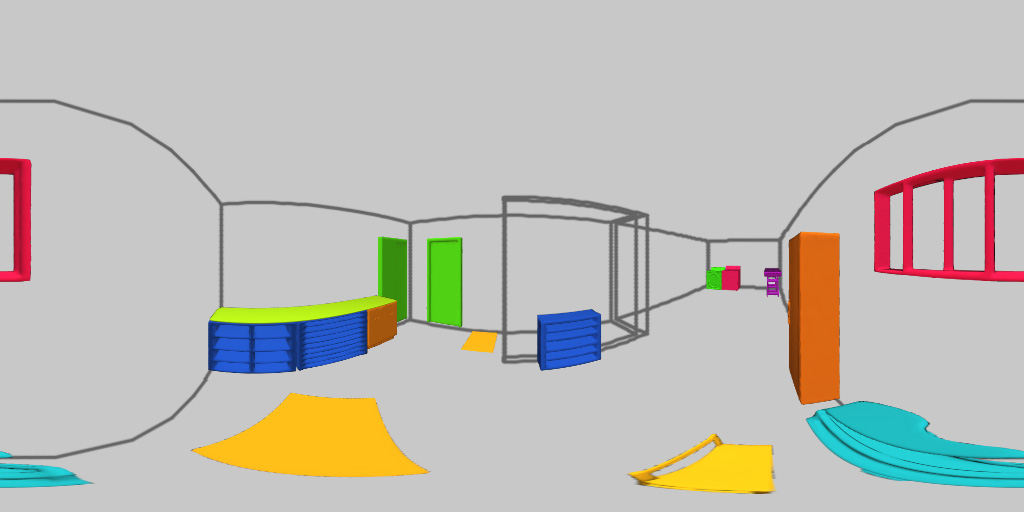}
        	           	    & \rgb{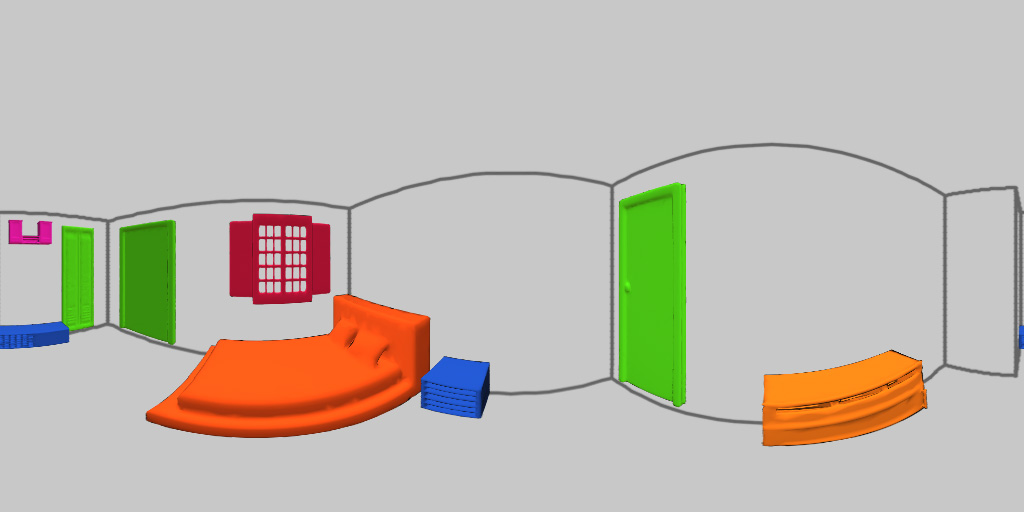}
        	    & \rgb{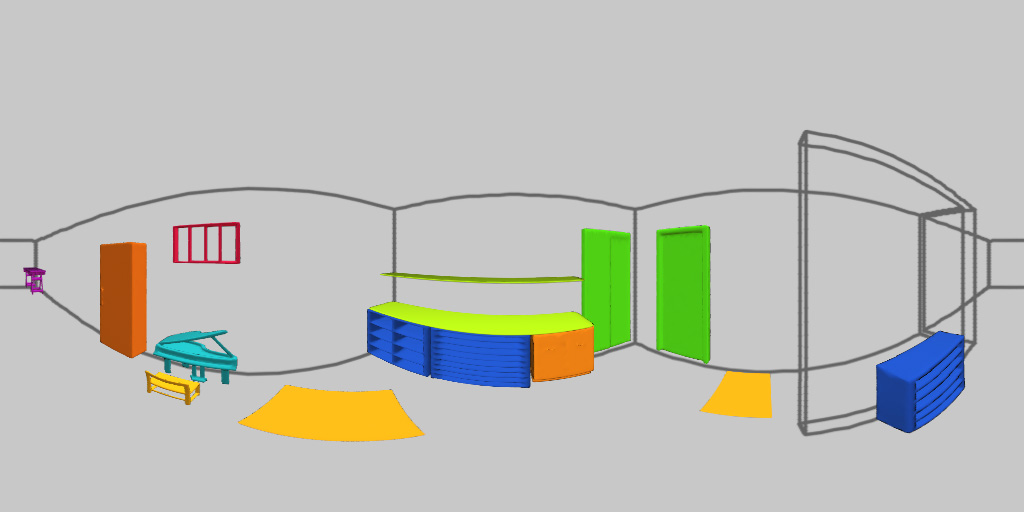}
        	    & \rgb{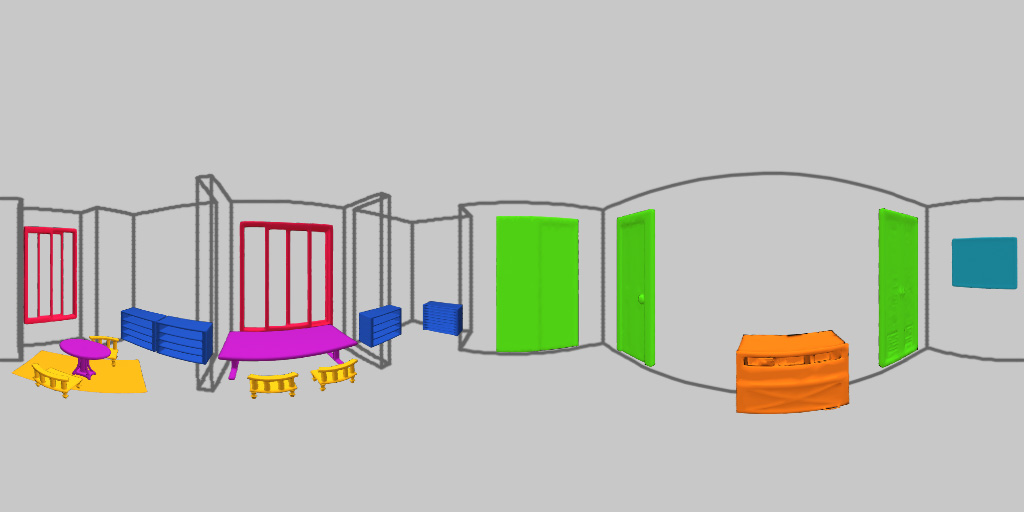}
        	           	    & \rgb{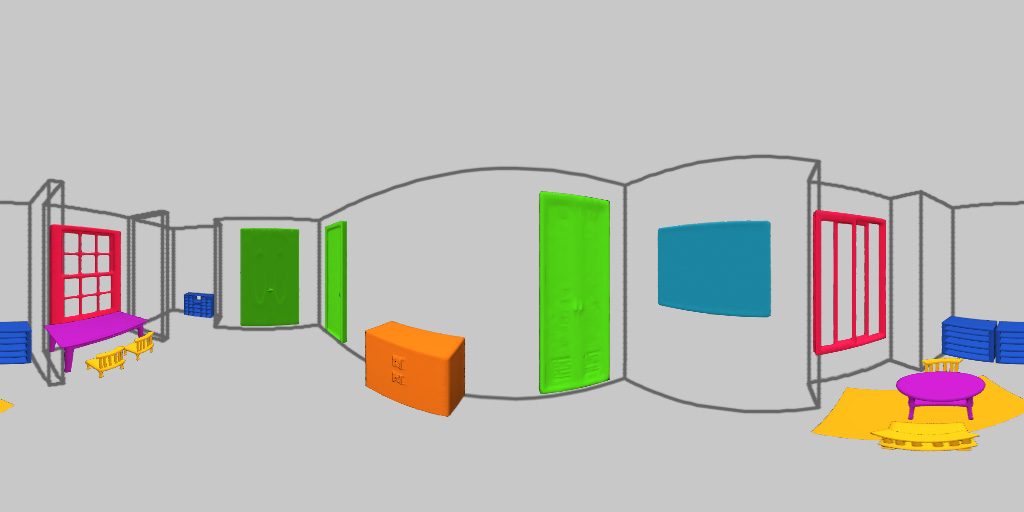}
        	    \\
    \end{tabular}
    \vspace{-0.8em}
	\caption{Qualitative comparison on 3D object detection and scene reconstruction. We compare object detection and compare scene reconstruction results with \yz{Total3D-Pers} and \yz{Im3D-Pers} in both bird's eye view and panorama format.}
	\label{fig:scnrecon}
	\vspace{-1.9em}
\end{figure*}

\subsection{Comparison with SoTA}

\begin{figure}[t]
    \vspace{-1.3em}
	\centering
	\begin{subfigure}[t]{0.23\textwidth}
		\includegraphics[width=\textwidth]  
		{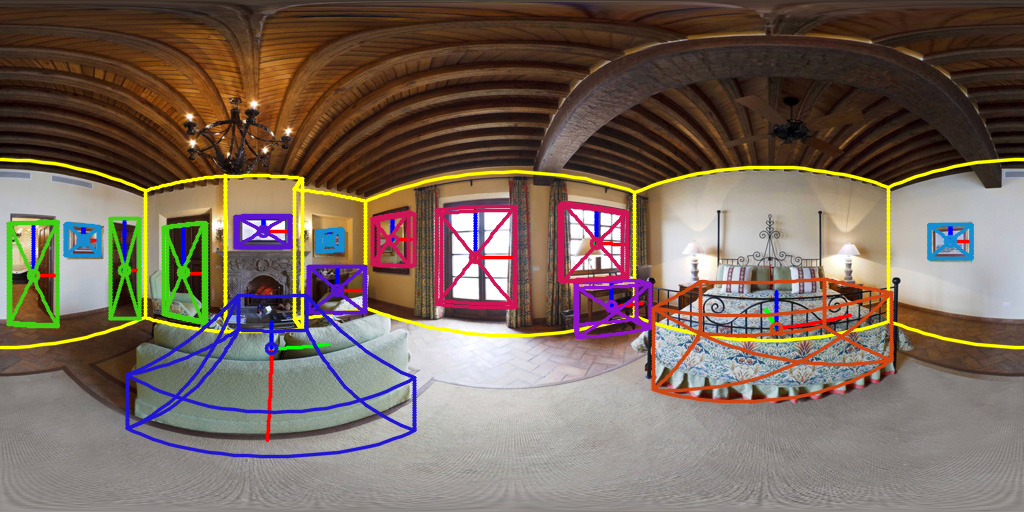}
		\includegraphics[width=\textwidth]
		{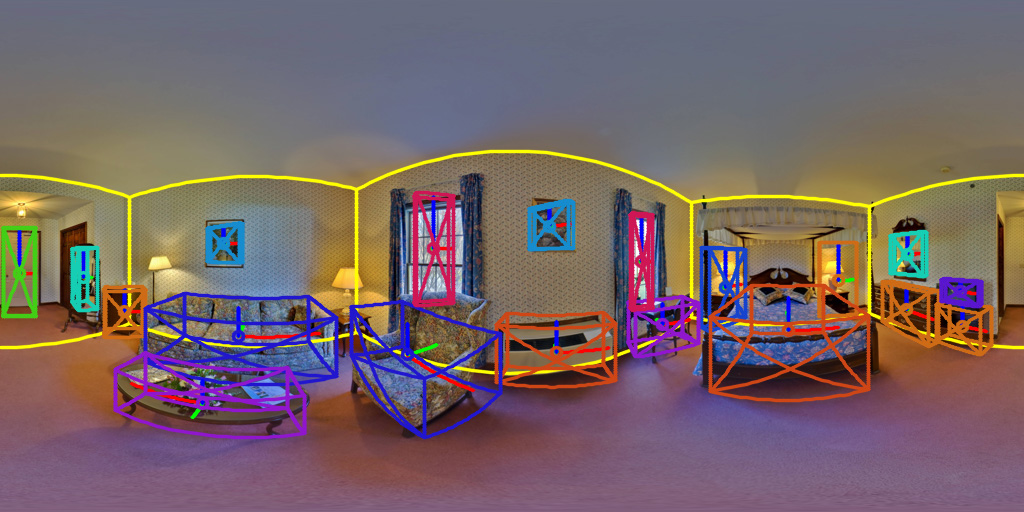}
	\end{subfigure}
	\begin{subfigure}[t]{0.23\textwidth}
		\includegraphics[width=\textwidth]  
		{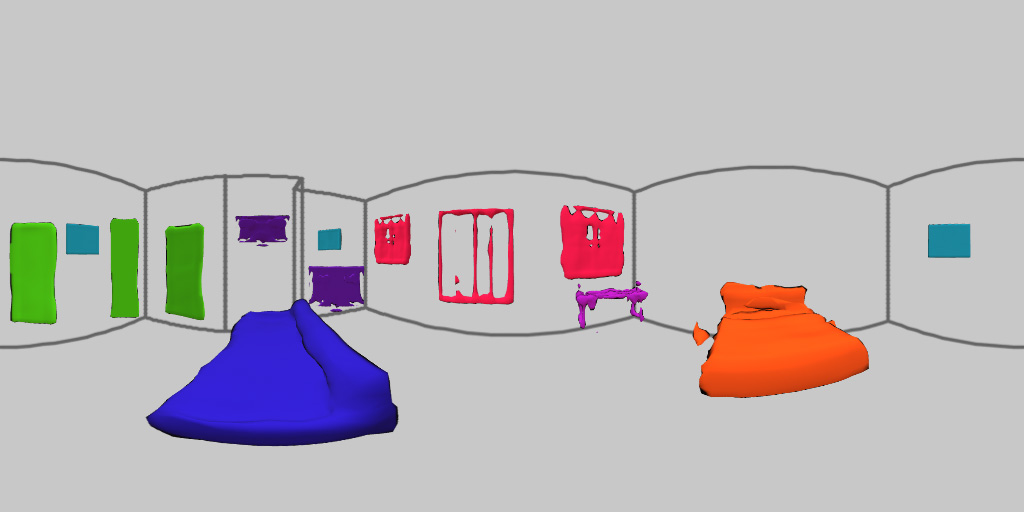}
		\includegraphics[width=\textwidth]
		{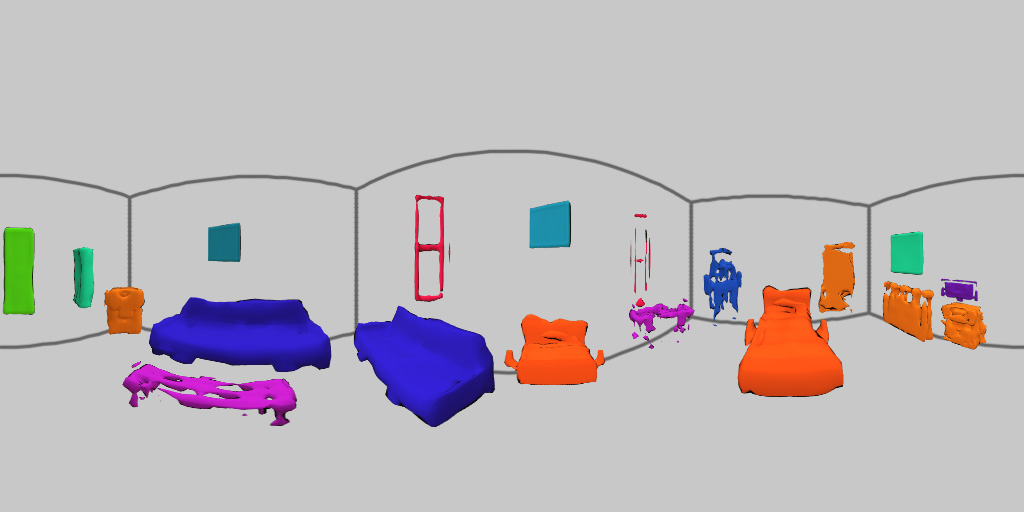}
	\end{subfigure}
	\vspace{-0.6em}
	\caption{Generalization examples on PanoContext dataset.}
	\label{fig:pano_context}
\end{figure}

\noindent \textbf{3D object detection}
We evaluate our method with 
mean average precision (mAP) on 3D object detection and scene understanding.
Following \cite{zhang2021holistic, nie2020total3dunderstanding, huang2018cooperative}, we consider a predicted 3D bounding box with IoU (with ground truth) more than 0.15 as a true detection.
As shown in \autoref{tbl:3d_detection}, our method has a great 
\yz{improvement} over the SoTA even without relation optimization, which mainly benefits from the novel geometric features extracted from initial estimates as well as the constraints between more objects.
To show the generalization ability of our model, we compare it with PanoContext \cite{zhang2014panocontext} on their proposed dataset in \autoref{tbl:sem_seg}.
Since the PanoContext dataset has no object orientation label, 
\zc{we fine-tune our Full model only up to 2D detector.}
The results show that our context model can also generalize to real data with only bottom-up models fine-tuned.
We also show corresponding qualitative results in \autoref{fig:scnrecon} and \autoref{fig:pano_context}.

\noindent \textbf{Physical violation}
To \yz{highlight the improvement benefit from relation optimization,}
\ie, collision avoidance, we \yz{calculate}
average collision times per scene and the average number of objects with collision.
\yz{We also report the number of collision of each kind between object to object/ceiling/floor/wall.}
The results shown in \autoref{tbl:collision} indicate that our method outperforms SoTA methods from all perspectives
\yz{in preventing physical collision,}
while the gap from the ablated version further illustrating the importance of relation optimization.
\yz{The analysis in ablation study (\autoref{sec:ablation}) further demonstrates the importance of relation optimization in delivering context plausible results.}

\noindent \textbf{Holistic Scene Reconstruction}
We compare the reconstruction results qualitatively with the \zc{\yz{Total3D-Pano}} and \zc{\yz{Im3D-Pano}} in \autoref{fig:scnrecon}.
\yz{Our method shows overall the best performance on object pose estimation and shape reconstruction.}
We also achieve more reasonable object-wall relations looking from the bird's eye view.
\yz{For example in the third column, our method places the bed in the bottom-left corner right next to the wall, while Im3D and Total3D all fail with considerable error.}

\begin{table}[t]
    \vspace{-0.8em}
	\begin{center}
	    \resizebox{1.\columnwidth}{!}{
    		\begin{tabular}{|l|c|c c c c|}
    			\hline
    			\multirow{2}{*}{Method} & collision times & \multicolumn{4}{c|}{objects having collision with} \\\cline{3-6}
    			& among objects & object & ceil & floor & wall \\
    			\hline
    			\yz{Total3D-Pers}  & 3.45 & 4.96 & 0.09 & 2.70 & 2.68\\     			\yz{Total3D-Pano}  & 3.41 & 4.87 & 0.14 & 2.81 & 2.66 \\     			\yz{Im3D-Pers}     & 3.16 & 4.54 & 0.03 & 1.79 & 2.42 \\     			\yz{Im3D-Pano}     & 2.62 & 3.98 & 0.02 & 2.36 & 2.26\\     			Ours (w/o. RO)  & 2.68 & 4.08 & \textbf{0.01} & 1.76 & 2.23 \\     			Ours (Full)       & \textbf{0.86} & \textbf{1.50} & 0.04 & \textbf{0.45} & \textbf{1.33} \\     			\hline
    		\end{tabular}
        }
	\end{center}
	\vspace{-1.5em}
	\caption{Physical violation. We compare our methods with average collision times per scene to verify the effect of the proposed relation optimization. The collision detection is done with a toleration of 0.1m.}
	\label{tbl:collision}
\end{table}

\subsection{Ablation Study}
\label{sec:ablation}

To evaluate the proposed relation and object features and different parts of the proposed relation optimization, we conduct ablation studies by removing different parts of our method and make comparisons on 3D object detection, collision, and relation estimation.
For binary relations, \eg, contact and relative distance, we compare Truth Positive Rate (TPR) and Truth Negative Rate (TNR).
For rotation relation, we compare mean absolute error in degrees.

\vspace{0.2em}
\noindent \textbf{Do the Proposed Features Matter?}
As described in \autoref{sec:rgcn}, we propose separate features for relation nodes and object nodes to 
\yz{encode} explicit collision information and 3D geometry priors to RGCN.
To \yz{show the necessity and effectiveness of these features,}
we remove the relation features (w/o. Fr) and the object features (w/o. Fo) respectively.
As shown in \autoref{tbl:abl_study}, 
\yz{removing any feature will cause a drop in object detection mAP, as well as attachment classification.}
\yz{This is as expected as both relation feature and object feature provides critical information to measure the distance/collision among objects and layouts.}

\vspace{0.2em}
\noindent \textbf{Does Relation Optimization Matter and How Each \yz{Loss Term} Contributes?}
Our proposed relation optimization provides an end-to-end solution to hard code collision, contact, and rotation constraints into \zc{RGCN},
with the purpose of obtaining more physically plausible and accurate detection results.
By removing it (w/o. RO), we observe a great drop on mAP and average collision times per scene.
\yz{We found our RO also improves Total3D and Im3D with predicted relation, and see supp. materials for more details.}
We also conduct study (w/o. ${E}^{c}$, w/o. ${E}^{r}$, w/o. ${E}^{o}$) on different \zc{terms} to see how they contributes to the final improvement.
The missing of collision \zc{term} hurts the average collision times most, which further illustrates its \yz{importance} on collision avoidance.
We also observe even greater drops on mAP when removing ${E}^{r}$ and ${E}^{o}$ independently, which shows that our proposed \zc{terms} collaborate together to improve 3D detection.

\begin{table}[!t]
    \vspace{-1.3em}
	\begin{center}
	    \resizebox{1.\columnwidth}{!}{
    		\begin{tabular}{|l|c|c|c|c|c|c|c|c|c|}
    			\hline
    			Method & obj \yz{attach} & wall \yz{attach} & obj rot (\degree) & wall rot (\degree) & mAP & avg col \\
    			\hline
    			w/o. Fr         & 0.30 & 0.71 & 64.21 & 52.03 & 33.22 & 0.74  \\     			w/o. Fo         & 0.46 & 0.74 & \textbf{62.83} & \textbf{43.46} & 33.32 & 0.83 \\     			w/o. RO         & - & - & - & - & 30.91 & 2.68 \\     			w/o. ${E}^{c}$  & - & - & - & - & 30.42 & 2.05 \\     			w/o. ${E}^{r}$  & - & - & - & - & 29.88 & 0.46 \\     			w/o. ${E}^{o}$  & - & - & - & - & 25.30 & \textbf{0.09} \\     			Full            & \textbf{0.47} & \textbf{0.76} & 62.97 & 43.72 & \textbf{33.59} & 0.86 \\     			\hline
    		\end{tabular}
    	}
	\end{center}
	\vspace{-1.5em}
	\caption{Ablation study. We compare F1 on binary-classified object-object attachment and object-wall attachment relations. For rotation relation classification, we compare mean absolute error in degrees. We evaluate 3D object detection with mAP of all 57 categories and physical violation with average collision times per scene.}
	\label{tbl:abl_study}
\end{table}

\begin{table}[!t]
    \vspace{-0.5em}
	\begin{center}
	    \resizebox{1.0\columnwidth}{!}{
    		\begin{tabular}{|l|c|c|c|c|c|c|}
    			\hline
    			FoV(\degree) & 360 & 180 & 120 & 90 & 60 & 30 \\
    			\hline
    			w/o. RO & 30.91 & 27.4 & 24.77 & 25.90 & 26.16 & 25.51 \\
    			Full & 33.92 & 26.09 & 24.34 & 23.01 & 21.72 & 18.52 \\
    			\hline
    		\end{tabular}
		}
	\end{center}
	\vspace{-1.5em}
	\caption{mAP vs FoV. By narrowing the FoV of our model, the performance drops greatly, especially for our full model.}
	\label{tbl:fov}
\end{table}

\vspace{0.5em}
\noindent \textbf{Is Panorama 360\degree FoV Helping \yz{RGCN} and RO?}
Following the same procedure of splitting detection results by horizontal FoVs when making Total3D and Im3D working on panorama, we conduct ablation experiments on our proposed method by narrowing the FoV of each split.
We compare our Full model, with or without RO, with different FoVs with mAP on full 57 categories in \autoref{tbl:fov}.
The results show that limiting the message flow within small FoVs hurts the performance, which means that our RGCN and RO are really taking advantage of the whole scene context to estimate relations and optimize object detection.

\section{Conclusion}
This paper presents a novel method for holistic 3D scene understanding from a single full-view panorama image, which recovers the 3D room layout and the shape, pose, position, and semantic category of each object in the scene.
To exploit the rich context information in the panorama image, we employ the graph neural network and design a novel context model to predict the relationship among objects and room layout, which will be further utilized by a novel differentiable relationship-based optimization module to refine the initial estimation. Due to the limitation of existing datasets for holistic 3D scene understanding, we present a new synthetic dataset. 
Experiments validate the effectiveness of each module in our method, and show that our method reaches the SoTA performance.
\zc{Future directions could include simplifying the terms of RO and unifying different modules into a single framework.}

\vspace{0.5em}
\noindent \textbf{Acknowledgement:} This research was supported in part by National Natural Science Foundation of China (NSFC) under grants No.61872067 and No.61720106004.

\clearpage
{\small
\bibliographystyle{ieee_fullname}
\bibliography{egbib}
}

\newpage\appendix

\renewcommand\thesection{\Alph{section}}
\renewcommand\thetable{\Alph{table}}
\renewcommand\thefigure{\Alph{figure}}
\setcounter{section}{0}
\setcounter{table}{0}
\setcounter{figure}{0}

\begin{center}
{\Large \textbf{Supplementary Material}}
\end{center}

In this supplementary material, we provide synthetic dataset examples, network architecture details, and implementation details. We also provide visualization \yz{of} relation optimization, 3D detection performance on all categories, more qualitative results, more comparison on Structured3D, and discussion of failure cases. 

\section{Dataset Examples}
\begin{figure*}[!ht]
	\centering
	\scriptsize
	\newcommand{\rgb}[1]{\raisebox{-0.5\height}{\includegraphics[width=.147\linewidth]{#1}}}
	\newcommand{\rot}[1]{\rotatebox[origin=c]{90}{#1}}
	\def\arraystretch{0.5}	\begin{tabular}{c*{6}{c@{\hspace{1px}}}}
        \rule{0pt}{5px}
            \rot{RGB}
            & \rgb{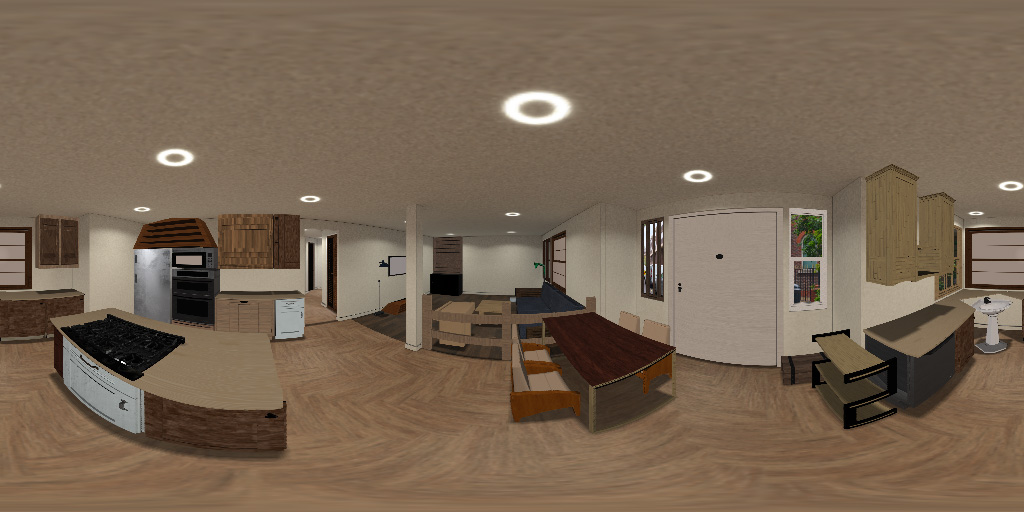}
    	    & \rgb{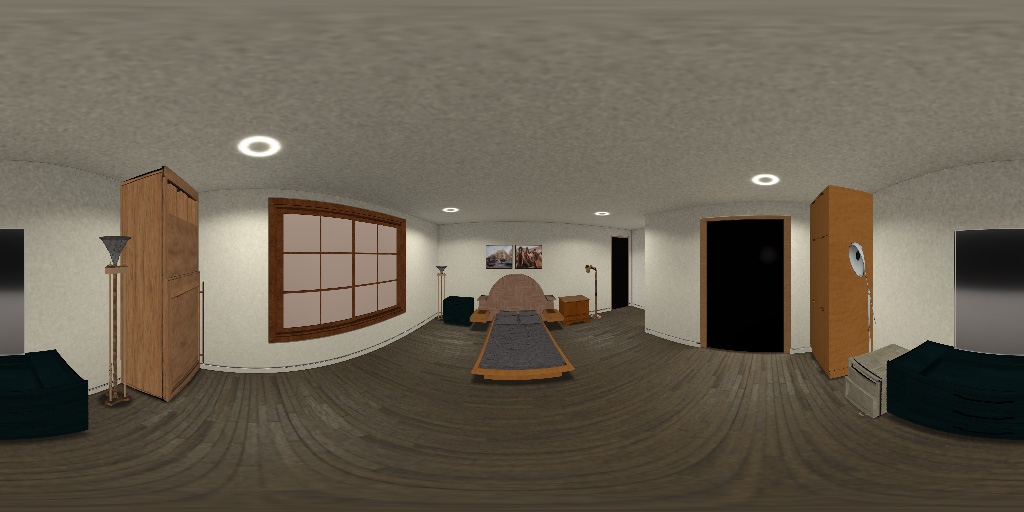}
    	    & \rgb{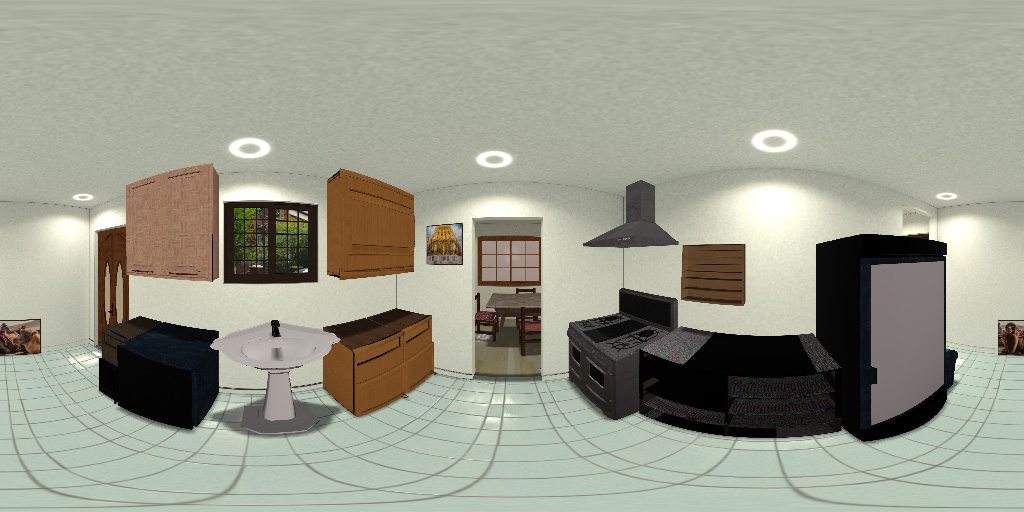}
    	    & \rgb{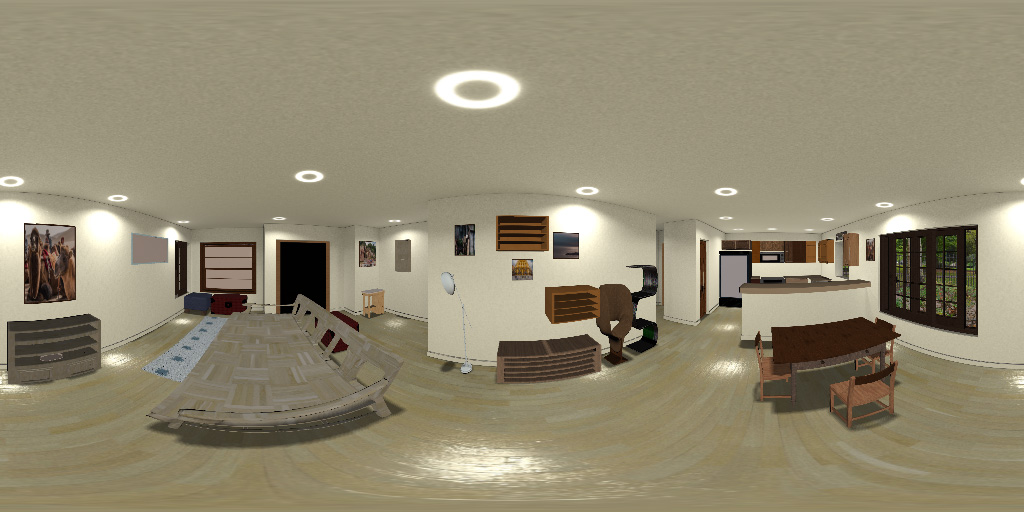}
    	    & \rgb{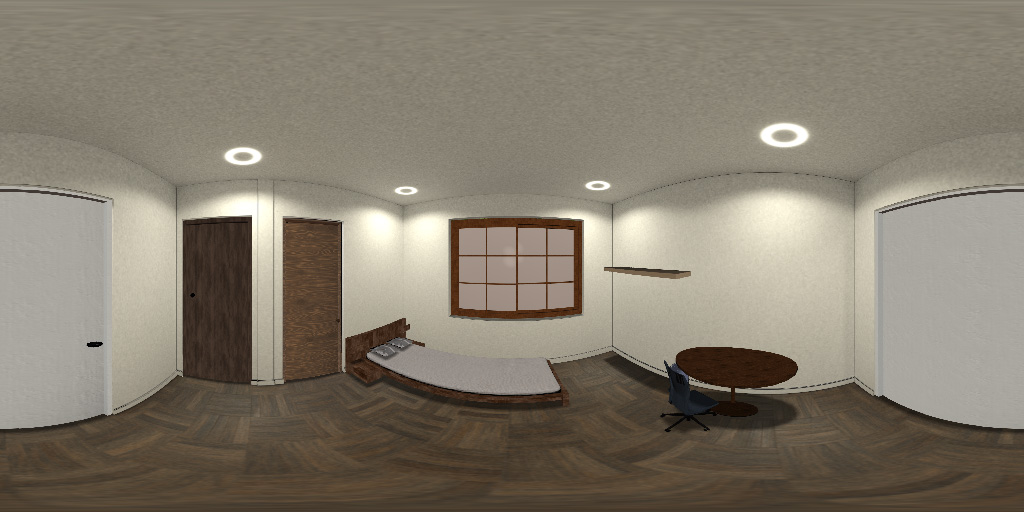}
    	    & \rgb{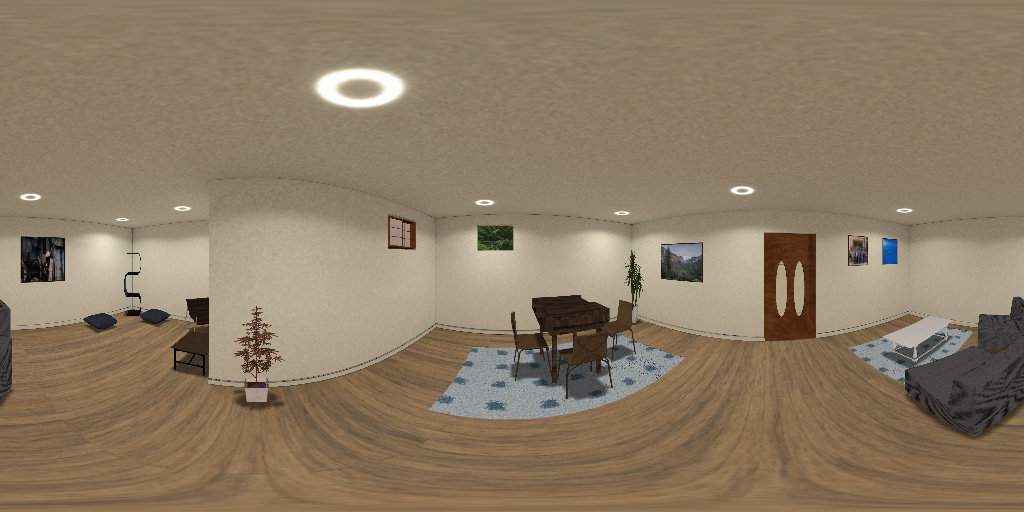}
    	    \\
    	    \rot{2D BBox}
            & \rgb{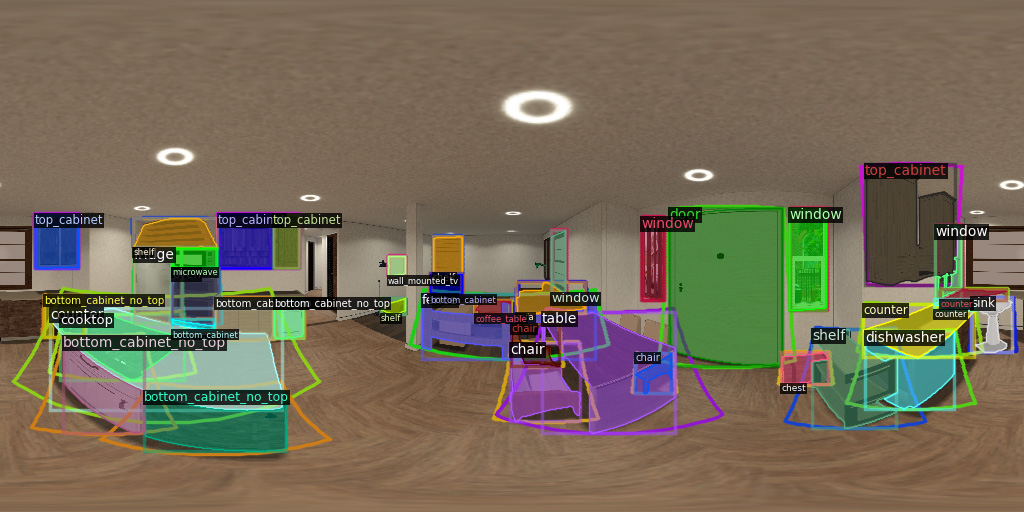}
    	    & \rgb{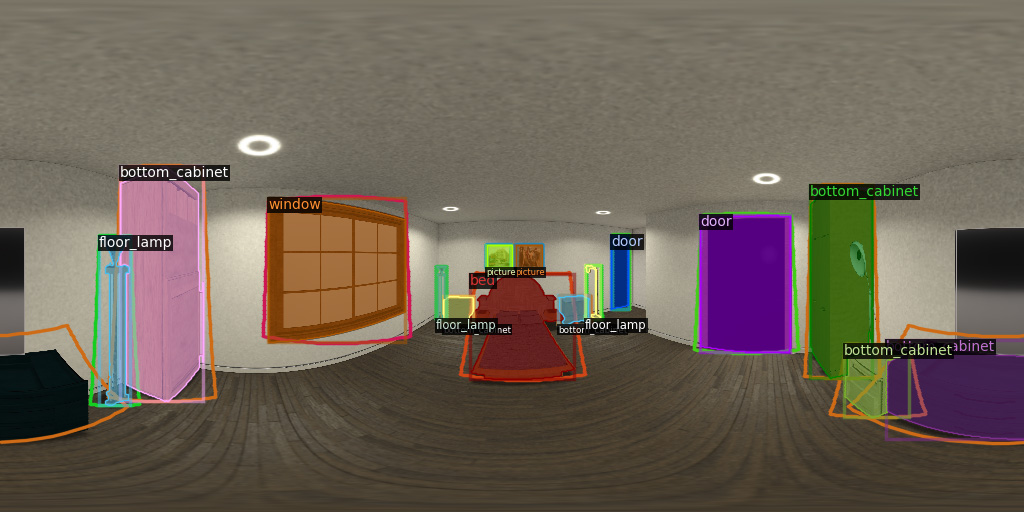}
    	    & \rgb{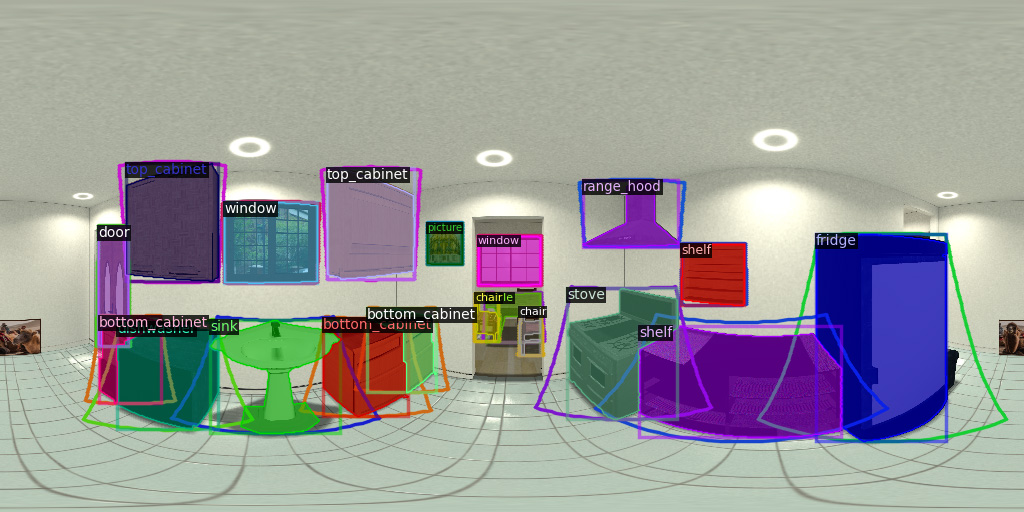}
    	    & \rgb{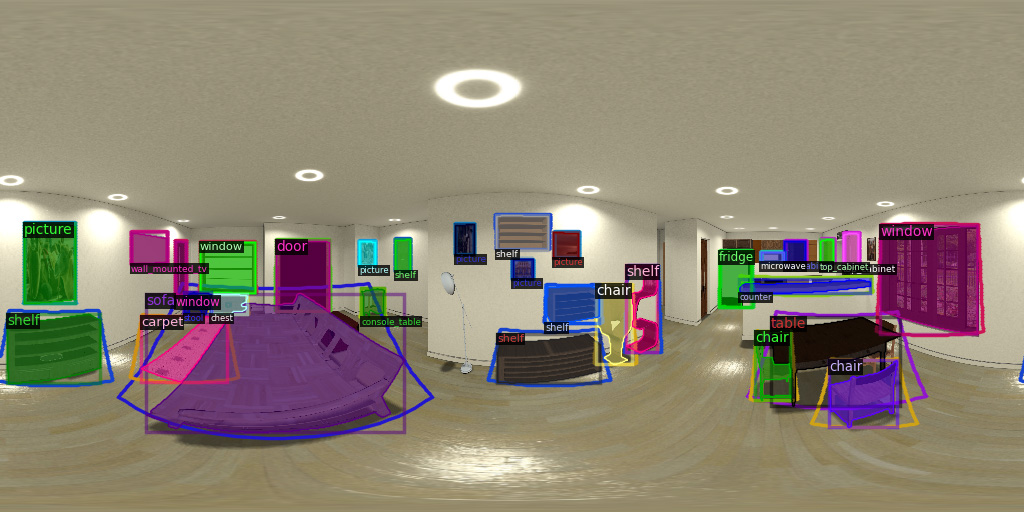}
    	    & \rgb{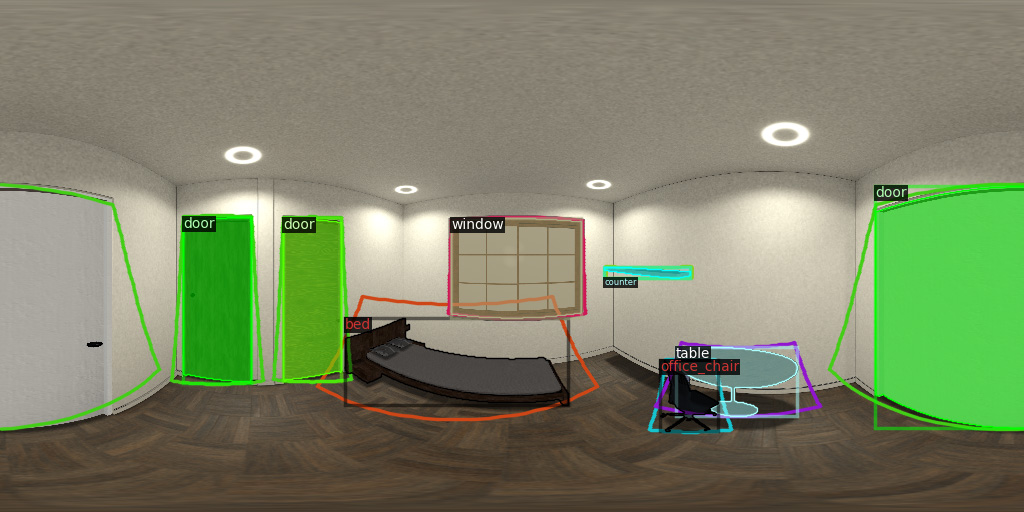}
    	    & \rgb{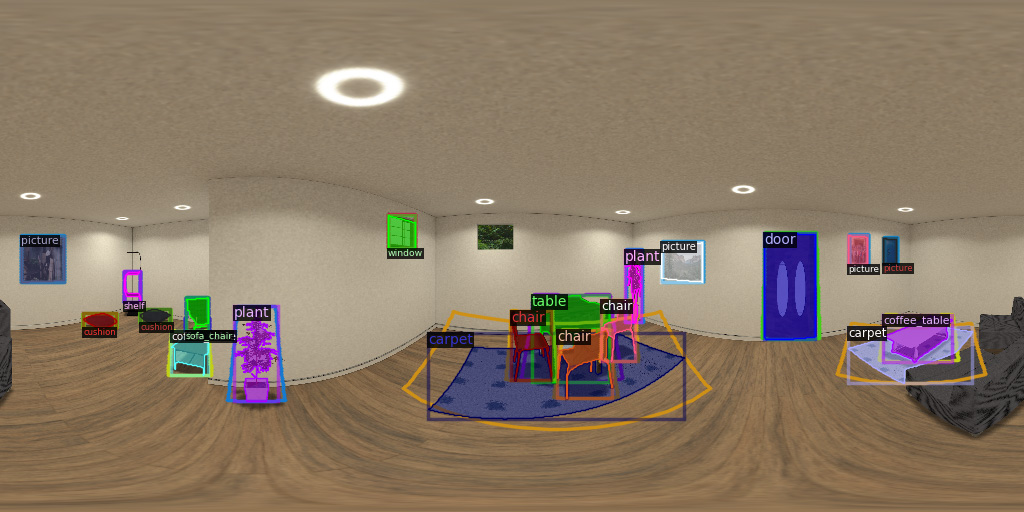}
    	    \\
    	    \rot{3D BBox}
            & \rgb{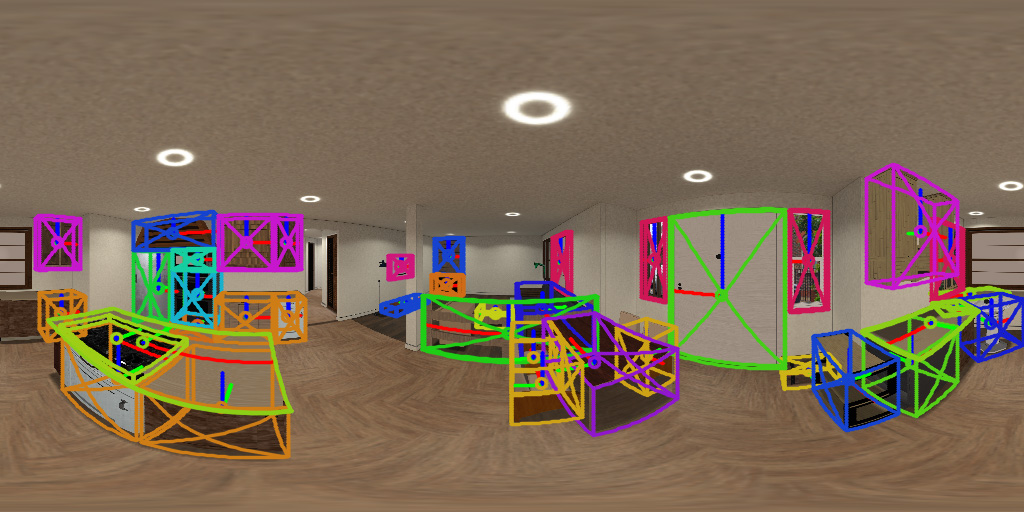}
    	    & \rgb{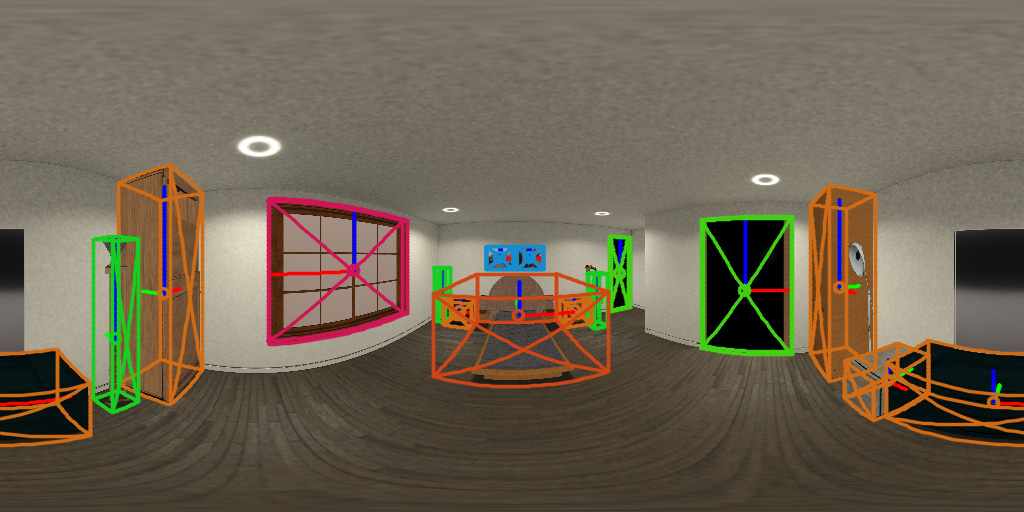}
    	    & \rgb{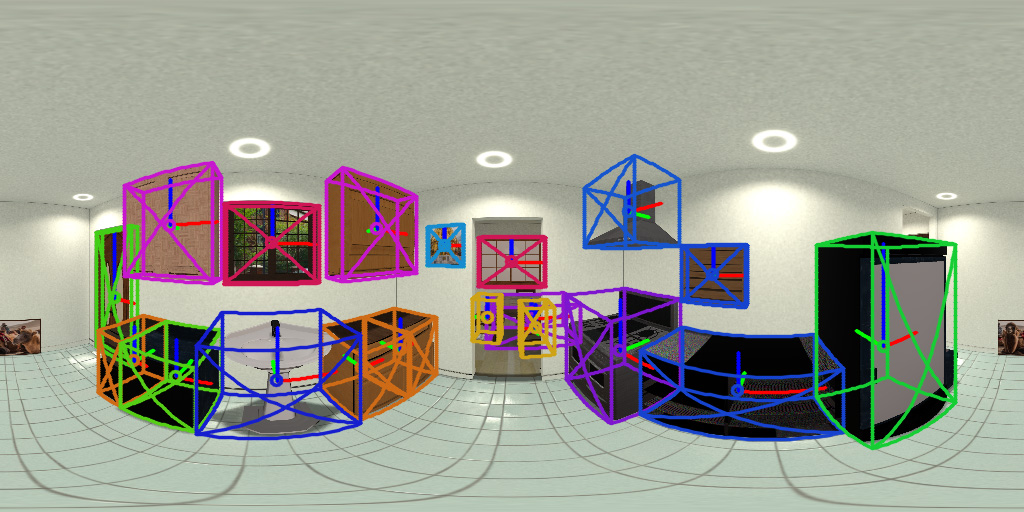}
    	    & \rgb{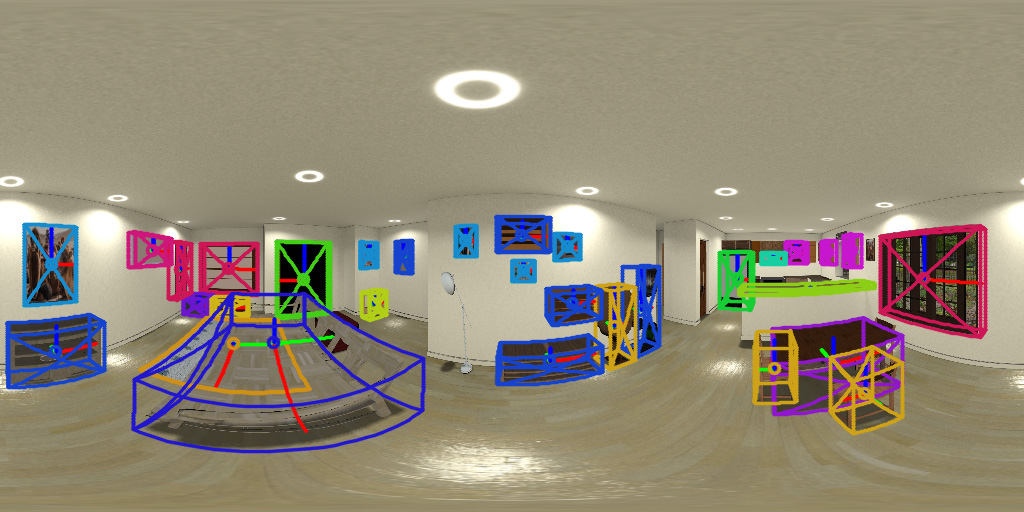}
    	    & \rgb{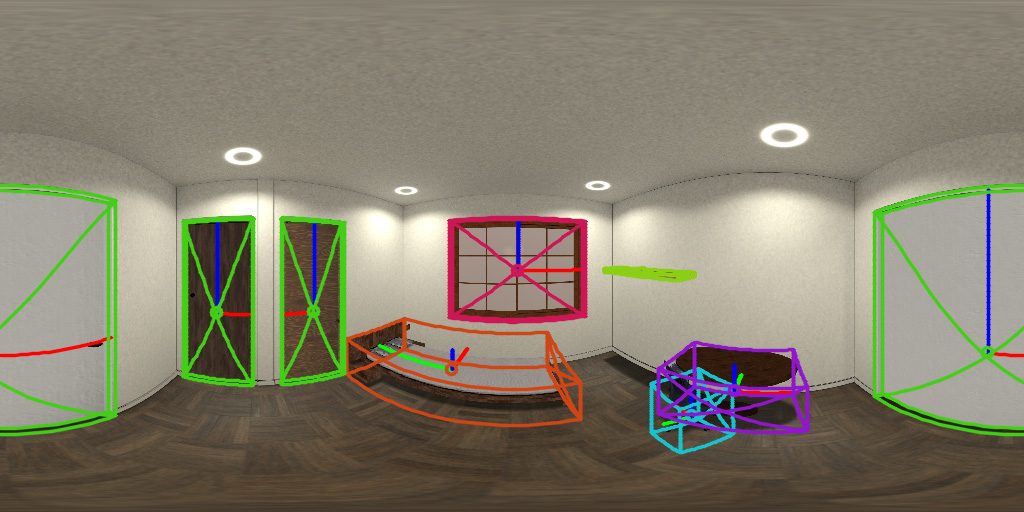}
    	    & \rgb{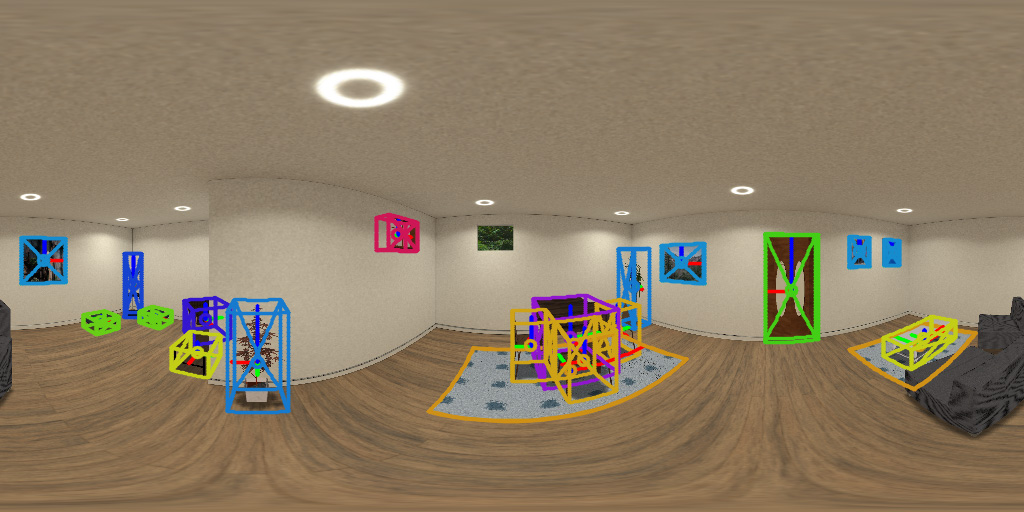}
    	    \\
    	    \rot{Mesh+Lo}
            & \rgb{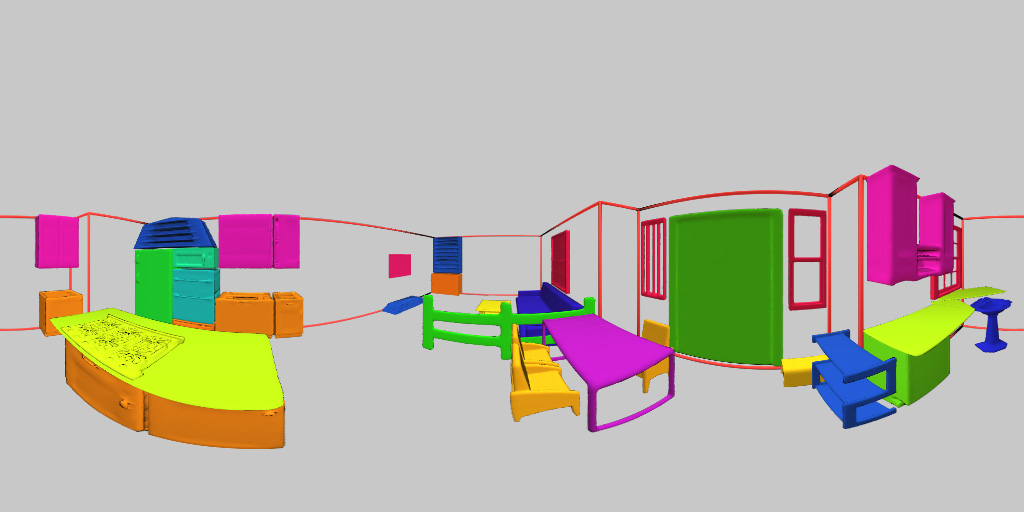}
    	    & \rgb{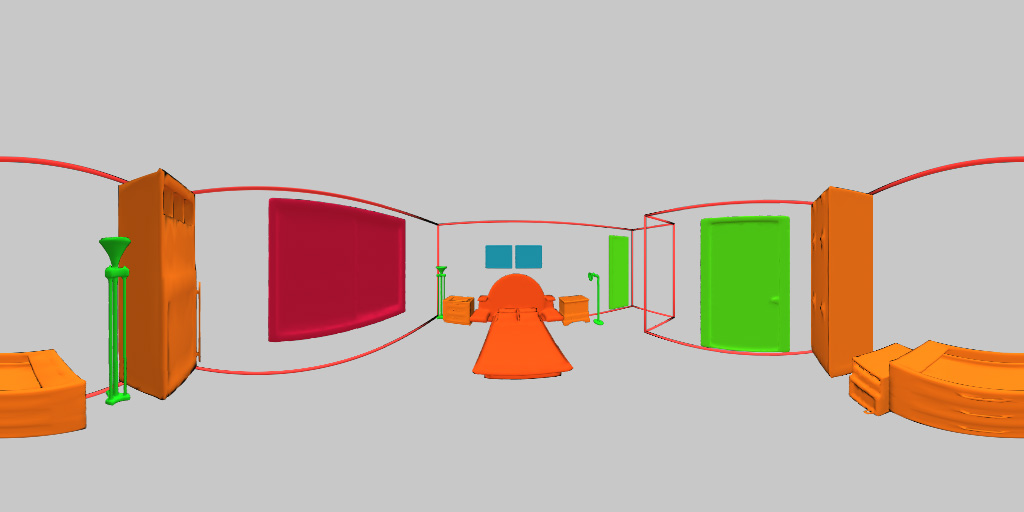}
    	    & \rgb{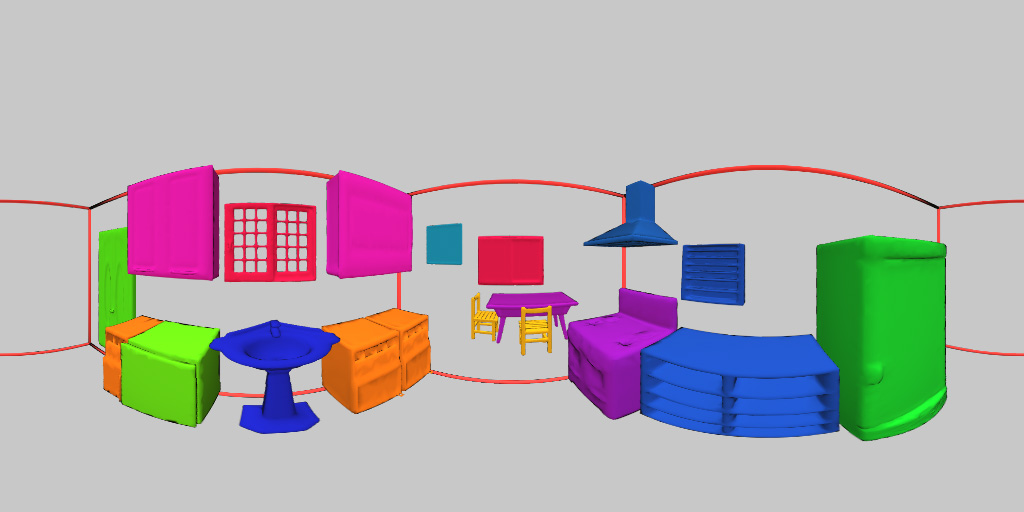}
    	    & \rgb{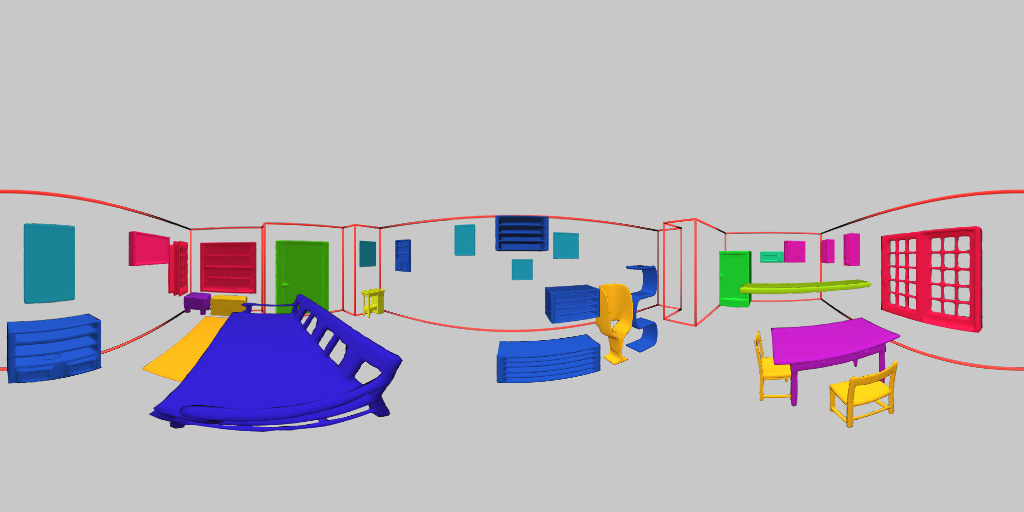}
    	    & \rgb{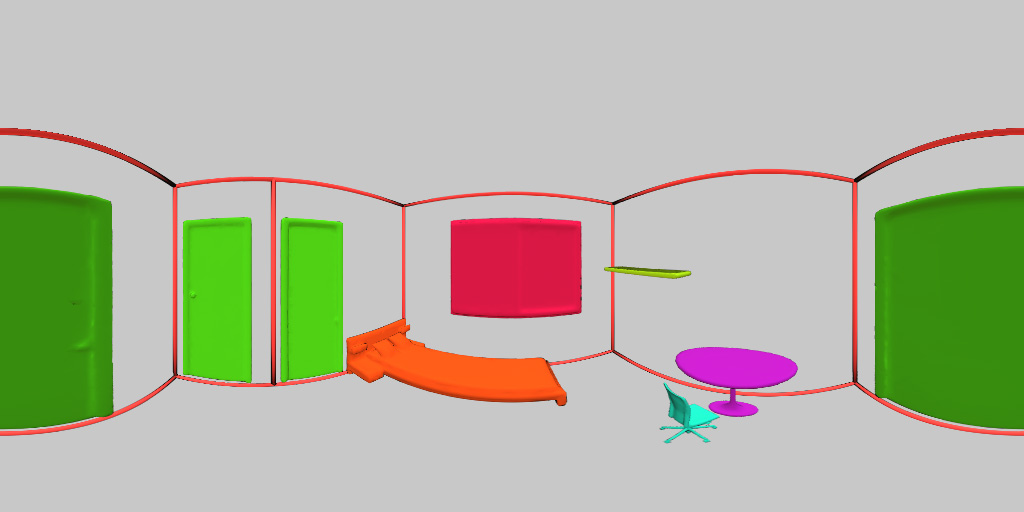}
    	    & \rgb{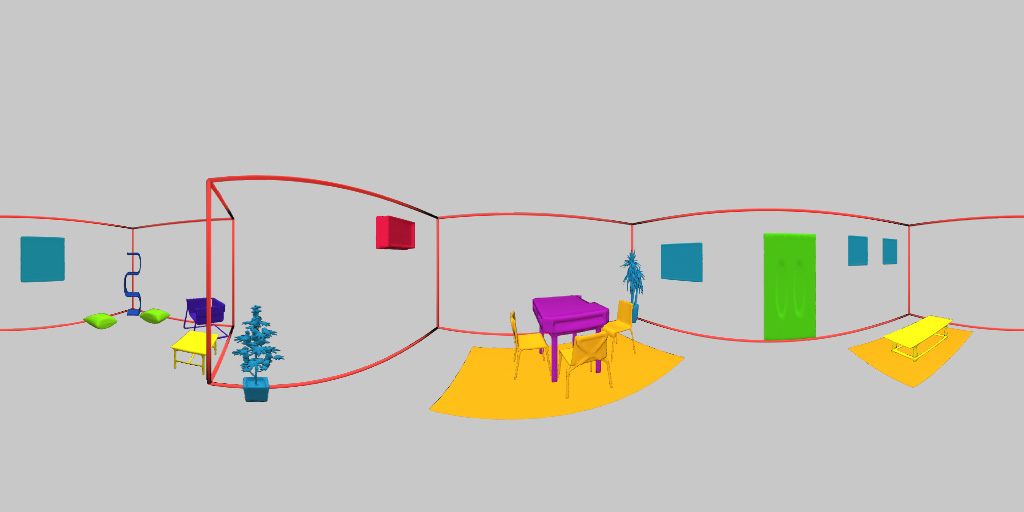}
    	    \\
        	\rot{Depth}
            & \rgb{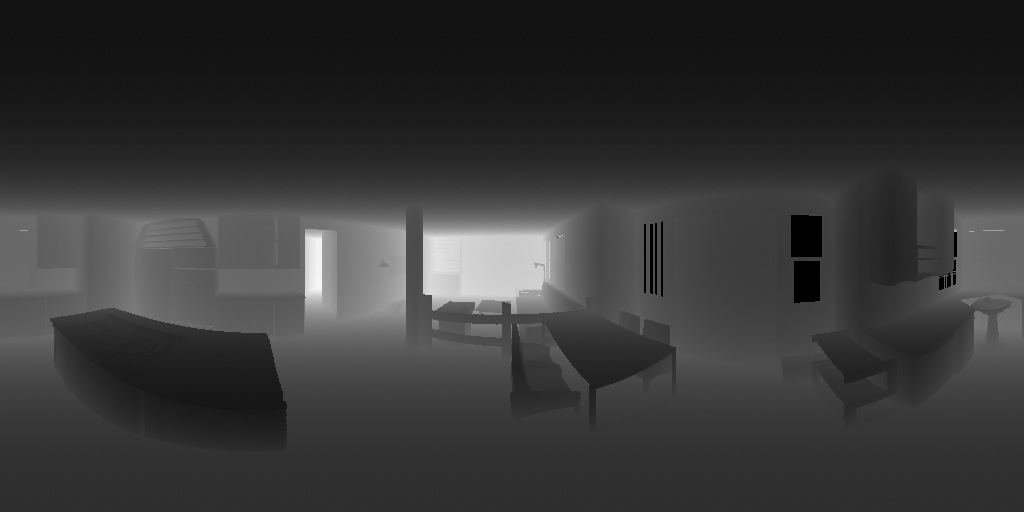}
    	    & \rgb{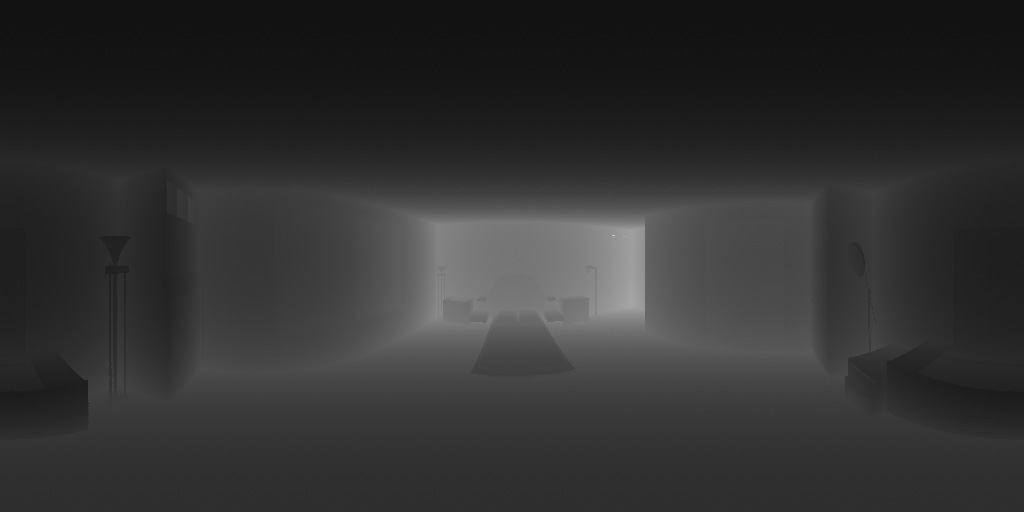}
    	    & \rgb{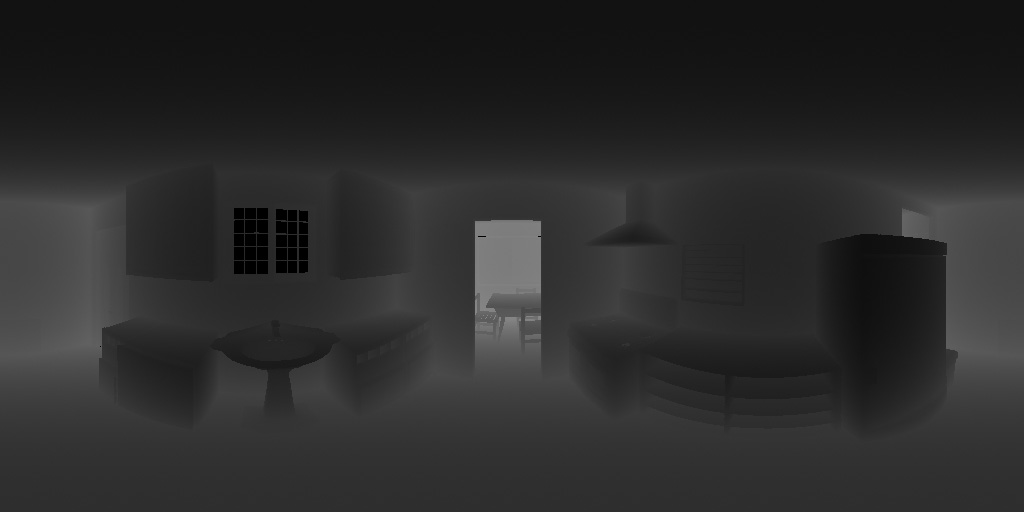}
    	    & \rgb{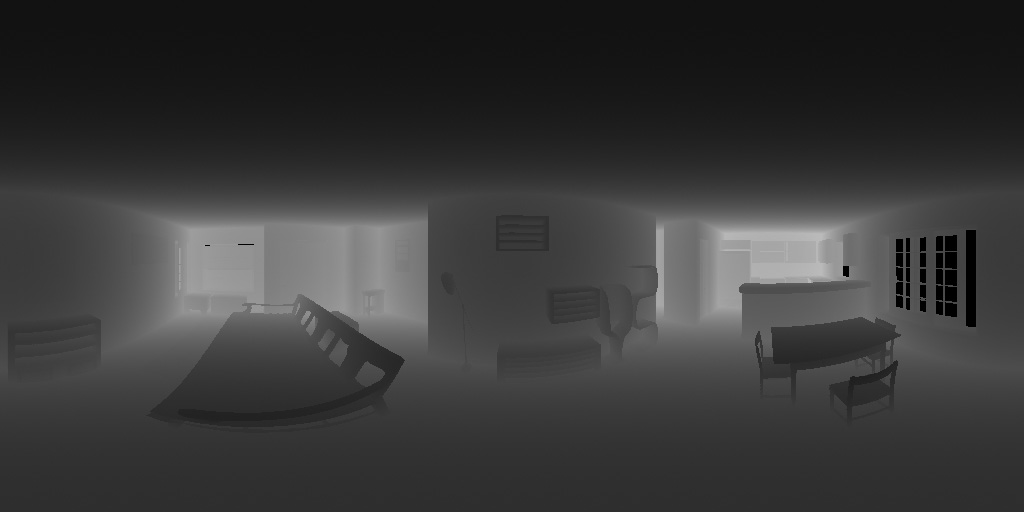}
    	    & \rgb{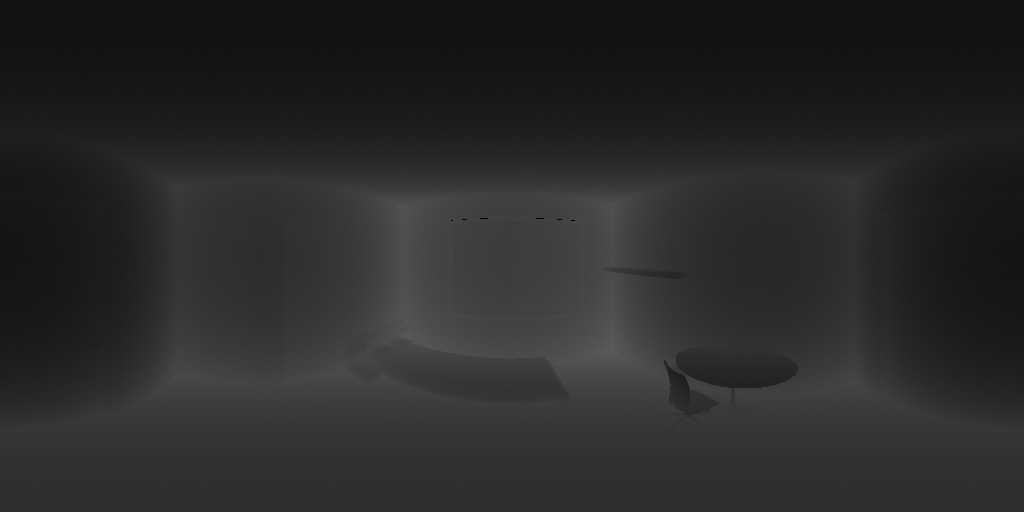}
    	    & \rgb{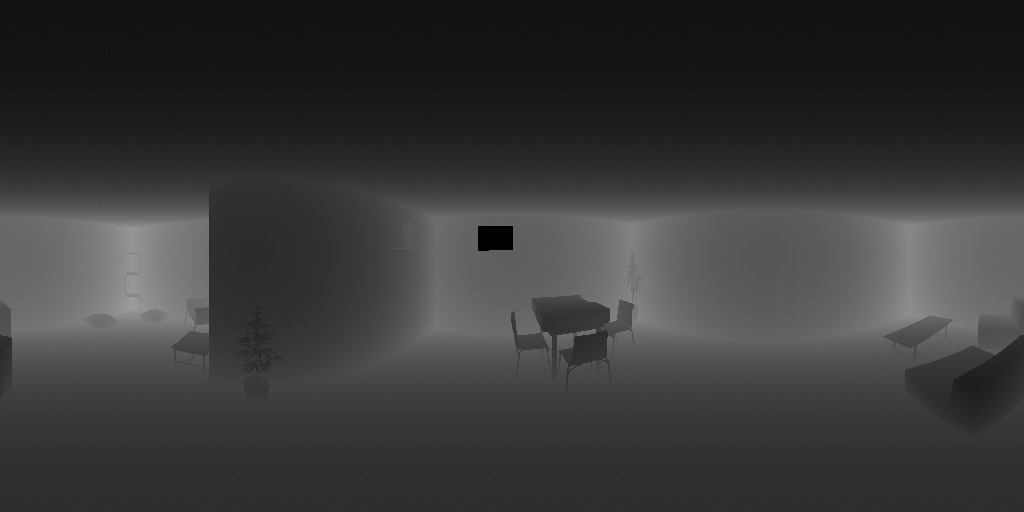}
    	    \\
        	\rot{Ins Seg}
            & \rgb{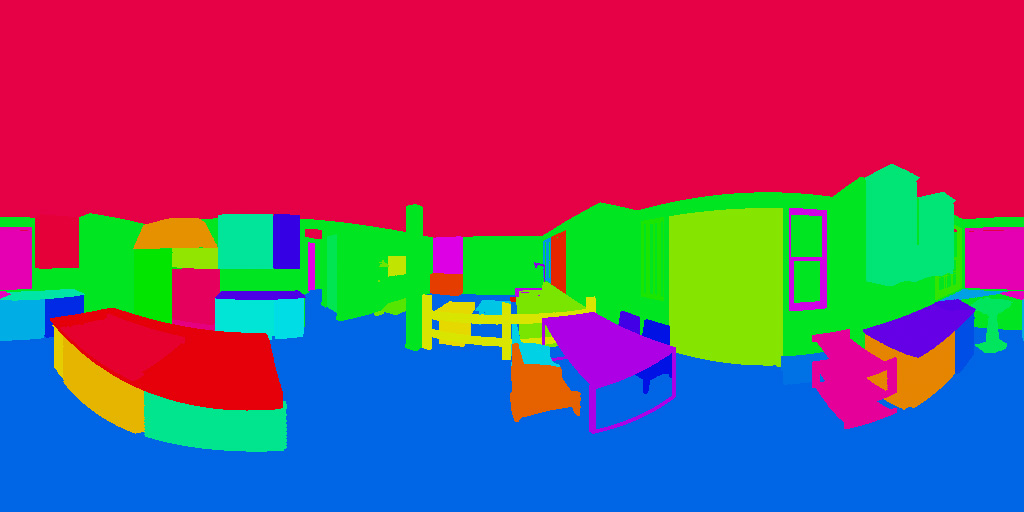}
    	    & \rgb{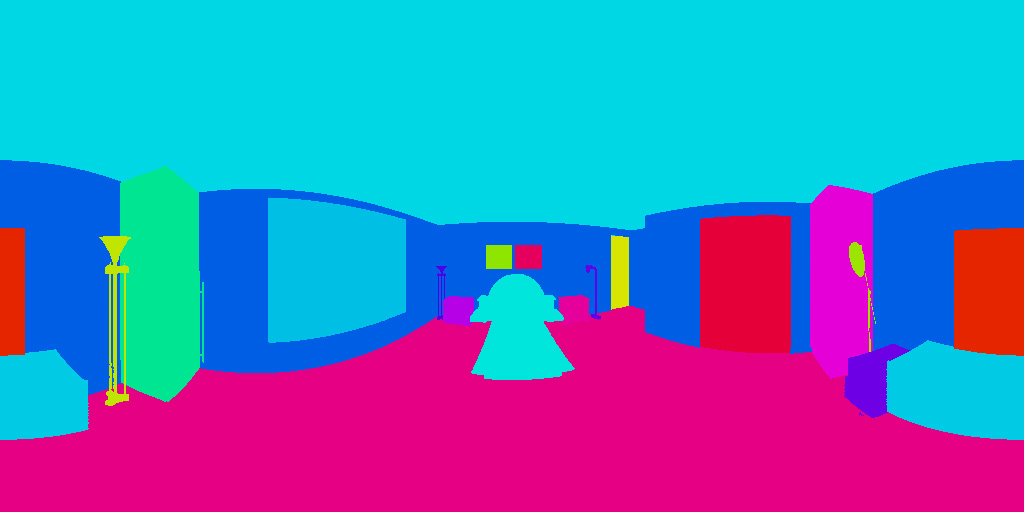}
    	    & \rgb{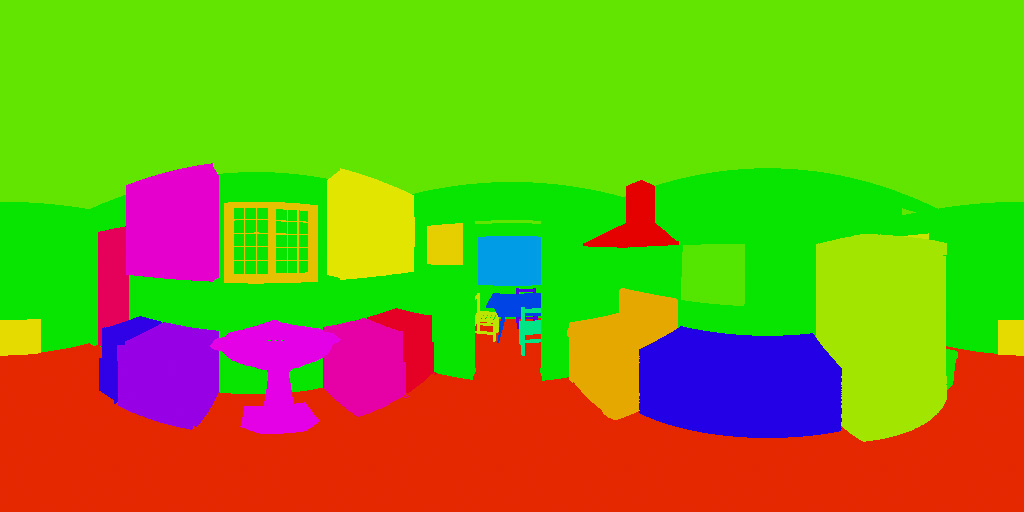}
    	    & \rgb{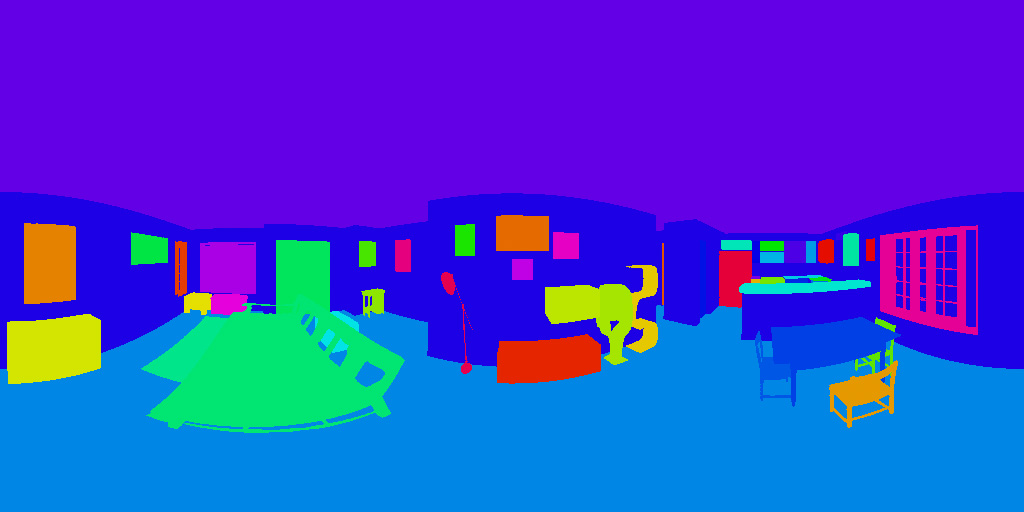}
    	    & \rgb{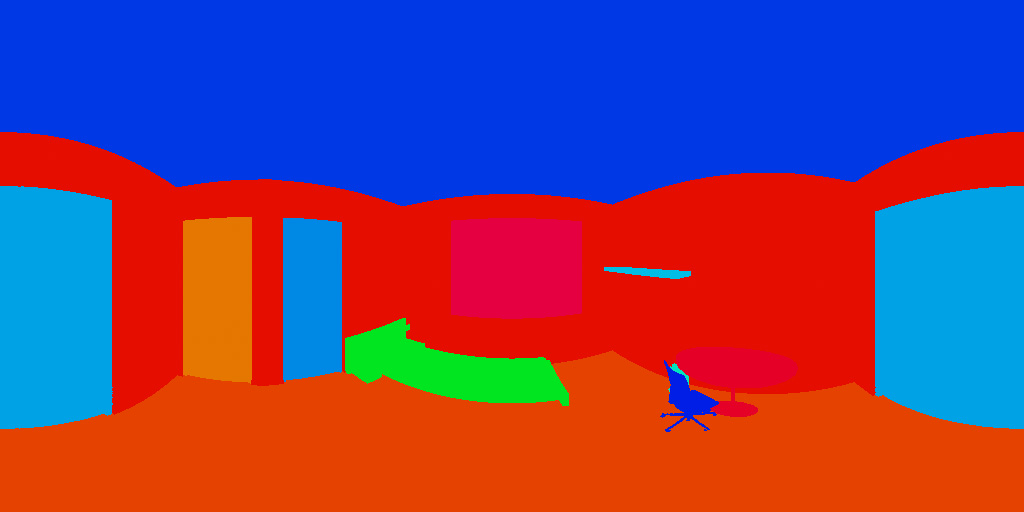}
    	    & \rgb{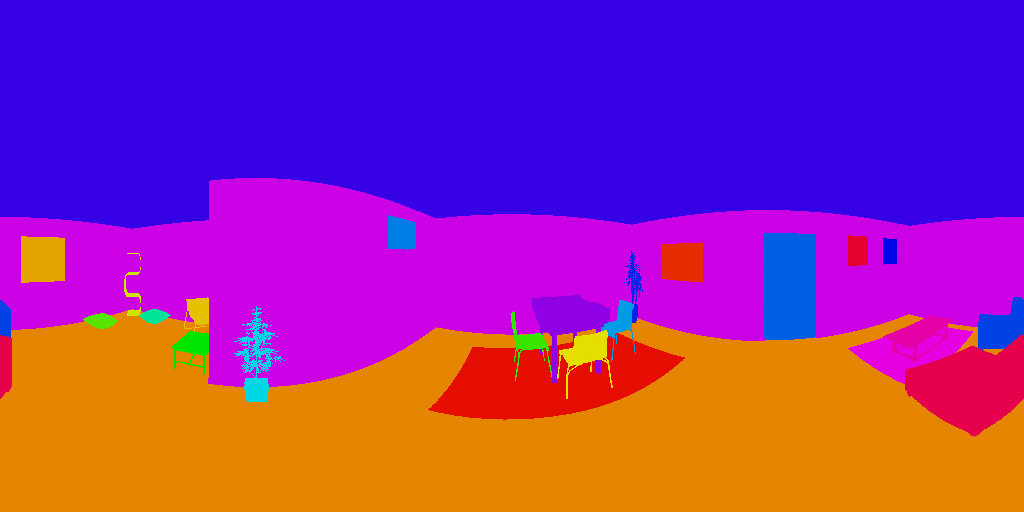}
    	    \\
    	    \rot{Sem Seg}
            & \rgb{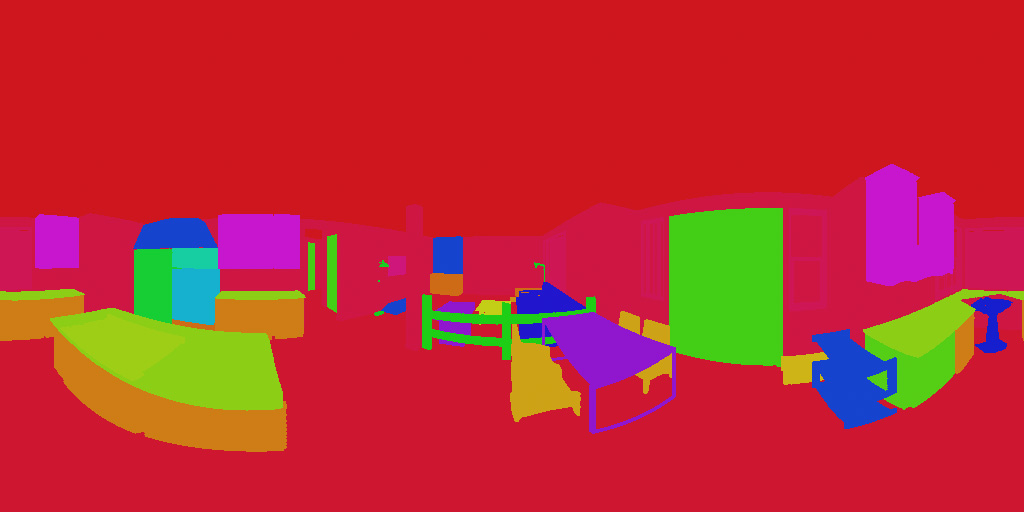}
    	    & \rgb{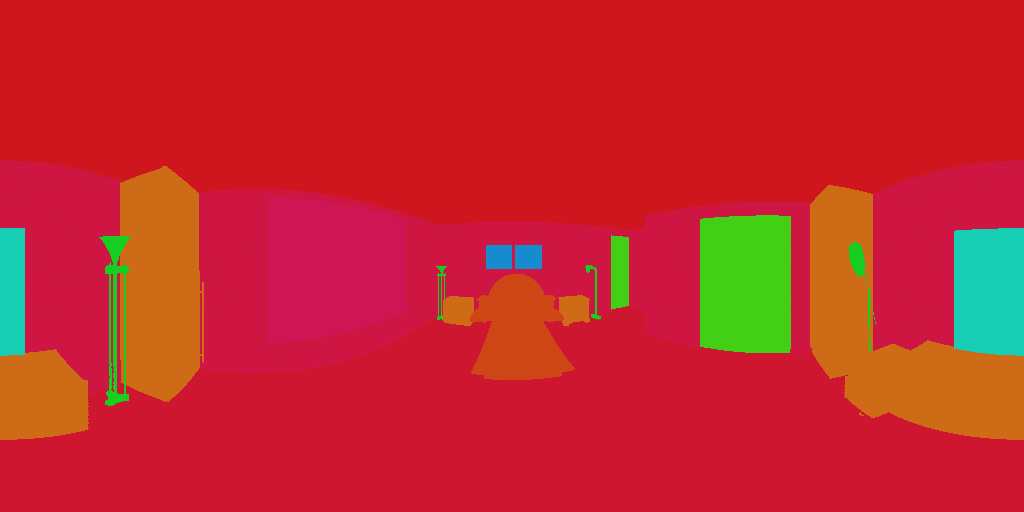}
    	    & \rgb{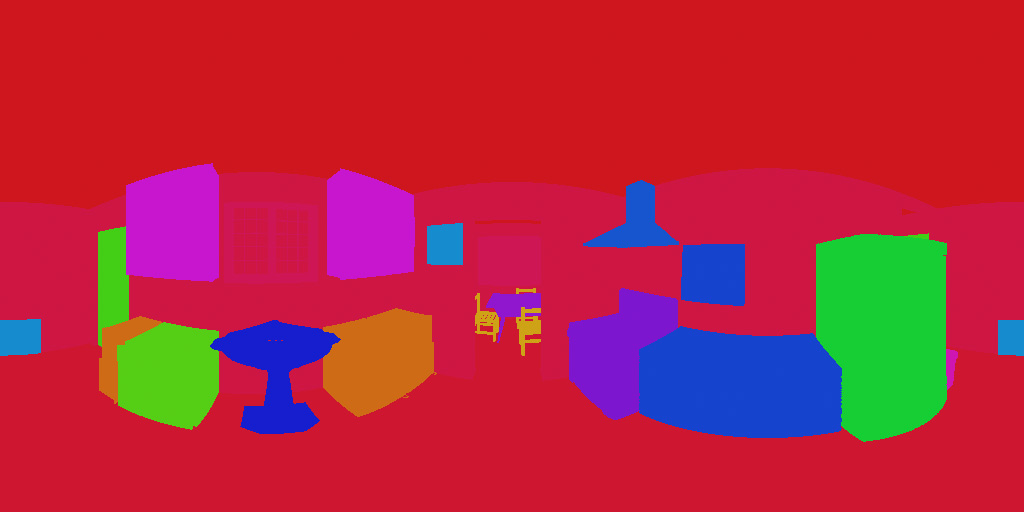}
    	    & \rgb{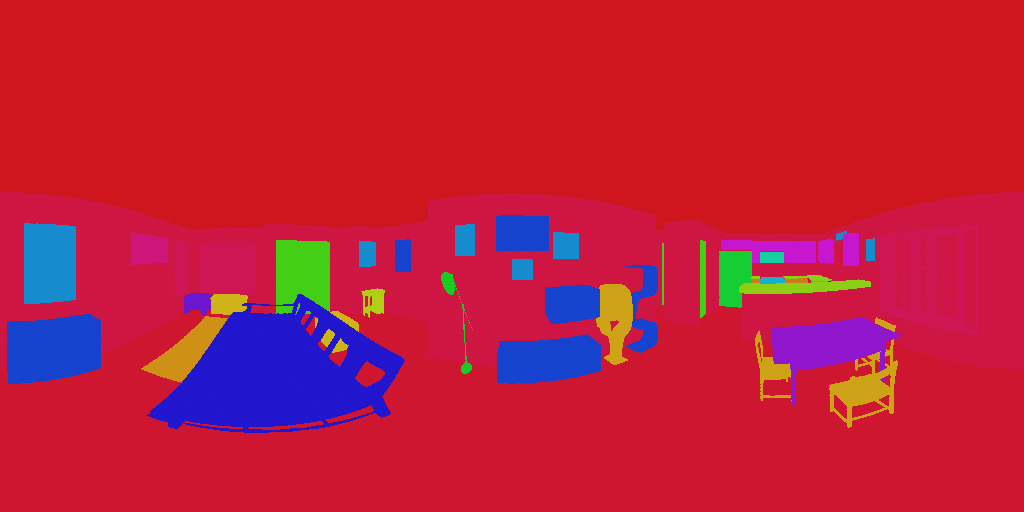}
    	    & \rgb{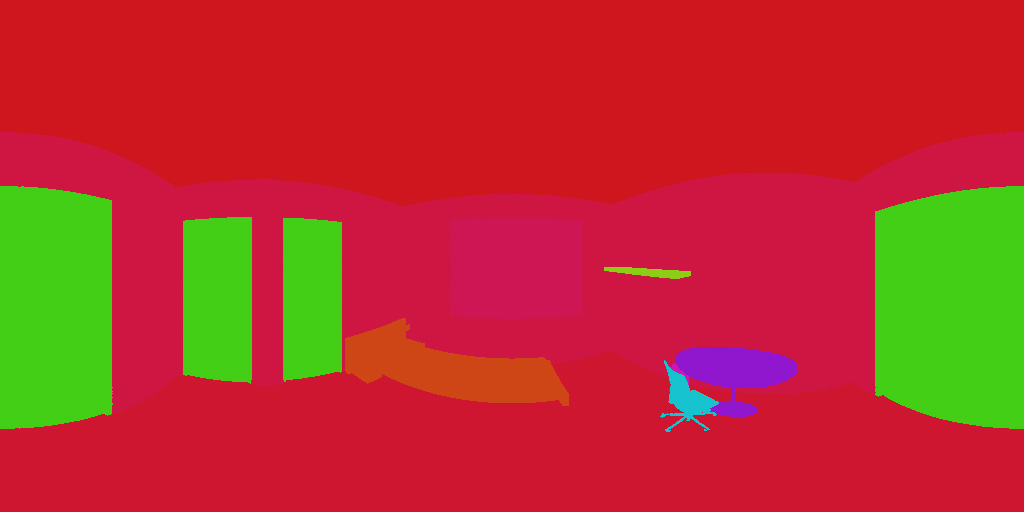}
    	    & \rgb{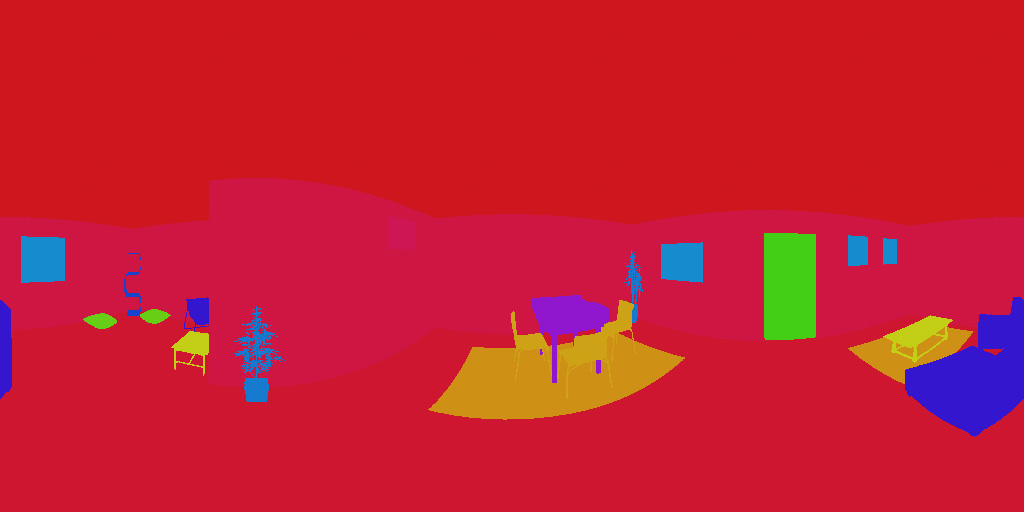}
    	    \\
    \end{tabular}
    	\caption{\zc{Samples of our proposed panorama 3D scene understanding dataset.}}
	\label{fig:dataset_scenes}
\end{figure*}

\begin{figure*}[!ht]
	\centering
	\scriptsize
	\newcommand{\rgb}[1]{\raisebox{-0.5\height}{\includegraphics[height=.08\linewidth]{#1}}}
	\newcommand{\rot}[1]{\rotatebox[origin=c]{90}{#1}}
	\def\arraystretch{0.5}	\begin{tabular}{c*{6}{c@{\hspace{1px}}}}
        \rule{0pt}{5px}
            \rot{Crop1}
            & \rgb{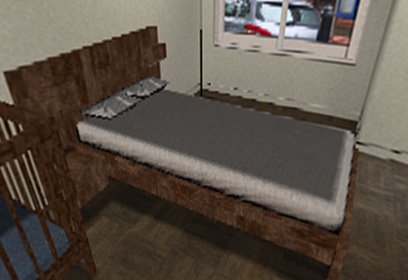}
    	    & \rgb{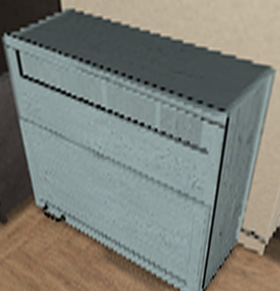}
    	    & \rgb{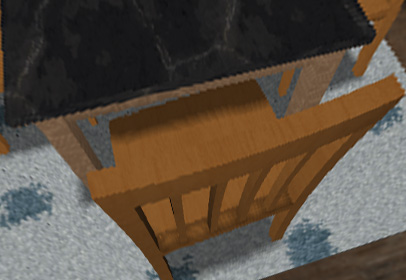}
    	    & \rgb{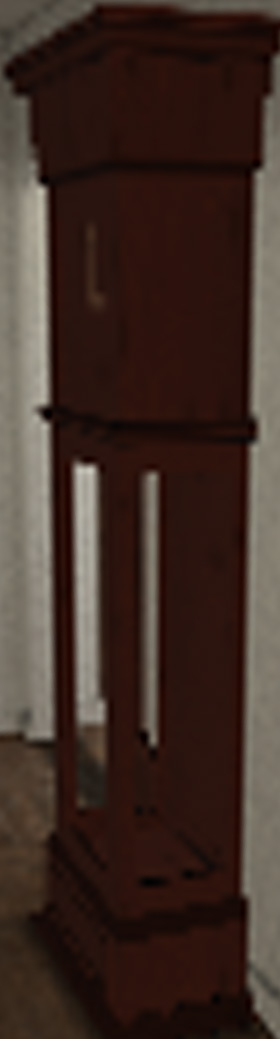}
    	    & \rgb{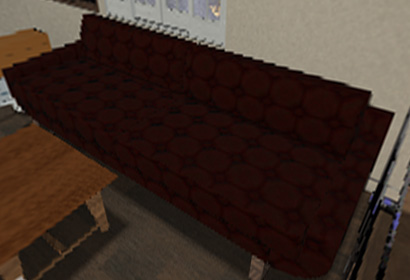}
    	    & \rgb{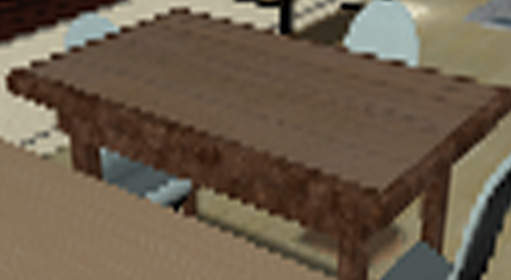}
    	    \vspace{2px}\\
    	    \rot{Crop2}
            & \rgb{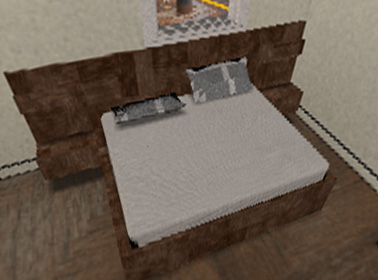}
    	    & \rgb{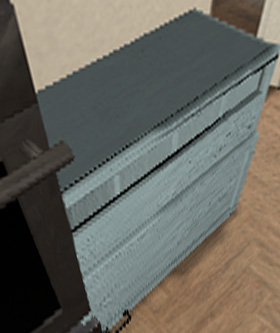}
    	    & \rgb{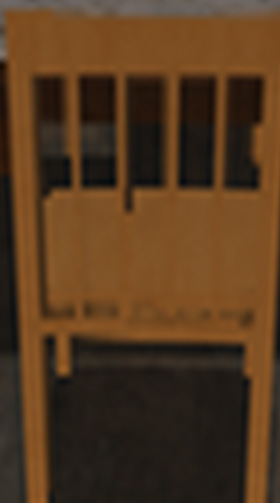}
    	    & \rgb{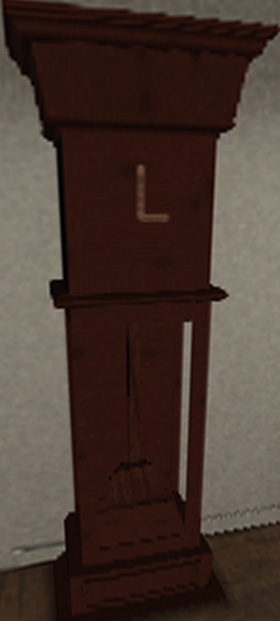}
    	    & \rgb{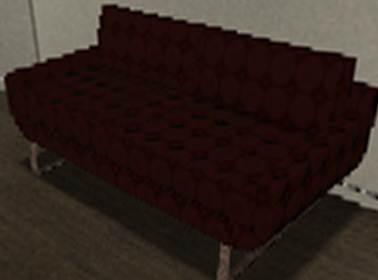}
    	    & \rgb{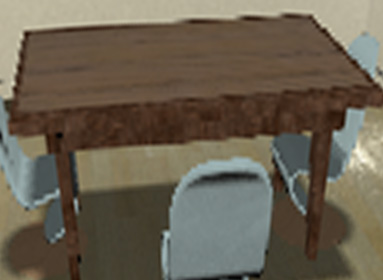}
    	    \vspace{2px}\\
    	    \rot{Crop3}
            & \rgb{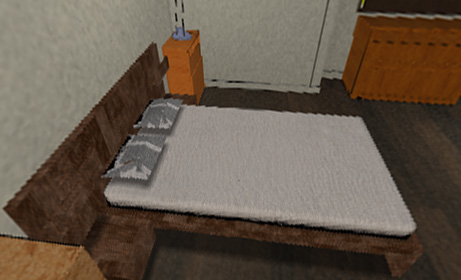}
    	    & \rgb{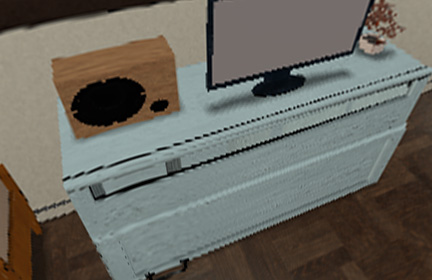}
    	    & \rgb{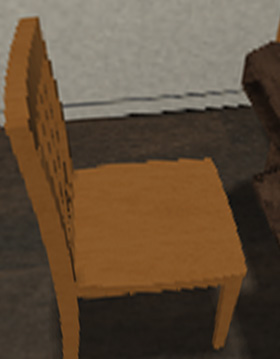}
    	    & \rgb{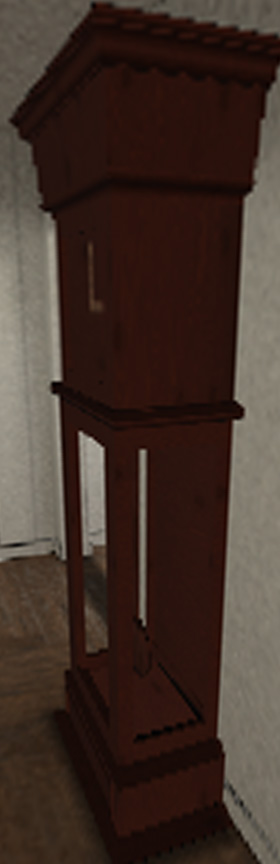}
    	    & \rgb{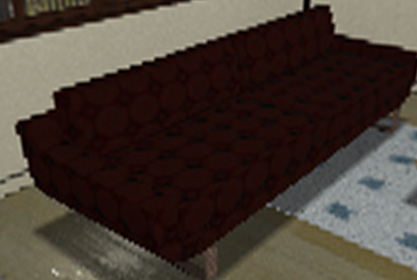}
    	    & \rgb{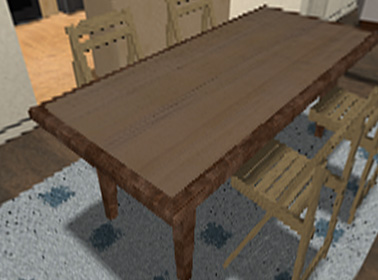}
    	    \vspace{2px}\\
    	    \rot{Render1}
            & \rgb{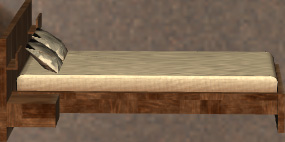}
    	    & \rgb{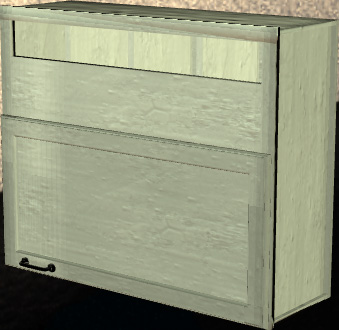}
    	    & \rgb{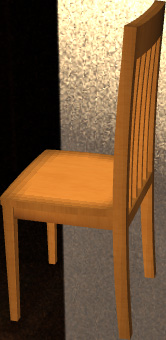}
    	    & \rgb{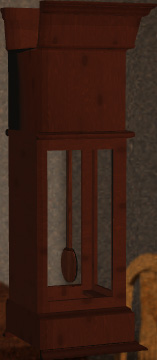}
    	    & \rgb{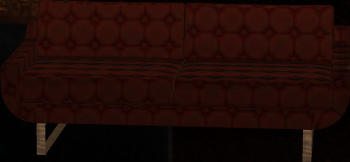}
    	    & \rgb{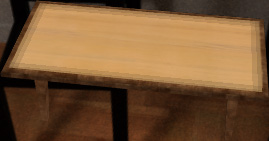}
    	    \vspace{2px}\\
        	\rot{Render2}
            & \rgb{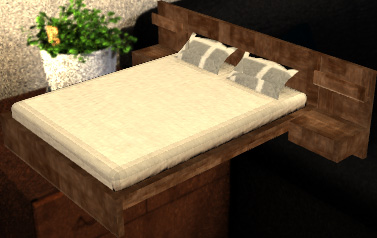}
    	    & \rgb{figure/dataset_objs/chair-28a0b2a5afc96922ba63bc389be1ed5a-render_00016}
    	    & \rgb{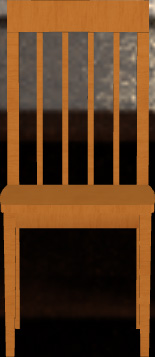}
    	    & \rgb{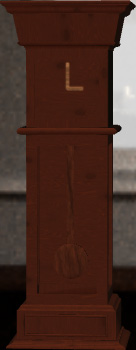}
    	    & \rgb{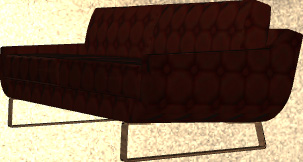}
    	    & \rgb{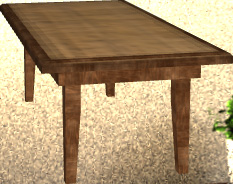}
    	    \vspace{2px}\\
        	\rot{Render3}
            & \rgb{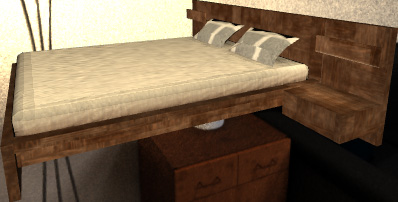}
    	    & \rgb{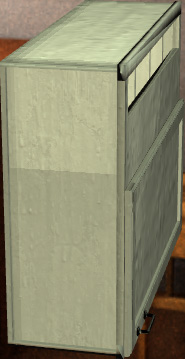}
    	    & \rgb{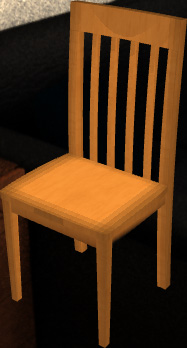}
    	    & \rgb{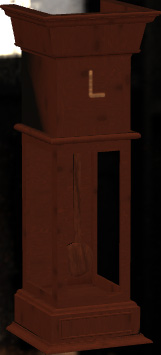}
    	    & \rgb{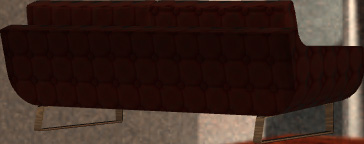}
    	    & \rgb{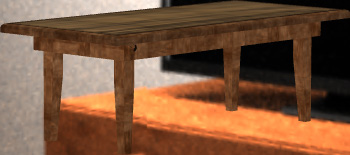}
    	    \vspace{2px}\\
    	    \rot{Mesh}
            & \rgb{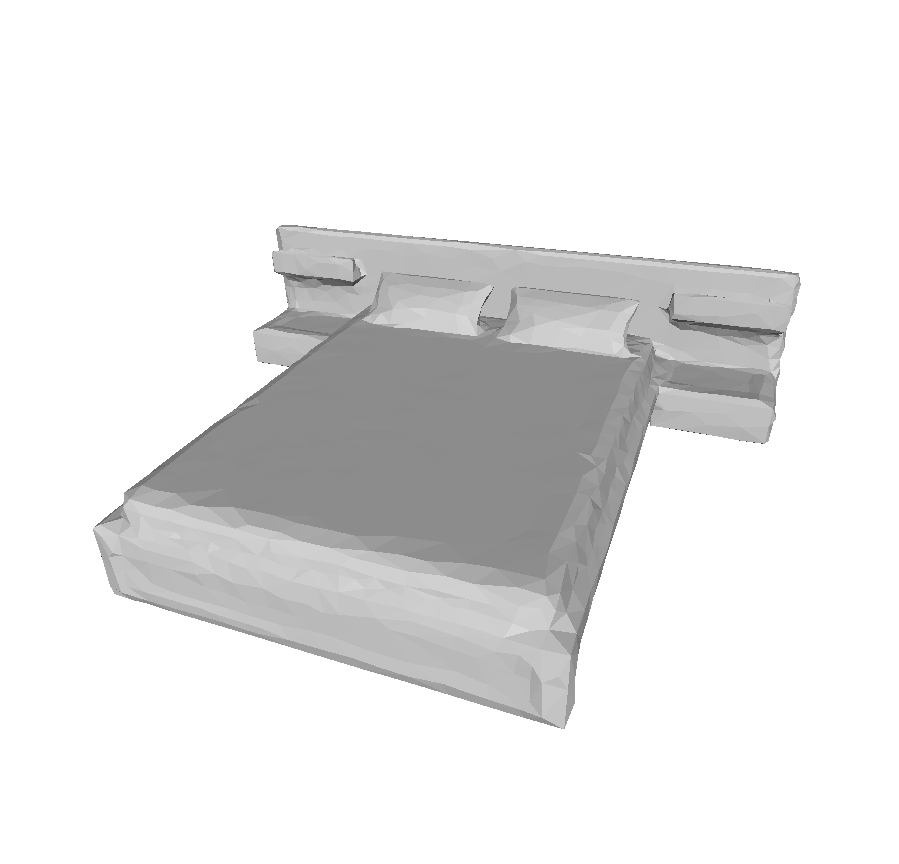}
    	    & \rgb{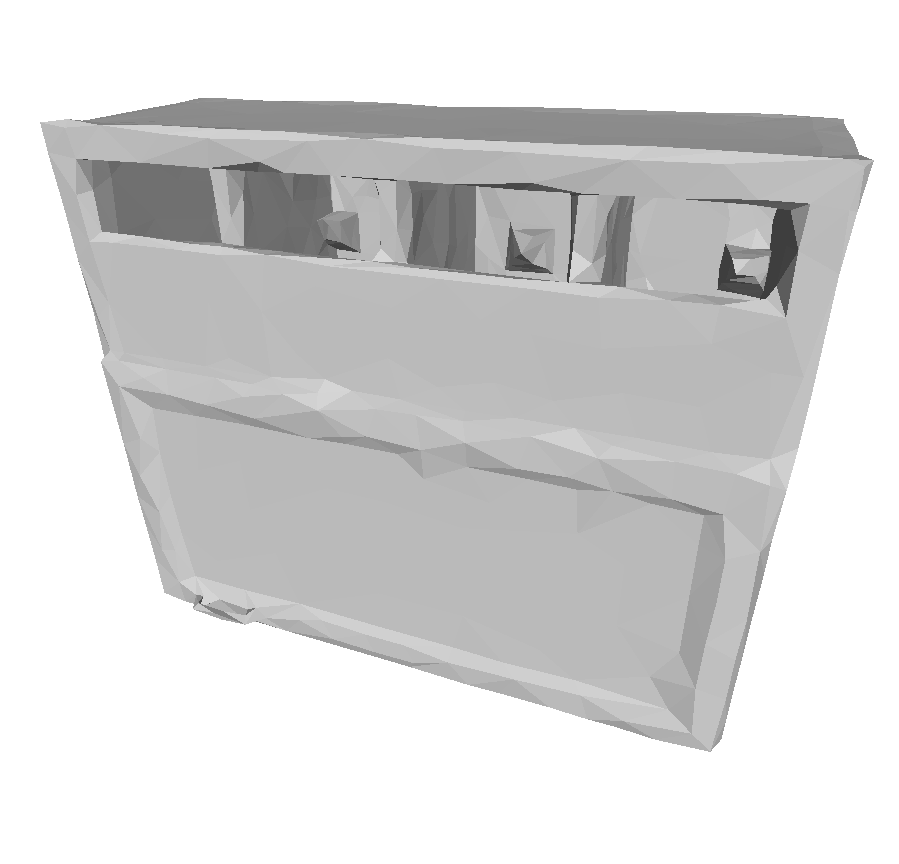}
    	    & \rgb{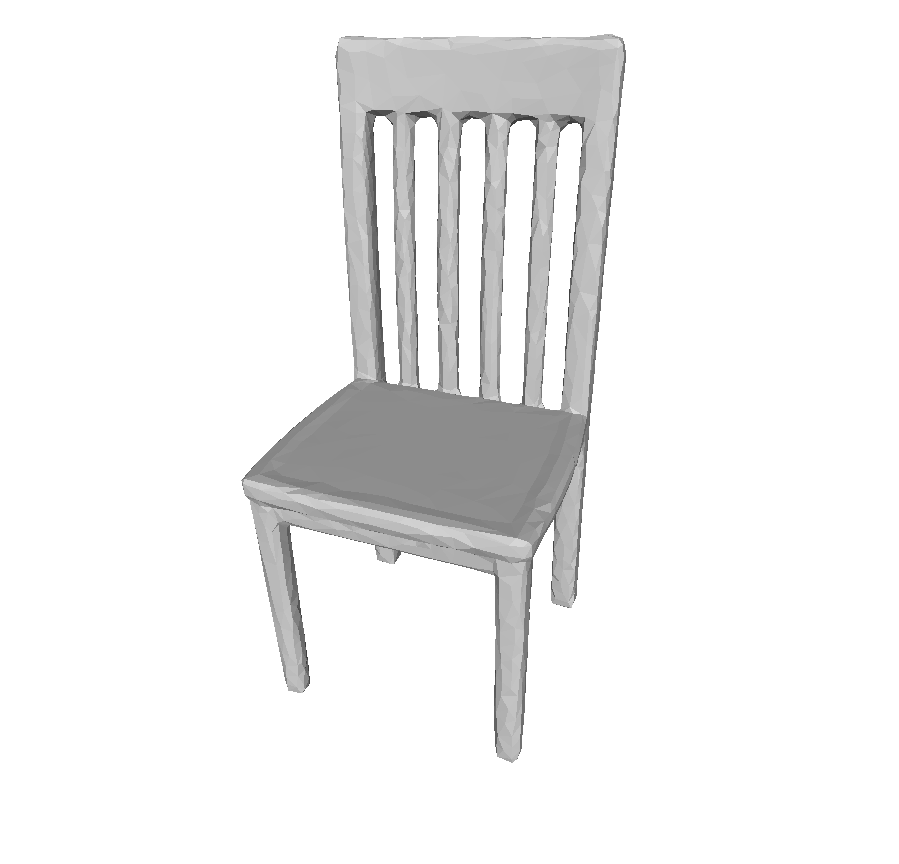}
    	    & \rgb{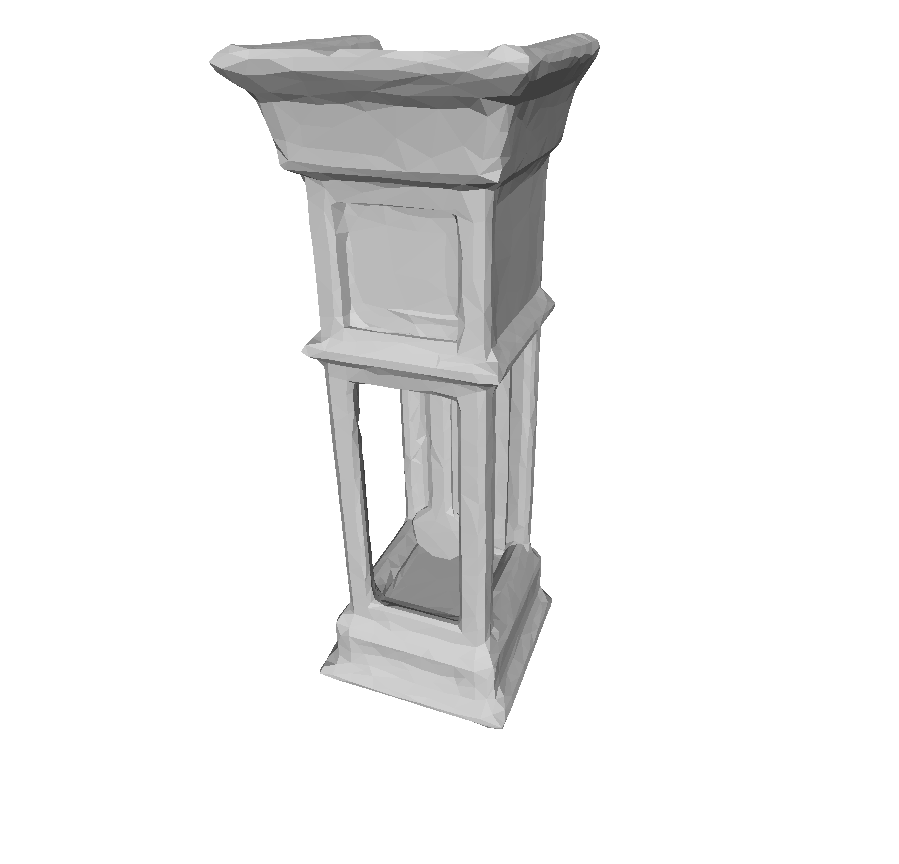}
    	    & \rgb{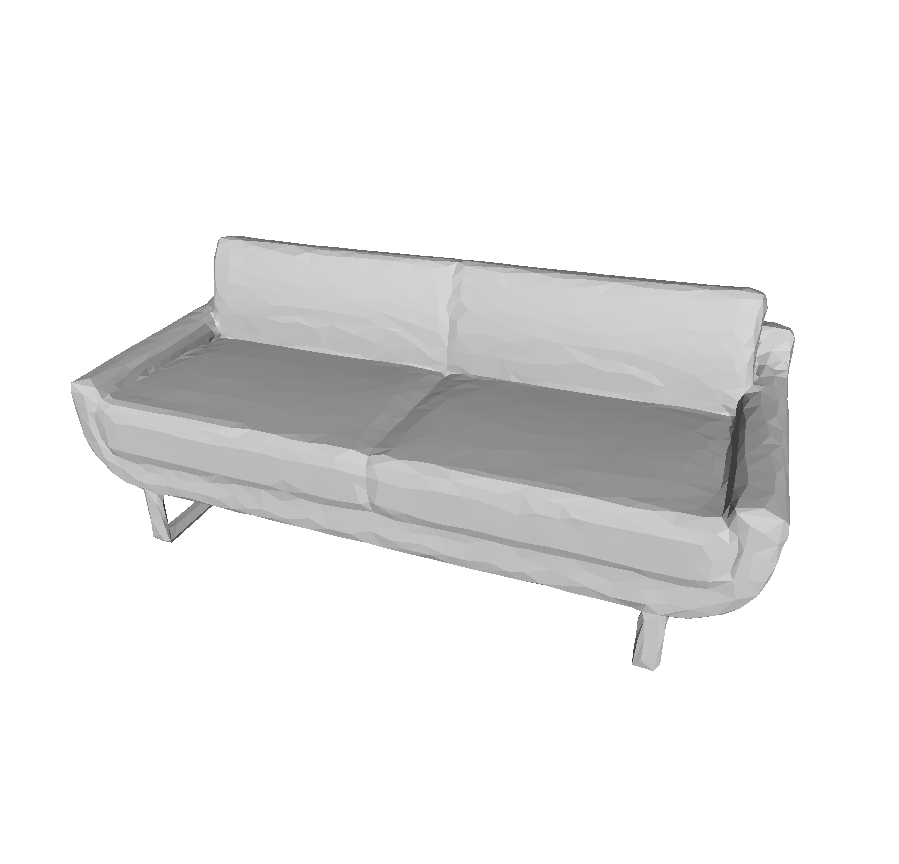}
    	    & \rgb{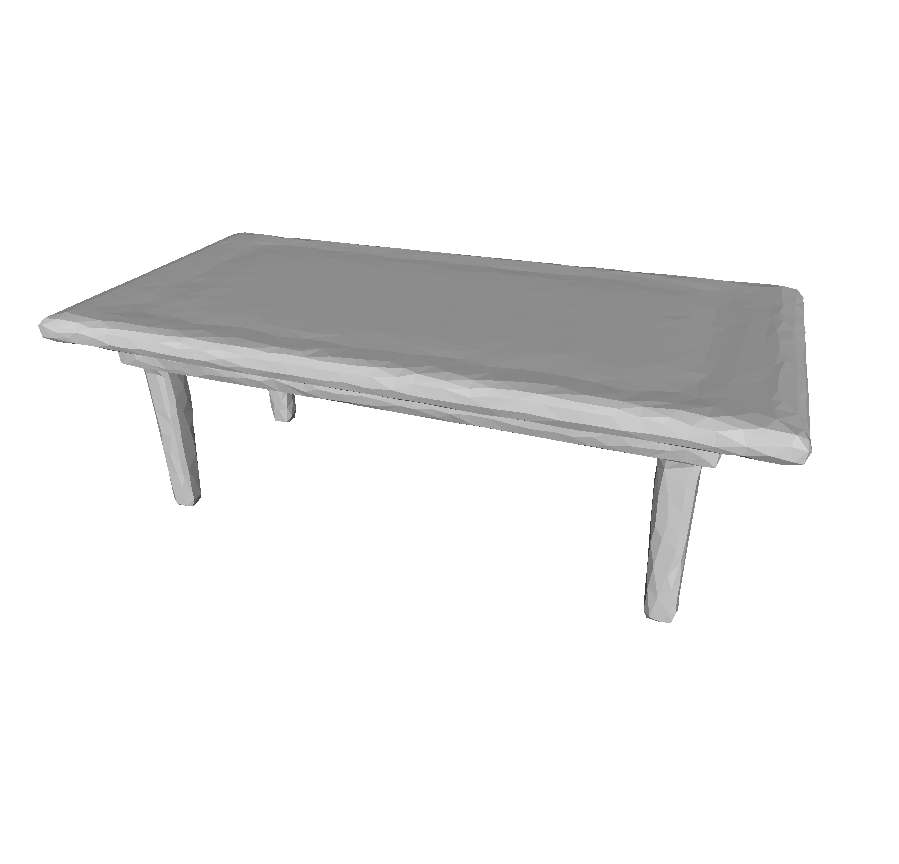}
    	    \vspace{2px}\\
    	    & bed & bottom cabinet & chair & chair & sofa & table
    	    \\
    \end{tabular}
        	\caption{\zc{Samples of dataset used for single image object reconstruction.}}
	\label{fig:dataset_objs}
\end{figure*}

Our synthetic dataset provides various ground truth along with the RGB panorama images, including 2D object bounding boxes/BFoVs, watertight scene/object meshes, oriented 3D object bounding boxes\yz{, and 3D room layout}. 
Our synthetic panorama scene understanding dataset also provides depth maps and semantic/instance segmentation images, which can be used by others.
Some examples of our panorama dataset are shown in \autoref{fig:dataset_scenes}.
\zc{We also show object crops collected from the panorama images and extra object-centric images rendered from object models used for single image object reconstruction in \autoref{fig:dataset_objs}.}
The data generation code is built upon iGibson \cite{shen2020igibson} and fully customized for panorama images. 

\section{Implementation Details}

\noindent \textbf{Dealing with Panorama Image}
As mentioned in the main paper, to deal with the continuity of panorama images, we parameterize the 2D bounding box with Bounding FoV (BFoV) \cite{chou2020360, yang2018object}, and extend the panorama \yz{boundary} before \yz{running} 2D detector.
Moreover, we change the object orientation ${\theta}$ to be the 
\zhpcui{yaw angle of the object in the cropped perspective image coordinate. Compared to directly}
estimating the orientation in the world frame as in Im3D \cite{zhang2021holistic} and Total3D \cite{nie2020total3dunderstanding}, 
\yz{our representation} is more intuitive because it 
\yz{explicitly} codes the transformation from the camera \yz{coordinates} to the world \yz{coordinates}.
When calculating bounding box projection \zc{term} ${e}^{bp}$ in relation optimization, we rotate the camera to each detected bounding box center then do the projection of 3D bounding boxes, which avoids cross-border situations.

\noindent \textbf{RGCN relation branch}
\zc{We design a relation branch for our RGCN} 
to facilitate the relation estimation from the 512-dim representation vectors of object/relation nodes.
We design the relation branch of RGCN as two-layer MLPs for each relation,
which consist of a 256-dim \yz{FC} layer, followed by a ReLU and Dropout layer with a drop factor of 0.5, and an output layer.
The output layer is 1-dim for binary relations (
\ie, object-object/wall/floor/ceiling contact, 
\yz{inside or outside room,}
\yz{closer and farther to camera center between a pair of objects}), and 8-dim for multi-class relations (\ie, the angular difference between object and object/wall).

\begin{table}[t]
    	\begin{center}
	    \resizebox{1.\columnwidth}{!}{
    		\begin{tabular}{|c|l|c|c|}
    			\hline
    			\multicolumn{2}{|c|}{RGCN Output} & \multicolumn{2}{c|}{Loss Weight} \\
    			\hline
    			Symbol & Description & Symbol & Value \\
    			\hline
    			${rr}$ & Object-object/wall relative rotation relation & ${\lambda}_{rr}$ & 10 \\
    			${oa}$ & Object-object/wall attachment relation & ${\lambda}_{oa}$ & 10 \\
    			${fa}$ & Object-floor attachment relation & ${\lambda}_{fa}$ & 10 \\
    			${ca}$ & Object-ceiling attachment relation & ${\lambda}_{ca}$ & 10 \\
    			${rd}$ & Object-object relative distance relation & ${\lambda}_{rd}$ & 10 \\
    			\hline
    			${\delta}$ & 3D bounding box center offset & ${\lambda}_{\delta}^{\prime}$ & 10 \\
    			${d}$ & 3D bounding box distance & ${\lambda}_{d}^{\prime}$ & 10 \\
    			${s}$ & 3D bounding box size & ${\lambda}_{s}^{\prime}$ & 10 \\
    			${\theta}$ & Object orientation & ${\lambda}_{\theta}^{\prime}$ & 10 \\
    			\hline
    		\end{tabular}
		}
	\end{center}
	\vspace{-1.0em}
	\caption{RGCN outputs and loss weights of $\mathcal{L}_{RGCN}$ and $\mathcal{L}$.}
	\label{tbl:rgcn_weights}
\end{table}

\begin{table}[t]
    	\begin{center}
	    \resizebox{1.\columnwidth}{!}{
    		\begin{tabular}{|c|l|c|c|}
    			\hline
        			\multicolumn{2}{|c|}{Term} & \multicolumn{2}{c|}{Weight} \\
    			\hline
    			Symbol & Description & Symbol & Value \\
    			\hline
    			${oc}$ & Object-object collision & ${\lambda}^{oc}$ & 1 \\
    			${wc}$ & Object-wall collision & ${\lambda}^{wc}$ & 1 \\
    			${fc}$ & Object-floor collision & ${\lambda}^{fc}$ & 1 \\
    			${cc}$ & Object-ceiling collision & ${\lambda}^{cc}$ & 1 \\
    			\hline
    			${rr}$ & Object-object/wall relative rotation relation & ${\lambda}^{rr}$ & 0.1 \\
    			${oa}$ & Object-object/wall attachment relation & ${\lambda}^{oa}$ & 1 \\
    			${fa}$ & Object-floor attachment relation & ${\lambda}^{fa}$ & 1 \\
    			${ca}$ & Object-ceiling attachment relation & ${\lambda}^{ca}$ & 1 \\
    			${rd}$ & Object-object relative distance relation & ${\lambda}^{rd}$ & 0.01 \\
    			\hline
    			${\delta}$ & 3D bounding box center offset & ${\lambda}^{rd}$ & 0.0001 \\
    			${d}$ & 3D bounding box distance & ${\lambda}^{d}$ & 0.01 \\
    			${s}$ & 3D bounding box size & ${\lambda}^{s}$ & 1 \\
    			${\theta}$ & Object orientation & ${\lambda}^{\theta}$ & 0.001 \\
    			${bp}$ & 3D bounding box projection & ${\lambda}^{bp}$ & 10 \\
    			\hline
    		\end{tabular}
		}
	\end{center}
	\vspace{-1.0em}
	\caption{\zc{Terms} in relation optimization and the weight of each \zc{term}.}
	\label{tbl:ro_weights}
\end{table}

\noindent \textbf{Hyper parameters}
For the weights of $\mathcal{L}_{ODN}$, we refer to Total3D \cite{nie2020total3dunderstanding} for detailed settings.
For the loss weights of $\mathcal{L}_{RGCN}$ and the joint loss $\mathcal{L}$, we show the description of each output and its corresponding loss weight in \autoref{tbl:rgcn_weights}.
For relation optimization, we weight each \yz{term} by its confidence and importance.
\yz{For example}, 2D observations should be more confident\yz{,} and collision \zc{terms} should be weighted more if we consider physically plausible object poses important.
We show the description of each \zc{term} and its corresponding weight in \autoref{tbl:ro_weights}.
\zc{In optimization, we use a gradient descend optimizer and set the learning rate to 1, steps to 100, and momentum to 0.9}.

\begin{figure}[t]
	\centering
    \animategraphics[autoplay,controls,loop,width=0.45\textwidth]{10}{figure/relation_optim/frame_}{0}{32}
	\caption{Visualization of our proposed relation optimization. A PDF reader like \textbf{Adobe Acrobat Reader / KDE Okular}  might be needed for displaying animated sequences. We also include more animation as a video along with this pdf file. The ground truth object bounding boxes are visualized with gray color for reference, while the current states are colorized. The attachment relations among objects, walls, floor, and ceiling are indicated by thick white lines, while the collisions are in red.}
	\label{fig:relation_optim}
\end{figure}

\noindent \textbf{Training}
All the borrowed networks (\ie, Mask-RCNN, HorizonNet, ODN, LIEN, LDIF) are fine-tuned individually on our proposed dataset.
Specifically, Mask-RCNN is fine-tuned from the weights pre-trained on COCO dataset, with batch size of 8 and learning rate of 2e-3 for 1e5 steps.
HorizonNet is fine-tuned from the weights pre-trained on Structured3D dataset, with batch size of 6 and learning rate of 1e-4 for 50 epochs.
ODN is fine-tuned from the weights pre-trained on SUN RGB-D, with batch size of 6 and learning rate of 1e-4 for 15 epochs.
LIEN and LDIF are fine-tuned from the weights pre-trained on Pix3D, with batch size of 24 and learning rate of 2e-4 for 100 epochs.
\zc{To make a fair comparison,}
\yz{all variation of Total3D \cite{nie2020total3dunderstanding} and Im3D \cite{zhang2021holistic} including the perspective and panorama version are also fine-tuned on our proposed dataset following the above process.}
\zc{For \yz{Total3D-Pers} and \yz{Im3D-Pers}, the ODN and \yz{Scene Graph Convolutional Network (SGCN)} are fine-tuned and tested with detection results}
\yz{obtained from split views.}
\zc{In addition, MGNet used by Total3D is fine-tuned from the weights pre-trained on Pix3D, with batch size of 16 and learning rate of 1e-4 for 100 epochs.}
To train our proposed RGCN, 
we generate the attachment relation ground truth by doing collision detection with a tolerance of 0.1m (\ie, before collision detection, we expand the bounding box by 0.05m) on the ground-truth 3D object bounding boxes and the estimated layout walls.
\yz{The} other relations are calculated according to their definition directly.
We first train our RGCN with only the pose refinement branch, with batch size of 16 and learning rate of 1e-4 for 35 epochs.
Then we fine-tune it with relation estimation branch for 20 epochs using the same settings.
Finally, we do an end-to-end training of ODN and RGCN with \yz{RO}, with batch size of 1 and learning rate of 1e-5 for 10 epochs.

\section{Visualization of Relation Optimization}
To visualize the process of our proposed relation optimization, we present \yz{an} animation in \autoref{fig:relation_optim}.
For demonstration, we add random noises to the ground truth object poses as the initial state, which 
\yz{simulates} the inaccuracy of the initial pose estimation.
\yz{We then use the relation generated from the ground truth poses to optimize the current poses (colorized) using our proposed method.}
We observe that as the optimization goes on, the position and orientation of the objects become closer to the ground truth, while the collisions are gradually resolved.

\begin{table}[t]
    	\begin{center}
	    \resizebox{1.\columnwidth}{!}{
    		\begin{tabular}{|l|c c|c c|}     			\hline
    			\multirow{2}{*}{Metric} & \multicolumn{2}{c|}{Total3D} & \multicolumn{2}{c|}{Im3D}     			\\
        			\cline{2-5}
    			& w/o. RO     			& w. RO     			& w/o. RO     			& w. RO     			\\
    			\hline
    			mAP (57 categories, $\uparrow$)  & 25.79 & 32.46 & 27.25 & 33.54 \\
			    avg col ($\downarrow$) & 3.41 & 0.89 & 2.62 & 0.90 \\
    			\hline
    		\end{tabular}
    	}
	\end{center}
	\vspace{-1em}
	\caption{\zc{The improvement of RO on different methods. The improvement of 3D object detection is evaluated with mAP of all 57 categories and physical violation is evaluated with average collision times per scene.}}
	\label{tbl:ro_improve}
\end{table}

\begin{table}[t]
    	\begin{center}
	    \resizebox{1.\columnwidth}{!}{
    		\begin{tabular}{|l|c|c|c|c|c|}
    			\hline
    			\makecell[c]{Methods \\ (Pano)} & Initial Estimation & \makecell[c]{Object \\ Reconstruction} & GCN & RO & Total\\
    			\hline
    			Total3D  & 0.66 & 0.23 (MGN) & - & - & 0.89 \\
    			\cline{1-1} \cline{3-6}
			    Im3D     & (Mask R-CNN, & 5.92 & 0.03 (Scene GCN) & - & 6.62 \\
			    \cline{1-1} \cline{4-6}
			    Ours     & HorizonNet, ODN) & (LIEN+LDIF) & 0.06 (Relation-based GCN) & 4.74 & 11.38 \\
    			\hline
    		\end{tabular}
    	}
	\end{center}
    \vspace{-1em}
	\caption{\zc{Efficiency comparison. We use average time per scene in seconds to compare efficiency of different methods and modules (tested on a single GTX 1080Ti).}}
	\label{tbl:eff_comp}
    \end{table}

\section{3D Detection mAP on all 57 categories}
\yz{In Tab. 1 of the main paper, we show 3D object detection results for 11 common categories. Here we show a complete quantitative evaluation on all 57 categories in \autoref{tbl:3d_detection_supp}.}
Same as the conclusion made in the main paper, our method outperforms the SoTA with a large margin.

\section{More Qualitative Comparison on 3D Detection and Scene Reconstruction}
\begin{figure*}[!ht]
	\centering
	\scriptsize
	\newcommand{\rgb}[1]{\raisebox{-0.5\height}{\includegraphics[width=.147\linewidth]{#1}}}
	\newcommand{\bird}[1]{\raisebox{-0.5\height}{\includegraphics[width=.147\linewidth,clip,trim=110 80 100 70]{#1}}}
	\newcommand{\rot}[1]{\rotatebox[origin=c]{90}{#1}}
	\def\arraystretch{0.5}	\begin{tabular}{c|c*{6}{c@{\hspace{1px}}}}
            \rule{0pt}{5px}&\rot{Input}
            & \rgb{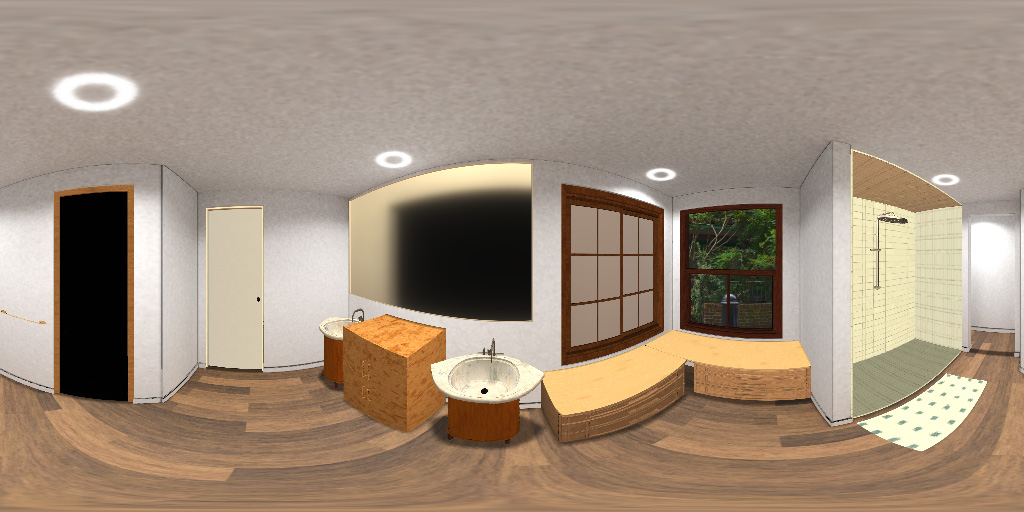}
            & \rgb{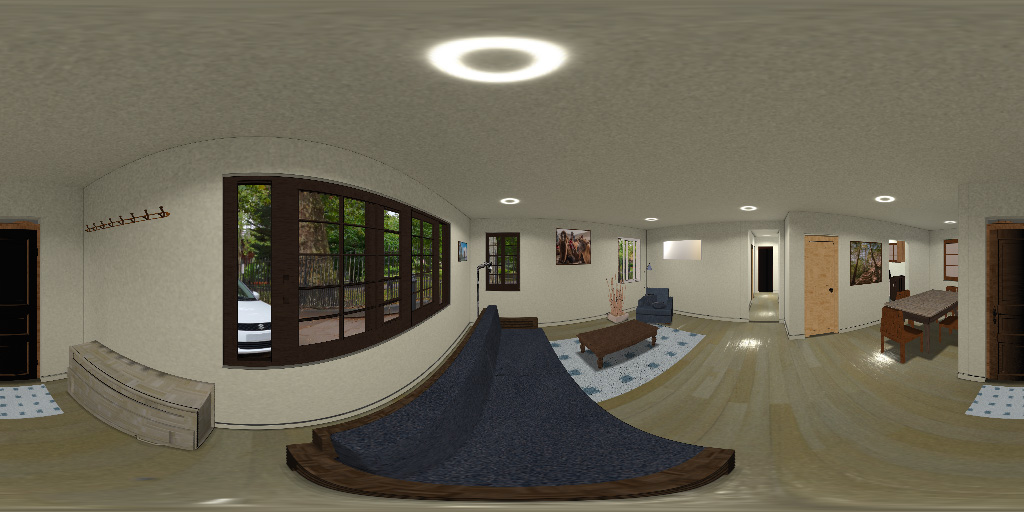}
    	    & \rgb{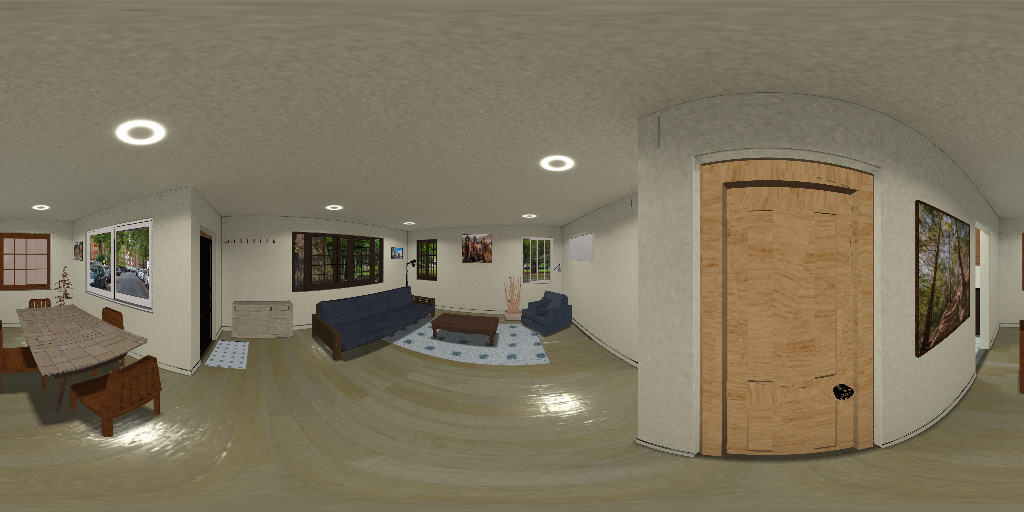}
    	    & \rgb{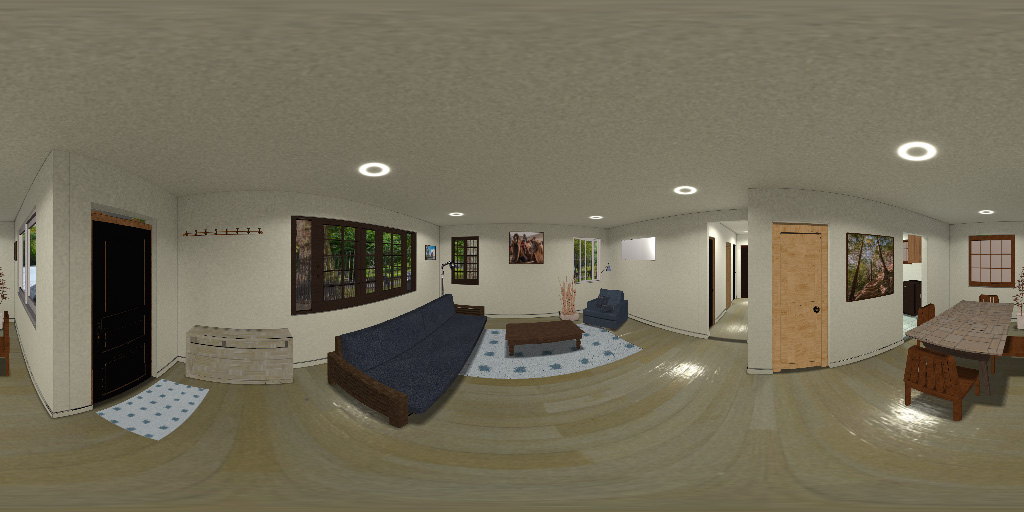}
    	    & \rgb{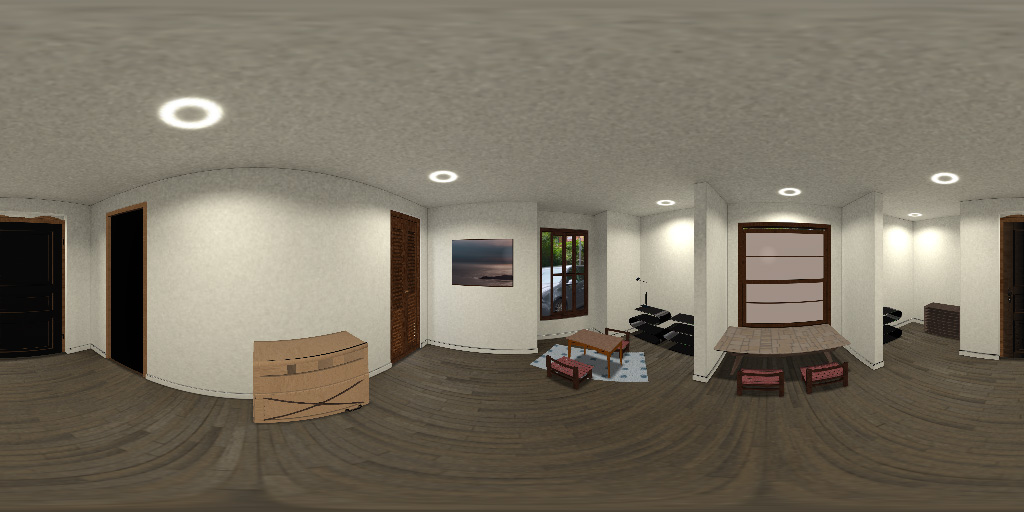}
    	    & \rgb{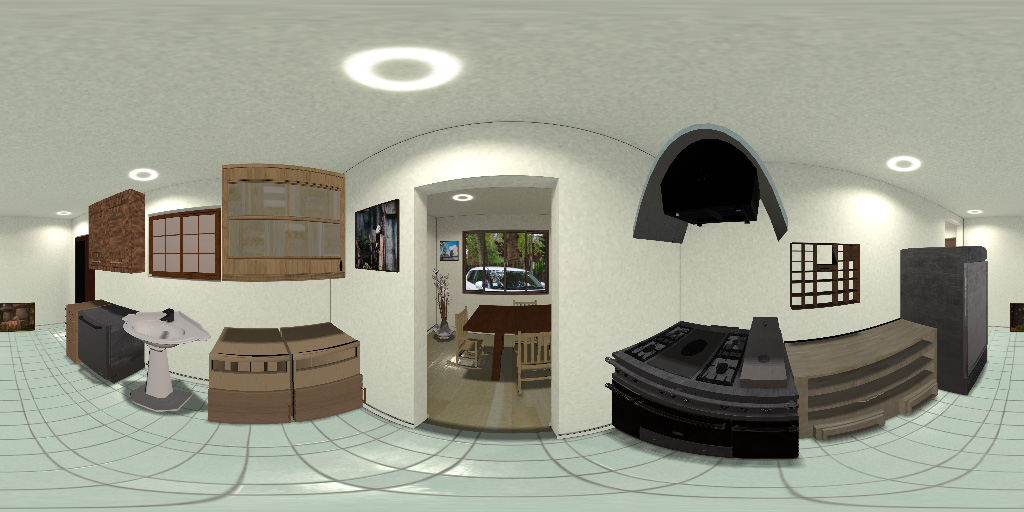}
    	       	       	       	    \\
    	\hline
    	    \multirow{4}{*}[-27ex]{\rot{Bird's Eye View}}
    	    &\rot{Total3D}
                & \raisebox{-0.5\height}{\includegraphics[width=.147\linewidth,clip,trim=80 60 80 50]{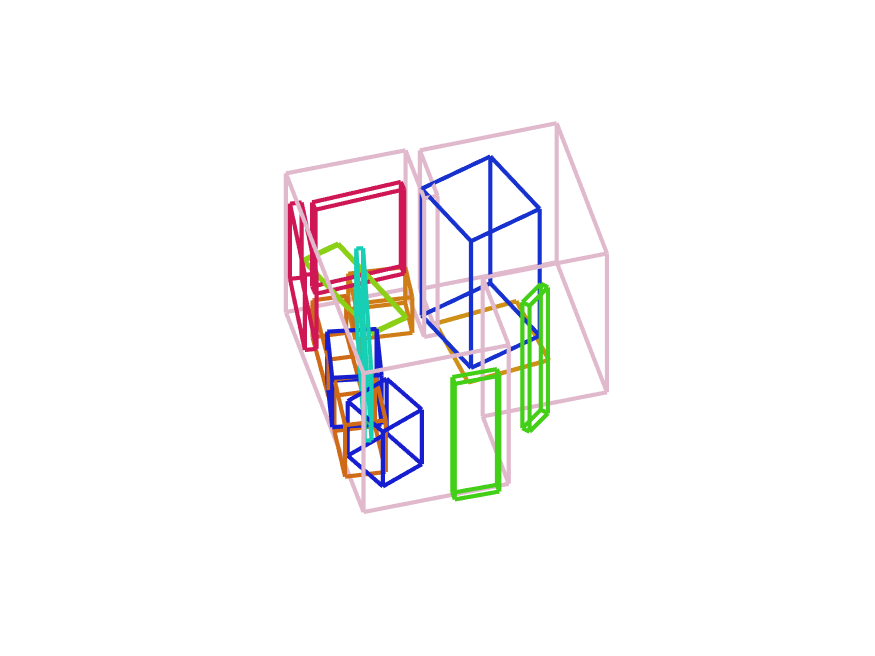}}
        	    & \bird{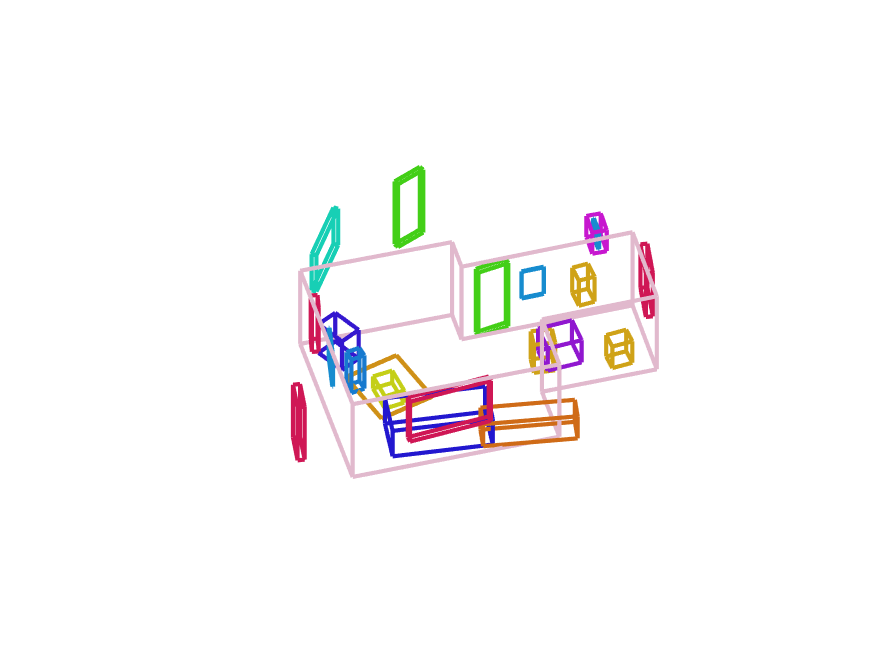}
        	    & \bird{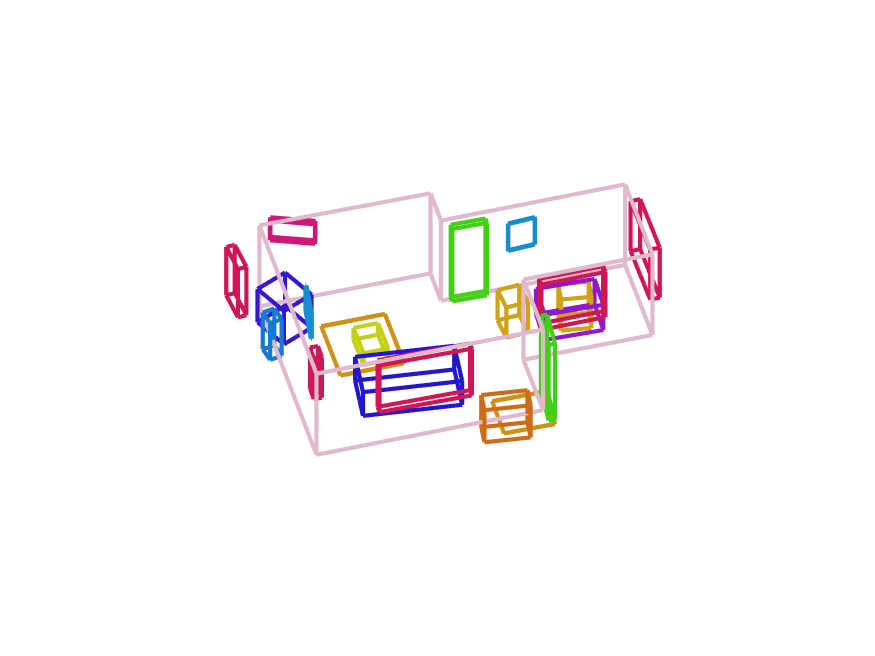}
        	    & \bird{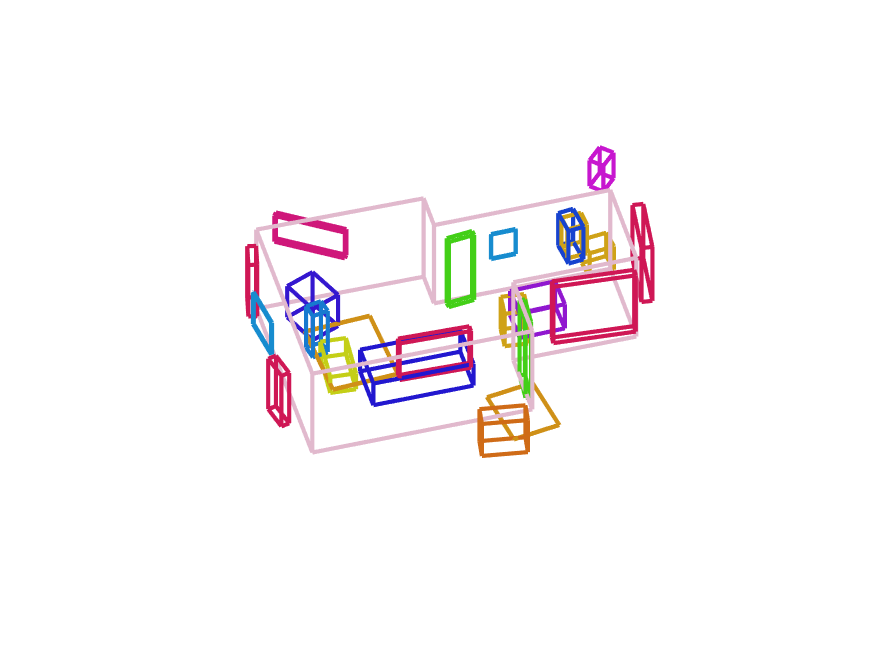}
        	    & \raisebox{-0.5\height}{\includegraphics[width=.147\linewidth,clip,trim=90 80 90 50]{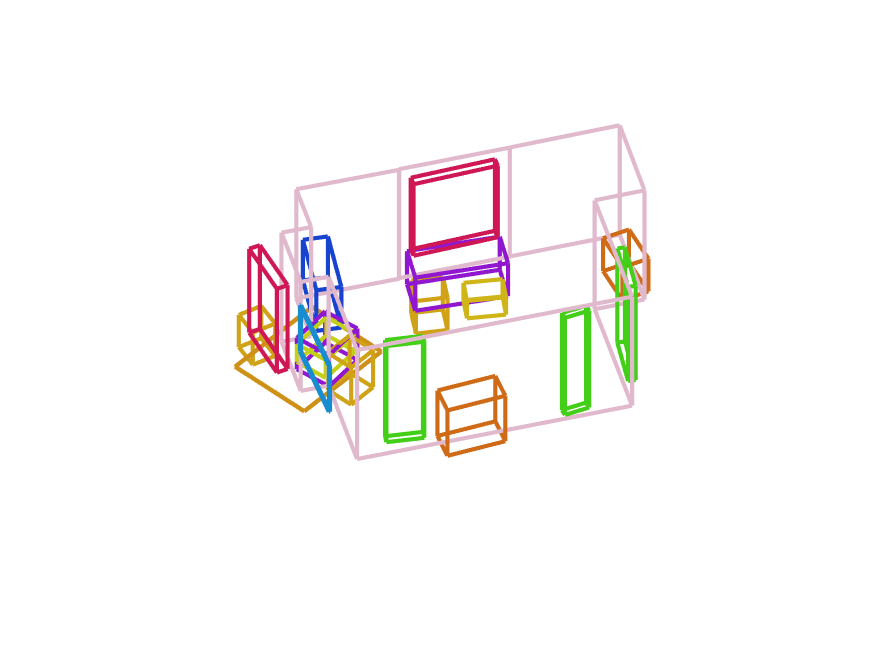}}
        	    & \raisebox{-0.5\height}{\includegraphics[width=.147\linewidth,clip,trim=120 90 80 70]{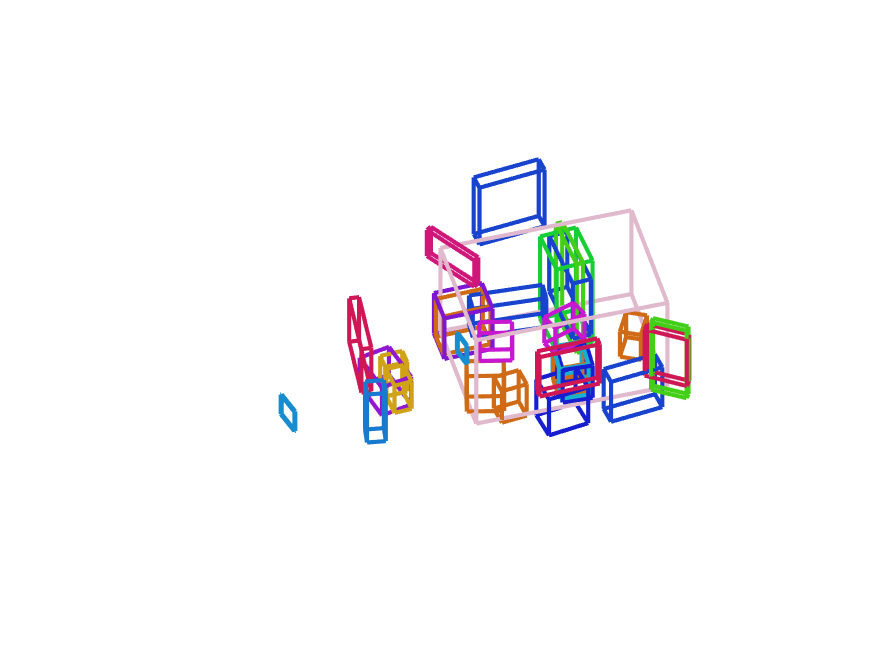}}
        	    \\
    	    &\rot{Im3D}
                & \raisebox{-0.5\height}{\includegraphics[width=.147\linewidth,clip,trim=80 60 80 50]{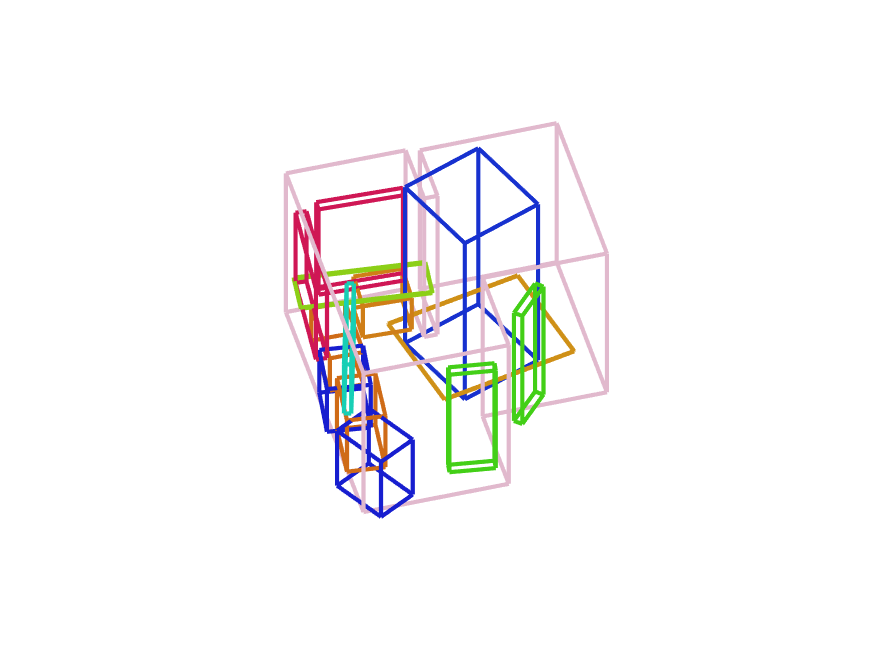}}
        	    & \bird{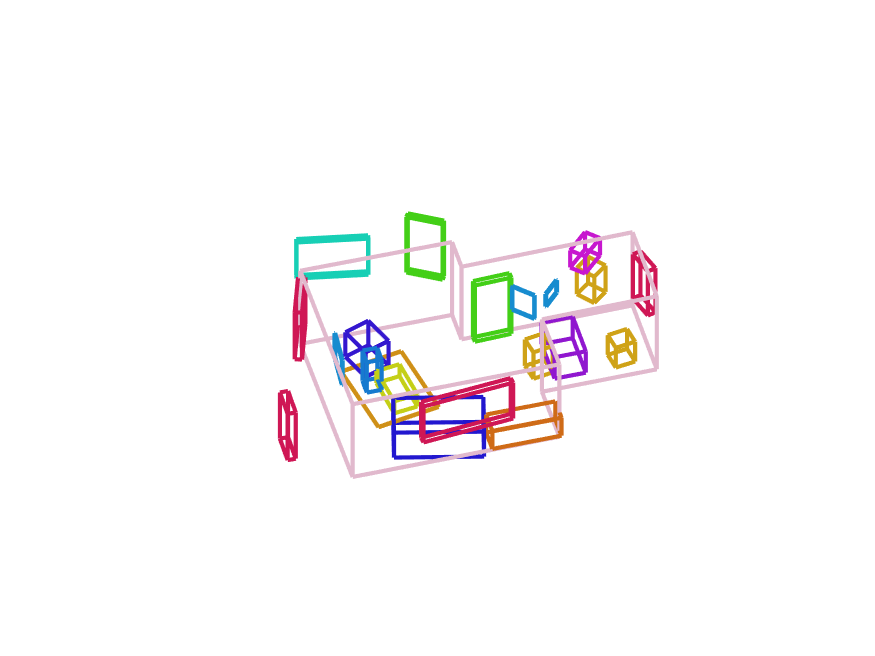}
        	    & \bird{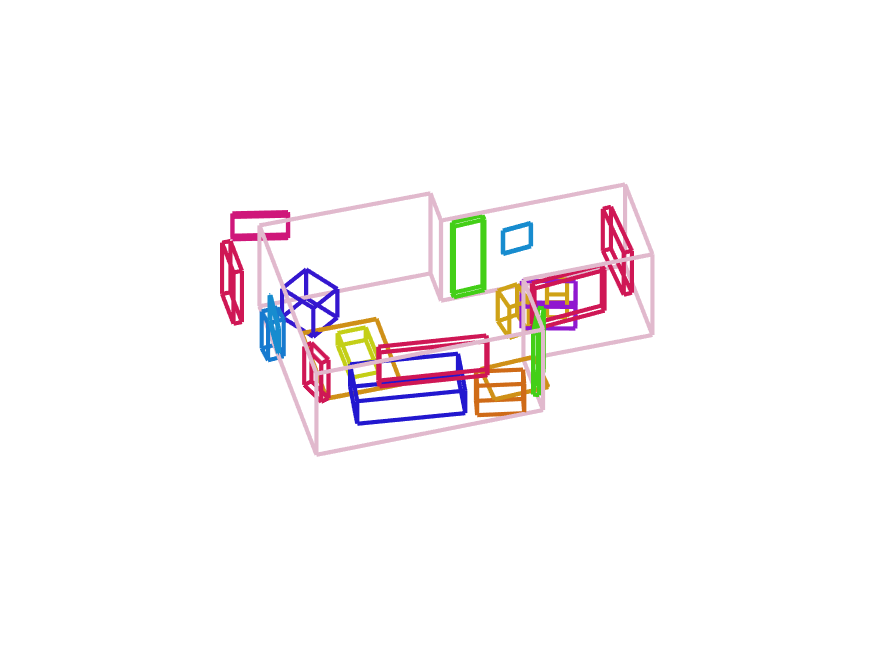}
        	    & \bird{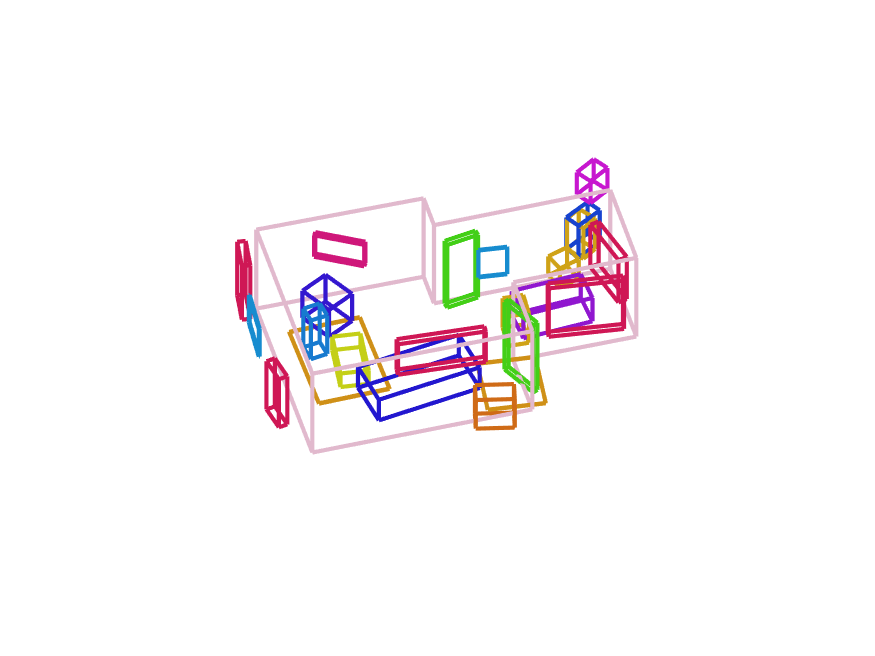}
        	    & \raisebox{-0.5\height}{\includegraphics[width=.147\linewidth,clip,trim=90 80 90 50]{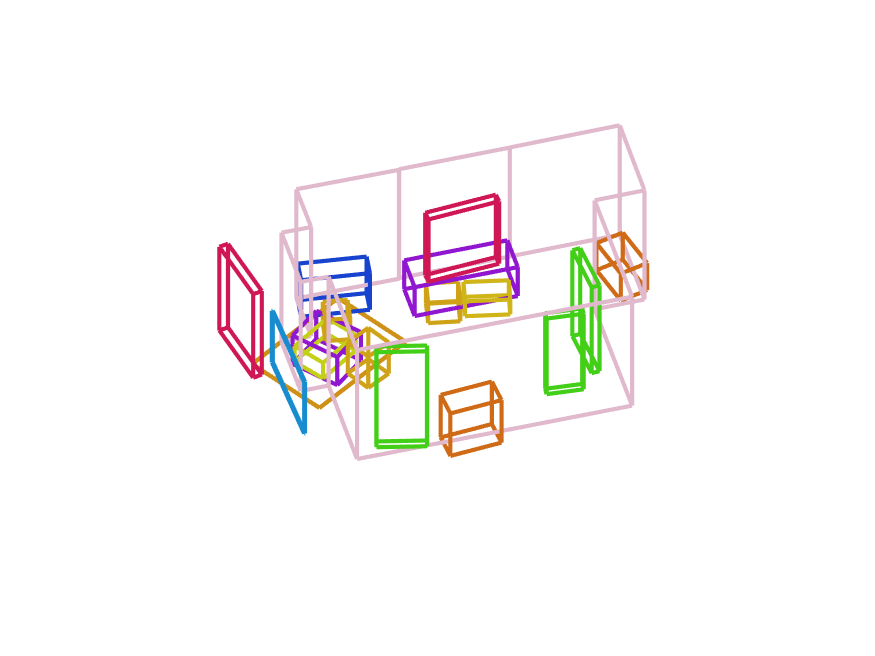}}
        	    & \raisebox{-0.5\height}{\includegraphics[width=.147\linewidth,clip,trim=120 90 80 70]{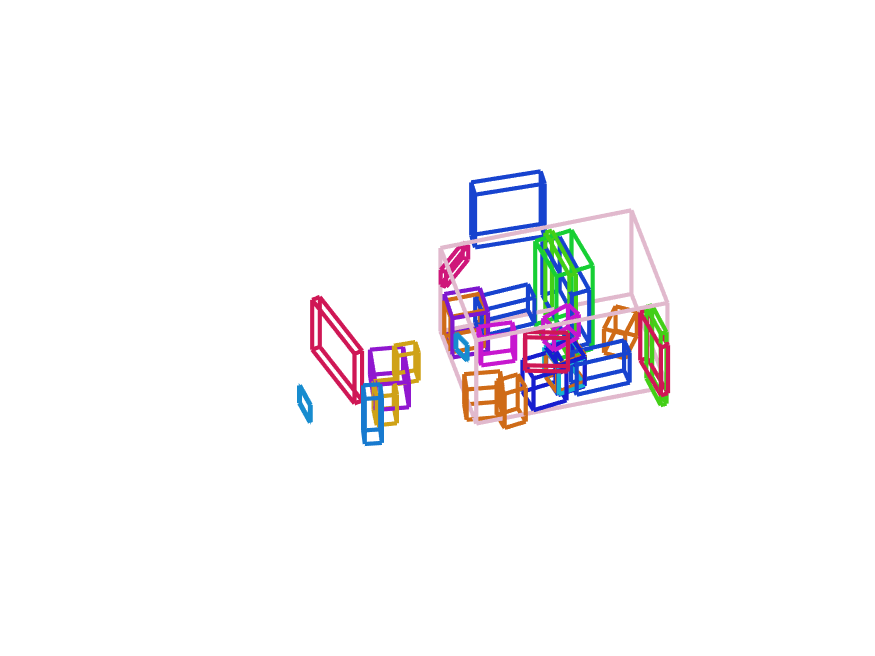}}
        	    \\
        	& \rot{Ours}
                & \raisebox{-0.5\height}{\includegraphics[width=.147\linewidth,clip,trim=80 60 80 50]{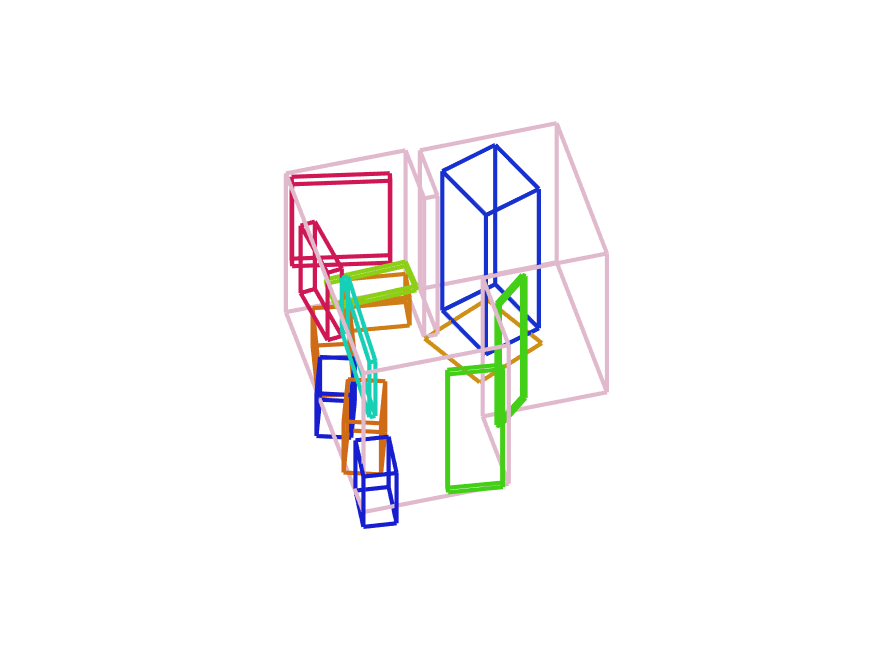}}
        	    & \bird{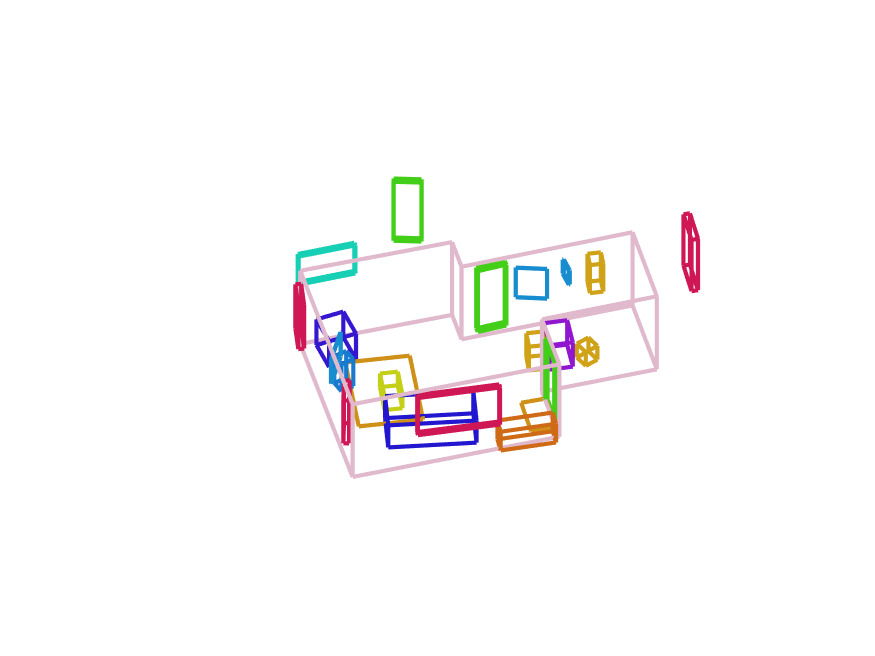}
        	    & \bird{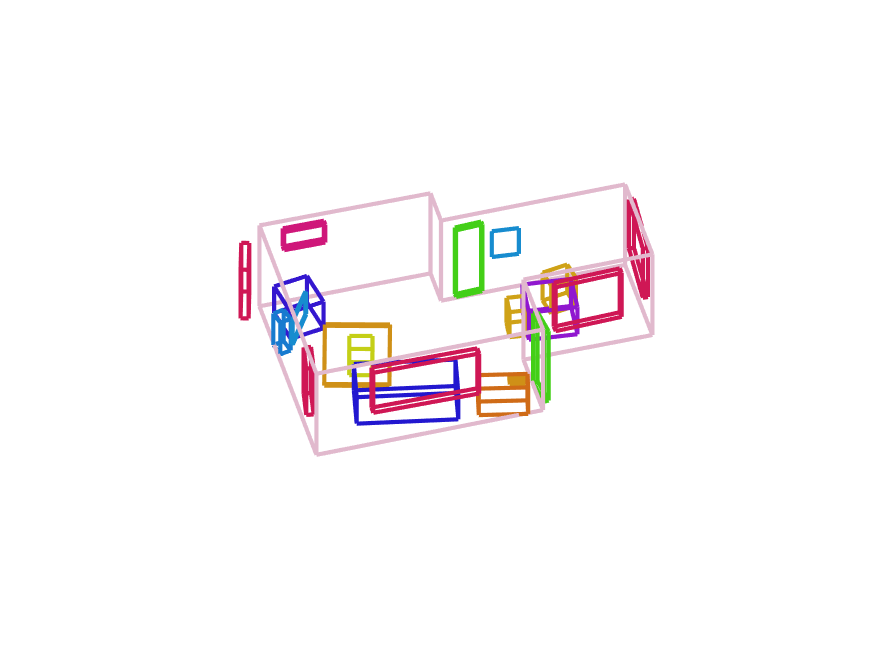}
        	    & \bird{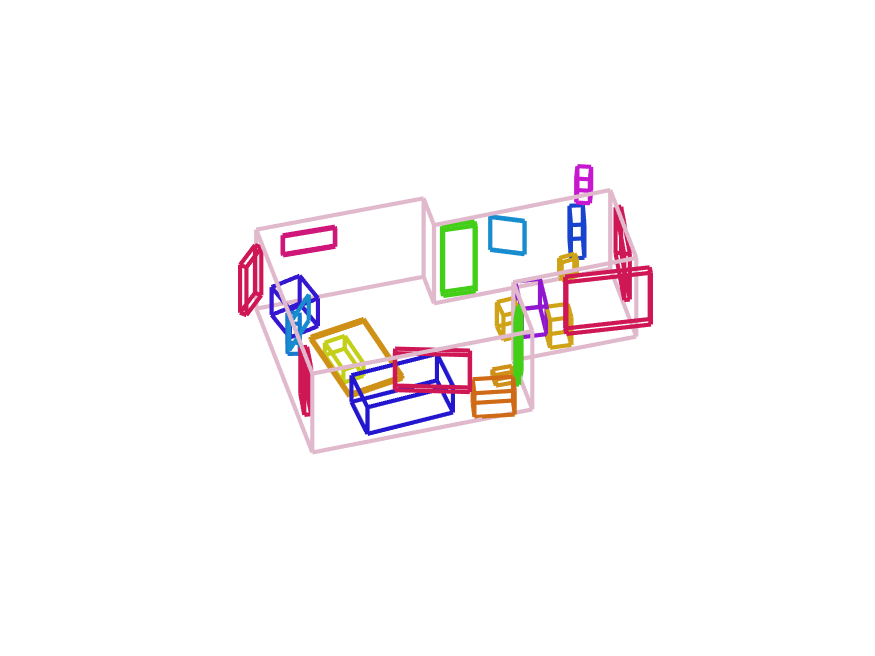}
        	    & \raisebox{-0.5\height}{\includegraphics[width=.147\linewidth,clip,trim=90 80 90 50]{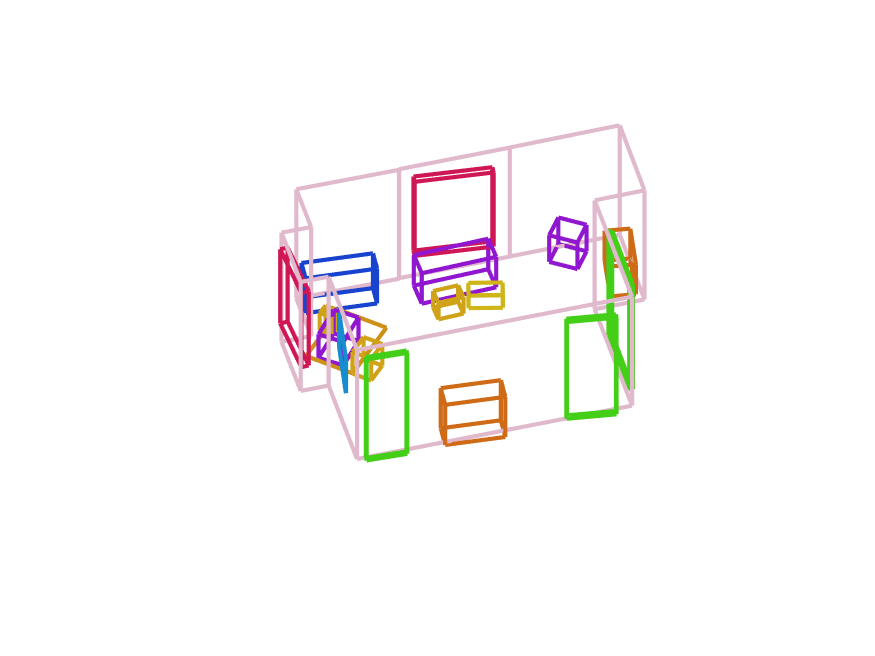}}
        	    & \raisebox{-0.5\height}{\includegraphics[width=.147\linewidth,clip,trim=120 90 80 70]{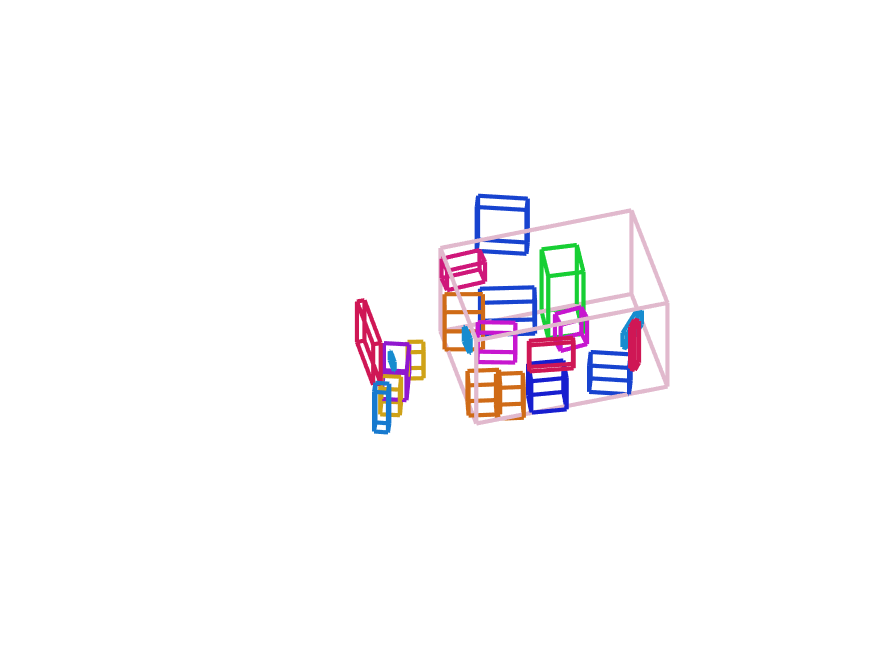}}
        	    \\
        	&\rot{GT}
                & \raisebox{-0.5\height}{\includegraphics[width=.147\linewidth,clip,trim=80 60 80 50]{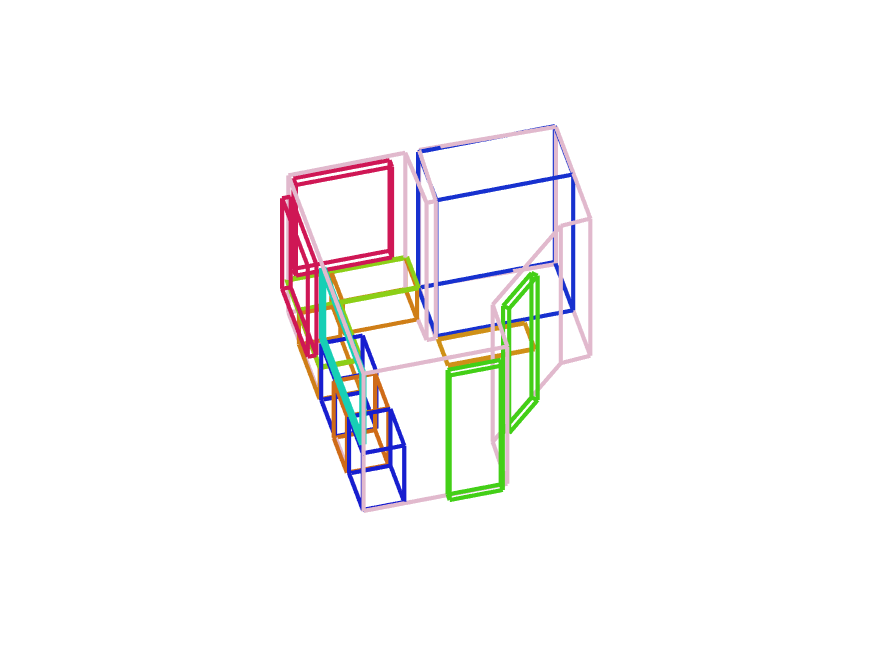}}
        	    & \bird{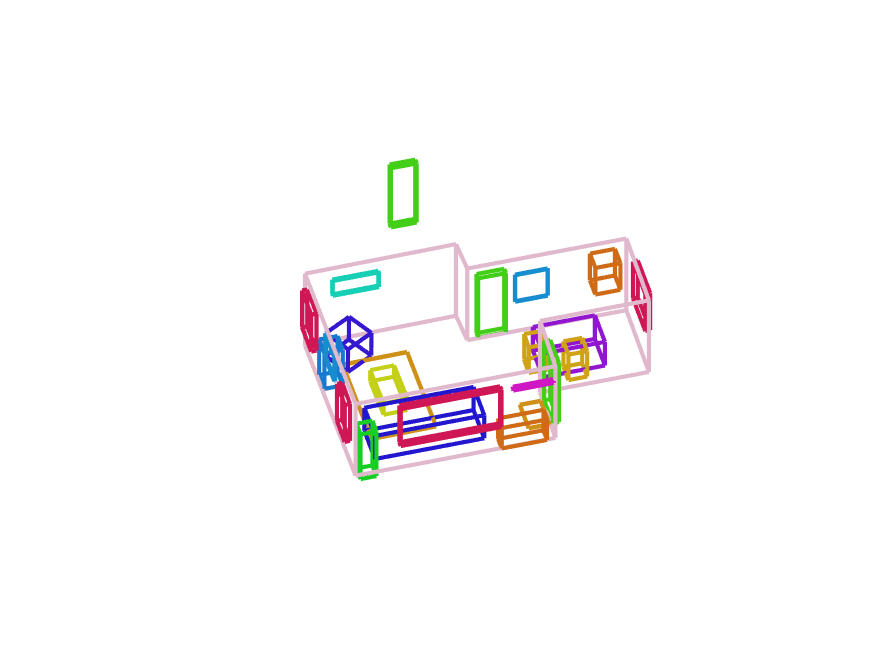}
        	    & \bird{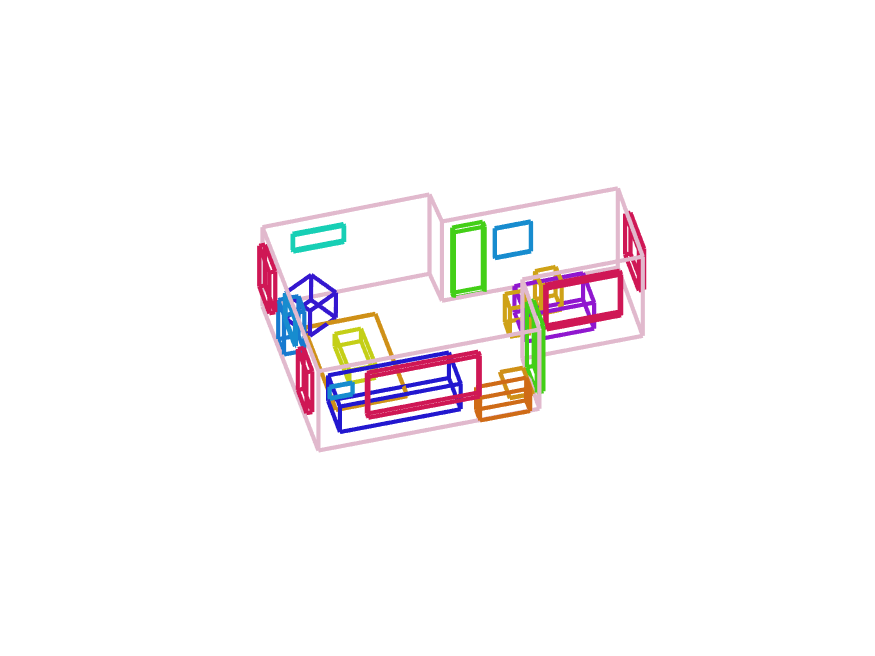}
        	    & \bird{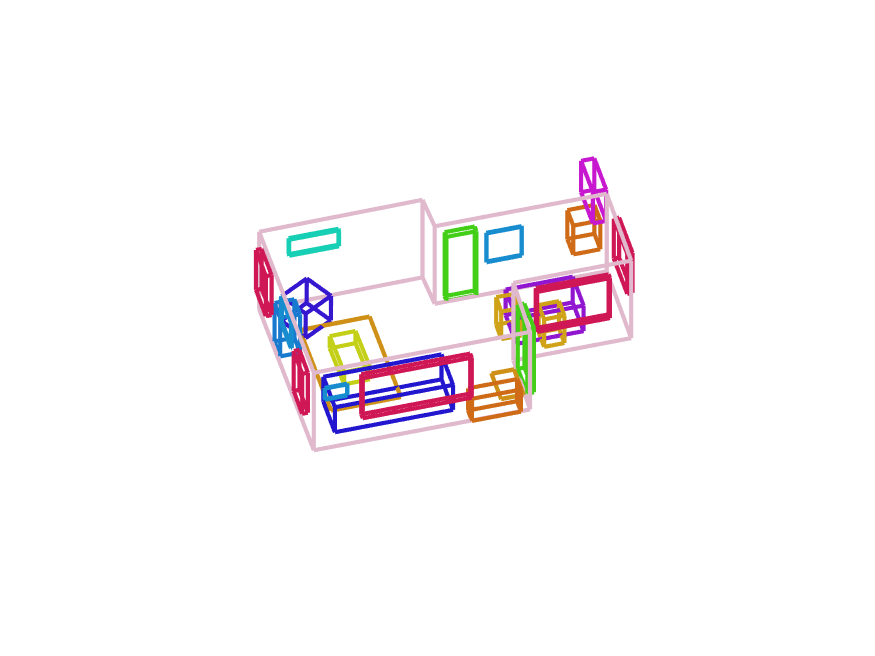}
        	    & \raisebox{-0.5\height}{\includegraphics[width=.147\linewidth,clip,trim=90 80 90 50]{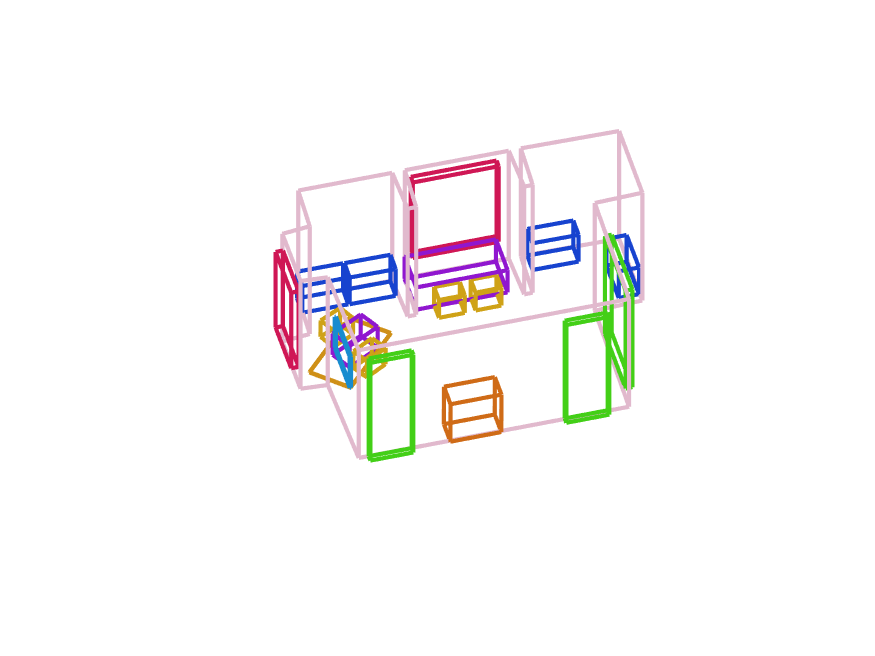}}
        	    & \raisebox{-0.5\height}{\includegraphics[width=.147\linewidth,clip,trim=120 90 80 70]{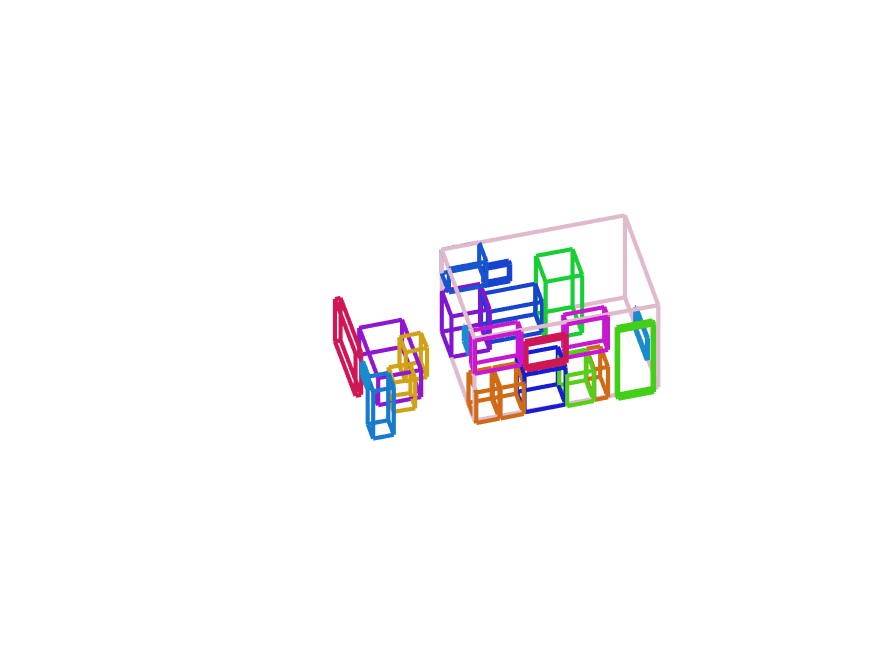}}
        	    \\
    	\hline
    	    \multirow{4}{*}[-10ex]{\rot{Panorama View}}
    	    &\rot{Total3D}
                & \rgb{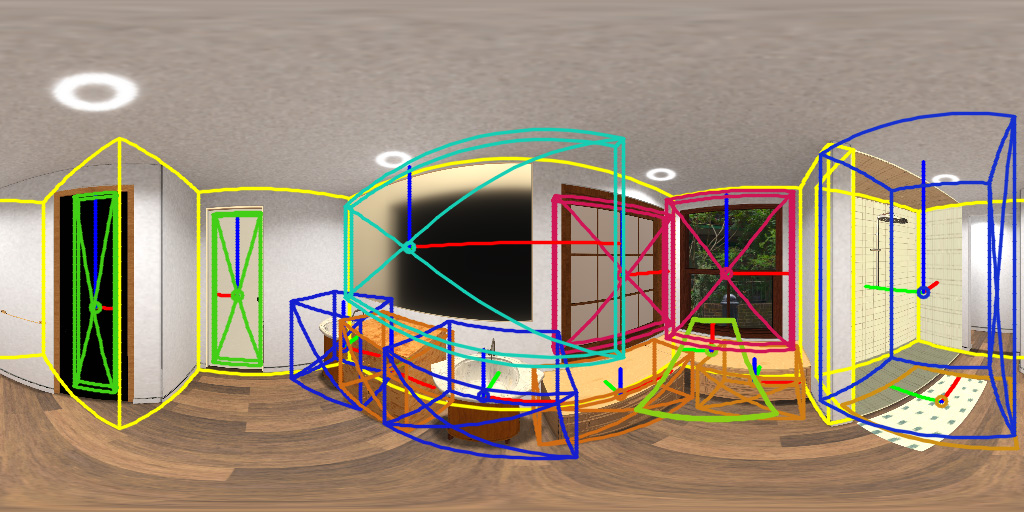}
        	    & \rgb{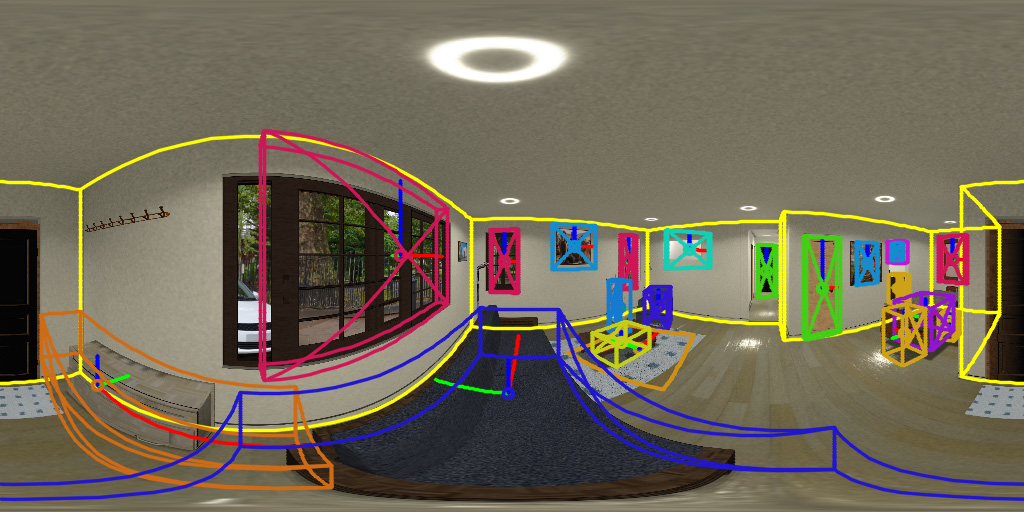}
        	    & \rgb{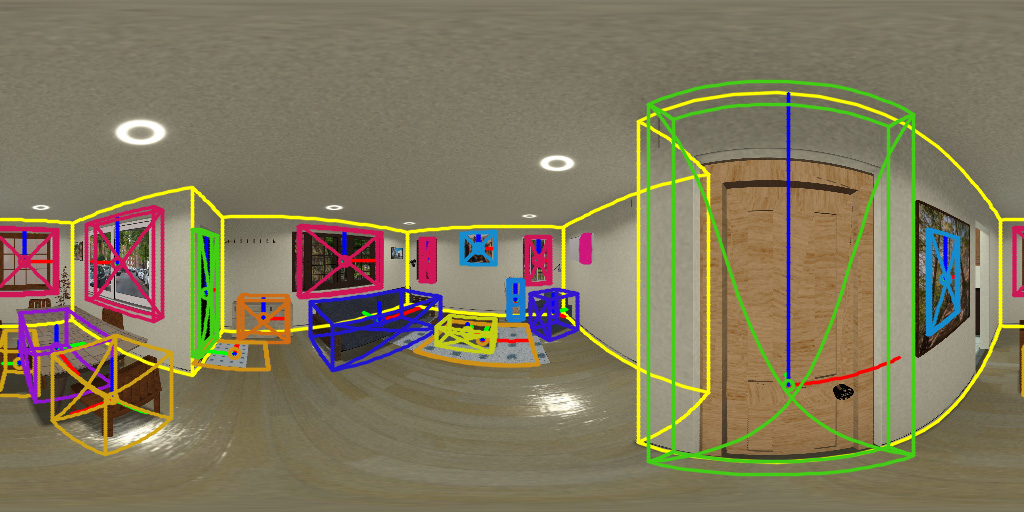}
        	    & \rgb{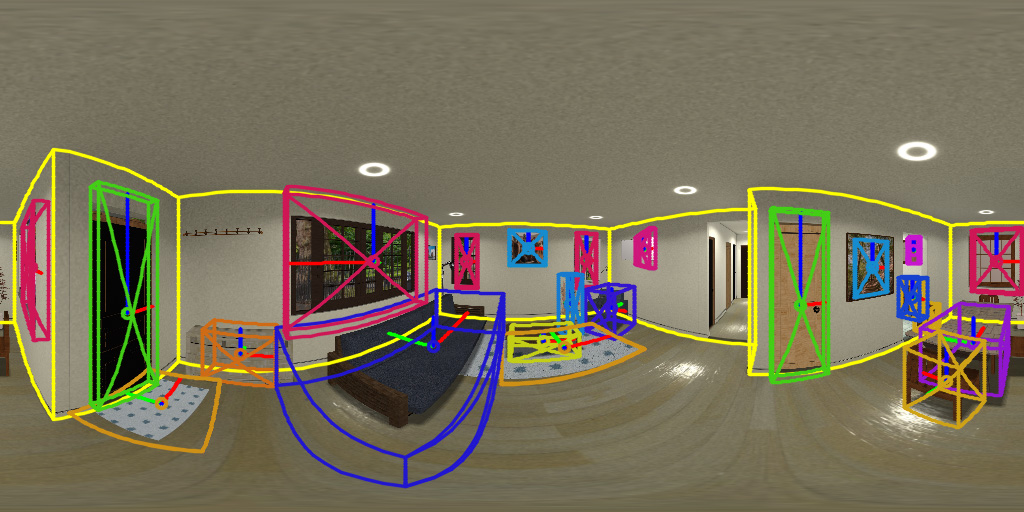}
        	    & \rgb{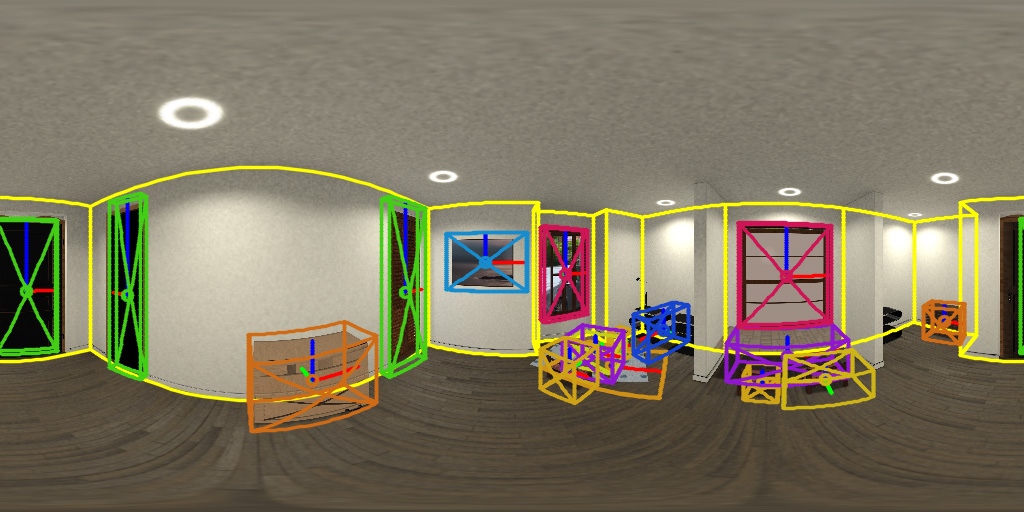}
        	    & \rgb{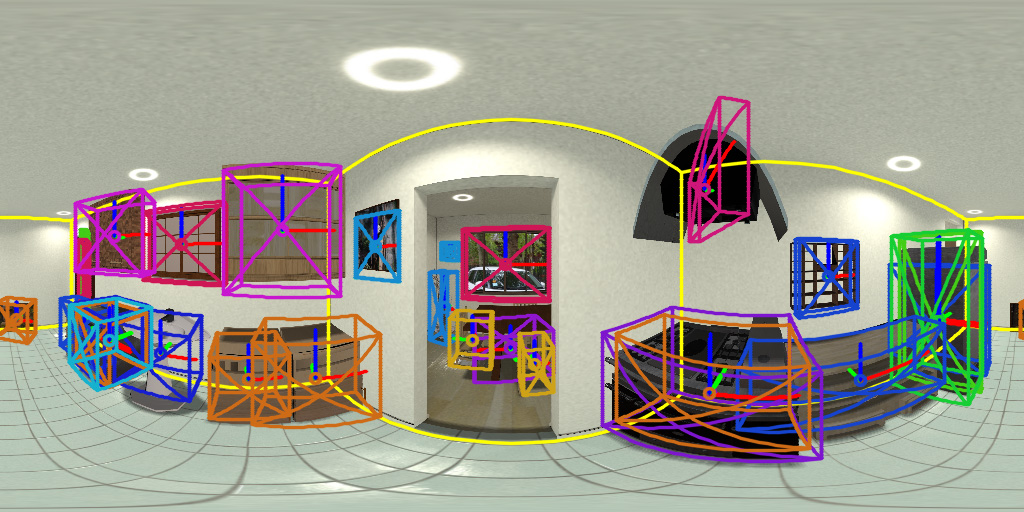}
        	    \\
    	    &\rot{Im3D}
                & \rgb{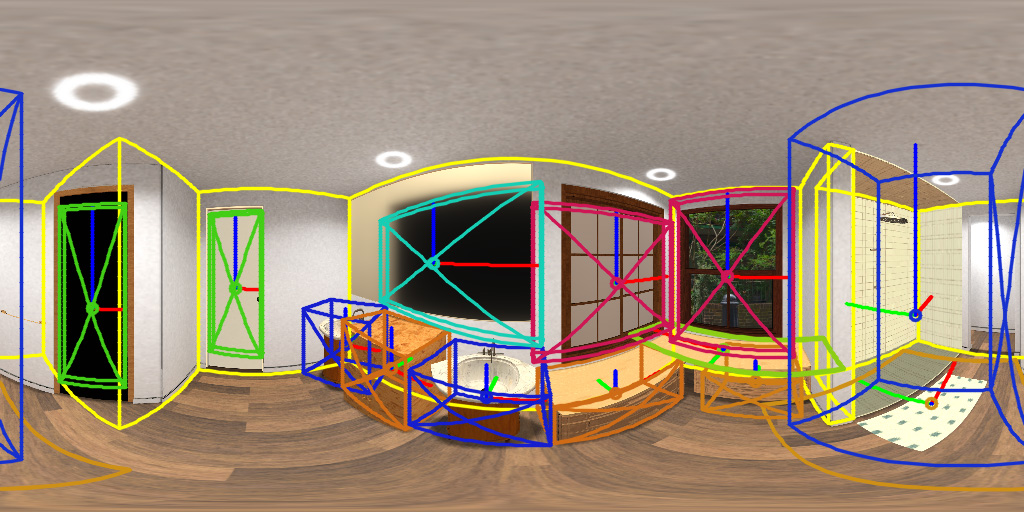}
        	    & \rgb{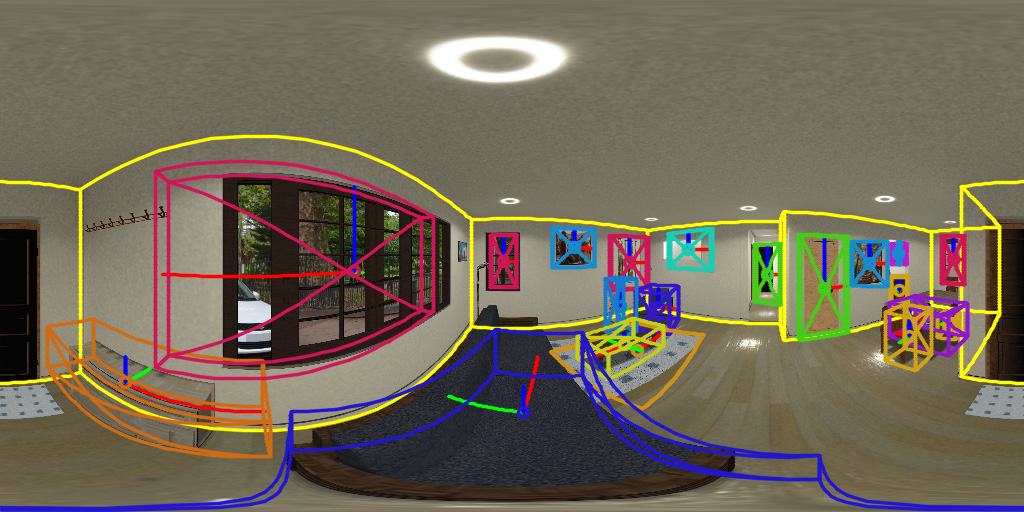}
        	    & \rgb{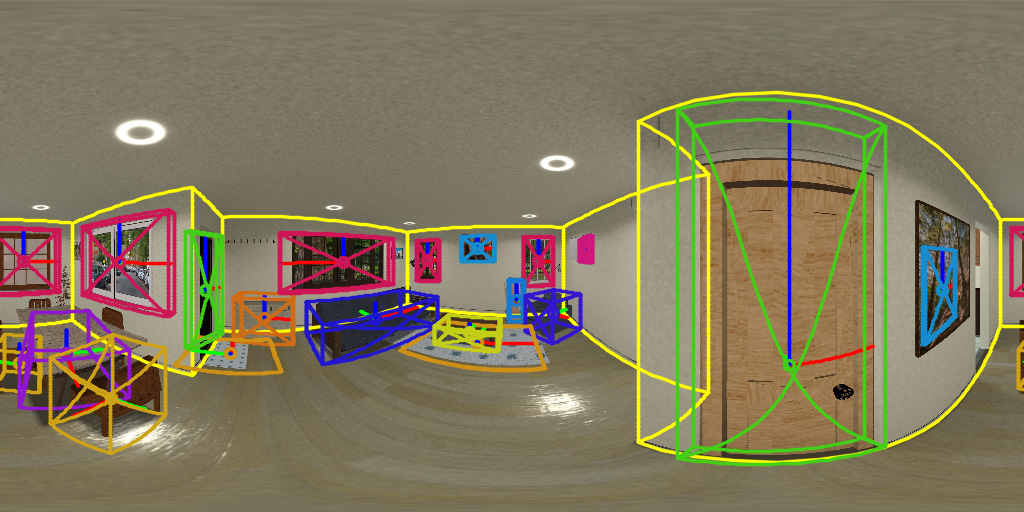}
        	    & \rgb{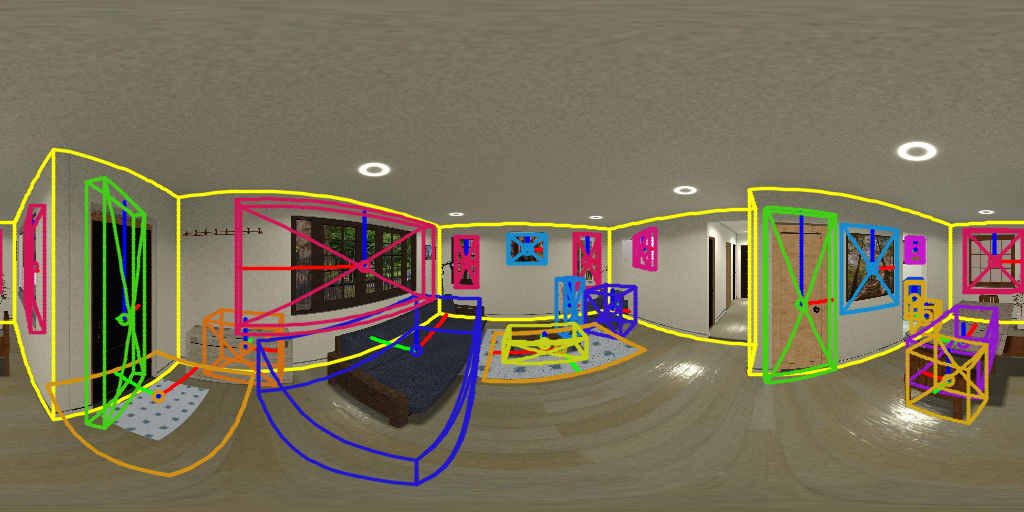}
        	    & \rgb{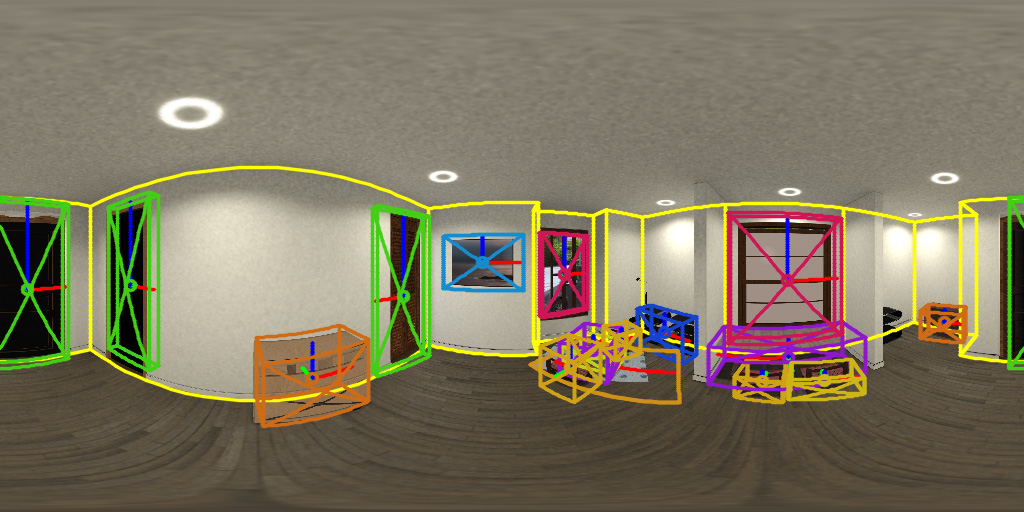}
        	    & \rgb{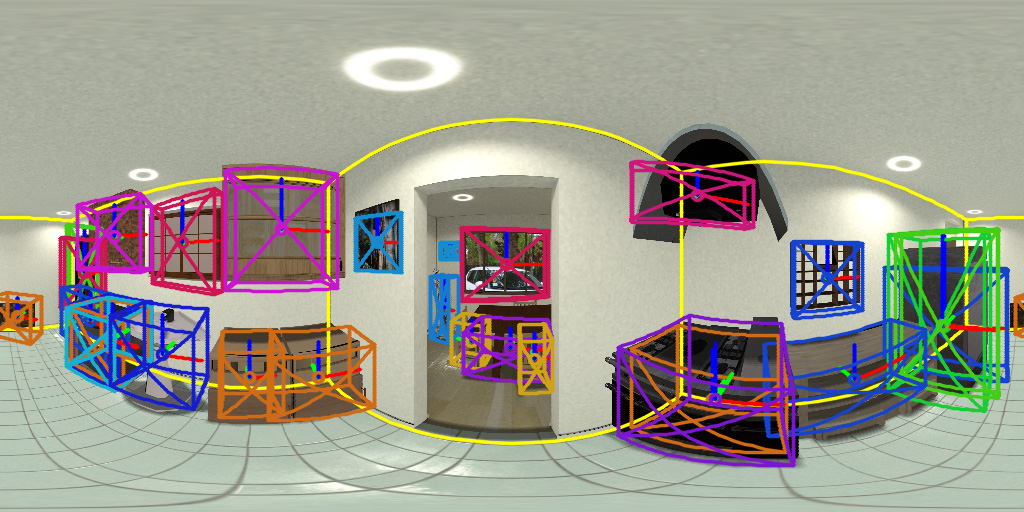}
        	    \\
            &\rot{Ours}
                & \rgb{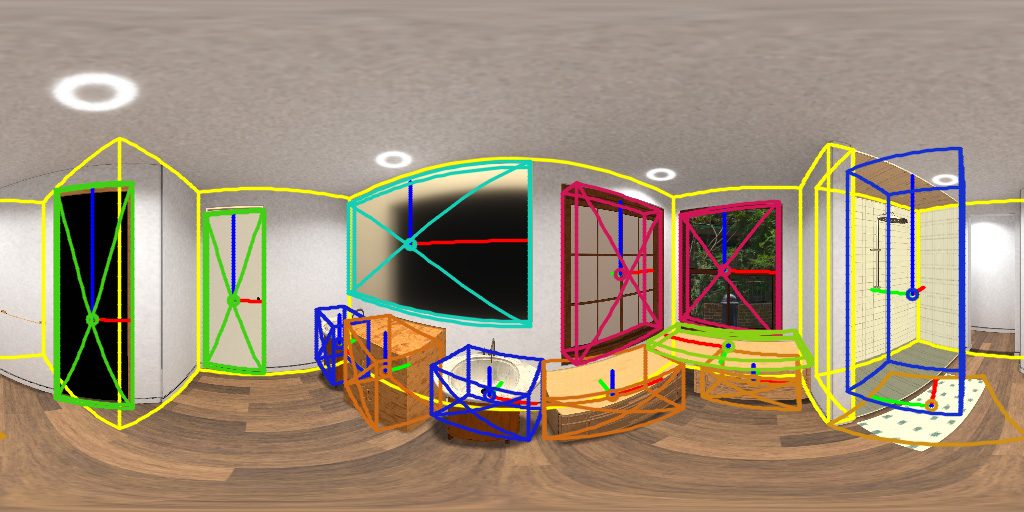}
        	    & \rgb{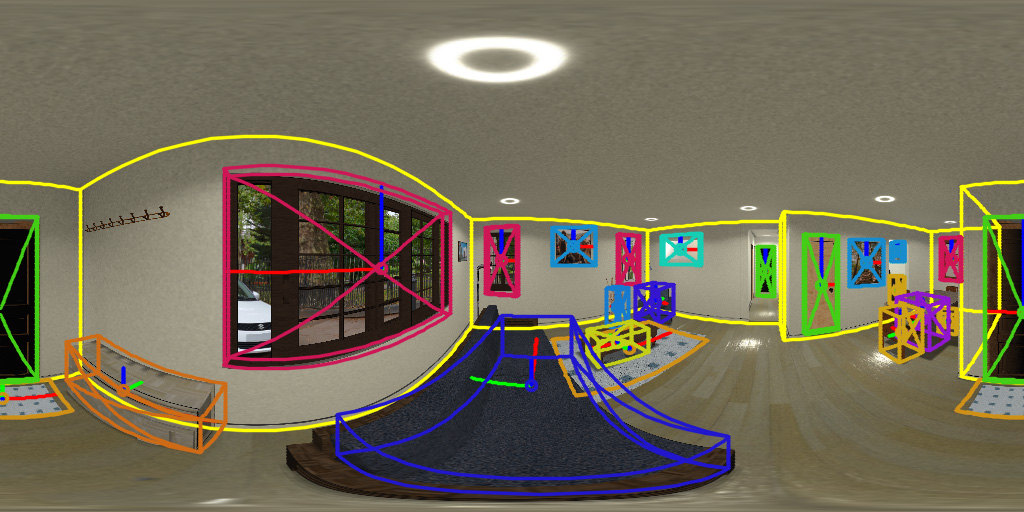}
        	    & \rgb{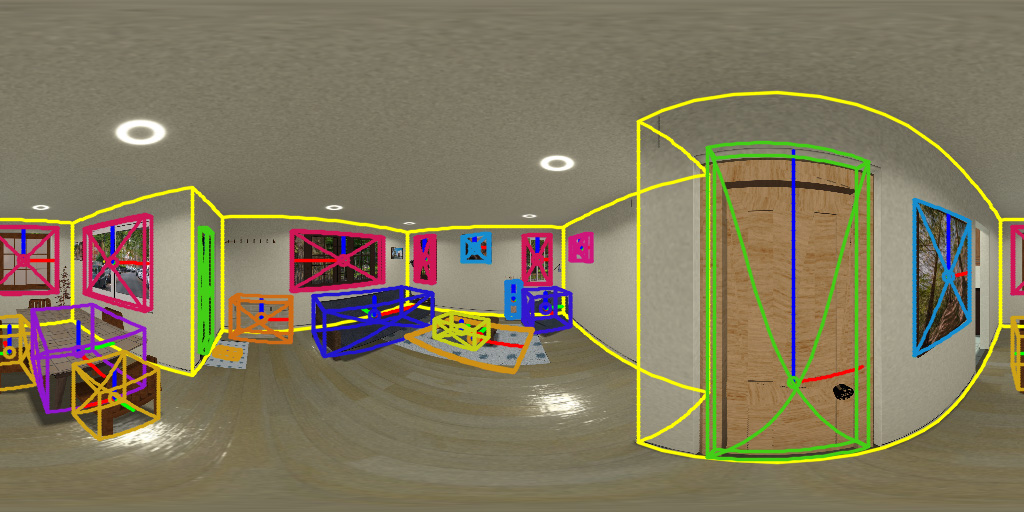}
        	    & \rgb{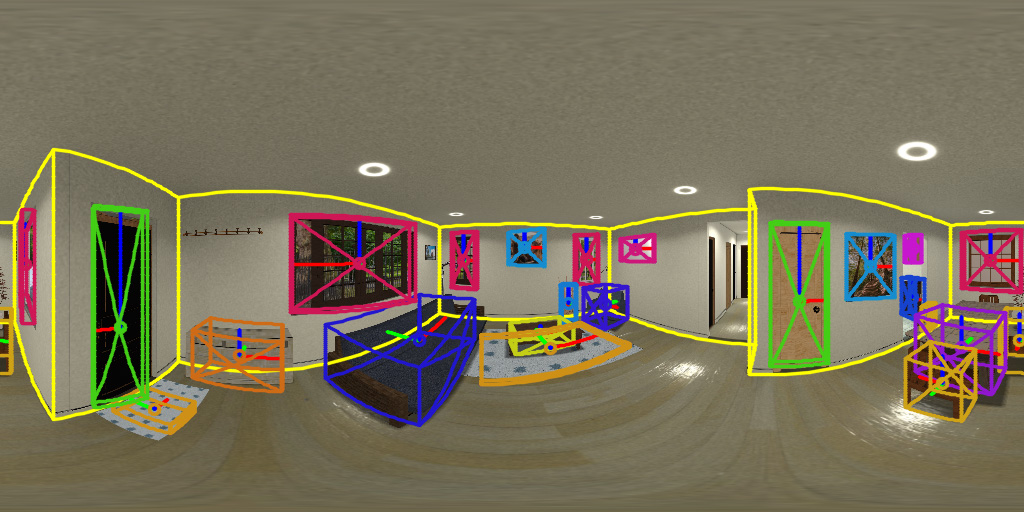}
        	    & \rgb{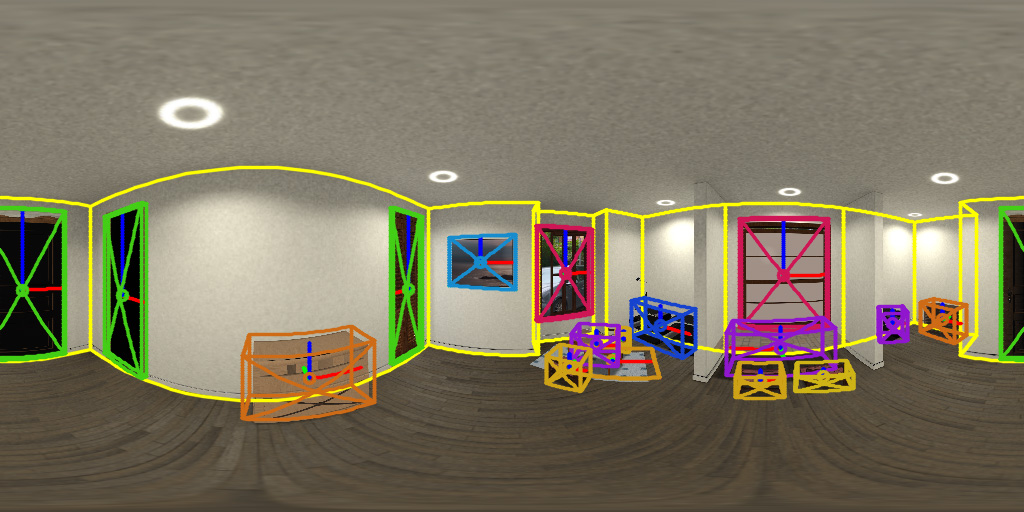}
        	    & \rgb{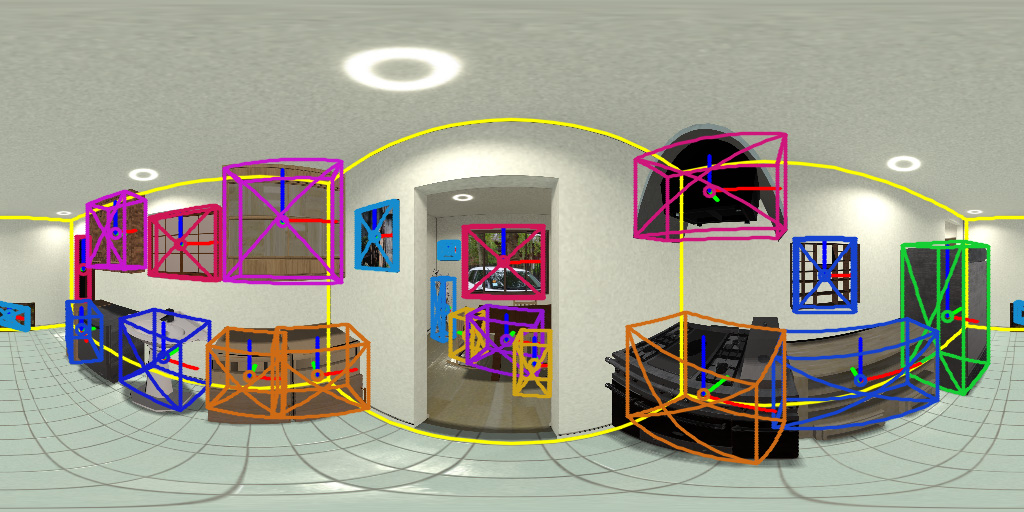}
        	    \\
        	&\rot{GT}
                & \rgb{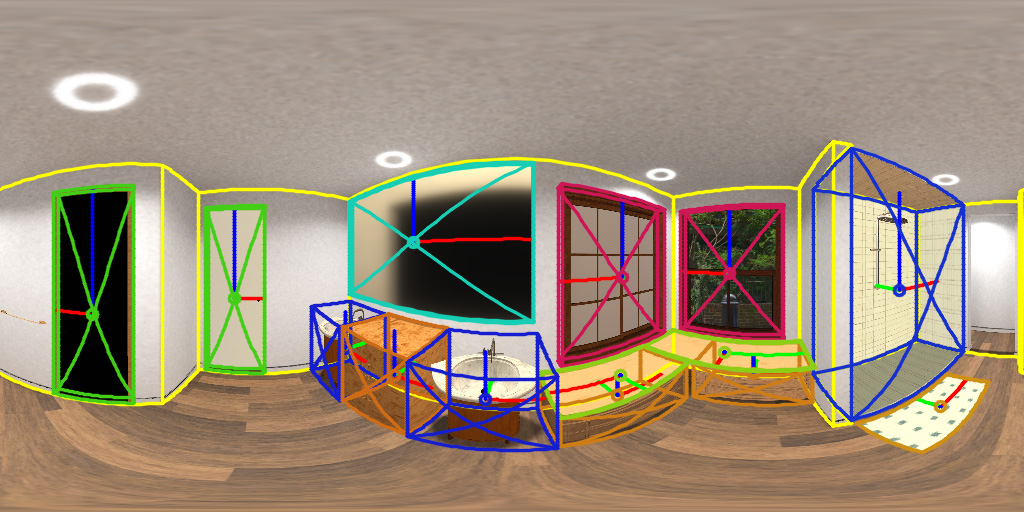}
        	    & \rgb{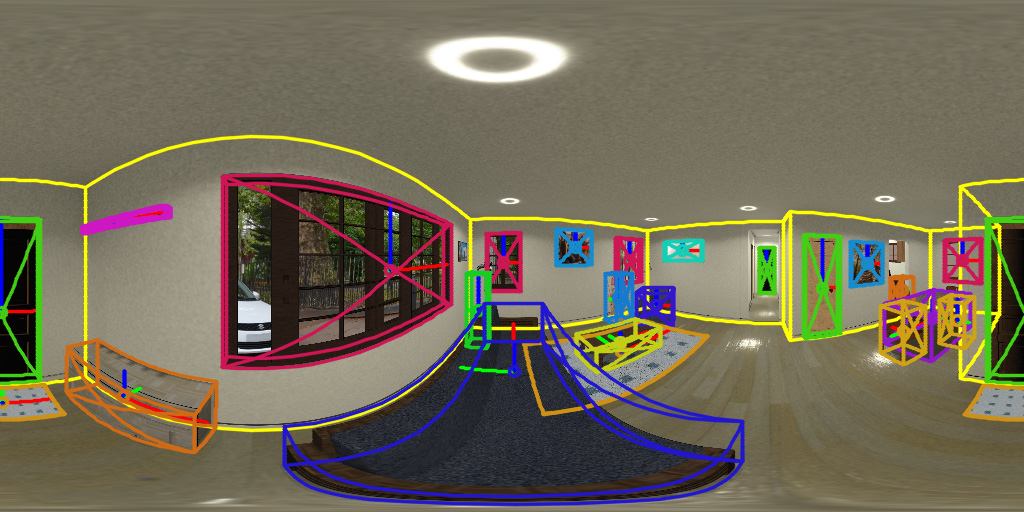}
        	    & \rgb{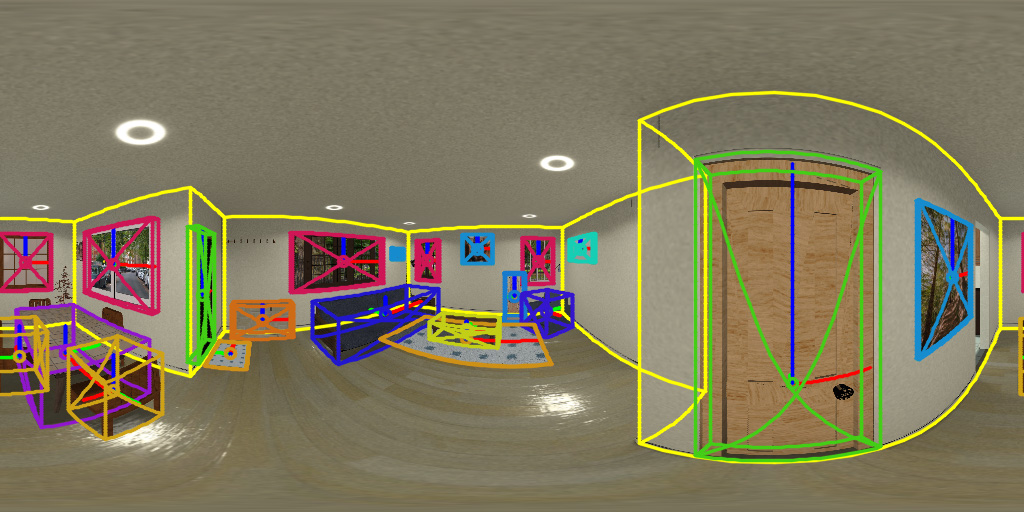}
        	    & \rgb{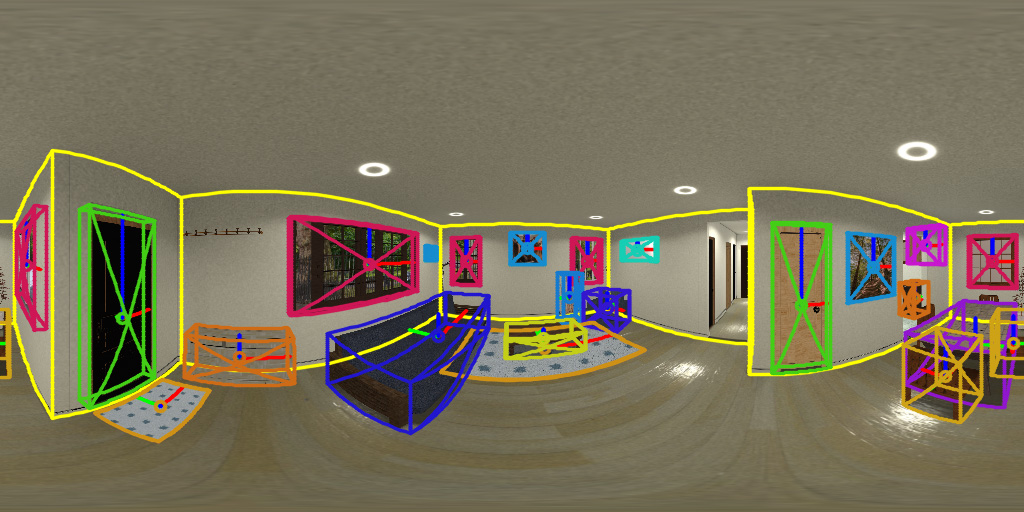}
        	    & \rgb{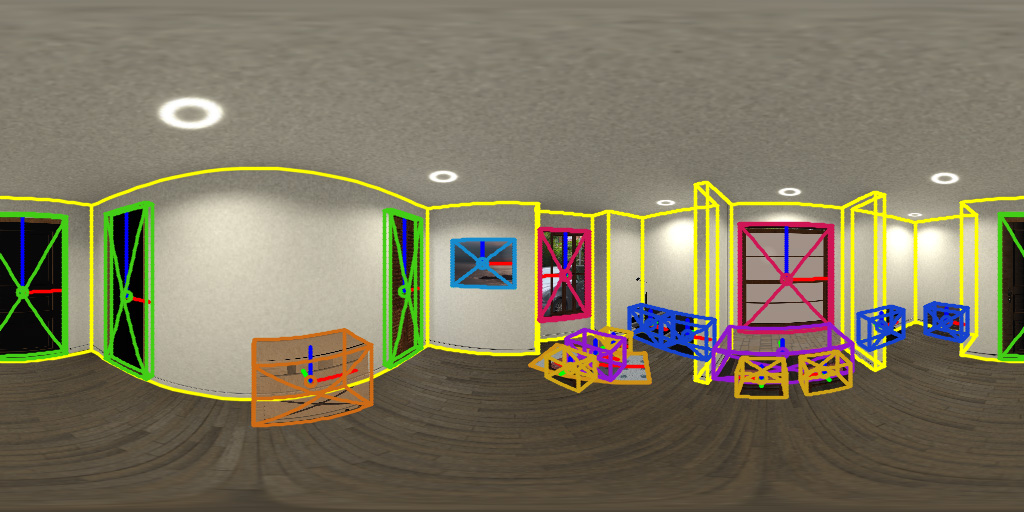}
        	    & \rgb{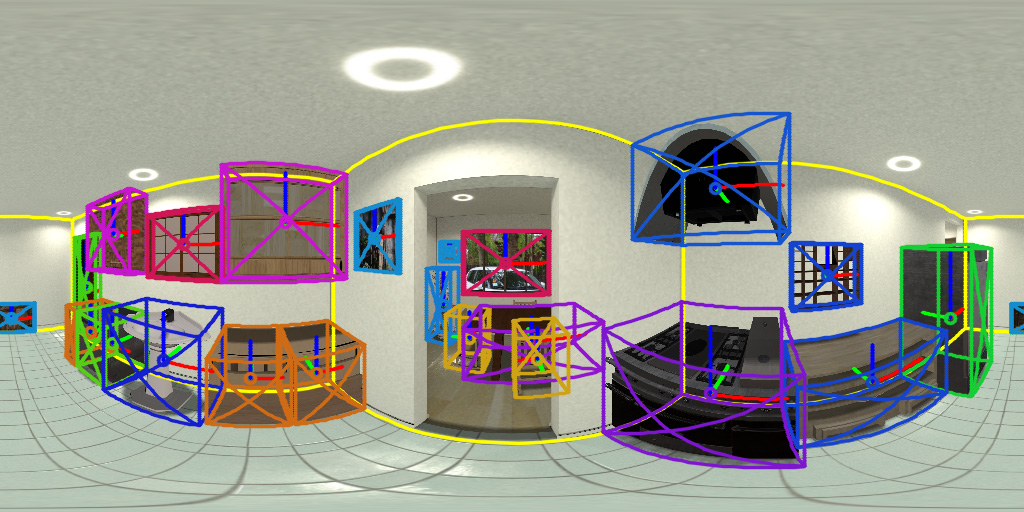}
        	    \\
    	\hline
    	    \multirow{4}{*}[-10ex]{\rot{Scene Reconstruction}}
    	    &\rot{Total3D}
                & \rgb{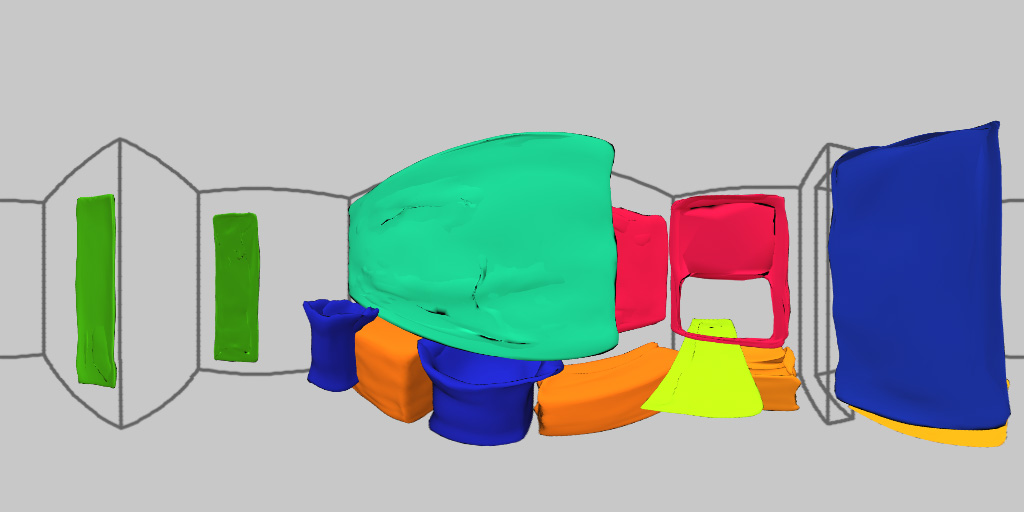}
        	    & \rgb{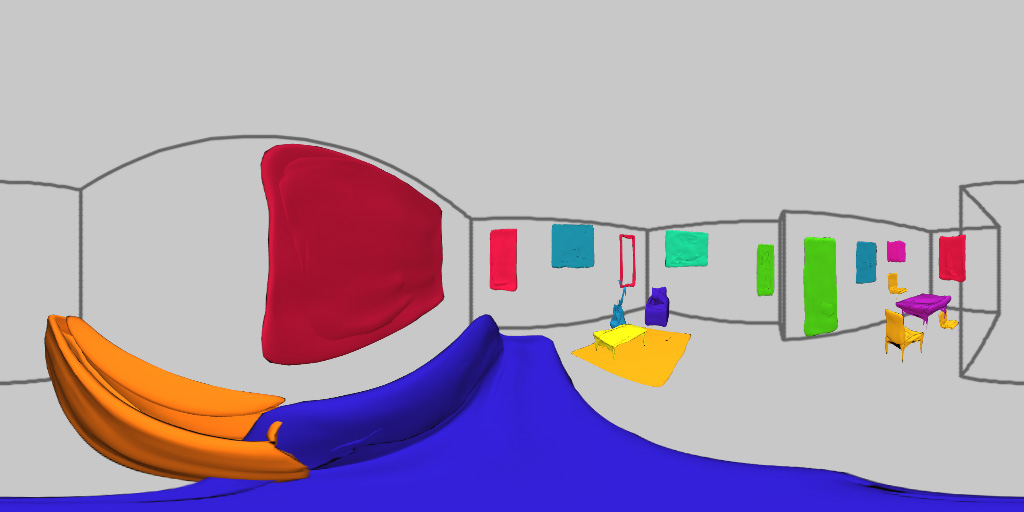}
        	    & \rgb{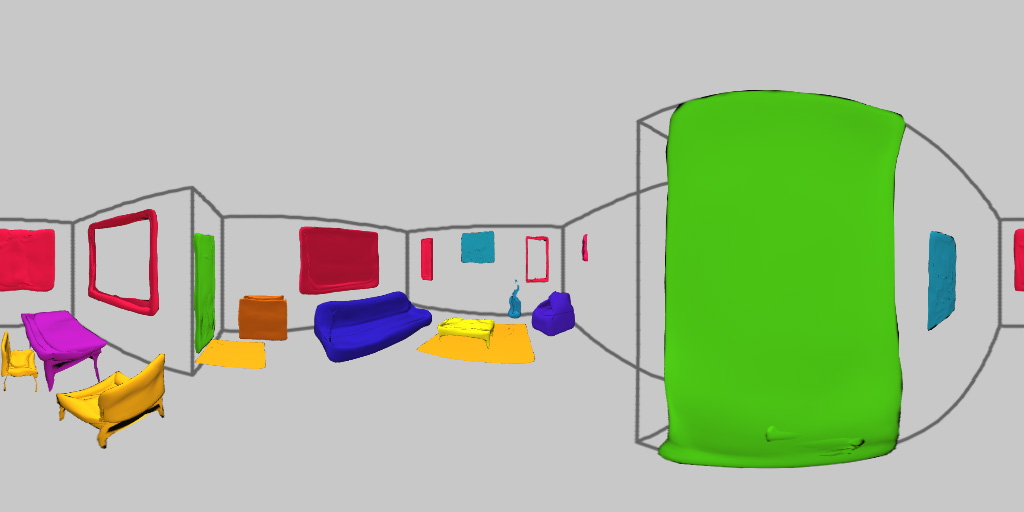}
        	    & \rgb{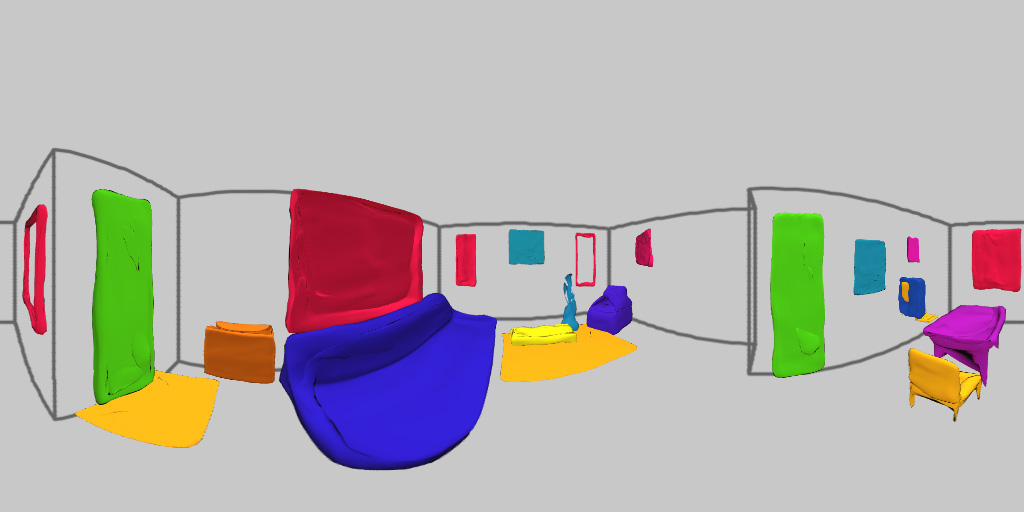}
        	    & \rgb{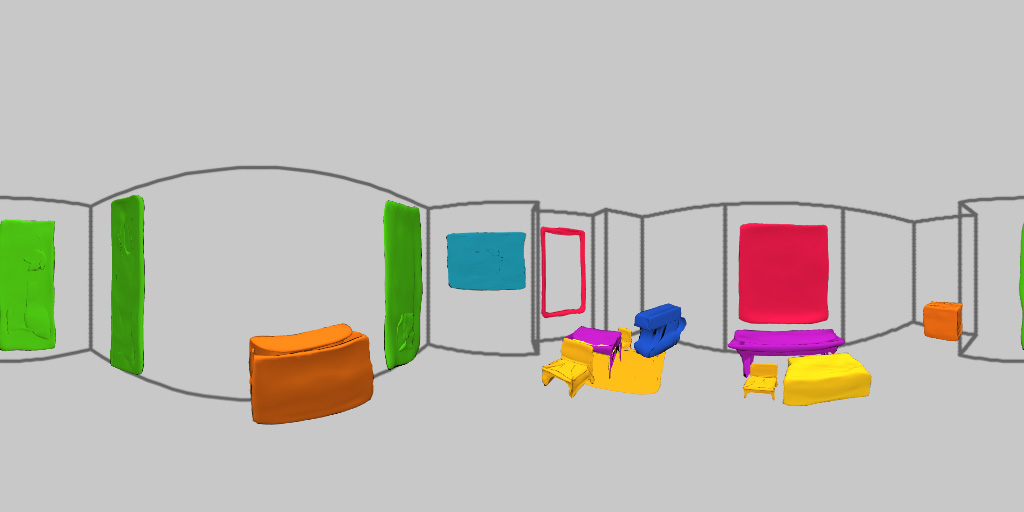}
        	    & \rgb{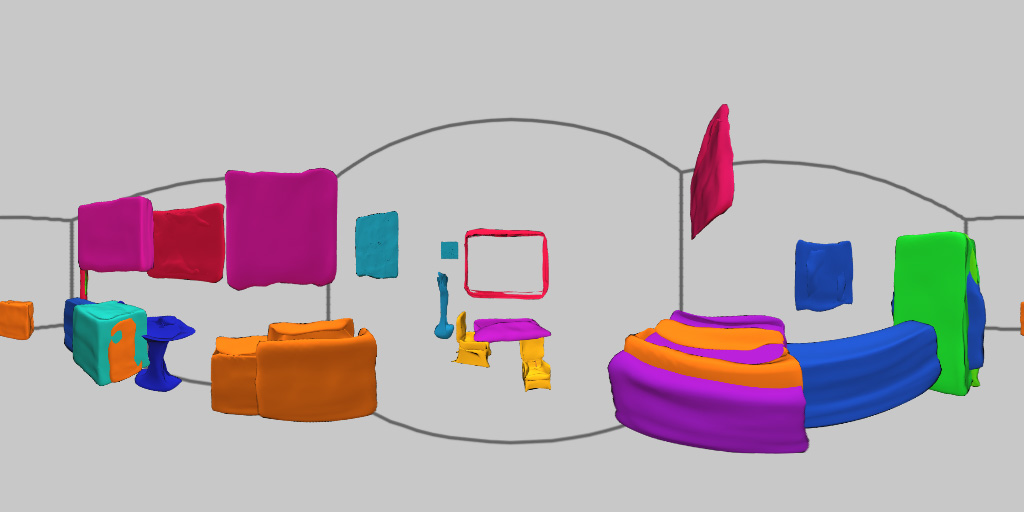}
        	    \\
        	&\rot{Im3D}
                & \rgb{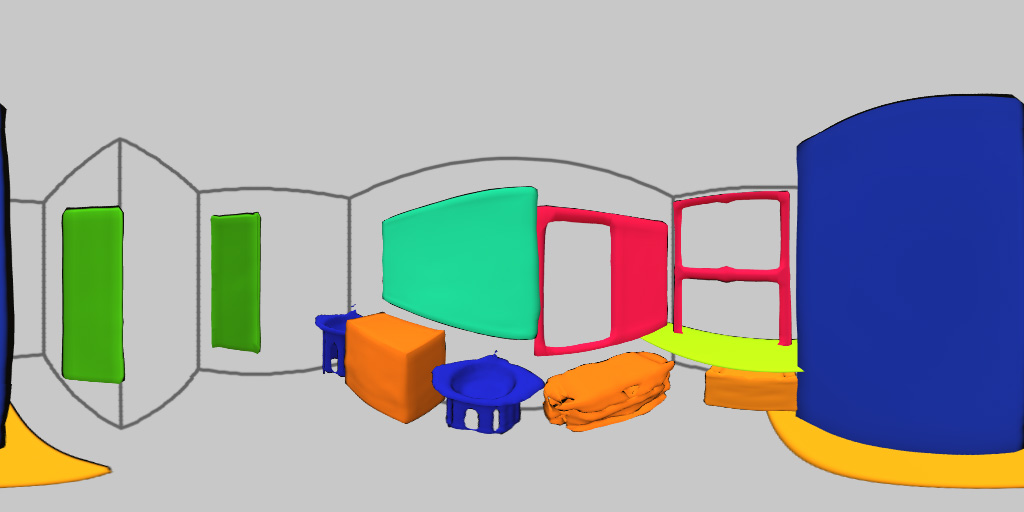}
        	    & \rgb{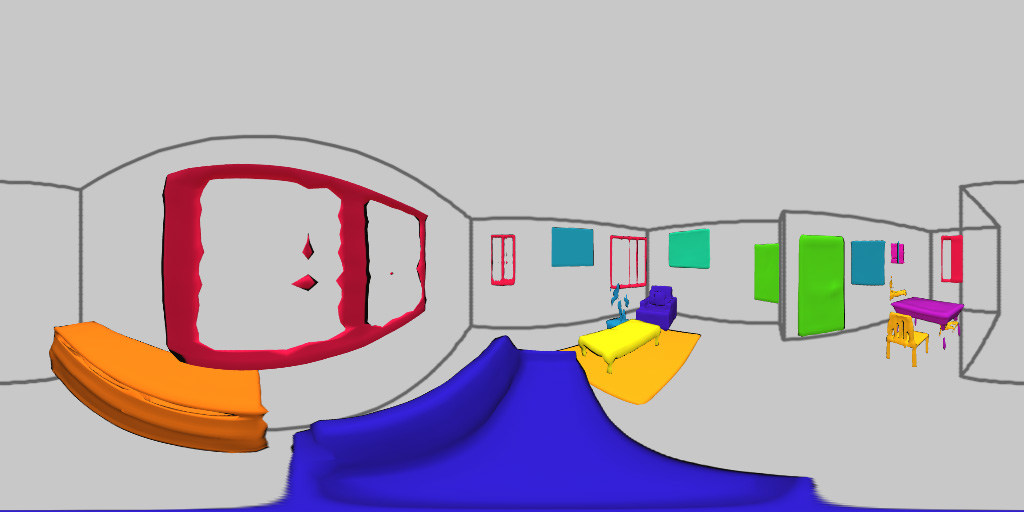}
        	    & \rgb{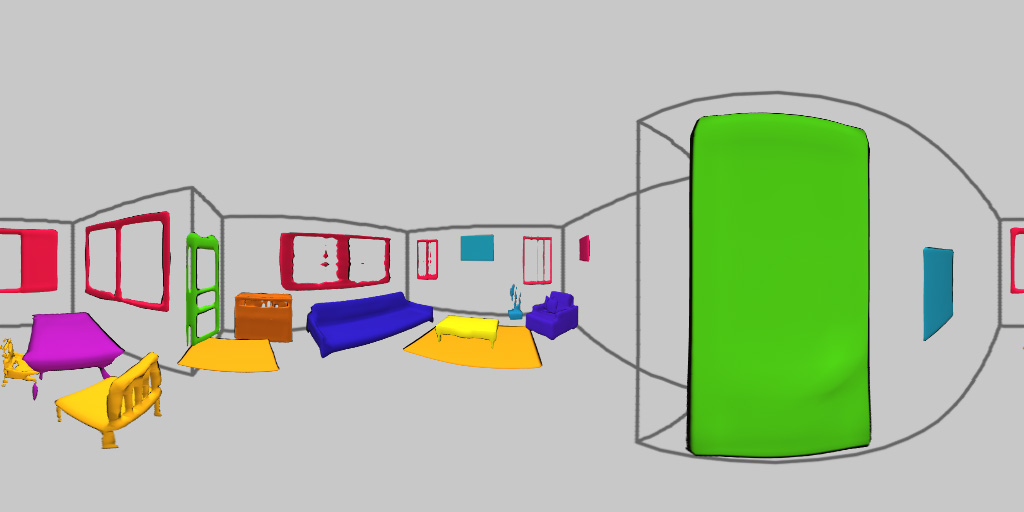}
        	    & \rgb{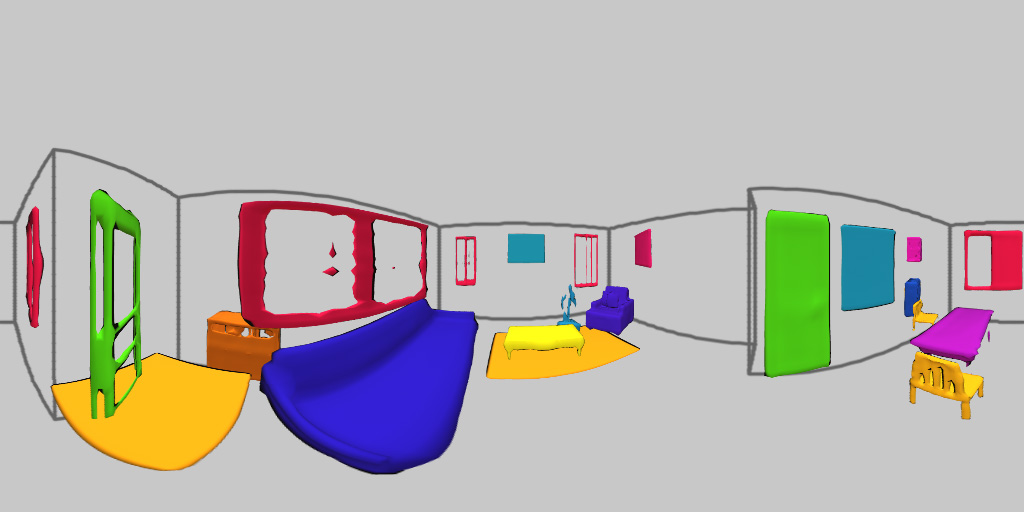}
        	    & \rgb{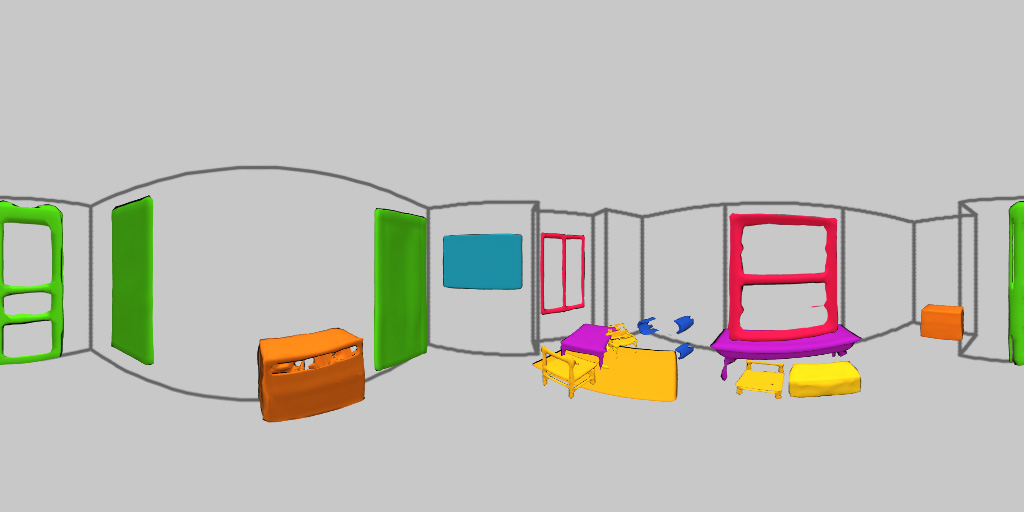}
        	    & \rgb{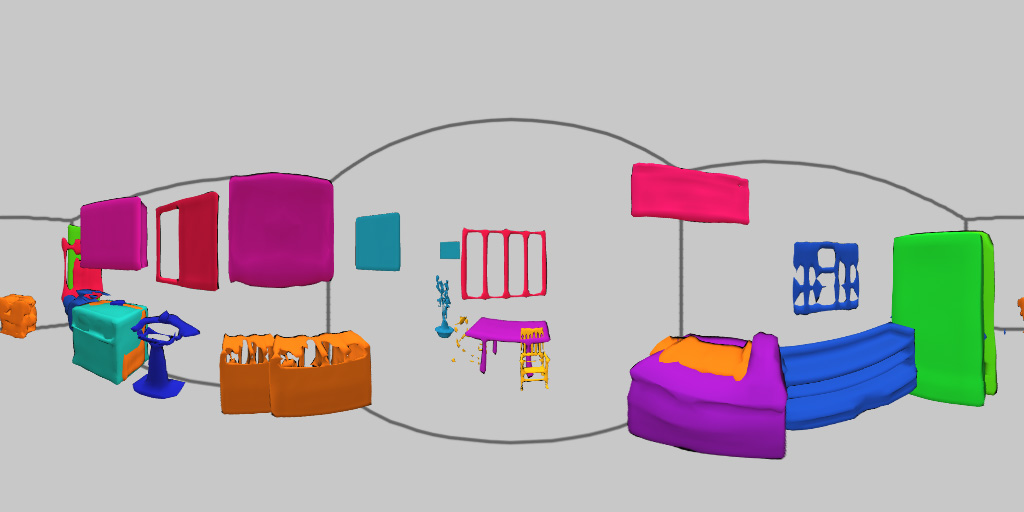}
        	    \\
            &\rot{Ours}
                & \rgb{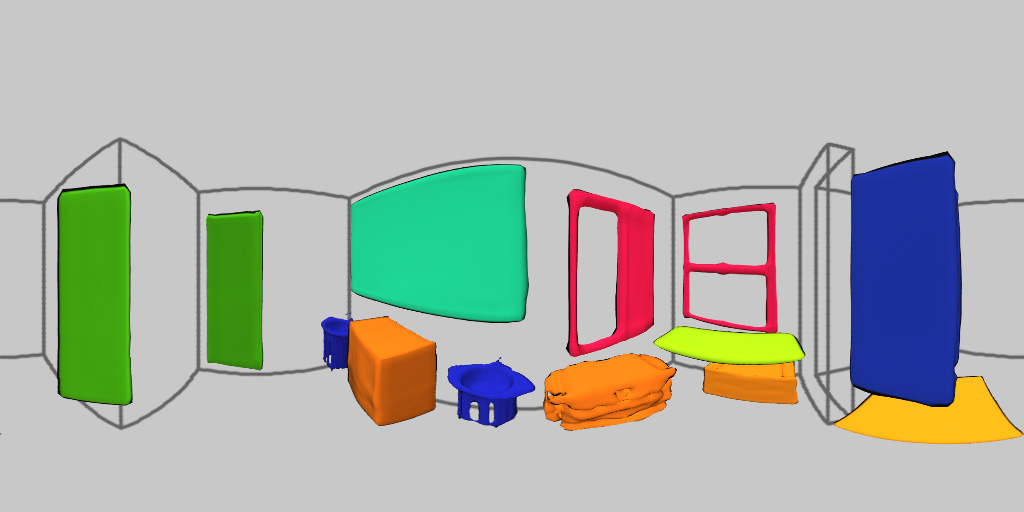}
        	    & \rgb{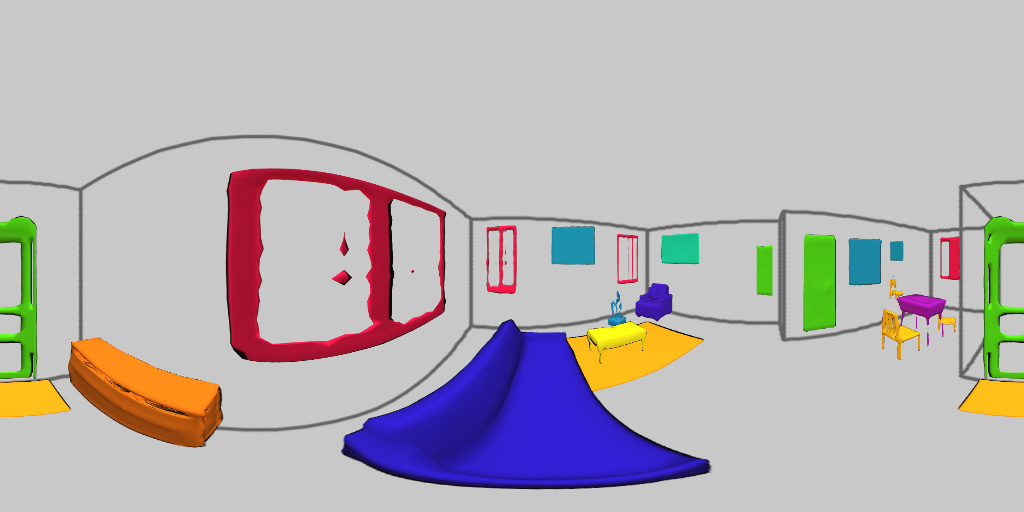}
        	    & \rgb{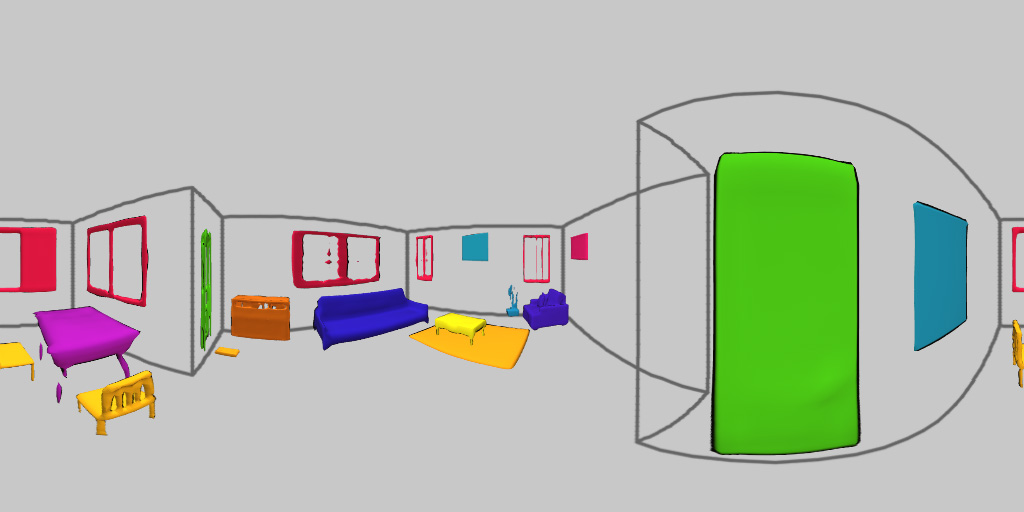}
        	    & \rgb{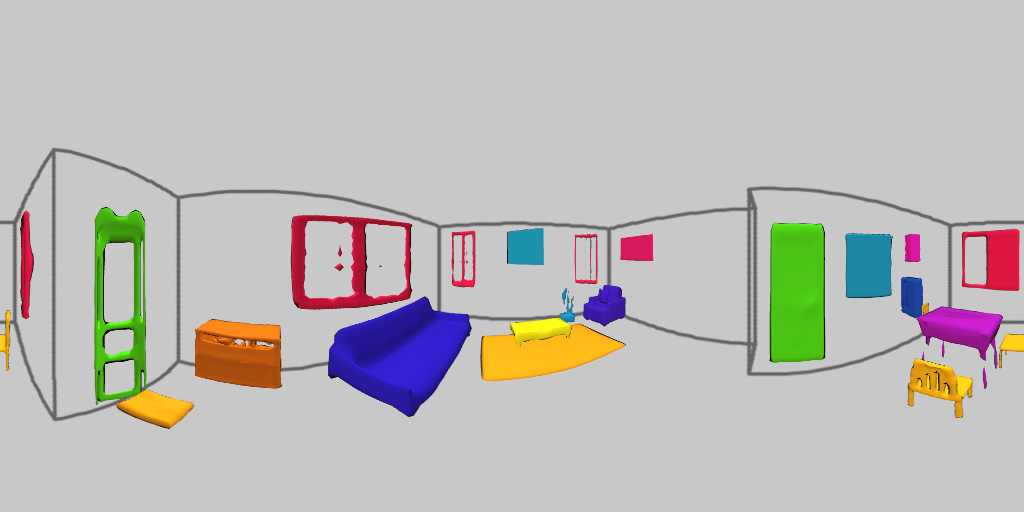}
        	    & \rgb{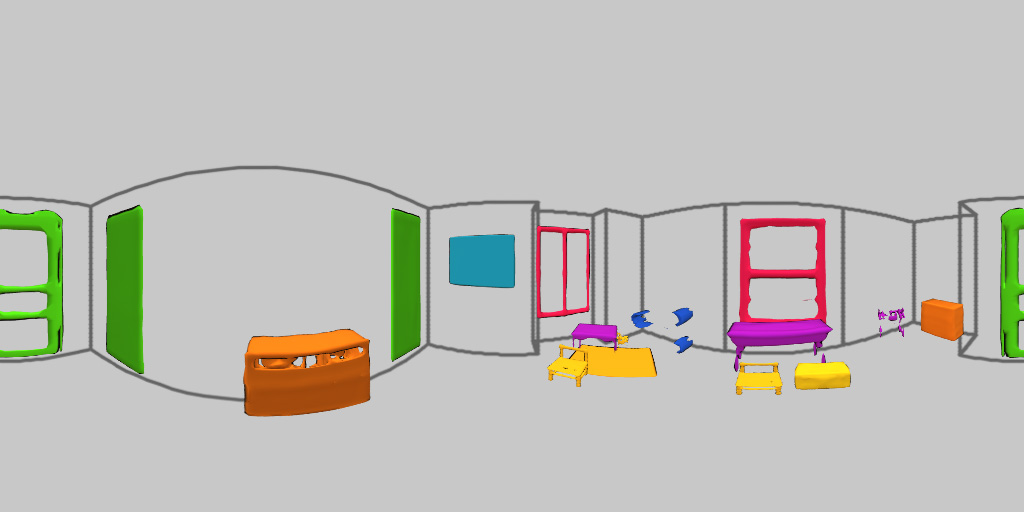}
        	    & \rgb{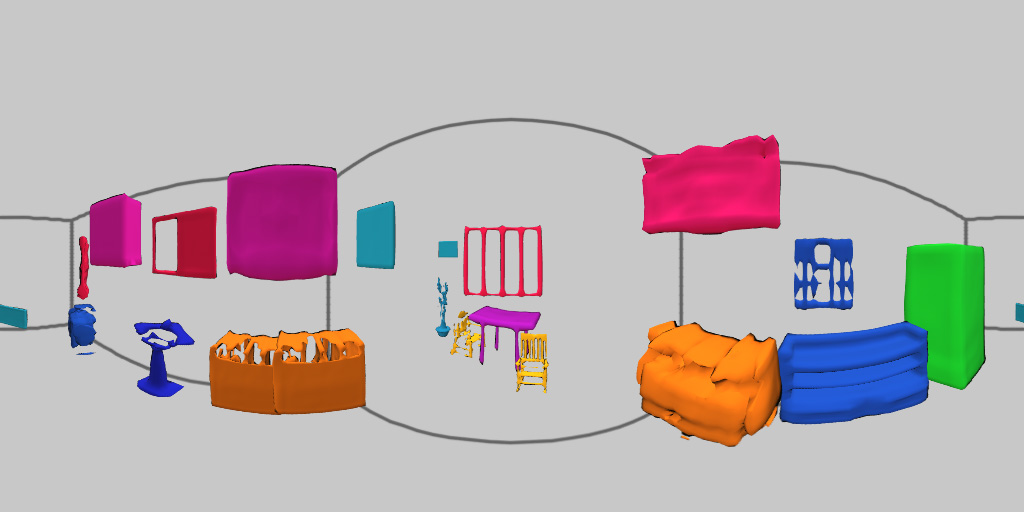}
        	    \\
        	&\rot{GT}
                & \rgb{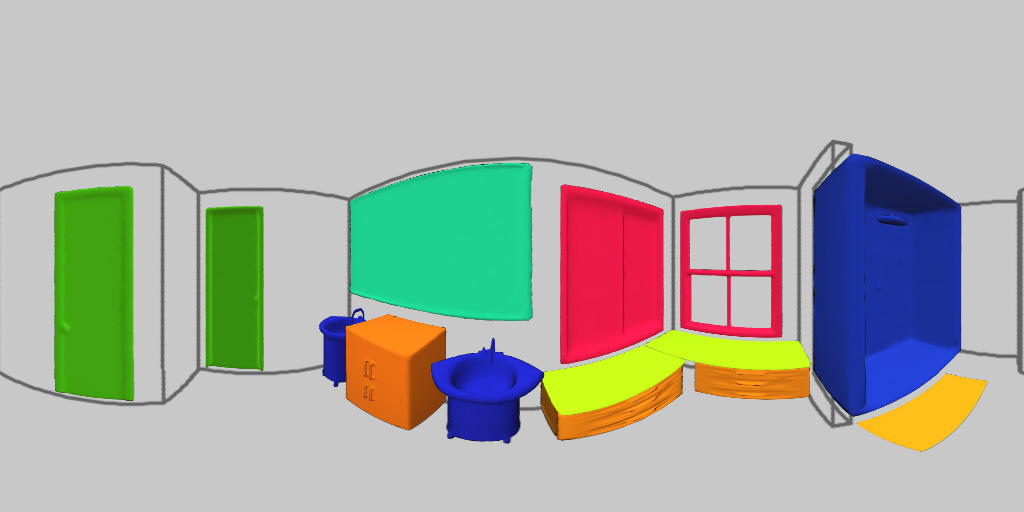}
        	    & \rgb{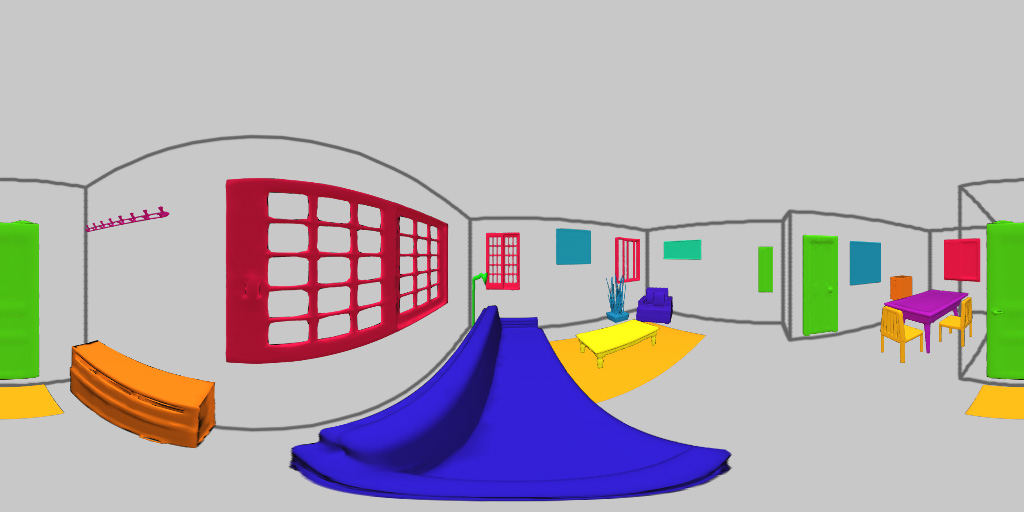}
        	    & \rgb{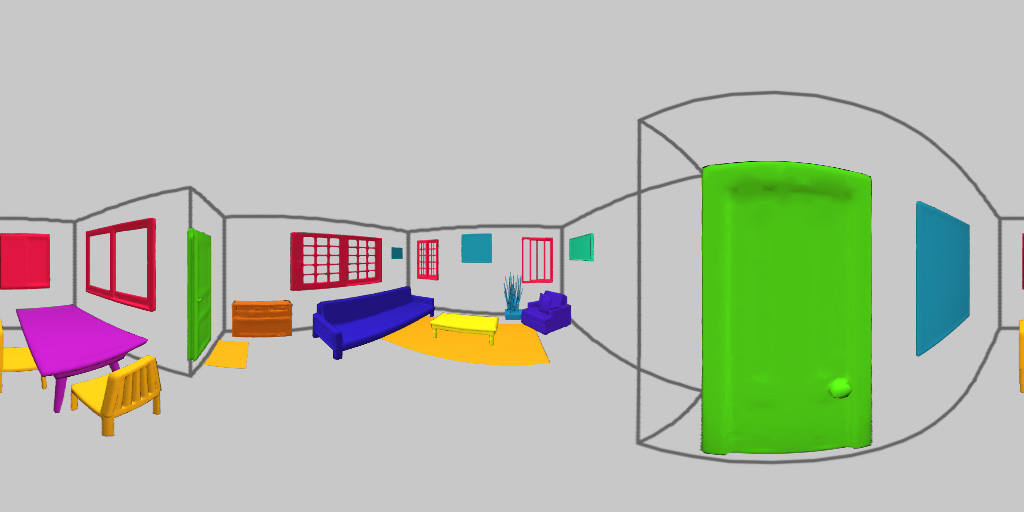}
        	    & \rgb{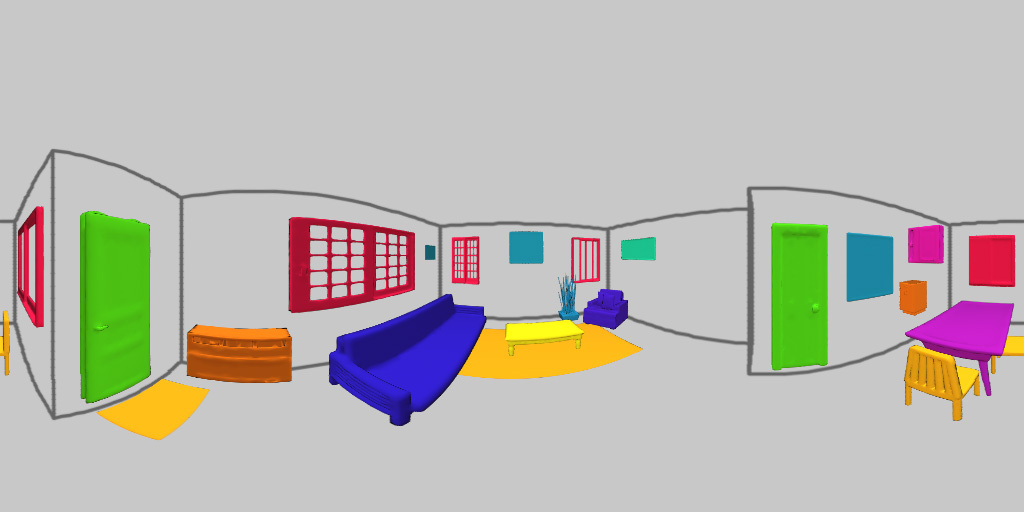}
        	    & \rgb{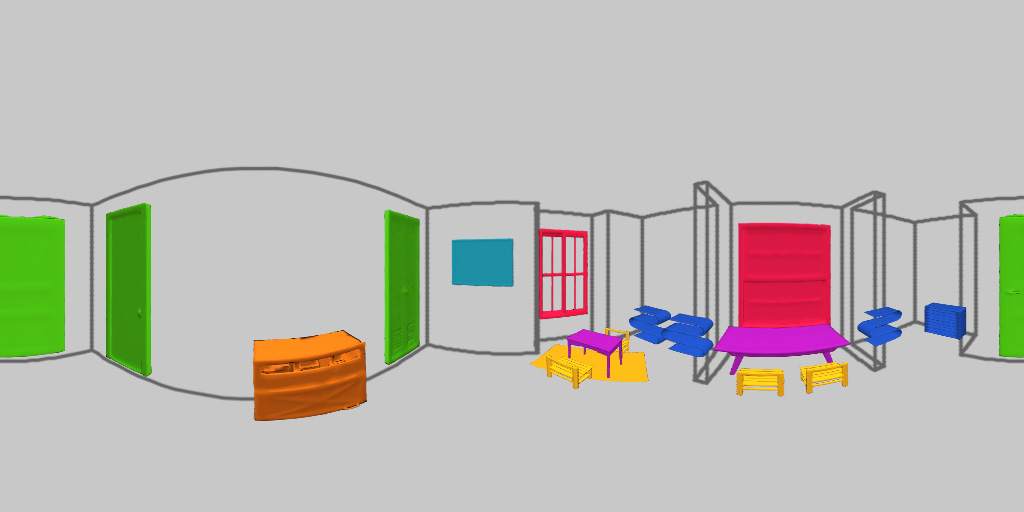}
        	    & \rgb{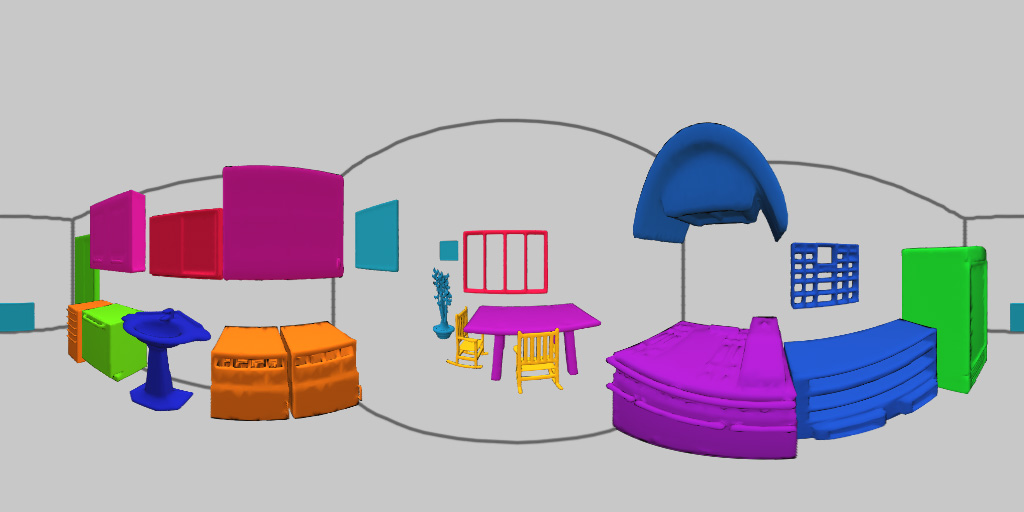}
        	    \vspace{2px}
        	    \\
        \multicolumn{2}{c}{}& (a) & (b) & (c) & (d) & (e) & (f) \\
    \end{tabular}
    	\caption{More Qualitative comparison on 3D object detection and scene reconstruction.}
	\label{fig:scnrecon_supp}
\end{figure*}

In Sec. 4.1 of the paper, we show qualitative comparisons on 3D detection and reconstruction. 
Here we provide more results in \autoref{fig:scnrecon_supp}.
Compared to the SoTA methods \cite{nie2020total3dunderstanding, zhang2021holistic}, our method produces significantly better 3D detection and reconstruction results.
From the 3D detection and reconstruction results in panorama view, we observe that our method generates more accurate projections of reconstructed objects
(\eg, the mirror of (a), the sofa of (b) and (d), the door of (c)).
From the 3D detection results in Bird's Eye View, we can see that our method generates more reasonable and physically plausible object poses
(\eg, (c), (e) have less object-wall collision and better rotation relations with walls).

\section{\yz{Would RO Improve Other Methods?}}
\zc{In order to further evaluate the proposed relation optimization, we apply our RO on Total3D and Im3D using our predicted relation and their final results, and show the results in \autoref{tbl:ro_improve}.
We can see that both methods still significantly benefit from the RO, which
demonstrates that our RO is effective and robust to different initial estimates.}

\begin{table*}[!ht]
    	\begin{center}
	    \resizebox{1.8\columnwidth}{!}{
    		\begin{tabular}{|l|c|c|c|c|c|c|c|c|c|}
    			\hline
    			Method (Pano)   & door & picture & table & sofa & chair & window & bed & bottom cabinet & chest \\
    			\hline
    			Total3D         & 28.65 & 0.06 & 38.83 & 31.64 & 23.71 & 4.78 & 74.09 & 37.08 & 62.07 \\     			Im3D            & 37.59 & 0.14 & \textbf{49.47} & \textbf{37.24} & 29.34 & 6.35 & 77.66 & 45.18 & 70.03 \\     			Ours (w/o. RO)  & 54.74 & 0.69 & 48.39 & 36.05 & 29.85 & \textbf{13.49} & 81.13 & 48.33 & 72.08 \\     			Ours (Full)     & \textbf{57.73} & \textbf{1.24} & 49.10 & 37.02 & \textbf{29.95} & 12.28 & \textbf{81.15} & \textbf{48.76} & \textbf{74.26} \\     			\hline\hline
    			Method (Pano)   & sink & fridge & bathtub & shelf & mirror & toilet & counter & standing tv & mean \\
    			\hline
    			Total3D         & 28.24 & 68.82 & 69.36 & 10.36 & 0.04 & 19.88 & 19.17 & 2.12 & 30.52 \\
    			Im3D            & \textbf{28.57} & 71.39 & \textbf{73.93} & 9.78 & 0.92 & 15.04 & 19.17 & 2.52 & 33.78 \\
    			Ours (w/o. RO)  & 27.43 & \textbf{73.35} & \textbf{73.93} & \textbf{15.84} & 1.47 & 32.87 & 19.17 & 9.61 & 37.55\\
    			Ours (Full)     & 27.93 & \textbf{73.35} & \textbf{73.93} & 15.76 & \textbf{3.19} & \textbf{65.54} & 19.17 & \textbf{13.20} & \textbf{40.21} \\
    			\hline
    		\end{tabular}
        }    	
	\end{center}
 	\vspace{-1em}
	\caption{3D object detection comparison on Structured3D. We evaluate on the 17 iGibson categories mapped from 20 Structured3D categories and use mean average precision (mAP) with the threshold of 3D bounding box IoU set at 0.15 as the evaluation metric.}
	\label{tbl:structured3d_det}
\end{table*}

\begin{figure}[t]
    	\centering
	\begin{subfigure}[t]{0.15\textwidth}
		\includegraphics[width=\textwidth]  
		{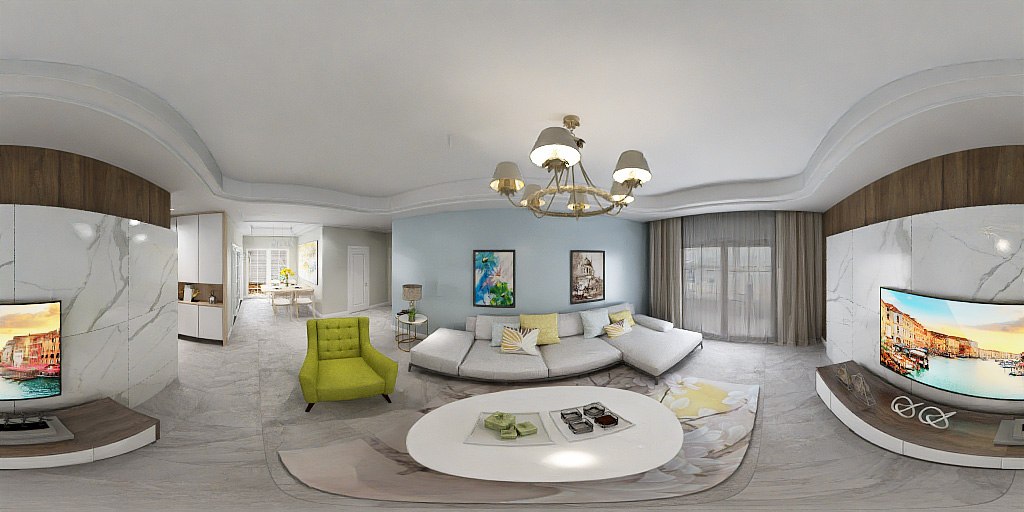}
		\includegraphics[width=\textwidth]  
		{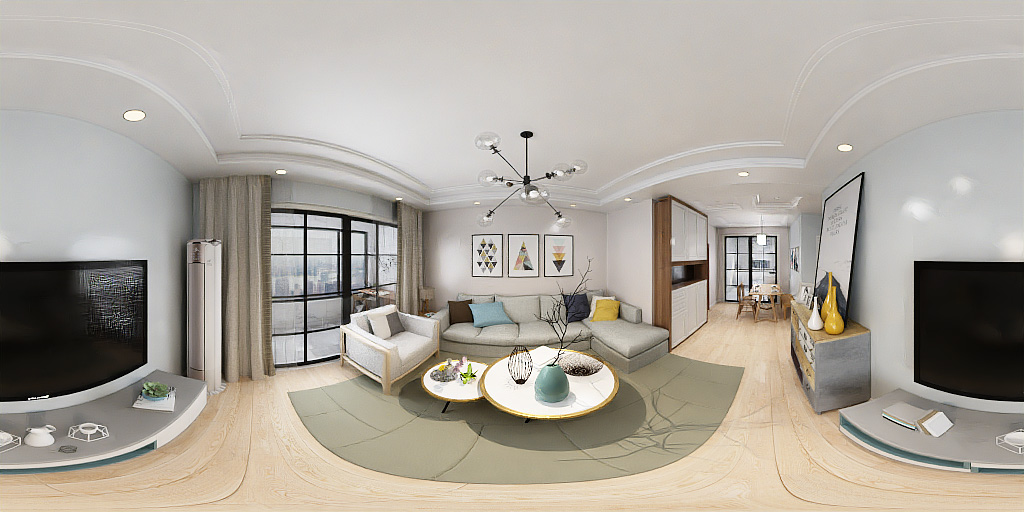}
		\caption{Input}
	\end{subfigure}
	\begin{subfigure}[t]{0.15\textwidth}
		\includegraphics[width=\textwidth]  
		{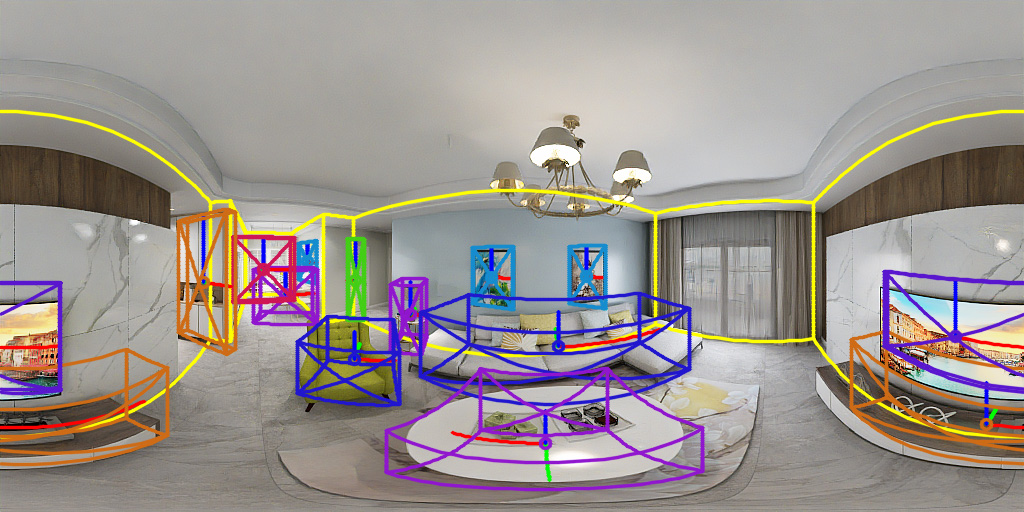}
		\includegraphics[width=\textwidth]  
		{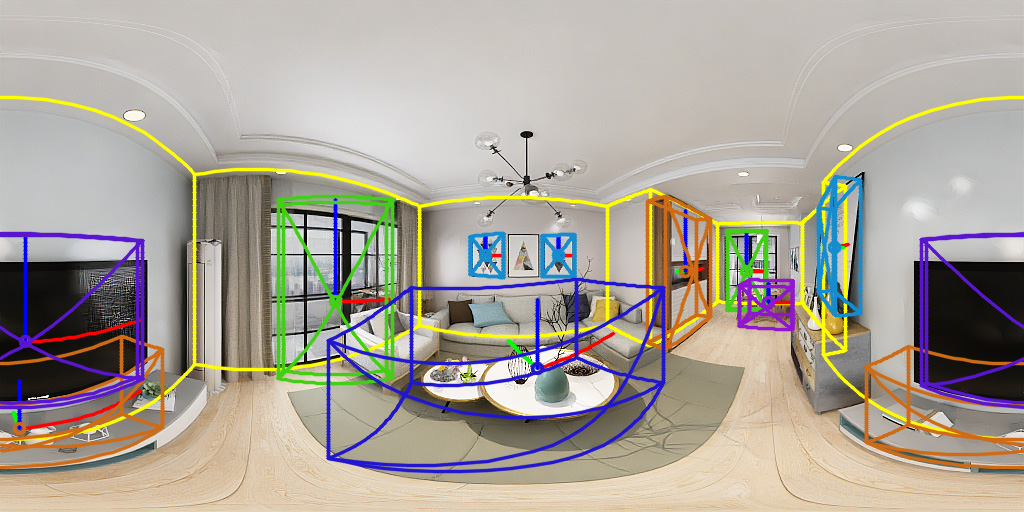}
		\caption{3D Detection}
	\end{subfigure}
	\begin{subfigure}[t]{0.15\textwidth}
		\includegraphics[width=\textwidth]
		{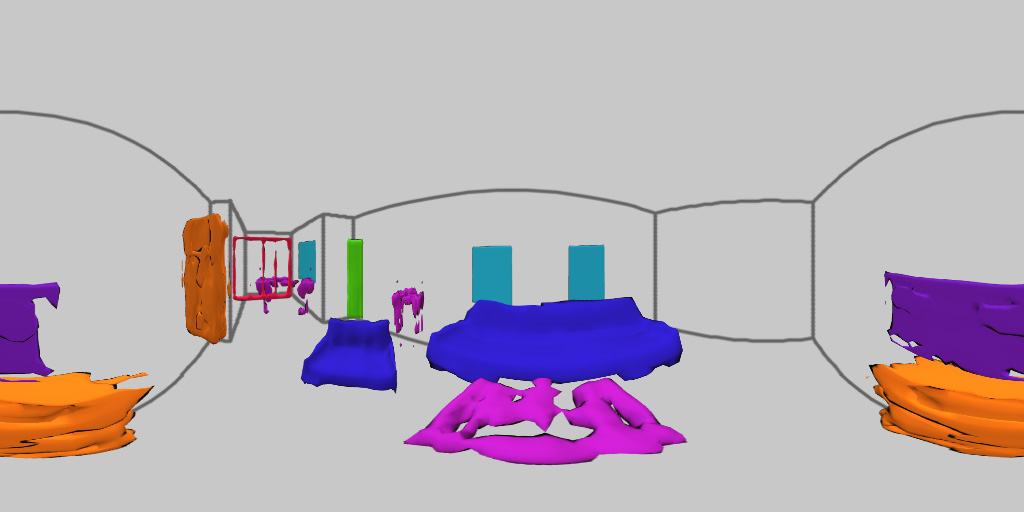}
		\includegraphics[width=\textwidth]
		{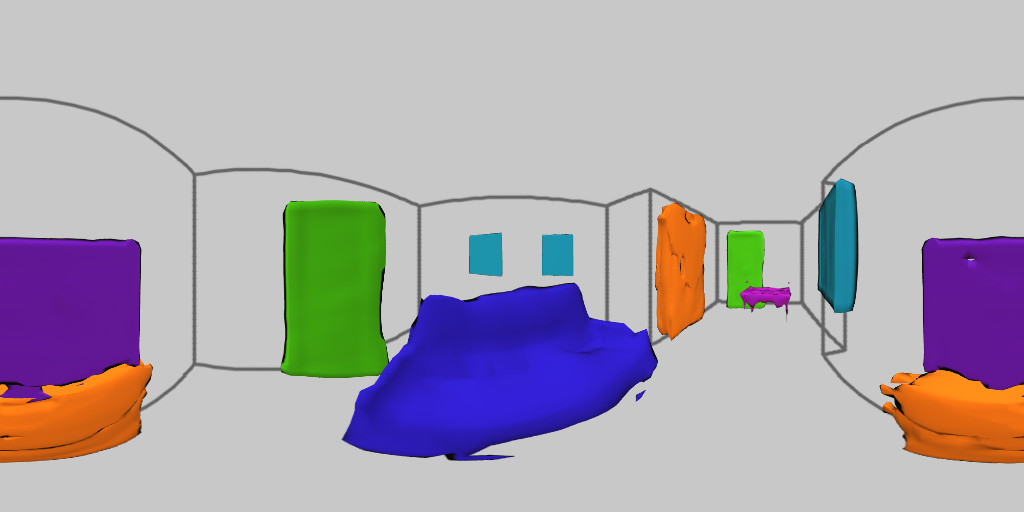}
		\caption{Reconstruction}
	\end{subfigure}
	\vspace{-0.5em}
	\caption{Qualitative results of our model on Structured3D.}
	\label{fig:structured3d_quali}
	\vspace{1em}
\end{figure}

\section{\yz{Run-time} Efficiency}
The efficiency comparison is shown in \autoref{tbl:eff_comp}. 
It is worth mentioning that implicit representation LDIF and RO are all implemented with PyTorch, and can be further optimized, \eg, using CUDA, to improve the efficiency.

\section{Experiment on Structured3D}

Since Structured3D provides the ground truth of object pose and layout, we can train our model up to \yz{R}GCN.
Due to the lack of mesh ground truth, 
we load the object reconstruction model with weights trained on iGibson.
Furthermore, since the object reconstruction model requires category label as input, we map the object categories from Structured3D to iGibson.
\yz{We found overlapping categories between two datasets, which ends up with 20 structure3D categories mapped to 17 iGibson categories. 
Specifically, ``cabinet'', ``bookshelf'', ``desk'', ``shelves'', ``dresser'', ``floor mat'', ``television'', ``box'', ``nightstand'' in Structure3D are mapped to ``bottom cabinet'', ``shelf'', ``table'', ``shelf'', ``bottom cabinet'', ``carpet'', ``standing tv'', ``chest'', ``chest'' in iGibson, 
and others are mapped with the same category name.
It is also worth mentioning that the bounding box GT of objects in Structured3D is not accurate or physical plausible, which makes it difficult to produce rich relation GT and to better refine the object poses with observation and collision terms.
So the weights of relation optimization terms need to be tuned to match the condition.
\zc{Specifically, we fix the weights of relation terms and auto-search the learning rate of gradient descend optimizer and other weights of relation optimization terms around the original settings used on iGibson.
In summary, we train object detection on overlapping categories and set weights of RO terms with auto-search \cite{wandb}.}
The auto-searched weights are shown in \autoref{tbl:ro_weights_struct}.}
\yz{Qualitative results are shown in \autoref{fig:structured3d_quali}.
We can see that our method performs well with good layout, pose and shape estimation although there is no ground truth for shapes. 
We compare 3D object detection against existing methods quantitatively in \autoref{tbl:structured3d_det}.
The results show that 
our method still outperforms SoTA methods significantly, and RO plays a big role in improving the mAP.}

\begin{table}[t]
    	\begin{center}
	    \resizebox{0.8\columnwidth}{!}{
    		\begin{tabular}{|l|c|l|c|}
    			\hline
    			Parameter & Value & Parameter & Value \\
    			\hline
    			${\lambda}^{rd}$ & 0.0040 & ${\lambda}^{oc}$ & 0.0157\\
    			${\lambda}^{d}$ & 0.1404 & ${\lambda}^{wc}$ & 0.2625\\
    			${\lambda}^{s}$ & 6.0502 & ${\lambda}^{fc}$ & 0.3182\\
    			${\lambda}^{\theta}$ & 0.0003 & ${\lambda}^{cc}$ & 0.2036\\
    			${\lambda}^{bp}$ & 0.2895 & learning rate & 0.0124\\
    			\hline
    		\end{tabular}
		}
	\end{center}
	\vspace{-1.0em}
	\caption{Auto-searched hyperparameters used on Structured3D, including weights of relation optimization terms and learning rate of relation optimization.}
	\vspace{1.2em}
	\label{tbl:ro_weights_struct}
\end{table}

\section{Failure Cases}
\begin{figure*}[!ht]
	\centering
	\scriptsize
	\newcommand{\rgb}[1]{\raisebox{-0.5\height}{\includegraphics[width=.147\linewidth]{#1}}}
	\newcommand{\bird}[1]{\raisebox{-0.5\height}{\includegraphics[width=.147\linewidth,clip,trim=90 50 90 45]{#1}}}
	\newcommand{\rot}[1]{\rotatebox[origin=c]{90}{#1}}
	\def\arraystretch{0.5}	\begin{tabular}{c|c*{6}{c@{\hspace{1px}}}}
        \rule{0pt}{5px}&\rot{Input}
                        & \rgb{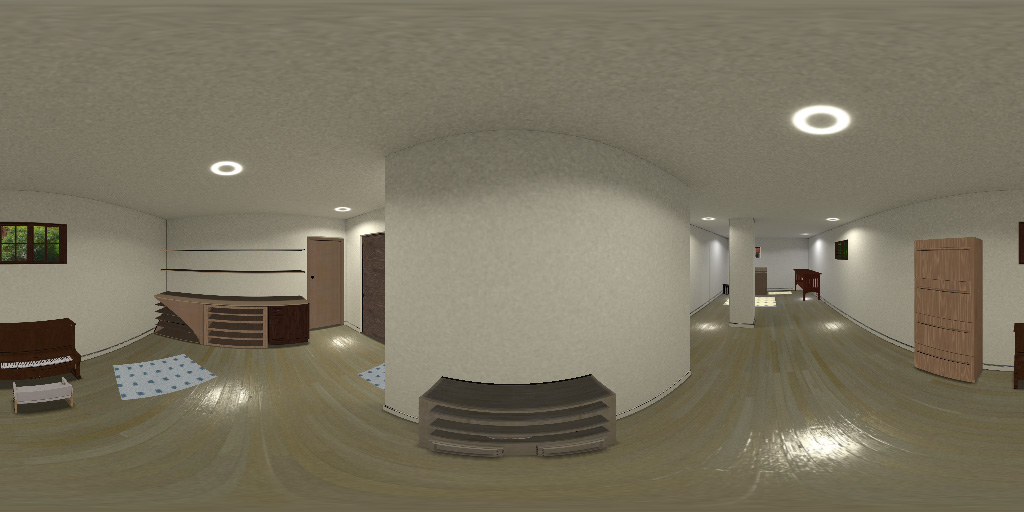}
    	    & \rgb{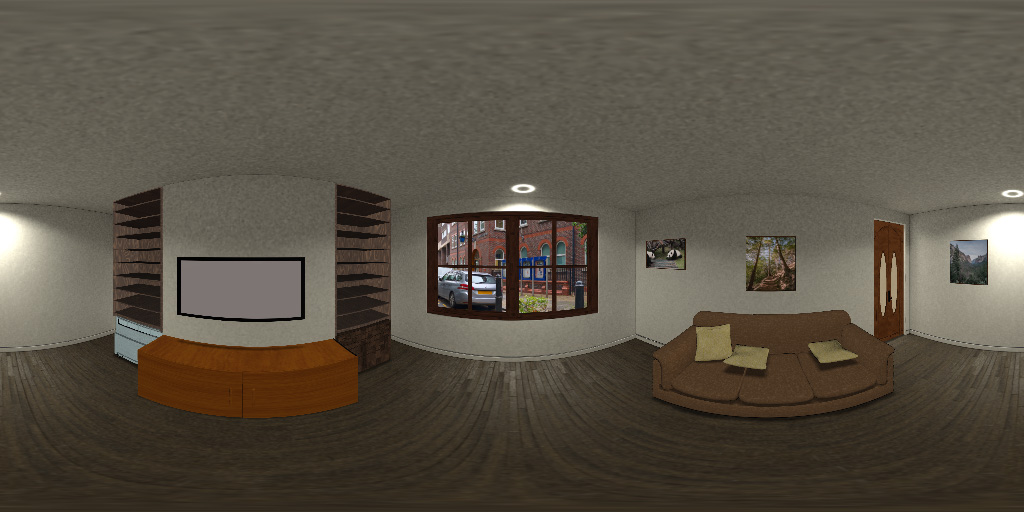}
    	    & \rgb{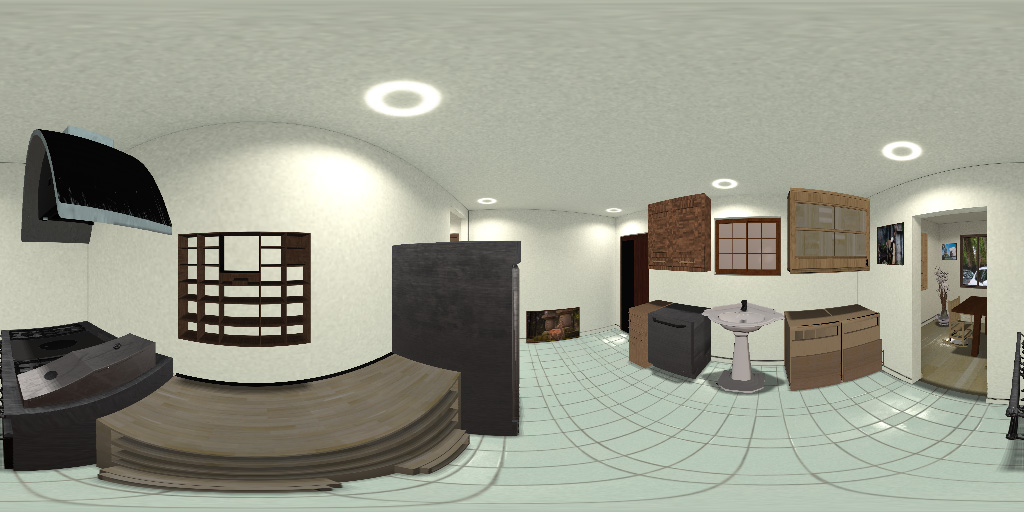}
    	    & \rgb{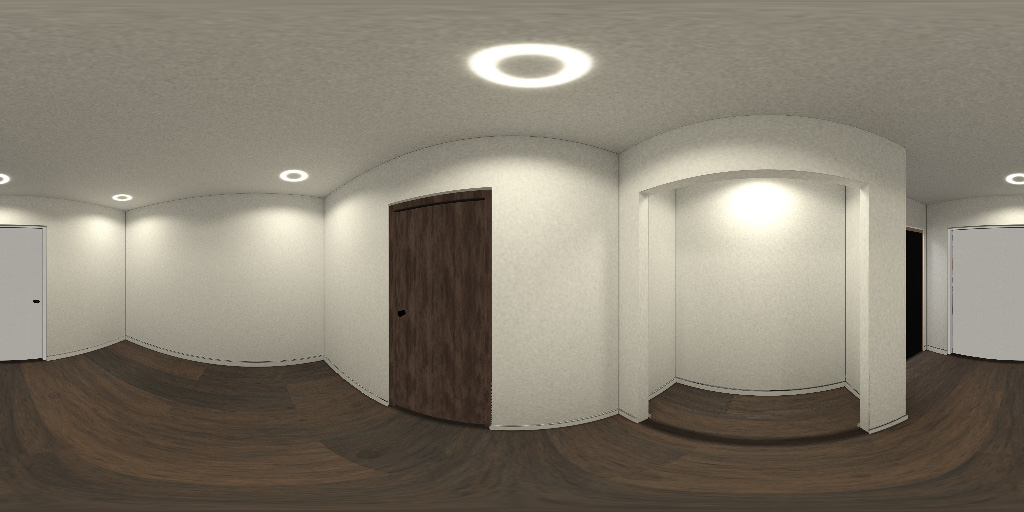}
    	    & \rgb{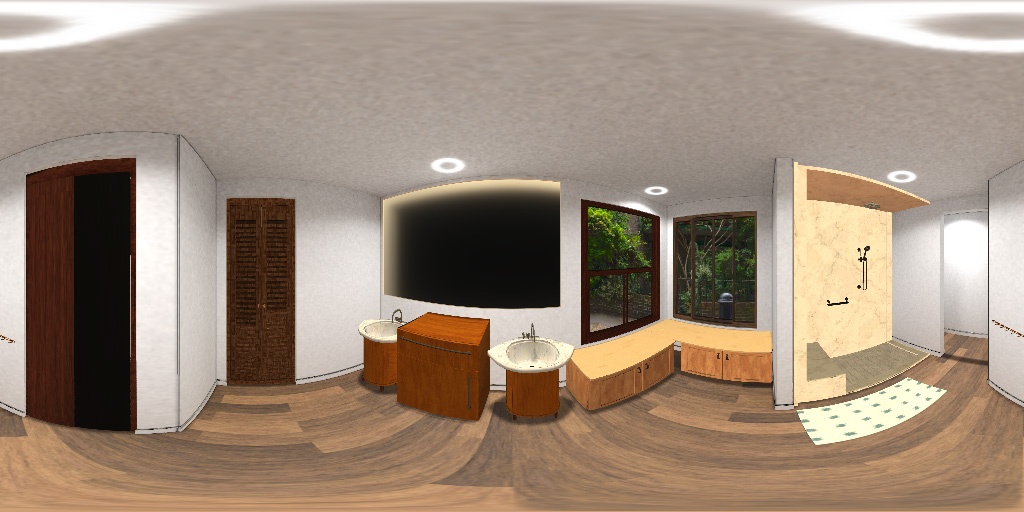}
    	       	       	    \\
    	\hline
    	    \multirow{2}{*}[-7ex]{\rot{Bird's Eye View}}
        	& \rot{Ours}
                & \raisebox{-0.5\height}{\includegraphics[width=.147\linewidth,clip,trim=130 110 120 90]{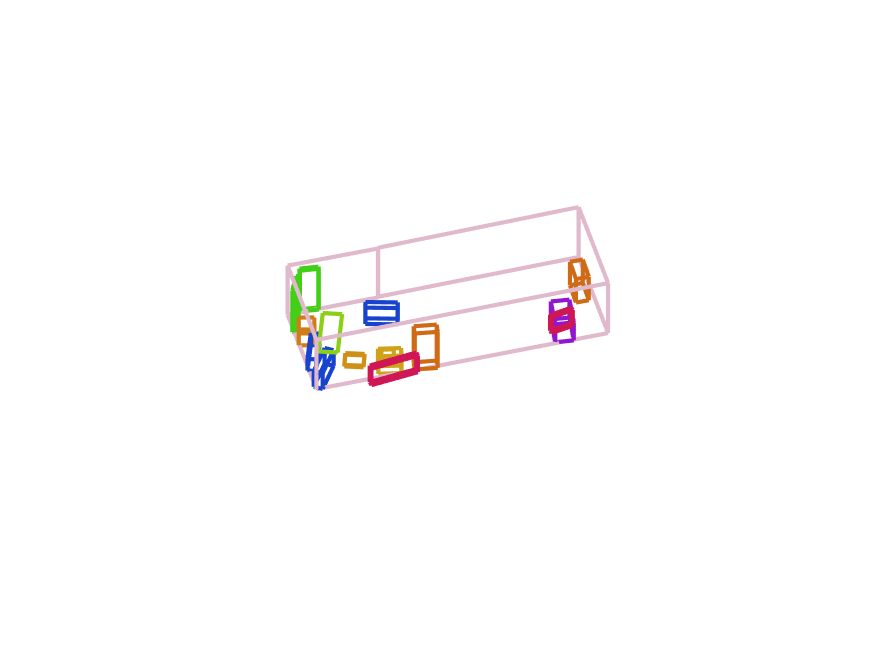}}
        	    & \bird{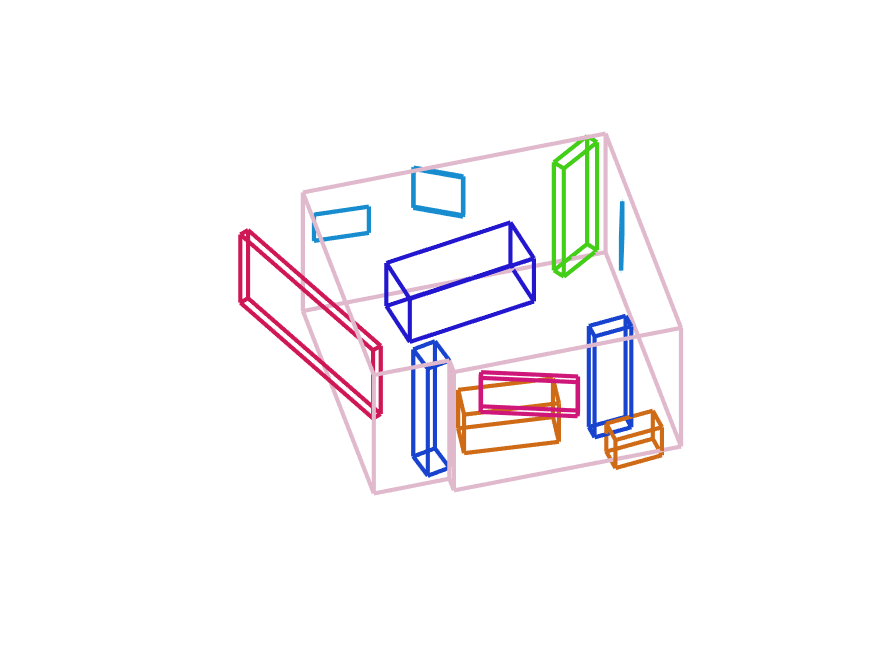}
        	    & \bird{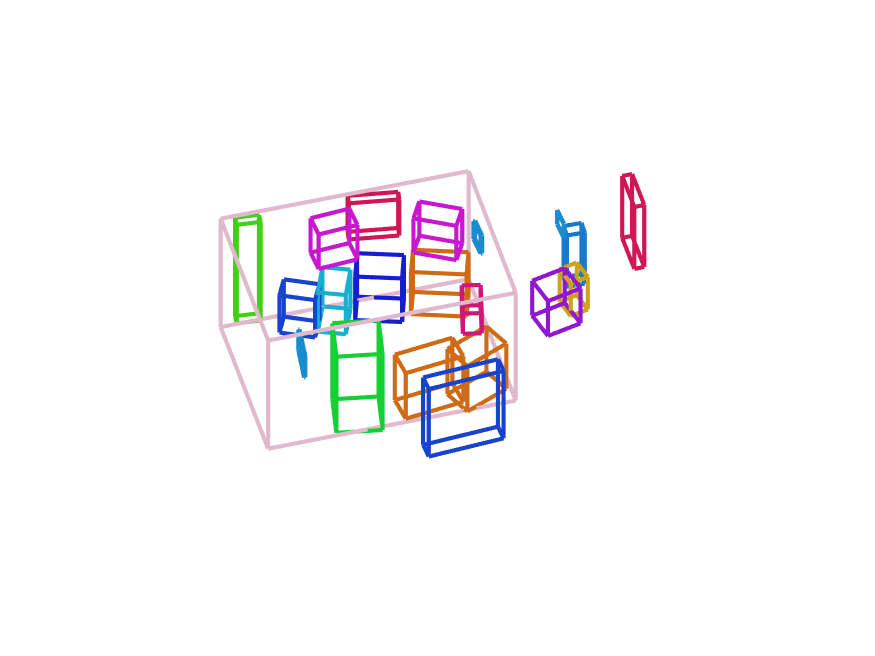}
        	    & \bird{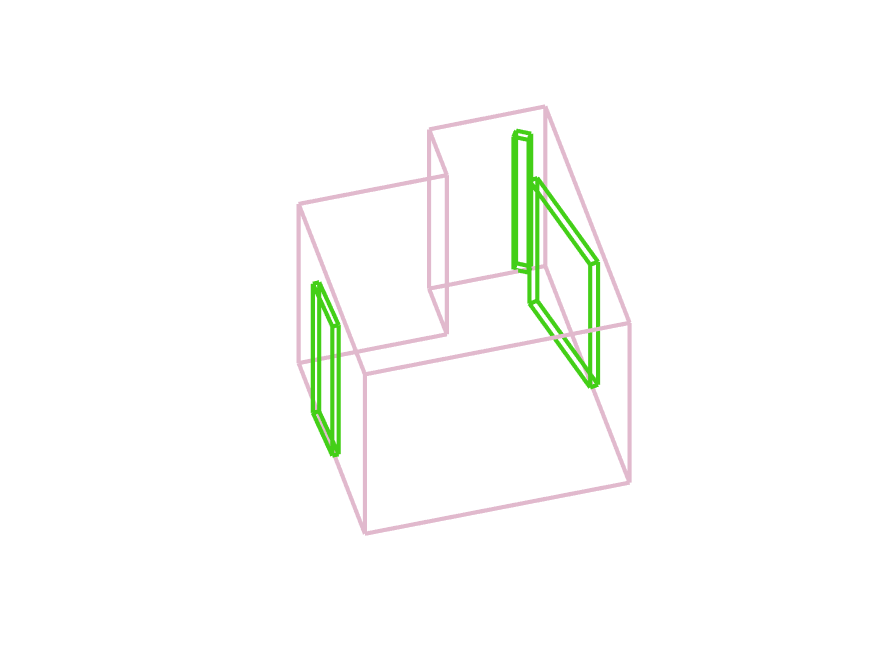}
        	    & \bird{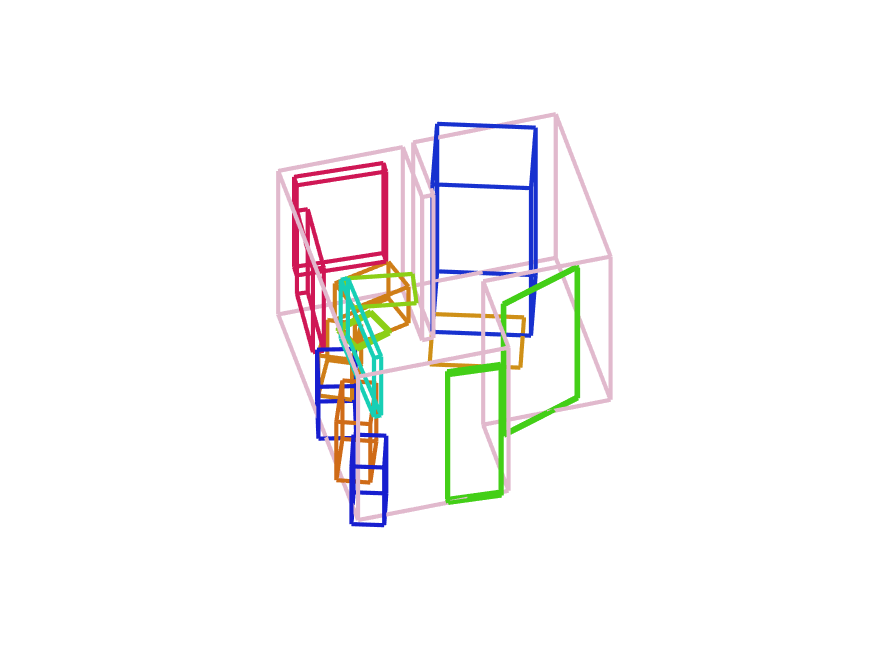}
        	           	    \\
        	&\rot{GT}
                & \raisebox{-0.5\height}{\includegraphics[width=.147\linewidth,clip,trim=130 110 120 90]{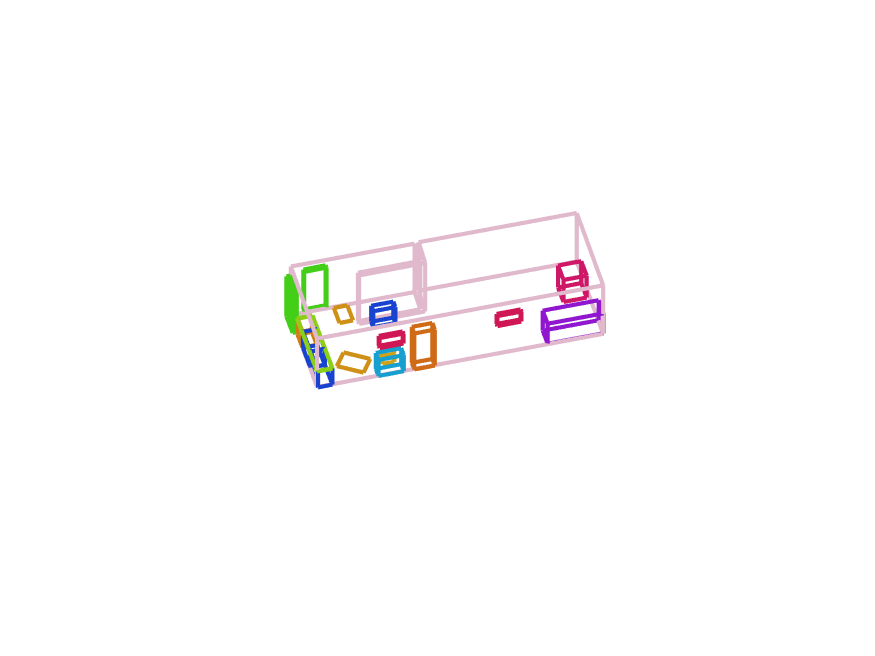}}
        	    & \bird{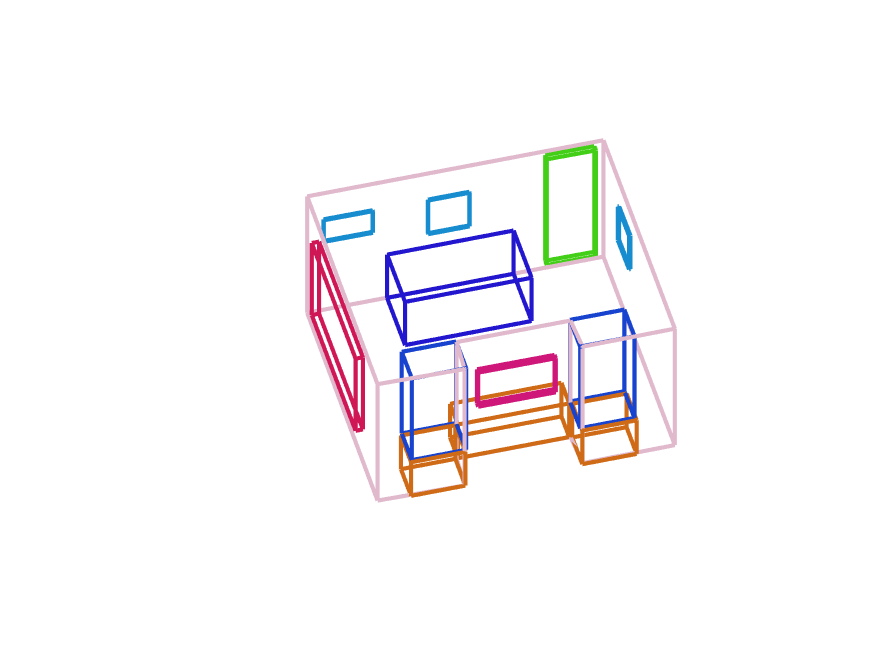}
        	    & \bird{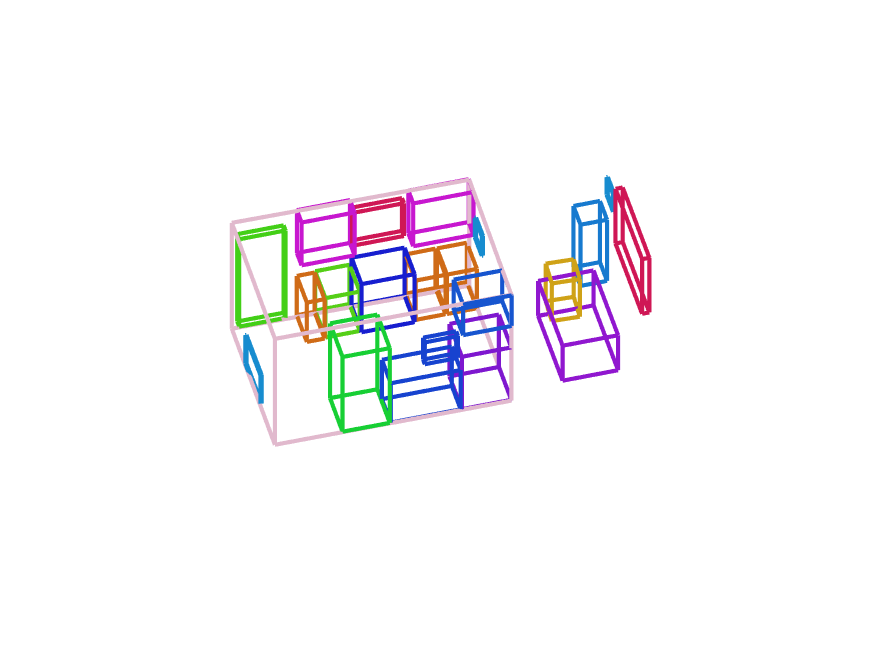}
        	    & \bird{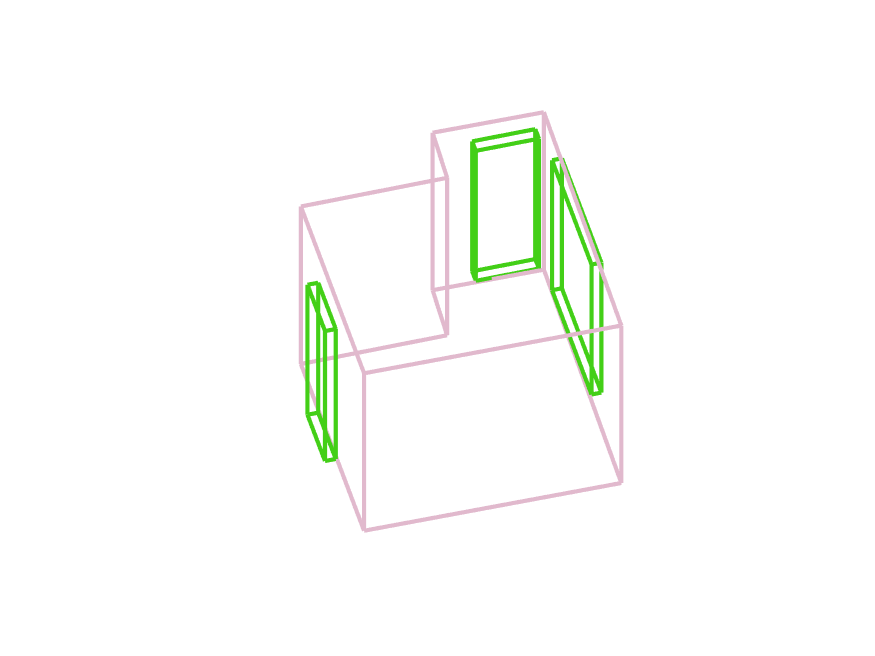}
        	    & \bird{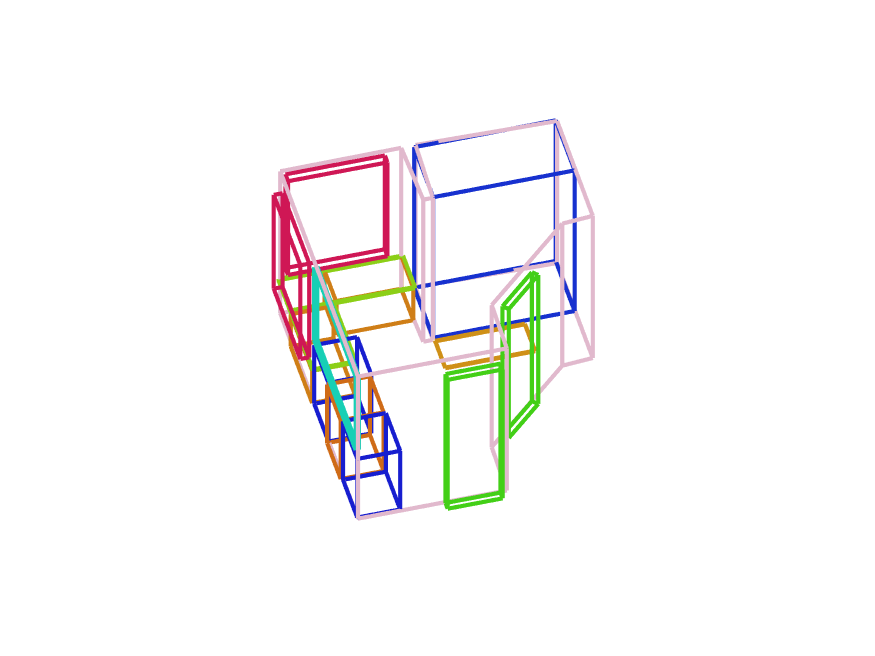}
        	           	    \\
    	\hline
    	    \multirow{2}{*}{\rot{Panorama View}}
            &\rot{Ours}
                & \rgb{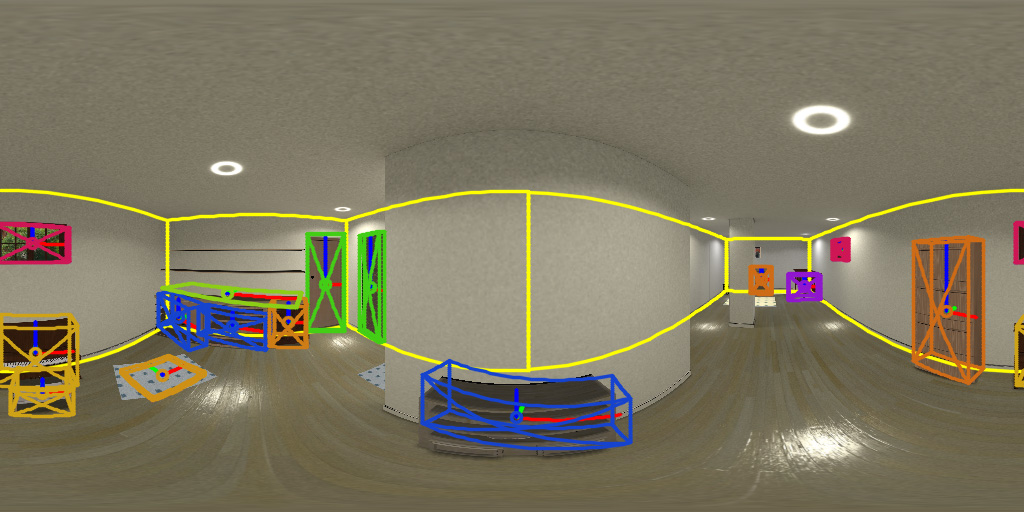}
        	    & \rgb{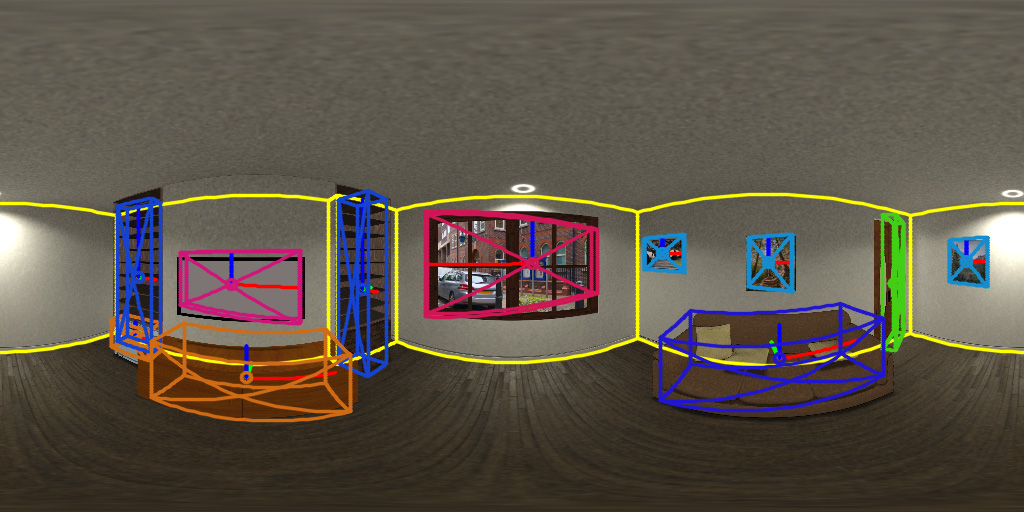}
        	    & \rgb{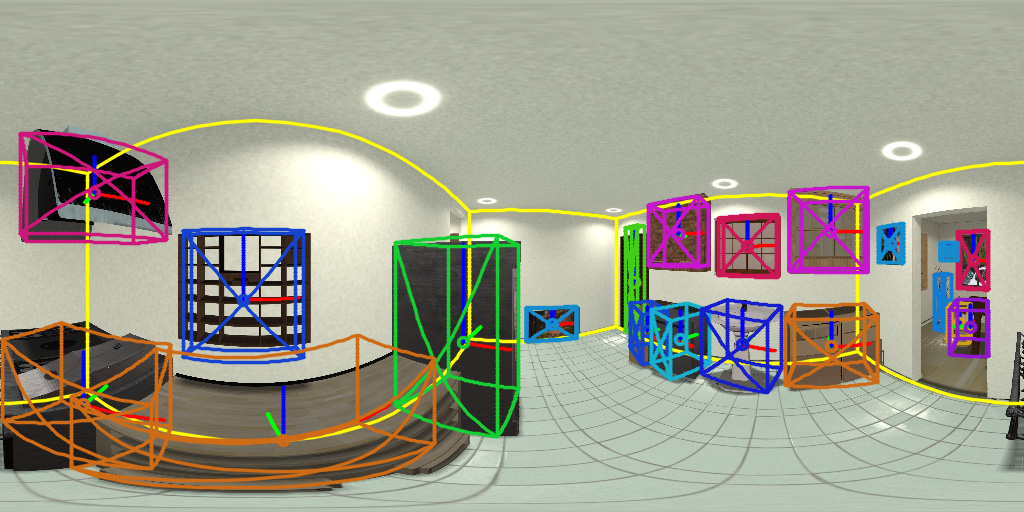}
        	    & \rgb{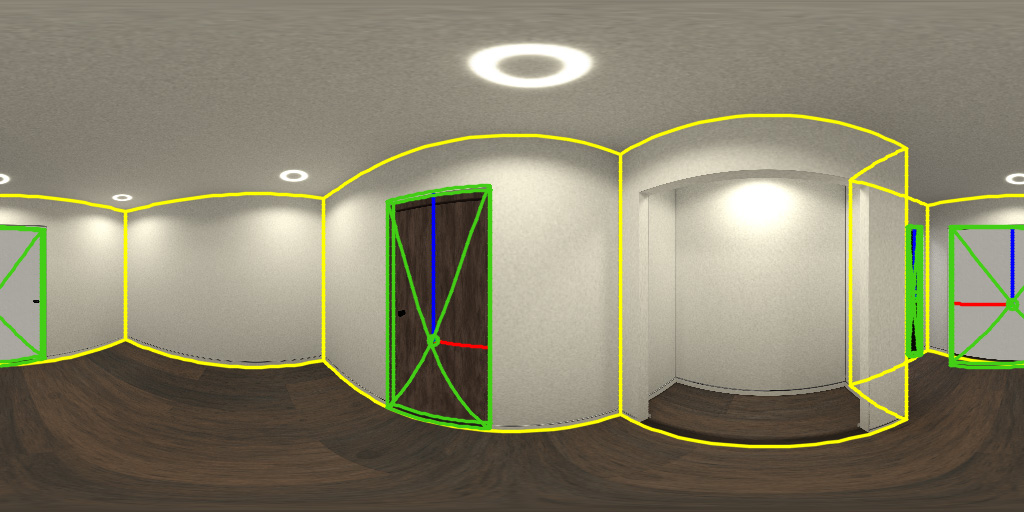}
        	    & \rgb{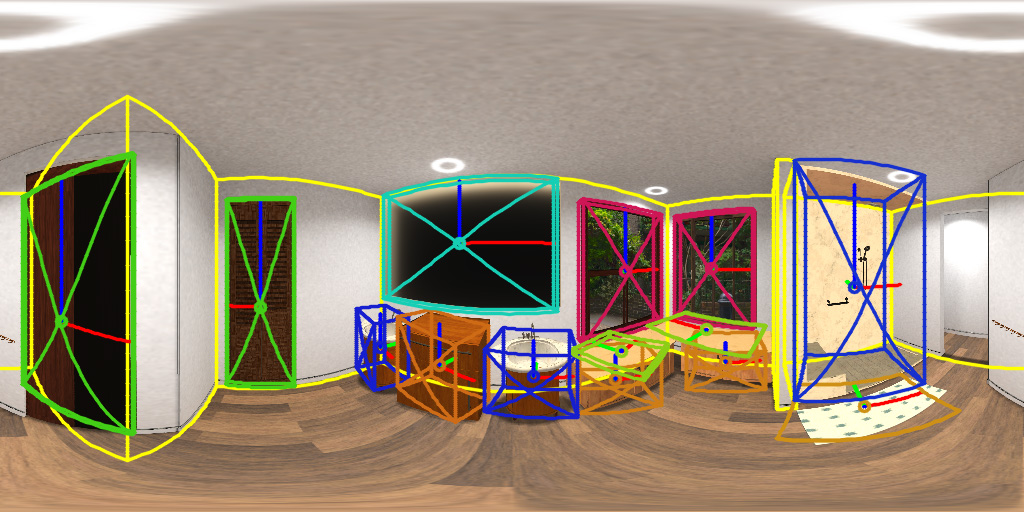}
        	           	    \\
        	&\rot{GT}
                & \rgb{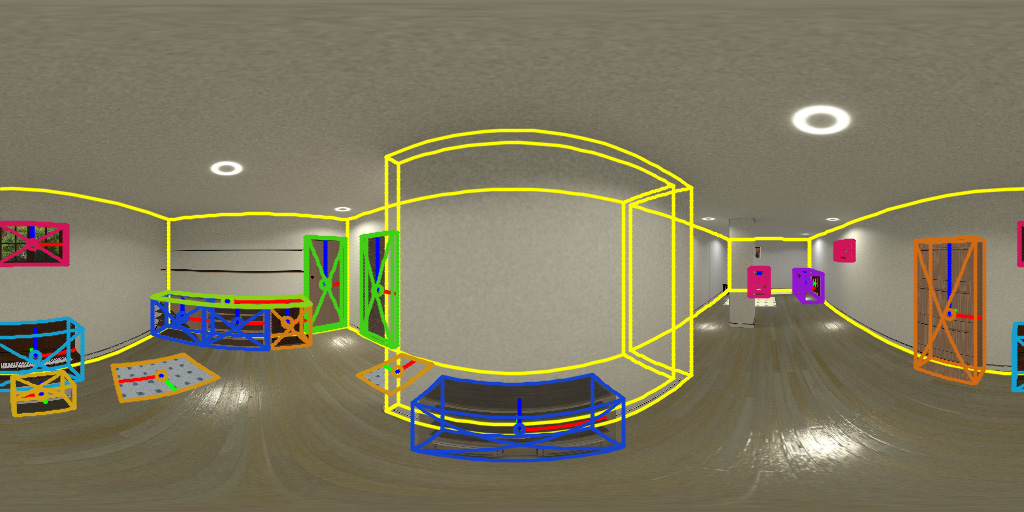}
        	    & \rgb{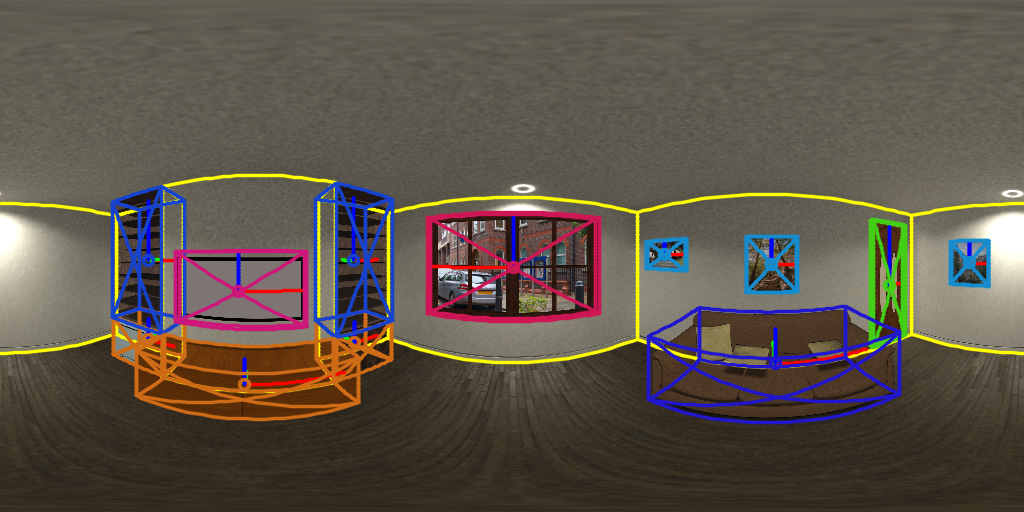}
        	    & \rgb{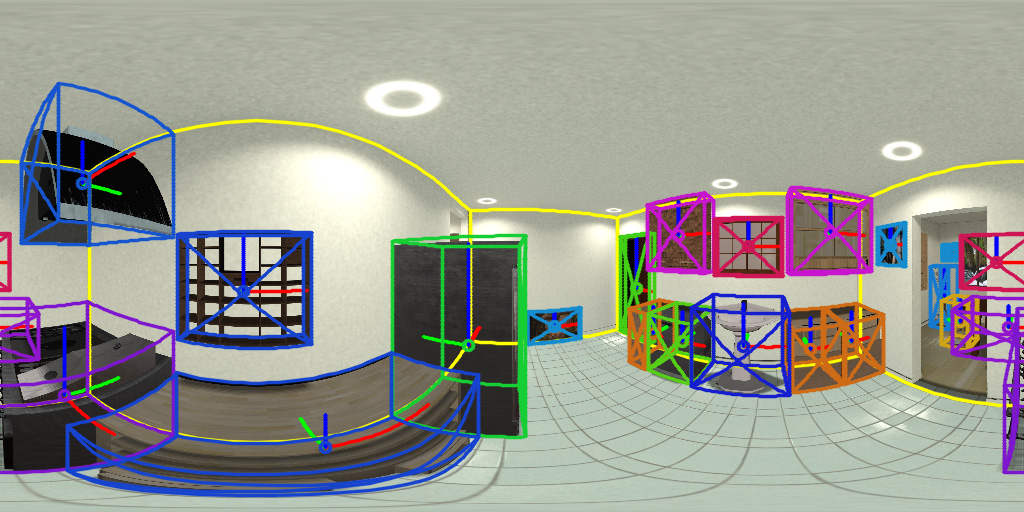}
        	    & \rgb{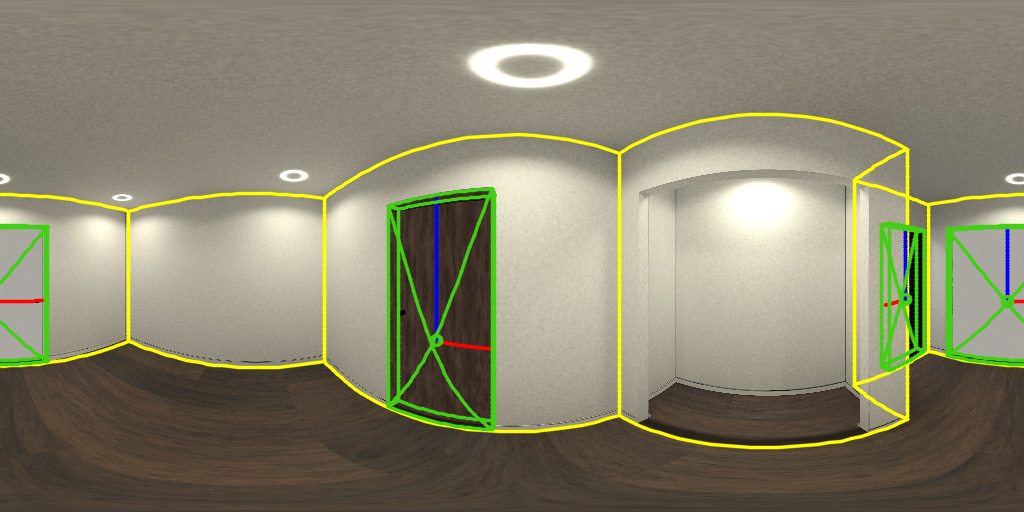}
        	    & \rgb{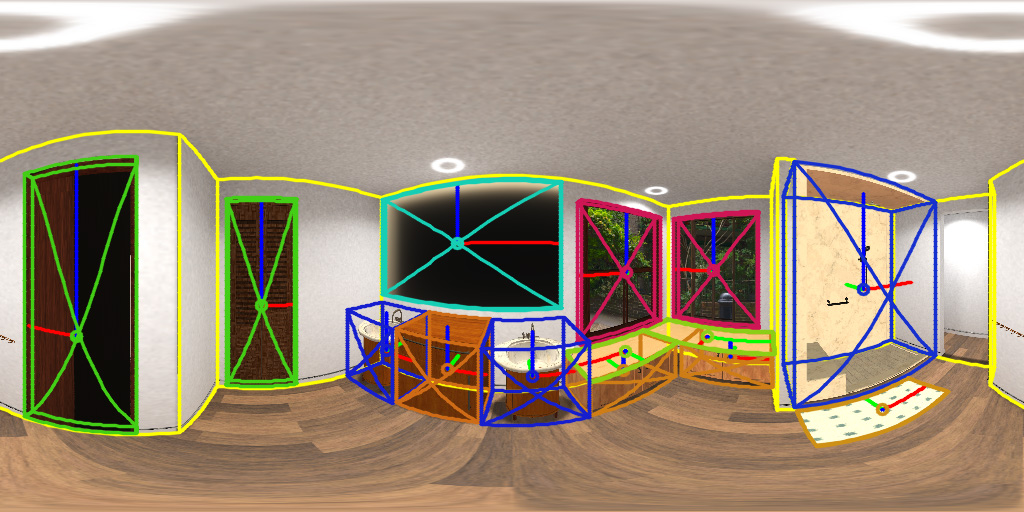}
        	           	    \\
    	\hline
    	    \multirow{2}{*}[3ex]{\rot{Scene Reconstruction}}
            &\rot{Ours}
                & \rgb{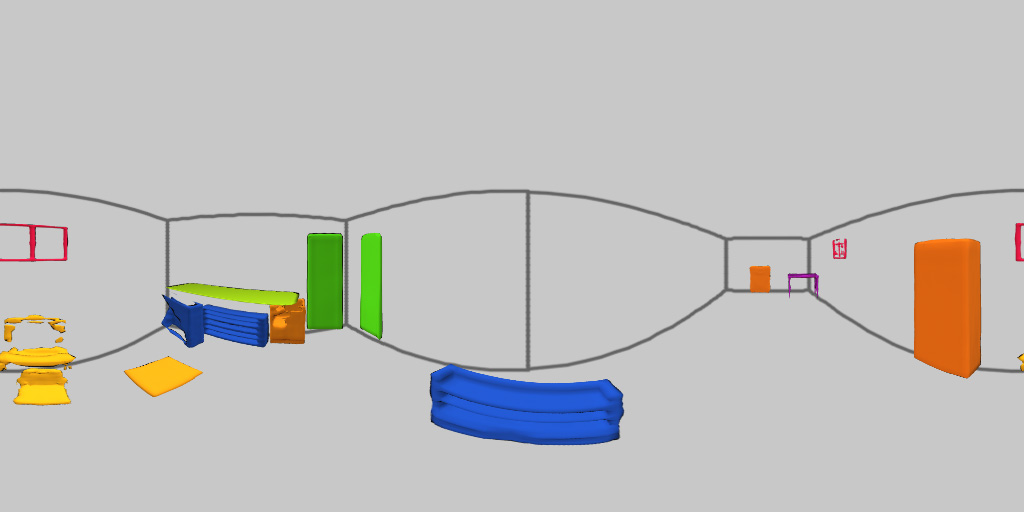}
        	    & \rgb{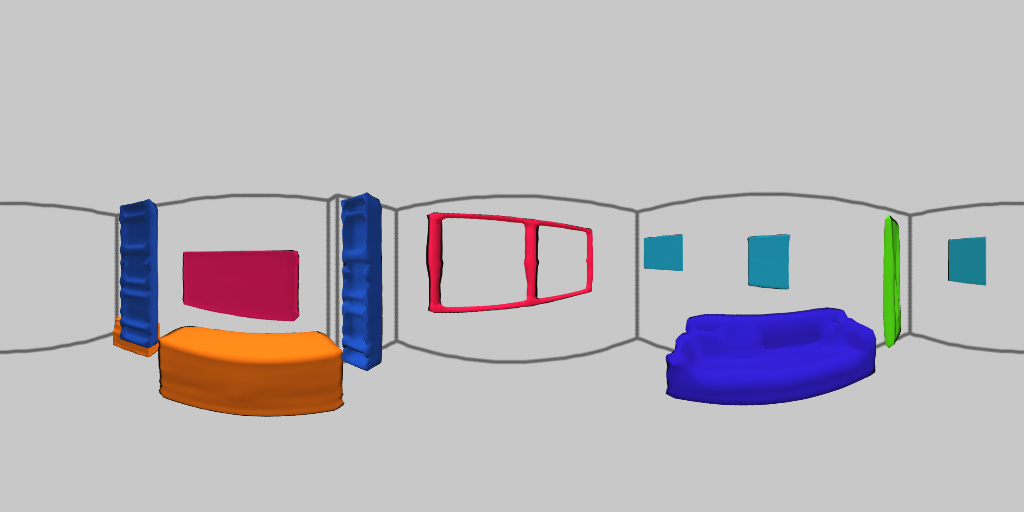}
        	    & \rgb{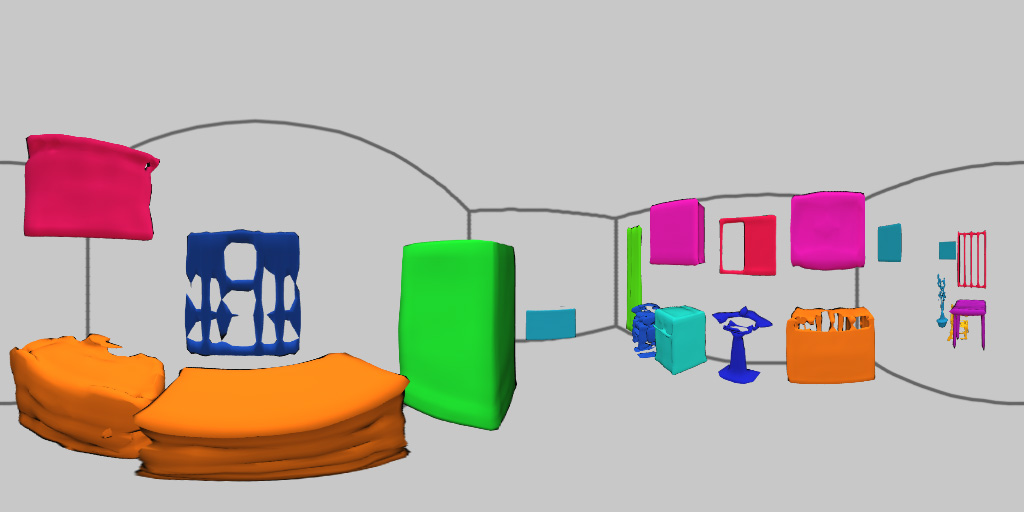}
        	    & \rgb{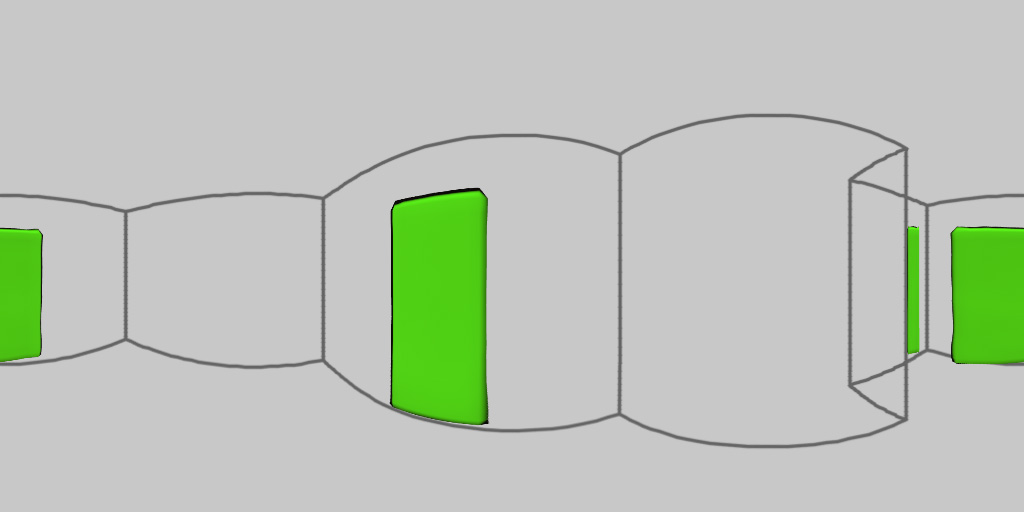}
        	    & \rgb{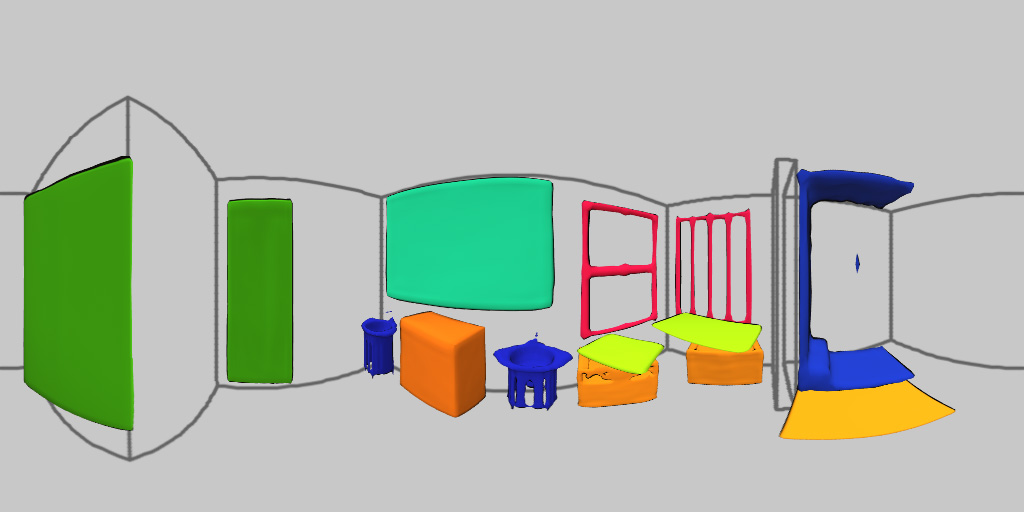}
        	           	    \\
        	&\rot{GT}
                & \rgb{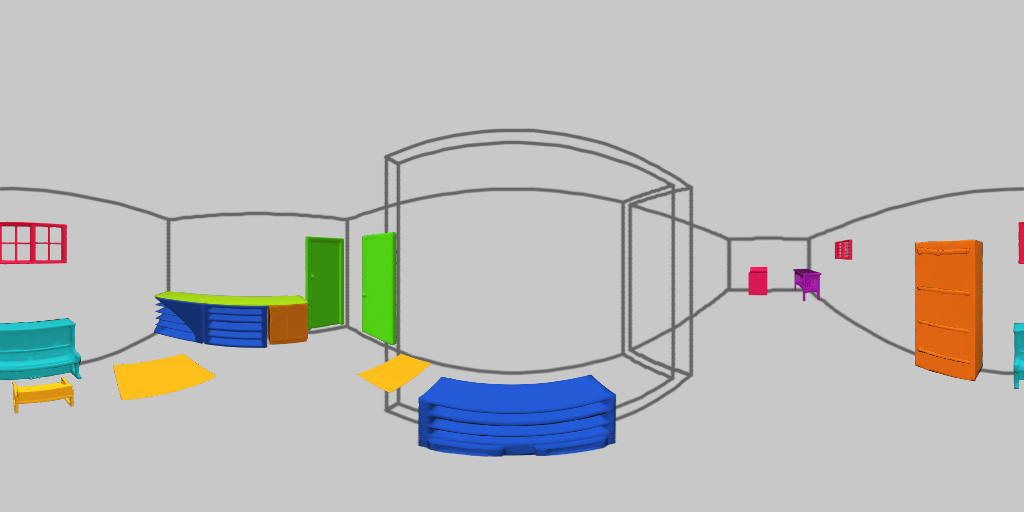}
        	    & \rgb{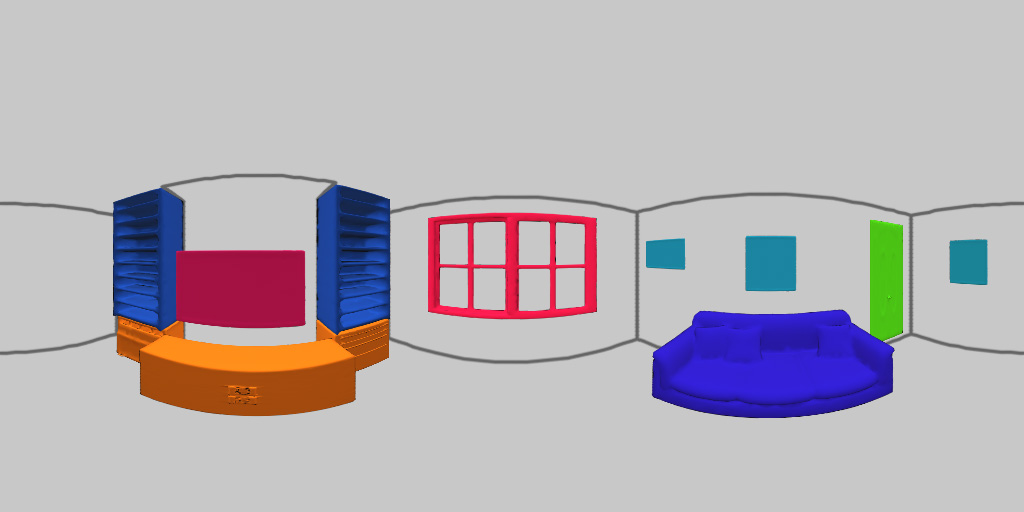}
        	    & \rgb{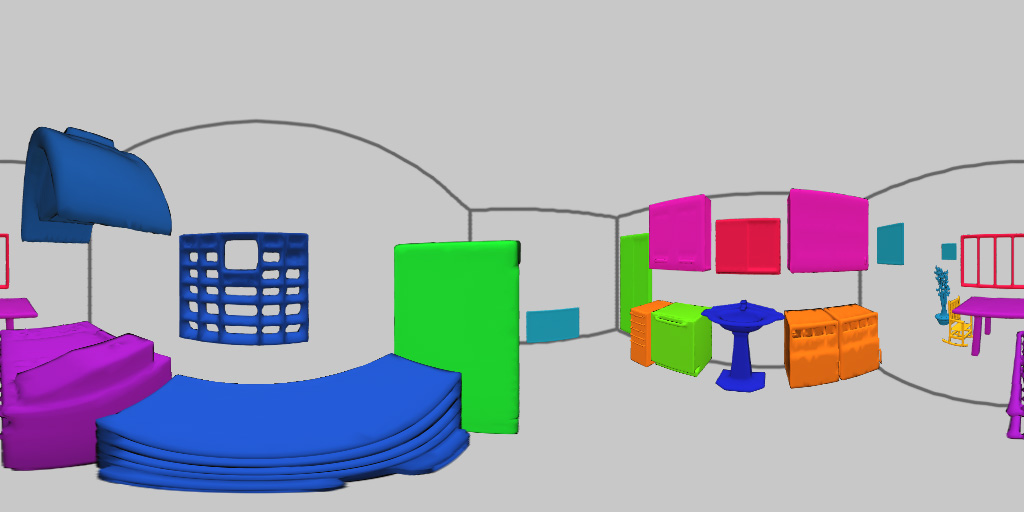}
        	    & \rgb{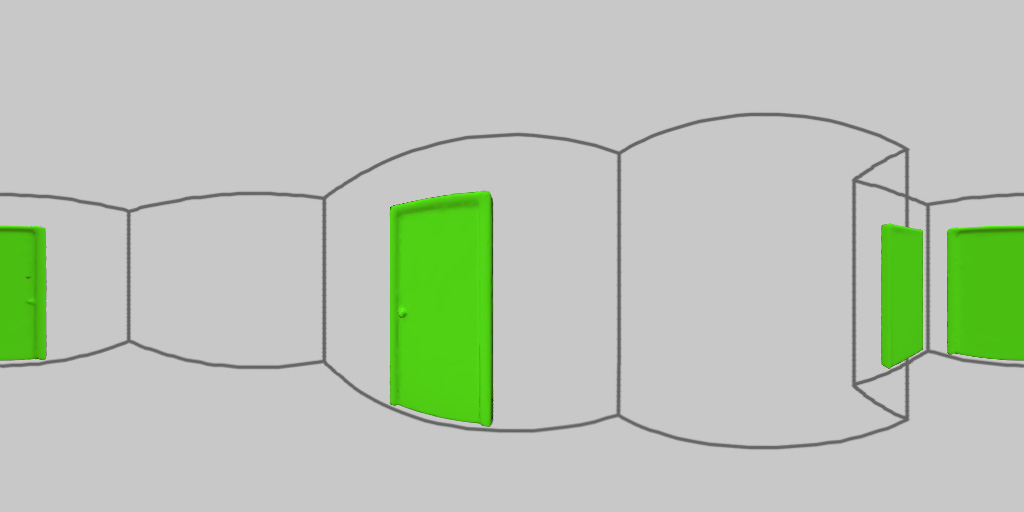}
        	    & \rgb{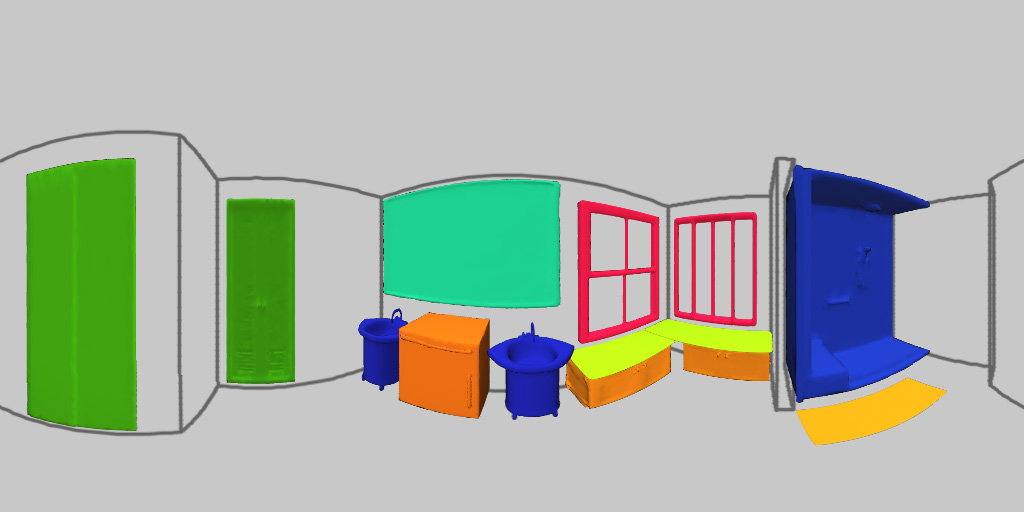}
        	           	    \vspace{2px}
        	    \\
        \multicolumn{2}{c}{}& (a) & (b) & (c) & (d) & (e) \\
    \end{tabular}
    	\caption{Failure cases.}
	\label{fig:failure}
\end{figure*}

We show failure cases in \autoref{fig:failure}.
One scenario that our pipeline fails is when heavy occlusion happens (\ie, one of the doors on the right in (d), the second door on the left in (a)), which
\yz{tends to shrink the size of the object in order to favor the projection term with the partial 2D observation.}
A possible solution might be to 
\yz{understand the occlusion and learn the mask behind occluder.}
Another scenario is when the 2D detector has multiple detection results on a single object (\ie, the wardrobe on the right in (a), the sofa on the right in (b), the drawer on the left in (c)), which lead to two overlapped object reconstructions in the same place \yz{but not sufficient to trigger non-maximum suppression}.
This might be solved by refining the category prediction of the 2D detector in the RGCN, which will presumably fix detected object categories with mistakes (or set reduplicated object to void) with a better understanding of the 3D scene context.
The last scenario is when HorizonNet fails to generate layouts for rooms that don't satisfy the Manhattan-world assumption (\ie, the wall on the left side in (e)), our pipeline will fail to optimize the object pose based on the wrong wall orientation.
Also when object-wall rotation relation is estimated badly (\ie, the window in (b)), the orientation cannot be optimized properly.

\begin{table*}[!h]
	\begin{center}
	    \resizebox{2.095\columnwidth}{!}{
    		\begin{tabular}{|l|c|c|c|c|c|c|c|c|c|c|c|c|c|}
    			\hline
    			Method          & chair & sofa & table & fridge & sink & door & \makecell[c]{floor \\ lamp} & \makecell[c]{bottom \\ cabinet} & \makecell[c]{top \\ cabinet} & \makecell[c]{sofa \\ chair} & dryer \\
    			\hline
    			\yz{Total3D-Pers}  & 13.71 & 68.06 & 30.55 & 36.02 & 69.84 & 11.88 & 12.57 & 35.56 & 19.19 & 64.29 & 41.36 \\     			\yz{Total3D-Pano}  & 20.84 & 69.65 & 31.79 & 43.13 & 68.42 & 10.27 & 16.42 & 34.42 & 20.83 & 62.38 & 33.78 \\     			\yz{Im3D-Pers}     & 30.23	& \textbf{75.23} & 44.16 & 52.56 & 76.46 & 14.91 & 9.99 & 45.51 & 23.37 & \textbf{80.11} & 53.28  \\     			\yz{Im3D-Pano}     & 33.08 & 72.15 & 37.43 & 70.45 & 75.20 & 11.58 & 6.06 & 43.28 & 18.99 & 78.46 & 41.02 \\     			Ours (w/o. RO)  & \textbf{33.57} & 75.18 & 38.65 & 71.97 & \textbf{80.66} & 19.94 & 18.29 & 50.67 & 29.05 & 79.42 & \textbf{60.07}\\     			Ours (Full)       & 27.78 & 73.96 & \textbf{46.85} & \textbf{74.22} & 75.29 & \textbf{21.43} & \textbf{20.69} & \textbf{52.03}  & \textbf{50.39} & 77.09 & 59.91 \\     			\hline\hline
    			Method          & window & carpet & picture & oven & \makecell[c]{bottom \\ cabinet \\ no top} & counter & \makecell[c]{dish \\ washer} & shelf & \makecell[c]{coffee \\ table} & mirror & toilet \\
    			\hline
    			\yz{Total3D-Pers}  & 2.92 & 0.05 & 0.01 & 31.33 & 34.40 & 0.78 & 43.54 & 10.93 & 39.72 & 0.11 & 90.00 \\
    			\yz{Total3D-Pano}  & 3.07 & 0.05 & 0.02 & 29.81 & 32.48 & 1.11 & 48.39 & 9.57 & 49.52 & 0.64 & 90.00 \\
    			\yz{Im3D-Pers}     & 3.52 & 0.12 & 0.00 & 31.28 & \textbf{47.45} & \textbf{2.60} & 51.47 & 15.01 & 59.02 & 0.81 & 90.00 \\
    			\yz{Im3D-Pano}     & 3.42 & 0.01 & 0.01 & 29.06 & 44.79 & 1.34 & 43.80 & 15.41 & 56.82 & 0.16 & 90.00 \\
    			Ours (w/o. RO)  & 6.94 & 0.12 & 0.03 & 32.52 & 46.42 & 1.83 & 59.78 & 15.58 & \textbf{61.17} & 2.42 & 90.00 \\
    			Ours (Full)     & \textbf{9.56} & \textbf{0.65} & \textbf{0.21} & \textbf{34.50} & 44.17 & 1.25 & \textbf{63.19} & \textbf{22.65} & 50.69 & \textbf{6.12} & 90.00 \\
    			\hline\hline
    			Method          & \makecell[c]{wall \\ mounted \\ tv} & \makecell[c]{loud \\ speaker} & \makecell[c]{console \\ table} & fence & chest & \makecell[c]{standing \\ tv} & \makecell[c]{table \\ lamp} & \makecell[c]{speaker \\ system} & bathtub & plant & treadmill\\
    			\hline
    			\yz{Total3D-Pers}  & \textbf{0.52} & 0.00 & 0.00 & 0.00 & 0.00 & 0.00 & 0.00 & 0.00 & 0.77 & 11.48 & 0.00 \\
    			\yz{Total3D-Pano}  & 0.24 & 0.00 & 0.00 & 0.00 & 0.00 & 0.00 & 0.00 & 0.00 & 0.77 & 3.10 & 0.00 \\
    			\yz{Im3D-Pers}     & 0.03 & 0.00 & 0.00 & 0.00 & 0.00 & 0.00 & 6.06 & 0.00 & 10.26 & 10.34 & 0.00 \\
    			\yz{Im3D-Pano}     & 0.08 & 0.00 & 0.00 & 0.00 & 0.00 & 0.00 & 3.17 & 0.00 & 10.26 & 12.69 & 0.00 \\
    			Ours (w/o. RO)  & 0.24 & 0.00 & 0.00 & 0.00 & 0.00 & 0.00 & \textbf{10.53} & 0.00 & 10.26 & 8.35 & 0.00 \\
    			Ours (Full)     & 0.14 & 0.00 & 0.00 & 0.00 & 0.00 & 0.00 & 2.79 & 0.00 & \textbf{41.02} & \textbf{16.46} & 0.00 \\
    			\hline\hline
    			Method          & washer & stool & \makecell[c]{trash \\ can} & stove & bed & \makecell[c]{office \\ chair} & shower & \makecell[c]{towel \\ rack} & piano & mAP & \\
    			\hline
    			\yz{Total3D-Pers}  & 35.06 & 29.09 & 24.45 & 44.44 & 71.87 & 0.00 & \textbf{100.00} & 25.00 & 55.56 & 25.11 & \\
    			\yz{Total3D-Pano}  & 32.21 & 29.09 & 25.84 & 44.44 & \textbf{73.22} & 0.00 & 72.73 & \textbf{50.00} & 75.00 & 25.79 & \\
    			\yz{Im3D-Pers}     & \textbf{36.50} & 29.09 & 22.02 & 44.44 & \textbf{73.22} & 0.00 & 81.82 & \textbf{50.00} & 83.33 & 29.86 & \\
    			\yz{Im3D-Pano}     & \textbf{36.50} & 29.09 & 39.13 & 44.44 & \textbf{73.22} & 0.00 & 80.17 & 0.00 & 43.33 & 27.25 & \\
    			Ours (w/o. RO)  & \textbf{36.50} & 29.09 & 31.15 & 44.44 & 71.57 & 0.00 & 81.82 & 0.00 & \textbf{100.00} & 30.91 & \\
    			Ours (Full)     & \textbf{36.50} & 29.09 & \textbf{66.23} & 44.44 & 71.57 & 0.00 & \textbf{100.00} & 0.00 & \textbf{100.00} & \textbf{33.59} & \\
    			\hline
    		\end{tabular}
        }    	
	\end{center}
	\caption{3D object detection comparison on full 57 categories. Some categories existing in training scenes do not exist in testing scenes, or vice versa, which is the main reason for some of the 0 mAP cases.}
	\label{tbl:3d_detection_supp}
\end{table*}

\end{document}